\documentclass{article}
\usepackage[margin=1in]{geometry}
\usepackage{authblk}

\usepackage{thmtools, thm-restate}  
\usepackage{hyperref}
\usepackage{url}

\usepackage{microtype}
\usepackage{graphicx}
\usepackage{float}
\usepackage{booktabs} 
\usepackage{bbding}

\usepackage{hyperref}
\usepackage{array}
\newcolumntype{P}[1]{>{\centering\arraybackslash}p{#1}}

\usepackage{amsmath}
\usepackage{amssymb}
\usepackage{mathtools}
\usepackage{amsthm}
\usepackage[authoryear]{natbib}
\usepackage{enumerate}
\usepackage{bbm}

\usepackage{caption}

\usepackage{booktabs} 
\usepackage{tikz} 

\usetikzlibrary{shapes.geometric, arrows}

\tikzstyle{startstop} = [rectangle, rounded corners, 
minimum width=1cm, 
minimum height=0.4cm,
text centered, 
draw=black ]

\tikzstyle{io} = [trapezium, 
trapezium stretches=true, 
trapezium left angle=70, 
trapezium right angle=110, 
minimum width=3.0cm, 
minimum height=1cm, text centered, 
draw=black]

\tikzstyle{process} = [rectangle, 
minimum width=3.0cm, 
minimum height=1cm, 
text centered, 
text width=3.0cm, 
draw=black, 
fill]

\tikzstyle{decision} = [diamond, 
minimum width=3.0cm, 
minimum height=1cm, 
text centered, 
draw=black, 
fill=green!30]
\tikzstyle{arrow} = [thick,->,>=stealth]

\usepackage[american]{babel}
\usepackage{amsmath}
\usepackage{amsthm}
\usepackage{amssymb} 
\usepackage[T1]{fontenc} 
\usepackage[capitalise]{cleveref}
\usepackage{ifthen}

\usepackage[ ]{algorithmic}
\usepackage[vlined,ruled, ]{algorithm2e}

\usetikzlibrary{shapes,decorations,arrows,calc,arrows.meta,fit,positioning}
\tikzset{
    -Latex,auto,node distance =1 cm and 1 cm,semithick,
    state/.style ={ellipse, draw, minimum width = 0.7 cm},
    point/.style = {circle, draw, inner sep=0.04cm,fill,node contents={}},
    bidirected/.style={Latex-Latex,dashed},
    el/.style = {inner sep=2pt, align=left, sloped}
}

\makeatletter
\newcommand*{\centernot}{%
  \mathpalette\@centernot
}
\def\@centernot#1#2{%
  \mathrel{%
    \rlap{%
      \settowidth\dimen@{$\m@th#1{#2}$}%
      \kern.5\dimen@
      \settowidth\dimen@{$\m@th#1=$}%
      \kern-.5\dimen@
      $\m@th#1\not$%
    }%
    {#2}%
  }%
}
\makeatother

\newtheorem{theorem}{Theorem}

\newtheorem{Proposition}{Proposition}
\newtheorem{Lemma}{Lemma}

\newtheorem{assumption}{Assumption}

\newcommand{\indep}{\perp \!\!\! \perp}

\usepackage{tikz}

\usepackage[textsize=tiny]{todonotes}

\usepackage[utf8]{inputenc} 
\usepackage[T1]{fontenc}    
\usepackage{hyperref}       
\usepackage{url}            
\usepackage{booktabs}       
\usepackage{amsfonts}       
\usepackage{nicefrac}       
\usepackage{microtype}      
\usepackage{lipsum}
\usepackage{fancyhdr}       
\usepackage{graphicx}       
\graphicspath{{media/}}     

\usepackage{caption}
\usepackage{subcaption}
\usepackage{comment}

\usepackage{amsmath} 
\usepackage{xspace}

\usepackage{amsmath}
\usepackage{amssymb}
\usepackage{mathtools}
\usepackage{amsthm}
\theoremstyle{plain}

\usepackage{lipsum,lineno}
\usepackage{tabularray}

\usepackage[utf8]{inputenc} 
\usepackage[T1]{fontenc}    
\usepackage{hyperref}       
\usepackage{url}            
\usepackage{booktabs}       
\usepackage{amsfonts}       
\usepackage{nicefrac}       
\usepackage{microtype}      

\usepackage{amsmath}
\usepackage{amssymb}
\usepackage{mathtools}
\usepackage{amsthm}

\usepackage{xargs}                      
\usepackage{xcolor}

\usepackage{tikz}
\usetikzlibrary{positioning, calc, shapes.geometric, shapes, shapes.multipart, arrows.meta, arrows, decorations.markings, external, trees}

\usepackage{scalefnt}
\usetikzlibrary{shapes.misc}
\usetikzlibrary{positioning, calc, shapes.geometric, shapes, shapes.multipart, arrows.meta, arrows, decorations.markings, external, trees, fit}
\tikzset{
	-Latex,auto,node distance =1 cm and 1 cm,semithick,
	state/.style ={ellipse, draw, minimum width = 0.7 cm},
	point/.style = {circle, draw, inner sep=0.04cm,fill,node contents={}},
	nnv/.style={
		rectangle, draw,thick,minimum width=0.7cm,minimum height=1.5cm
	},
        nnh/.style={
            rectangle, draw,thick,minimum width=1.5cm,minimum height=1.0cm
          },
	outer/.style={draw=gray,dashed,thick, inner sep=3pt
	},
	XOR/.style={draw,circle,append after command={
			[shorten >=\pgflinewidth, shorten <=\pgflinewidth,]
			(\tikzlastnode.north) edge (\tikzlastnode.south)
			(\tikzlastnode.east) edge (\tikzlastnode.west)
		}
	},
	bidirected/.style={Latex-Latex,dashed},
	el/.style = {inner sep=2pt, align=left, sloped},
	cross/.style={cross out, draw=black, minimum size=2*(#1-\pgflinewidth), inner sep=0pt, outer sep=0pt},
	cross/.default={1pt}
}

\usepackage{natbib}
\usepackage{microtype}

\title{
Information-Theoretic Analysis of Multi-Objective Bandits: Pareto Violations \& Gap-Based Regret
}

\author{%
  Muhammad Qasim Elahi$^1$, Murat Kocaoglu$^2$, Mahsa Ghasemi$^1$ \\
       School of Electrical and Computer Engineering, Purdue University$^1$\\
       School of Computer Science, Johns Hopkins University$^2$\\

  \texttt{elahi0@purdue.edu, mkocaoglu@jhu.edu, mahsa@purdue.edu } \\
}

\usepackage{fancyhdr}
\begin{document}
\maketitle

\begin{abstract}
We study Bayesian multi-objective bandits and ask a basic information-theoretic question: what aspects of an unknown environment must be resolved to achieve low regret? We show that the answer depends on the regret criterion and on the resolution of information needed to represent the corresponding loss. We first consider Pareto-violation regret, which incurs unit loss whenever the selected action is dominated. Because this loss depends only on Pareto membership, the Pareto-optimal set provides a natural information target. We introduce a separation-based measure of the difficulty of distinguishing among competing candidate Pareto sets, relate it to information gain about the unknown Pareto-optimal set, and use this connection to analyze Thompson Sampling and develop an Information-Directed Sampling strategy. We establish sublinear Bayesian regret under uniform and polynomially decaying separation and prove a lower bound showing that weak separability can impose an unavoidable statistical cost. We then consider gap-based Pareto regret, which measures the magnitude of Pareto suboptimality. Here, Pareto-set information is generally too coarse for a direct magnitude-sensitive analysis: distinct environments can share the same Pareto set while inducing different reward geometries and hence different Pareto gaps, and averaging over environments consistent with the same Pareto set need not preserve the Pareto dominance relations. This motivates retaining finer information about the environment that preserves the relevant reward geometry. We develop parameter-level information-theoretic analyses of Thompson Sampling and Information-Directed Sampling to derive gap-based Pareto regret guarantees. Together, our results reveal a hierarchy in the information needed for multi-objective learning: binary Pareto membership can be controlled through information about the Pareto set, whereas quantitative Pareto-gap suboptimality can require a finer representation of reward geometry.
\end{abstract}

\section{Introduction}

Many real-world decision-making problems require balancing competing objectives, such as engagement, fairness, and diversity in recommendation systems \citep{abdollahpouri2020multistakeholder,ge2021towards,rodriguez2012multiple}, or treatment effectiveness, risks, and adverse outcomes in healthcare \citep{char2018implementing,komorowski2018artificial,durand2018contextual}. These settings are naturally modeled as multi-objective bandits, where each action yields a vector of rewards. The Pareto set consists of undominated actions. One action dominates another if it is at least as good in every objective and strictly better in at least one. Learning therefore involves navigating trade-offs across objectives \citep{turgay2018multi,roijers2013survey}.

Much of the work on multi-objective bandits evaluates performance through gap-based or related notions of Pareto regret, which quantify suboptimality relative to the Pareto frontier \citep{drugan2013designing,xu2023pareto,oner2018combinatorial,davoodi2025stochastic}. This line of work includes Pareto-UCB \citep{drugan2013designing} and subsequent analyses in stochastic, adversarial, combinatorial, and contextual settings \citep{xu2023pareto,oner2018combinatorial,park2025thompson}. For gap-based Pareto regret, the per-round Pareto gap is bounded above by the one-step regret under any normalized nonnegative linear scalarization. \citet{xu2023pareto} use a related bound for their Pareto regret on cumulative reward vectors and apply UCB to a fixed coordinate in the stochastic setting. Consider three arms with mean rewards $(1,0)$, $(0,1)$, and $(0.4,0.4)$. All three are Pareto optimal and have zero Pareto gap. Every normalized nonnegative linear scalarization ranks at least one extreme arm above the balanced arm \citep{drugan2013designing}. The balanced arm is also suboptimal in either coordinate, so fixed-coordinate UCB favors an extreme arm and may sample the balanced arm during exploration. If deployment later requires at least $0.3$ in both objectives, the balanced arm is the only feasible choice. We therefore seek to learn the full Pareto set while keeping gap-based Pareto regret low. Much less is understood about cumulative Bayesian Pareto-violation regret, which counts dominated pulls regardless of their gaps. We draw on the theory of Thompson Sampling for single-objective bandits \citep{thompson1933likelihood,agrawal2012analysis,agrawal2017near}. In that setting, information-theoretic analyses of Thompson Sampling and Information-Directed Sampling (IDS) relate regret to information about the unknown optimal action \citep{russo2014learning,russo2016information}. With vector rewards, a Pareto set represents optimality.

We study how the information relevant to low regret depends on the criterion. For Pareto-violation regret, the Pareto set provides a natural information target. We link predictive separation between competing Pareto sets to information gain about the true set. Building on this connection, we analyze Thompson Sampling and develop an IDS strategy that balances expected Pareto-violation loss against information gain about the set. Under uniform separation, both achieve sublinear Bayesian regret. When separation decays polynomially as $t^{-\beta}$, the default Thompson Sampling schedule gives sublinear regret for $\beta<1/2$, while IDS and separation-tuned Thompson Sampling extend the guarantee to $\beta<1$. We also establish an information-theoretic lower bound showing that weak predictive separation can force large finite-horizon violation regret for every adaptive policy.

For gap-based Pareto regret, where loss measures the magnitude of Pareto suboptimality, Pareto-set
identity alone does not determine the reward geometry governing the gaps. Different models can
share the same Pareto set yet have different reward geometries and Pareto gaps, and averaging across
such models need not preserve Pareto optimality. Thus, gap-based IDS trades expected Pareto gap
against information about the underlying model parameter rather than only about the Pareto set. We
develop a parameter-level analysis for Thompson Sampling and IDS and obtain gap-based Pareto
regret guarantees for standard finite-action Bayesian models. Together, these results show that Pareto
regret criteria can require different resolutions of information. The main contributions are as follow
\begin{itemize}

\item We study Pareto-violation regret and gap-based Pareto regret in Bayesian multi-objective bandits and show how the choice of loss changes the information targets and statistical difficulty of the learning problem.

\item For Pareto-violation regret, we introduce a predictive separation measure for competing Pareto sets and connect it to information gain. We establish sublinear Bayesian regret for Thompson Sampling and IDS under uniform and polynomially decaying separation and prove an information-theoretic lower bound in the weak-separation regime. We also establish generic pointwise separability under the stated conditions for finite discrete Bayesian models and give sufficient conditions based on distinguishable witness actions.

\item For gap-based Pareto regret, we derive parameter-level information-ratio bounds for Pareto Thompson Sampling and gap-based IDS. For bounded rewards and logarithmic information budgets, these bounds give near-$\sqrt{T}$ Bayesian gap regret at fixed model sizes.  Under bounded rewards and a sublinear parameter-information budget, a single run of Pareto Thompson Sampling achieves sublinear Bayesian gap regret and concentrates its posterior on the full true Pareto set. The joint guarantee extends to a fixed mixture with gap-based IDS that uses Pareto Thompson Sampling with positive probability.

\end{itemize}

\section{Problem Setup and Preliminaries}

We consider a stochastic multi-objective bandit in a Bayesian setting with a finite action set $\mathcal{A}$. Let $\boldsymbol{\theta}$ be an unknown possibly vector-valued parameter drawn from a known prior representing uncertainty about the environment. Conditioned on $\boldsymbol{\theta}$, at each round $t\in\mathbb{N}$ the learner selects an action $A_t\in\mathcal{A}$ and observes only the corresponding reward vector $\mathbf{Y}_{t,A_t}\in\mathbb{R}^d$. The sequence $(\mathbf{Y}_t)_{t\in\mathbb{N}}$, where $\mathbf{Y}_t=(\mathbf{Y}_{t,a})_{a\in\mathcal{A}}$, is i.i.d. conditional on $\boldsymbol{\theta}$. Define $\boldsymbol{\mu}_{\boldsymbol{\theta}}(a):=\mathbb{E}_{\boldsymbol{\theta}}[\mathbf{Y}_{t,a}]\in\mathbb{R}^d$. For $\mathbf{x},\mathbf{y}\in\mathbb{R}^d$, write $\mathbf{x}\succeq\mathbf{y}$ if $x_i\ge y_i$ for all $i\in[d]$. The Pareto-optimal set is $\mathcal{S}^*(\boldsymbol{\theta}):=\{a\in\mathcal{A}\mid \nexists\,a'\in\mathcal{A}\text{ such that }\boldsymbol{\mu}_{\boldsymbol{\theta}}(a')\succeq\boldsymbol{\mu}_{\boldsymbol{\theta}}(a)\text{ and }\boldsymbol{\mu}_{\boldsymbol{\theta}}(a')\neq\boldsymbol{\mu}_{\boldsymbol{\theta}}(a)\}$. We write $\mathcal{S}^*:=\mathcal{S}^*(\boldsymbol{\theta})$ for the random Pareto set induced by the prior over $\boldsymbol{\theta}$. This is the multi-objective analogue of the unknown optimal-action target $A^*$ used in single-objective information-theoretic bandit analyses \citep{russo2014learning}.

 We study two notions of regret. The first is Pareto-violation regret, with loss $\ell(a,\boldsymbol{\theta})=\mathbbm{1}\{a\notin\mathcal{S}^*(\boldsymbol{\theta})\}$. Its cumulative Bayesian regret over $T$ rounds is $\mathbb{E}\big{[}\sum_{t=1}^T\mathbbm{1}\{A_t\notin\mathcal{S}^*(\boldsymbol{\theta})\}\big{]}$, i.e., the expected number of times a dominated action is selected. The expectation is over the prior on $\boldsymbol{\theta}$, the observations, and any randomness in the policy. The second is gap-based Pareto regret using standard notion of uniform coordinate-wise Pareto gap~\citep{drugan2013designing,auer2016pareto,xu2023pareto}.  For each $a\in\mathcal{A}$, we define $\Delta_{\boldsymbol{\theta}}(a):=\sup\{\epsilon\ge0:\exists a'\in\mathcal{A}\text{ such that }\boldsymbol{\mu}_{\boldsymbol{\theta}}(a')\succeq\boldsymbol{\mu}_{\boldsymbol{\theta}}(a)+\epsilon\mathbf{1}\}$, where $\mathbf{1}\in\mathbb{R}^d$ denotes the all-ones vector.
 Its cumulative Bayesian regret is $\mathbb{E}\!\left[\sum_{t=1}^T\Delta_{\boldsymbol{\theta}}(A_t)\right]$. Every $a\in\mathcal{S}^*(\boldsymbol{\theta})$ satisfies $\Delta_{\boldsymbol{\theta}}(a)=0$, while $\Delta_{\boldsymbol{\theta}}(a)>0$ implies $a\notin\mathcal{S}^*(\boldsymbol{\theta})$. The converse need not hold because Pareto-violation regret records whether the selected action is dominated, whereas gap-based regret quantifies the magnitude of the largest simultaneous coordinate-wise improvement achievable over that action.  The action $A_t$ is chosen based on the history $\mathcal{H}_t=(A_1,\mathbf{Y}_{1,A_1},\ldots,A_{t-1},\mathbf{Y}_{t-1,A_{t-1}})$. A randomized policy $\pi=(\pi_t)_{t\in\mathbb{N}}$ maps each history $\mathcal{H}_t$ to a distribution over $\mathcal{A}$. Let $\Pi(\mathcal{A}):=\{p\in\mathbb{R}^{|\mathcal{A}|}_{+}:\sum_{a\in\mathcal{A}}p(a)=1\}$ denote the probability simplex over $\mathcal{A}$. With slight abuse of notation, we write $\pi_t(a):=\mathbb{P}(A_t=a\mid\mathcal{H}_t)$. We denote the posterior distribution of $\boldsymbol{\theta}$ given $\mathcal{H}_t$ by $p_t$, which induces the posterior $p_t(s):=\mathbb{P}(\mathcal{S}^*=s\mid\mathcal{H}_t)$ over the random Pareto set for $s\in2^{\mathcal{A}}$.

\section{Learning to Minimize Pareto-Violation Regret}
\label{sec:pareto_violation}

We first study sequential decision-making under Pareto violations, where selecting an action outside the Pareto-optimal set incurs unit loss. Since Pareto-violation loss depends only on Pareto membership, we use the unknown Pareto set $\mathcal{S}^*:=\mathcal{S}^*(\boldsymbol{\theta})$ as the information target and track its posterior uncertainty. The learner maintains $p_t(s):=\mathbb{P}(\mathcal{S}^*=s\mid\mathcal{H}_t)$ over candidate Pareto sets $s\in2^{\mathcal{A}}$. Define the disagreement probability $G_t:=\mathbb{P}(S\neq S'\mid\mathcal{H}_t)=1-\sum_{s\in2^{\mathcal{A}}}p_t(s)^2$, where $S,S'\stackrel{\mathrm{i.i.d.}}{\sim}p_t$ conditional on $\mathcal{H}_t$; hence $G_t>0$ precisely when multiple Pareto-set candidates remain plausible. For \(a\in\mathcal{A}\), define \(I_t(a):=I(\mathcal{S}^*;\mathbf{Y}_{t,a}\mid\mathcal{H}_t,A_t=a)\), and let \(I_t:=\sum_{a\in\mathcal{A}}\pi_t(a)I_t(a)\) denote the one-step information gain under the sampling distribution \(\pi_t\).
For $p_t(s)>0$, let $P_{t,s}^a:=\mathcal{L}(\mathbf{Y}_{t,a}\mid\mathcal{H}_t,A_t=a,\mathcal{S}^*=s)$ denote the posterior-predictive reward law under action $a$ and Pareto-set candidate $s$, where $\mathcal{L}(X\mid\mathcal{G})$ denotes the conditional law of $X$ given $\mathcal{G}$. Lemma \ref{lem:info_gain_tv} connects predictive distinguishability of competing candidate Pareto sets  to information gain about $\mathcal{S}^*$.

\begin{restatable}{Lemma}{lemInfoGainTV}
\label{lem:info_gain_tv}
\textbf{(Lower bound on information gain via average pairwise distinguishability)}
Fix a round $t$, any realizable history $\mathcal{H}_t$, and an action $a\in\mathcal{A}$. Let $S,S'\stackrel{\mathrm{i.i.d.}}{\sim}p_t$. Then
\begin{equation}
I(\mathcal{S}^*;\mathbf{Y}_{t,a}\mid\mathcal{H}_t,A_t=a)
\;\ge\;
\frac{1}{2}\,
\mathbb{E}_{S,S'\stackrel{\mathrm{i.i.d.}}{\sim}p_t}
\!\left[
\mathrm{TV}\!\left(P_{t,S}^a,P_{t,S'}^a\right)^2
\right].
\label{eq:info_gain_tv_lower_bound}
\end{equation}
\end{restatable}

Lemma~\ref{lem:info_gain_tv} shows that actions whose predictive reward laws better distinguish plausible Pareto-set candidates provide more information about the unknown Pareto set. To summarize this distinguishability across actions using a policy-independent reference measure, whenever $G_t>0$ define $\kappa_t(\mathcal{H}_t):=\mathbb{E}\!\left[\mathbb{E}_{a\sim\mathrm{Unif}(\mathcal{A})}\!\left[\mathrm{TV}(P^a_{t,S},P^a_{t,S'})^2\right]\mid S\neq S',\mathcal{H}_t\right]$. Thus, $\kappa_t(\mathcal{H}_t)$ captures the average squared predictive separation between competing Pareto-set candidates at history $\mathcal{H}_t$ and provides the separation measure used for the analyses below. Averaging Lemma~\ref{lem:info_gain_tv} uniformly over actions yields $\mathbb{E}_{a\sim\mathrm{Unif}(\mathcal{A})}[I_t(a)]\ge\frac{1}{2}G_t\kappa_t(\mathcal{H}_t)$, directly linking predictive separation to information gain. We next prove that $\kappa_t(\mathcal{H}_t)$ is generically nonzero in the finite discrete model.

\begin{theorem}
\label{thm:avgsep-allt-single}
Consider a multi-objective bandit model with finite discrete reward distributions and finite action set $\mathcal{A}$, leading to Dirichlet count updates in the posterior over $\boldsymbol{\theta}$. Let $S,S'\stackrel{\mathrm{i.i.d.}}{\sim}p_t$, where $p_t(s):=\mathbb{P}(\mathcal{S}^*=s\mid\mathcal{H}_t)$, and let $q\in\Pi(\mathcal{A})$ satisfy $q(a)>0$ for all $a\in\mathcal{A}$. Then, for all but a Lebesgue measure-zero set of Dirichlet prior parameters, for all $t$ and all realized histories $\mathcal{H}_t$,
\[
\mathbb{E}\!\left[
\mathbb{E}_{a \sim q}\left[
\mathrm{TV}\!\left(P^a_{t,S}, P^a_{t,S'}\right)^2
\right]
\;\middle|\; S \neq S',\mathcal{H}_t
\right]
> 0,
\]
whenever $\mathbb{P}(S \neq S' \mid \mathcal{H}_t) > 0$, under mild regularity and non-degeneracy conditions.
\end{theorem}

Theorem~\ref{thm:avgsep-allt-single} establishes generic predictive separability for finite discrete rewards with Dirichlet count updates: outside a measure-zero set of prior parameters, whenever multiple Pareto-set candidates remain plausible, their posterior-predictive reward laws cannot be indistinguishable under every action. In particular, taking $q=\mathrm{Unif}(\mathcal{A})$ gives $\kappa_t(\mathcal{H}_t)>0$, so Lemma~\ref{lem:info_gain_tv} implies that some action has strictly positive information gain. For bounded rewards, a related finite-observation construction follows by Bernoulliization: for $R_{\max}>0$ and $R\in[0,R_{\max}]^d$, generate $Z_j\mid R_j\sim\mathrm{Bernoulli}(R_j/R_{\max})$ independently across coordinates conditional on $R$. Since $\mathbb{E}[Z]=\mathbb{E}[R]/R_{\max}$, Pareto ordering of mean rewards is preserved up to positive scaling. Moreover, Bernoulliization is a fixed Markov kernel from $R$ to $Z$, so distinct posterior-predictive laws for $Z$ imply distinct laws for $R$. Hence, if the induced finite-output Bayesian model satisfies Theorem~\ref{thm:avgsep-allt-single}, its predictive-separability conclusion also certifies separation of the underlying reward laws. The formal regularity and non-degeneracy conditions and proof are deferred to the supplementary material, where Appendix~\ref{app:beta-bernoulli-example} verifies them explicitly for a two-action, two-objective Beta--Bernoulli model. Theorem~\ref{thm:avgsep-allt-single} guarantees only pointwise positivity of $\kappa_t(\mathcal{H}_t)$. Define the finite-horizon separation $\underline{\kappa}_T:=\inf_{\substack{1\le t\le T,\ \mathcal{H}_t\ \text{realizable}\\ G_t>0}}\kappa_t(\mathcal{H}_t)$. In the finite-action discrete model, only finitely many posterior count states are realizable up to any fixed $T$, so outside the Lebesgue measure-zero exceptional set, $\underline{\kappa}_T>0$ for every fixed $T$. The assumptions below quantify how this separation behaves over time.
\vspace{0.05em}
\begin{assumption}\textbf{(Uniform separation)}
\label{ass:uniform_sep}
There exists a constant $\kappa>0$ such that $\kappa_t(\mathcal{H}_t)\ge\kappa$ for every realizable history $\mathcal{H}_t$ satisfying $G_t>0$ and every $t\in[T]$.
\end{assumption}

\begin{assumption}\textbf{(Decaying separation)}
\label{ass:decay_sep}
There exist constants $c>0$ and $\beta\ge0$ such that $\kappa_t(\mathcal{H}_t)\ge c\,t^{-\beta}$ for every realizable history $\mathcal{H}_t$ satisfying $G_t>0$ and every $t\in[T]$.
\end{assumption}

Although $\kappa_t(\mathcal{H}_t)$ is history-dependent, Assumptions~\ref{ass:uniform_sep}--\ref{ass:decay_sep} require their bounds to hold over every realizable history $\mathcal{H}_t$ at every $t\in[T]:=\{1,\ldots,T\}$ for a finite horizon $T$. They are therefore model-level assumptions, independent of the particular algorithm. The quantities $\kappa_t(\mathcal{H}_t)$, $\kappa$, $c$, and $\beta$ are instance-dependent and need not be known by Algorithms~\ref{alg:ts_pareto}/\ref{alg:IDS}. The assumptions admit interpretable sufficient conditions. For bounded rewards, separation in one posterior-predictive mean coordinate yields an explicit lower bound on $\kappa_t(\mathcal{H}_t)$; see Proposition~\ref{prop:mean-separation} in Appendix~\ref{app:mean-separation}. More generally, if every competing posterior-supported Pareto-set pair admits, at each realizable $\mathcal{H}_t$ with $G_t>0$, a witness action whose predictive laws have TV distance atleast $\delta_t>0$, then $\kappa_t(\mathcal{H}_t)\ge\delta_t^2/|\mathcal{A}|$. Thus, $\delta_t\ge\delta>0$ implies Assumption~\ref{ass:uniform_sep}, while $\delta_t\ge\delta t^{-\gamma}$ implies Assumption~\ref{ass:decay_sep} with $c=\delta^2/|\mathcal{A}|$ and $\beta=2\gamma$. For example, consider a two-point prior over two environments such that action $a_0$ has deterministic reward $(1/2,1/2)$ in both, whereas $a_1$ yields $(Z,Z)$ with $Z\sim\mathrm{Bernoulli}(1/2-\delta)$ in one and $Z\sim\mathrm{Bernoulli}(1/2+\delta)$ in the other. The two environments induce different Pareto sets. For $a_0$, the predictive laws coincide and have TV distance $0$, while for $a_1$ their TV distance is $2\delta$. Uniformly averaging over two actions gives $\kappa_t(\mathcal{H}_t)=2\delta^2$ whenever both candidates remain plausible. Since $0<1/2\pm\delta<1$, neither environment is ruled out after any finite realizable history. Thus, Assumption~\ref{ass:uniform_sep} holds with $\kappa=2\delta^2$, and Assumption~\ref{ass:decay_sep} holds with $c=2\delta^2$ \& $\beta=0$.

Uniform separation keeps competing Pareto-set candidates statistically separated, whereas decaying separation allows their predictive laws to approach one another at a controlled rate.  These separation conditions characterize the model, but Algorithm~\ref{alg:ts_pareto} must still sample informative actions sufficiently often. At round $t$, Algorithm~\ref{alg:ts_pareto} draws $B_t\sim\mathrm{Bernoulli}(1-\varepsilon_t)$. If $B_t=1$, it samples $\boldsymbol{\theta}_t\sim p_t$, sets $S_t=\mathcal{S}^*(\boldsymbol{\theta}_t)$, and selects $A_t\sim\mathrm{Unif}(S_t)$; otherwise, it sets $S_t=\mathcal{A}$ and explores uniformly. Thus, with probability $1-\varepsilon_t$ it follows a posterior Pareto-set draw, and with probability $\varepsilon_t$ it explores uniformly. We analyze Bayesian cumulative Pareto-violation regret under Assumptions~\ref{ass:uniform_sep} and~\ref{ass:decay_sep}.

\begin{restatable}{theorem}{thmTS}
\label{thm:ts_regret}
The Bayesian cumulative Pareto-violation regret of Algorithm~\ref{alg:ts_pareto} satisfies:
\begin{itemize}
\item Under Assumption~\ref{ass:uniform_sep} (Uniform separation)
$\mathbb{E}\left[\sum_{t=1}^T \mathbbm{1}\{A_t \notin \mathcal{S}^*\}\right]
\le
\left(
2+\frac{2H(\mathcal{S}^*)}{\kappa}
\right)\sqrt{T}$.
\item Under Assumption~\ref{ass:decay_sep} (Decaying separation)
$\mathbb{E}\!\left[\sum_{t=1}^T \mathbbm{1}\{A_t \notin \mathcal{S}^*\}\right]
\le
2\sqrt{T}
+
\frac{2H(\mathcal{S}^*)}{c}\,T^{\beta+\frac12}$.
In particular, under Assumption~\ref{ass:decay_sep}, the regret is sublinear whenever $\beta<\tfrac12$.
\end{itemize}
\end{restatable}

Theorem~\ref{thm:ts_regret} shows that the schedule \(\varepsilon_t=t^{-1/2}\) yields \(O(\sqrt{T})\) expected Pareto-violation regret under uniform separation and sublinear regret under decaying separation whenever \(\beta<\tfrac12\). The proof bounds the one-step expected violation loss using the posterior disagreement \(G_t=\mathbb{P}(S\neq S'\mid\mathcal{H}_t)\) and relates it to information gain using Lemma~\ref{lem:info_gain_tv}, the separation condition, and explicit exploration. By the mutual-information chain rule, the cumulative expected information gain equals \(I(\mathcal{S}^*;\mathcal{H}_{T+1})\le H(\mathcal{S}^*)\), where \(H(\mathcal{S}^*)\) is the entropy of the random Pareto set induced by the prior through \(\boldsymbol{\theta}\mapsto\mathcal{S}^*(\boldsymbol{\theta})\); since \(\mathcal{S}^*\in2^{\mathcal{A}}\), \(H(\mathcal{S}^*)\le|\mathcal{A}|\log 2\). This information budget is independent of \(T\). Proposition~\ref{prop_ts_app} further shows that, when the separation profile is known, a tuned exploration schedule improves the rate to \(T^{(1+\beta)/2}\), which is sublinear for every \(\beta<1\); thus, the \(\beta<\tfrac12\) threshold is a consequence of the fixed exploration schedule.  Proposition~\ref{prop:ts_set_mismatch} strengthens the regret guarantee by showing that Algorithm~\ref{alg:ts_pareto} also asymptotically tracks the true Pareto set, i.e., \(\mathbb{E} [\sum_{t=1}^T\mathbbm{1}\{S_t\neq\mathcal{S}^*\}]=O(\sqrt{T})\) under Assumption~\ref{ass:uniform_sep} and \(O(T^{\beta+\frac12})\) under Assumption~\ref{ass:decay_sep}, which is sublinear for \(\beta<\tfrac12\). On TS rounds, \(S_t\) is a posterior draw of the Pareto set, while exploration rounds set \(S_t=\mathcal{A}\) and can themselves cause mismatch. Overall, however, the sampled set is incorrect on only a vanishing fraction of rounds. Full proofs are deferred to the supplementary material.

\begin{restatable}{Proposition}{propTSMismatch}
\label{prop:ts_set_mismatch}
For Algorithm~\ref{alg:ts_pareto}, the expected cumulative Pareto-set mismatch, defined as $\mathbb{E}\!\left[\sum_{t=1}^T\mathbbm{1}\{S_t\neq\mathcal{S}^*\}\right]$, satisfies $\mathbb{E}\!\left[\sum_{t=1}^T\mathbbm{1}\{S_t\neq\mathcal{S}^*\}\right]\le\left(2+\frac{2H(\mathcal{S}^*)}{\kappa}\right)\sqrt{T}$ under Assumption~\ref{ass:uniform_sep}, and $\mathbb{E}\!\left[\sum_{t=1}^T\mathbbm{1}\{S_t\neq\mathcal{S}^*\}\right]\le2\sqrt{T}+\frac{2H(\mathcal{S}^*)}{c}\,T^{\beta+\frac12}$ under Assumption~\ref{ass:decay_sep}.
\end{restatable}

\begin{algorithm}[h]
\caption{TS-Pareto with Uniform Exploration}
\label{alg:ts_pareto}
\KwIn{Prior $p_1$ over $\boldsymbol{\theta}$ and exploration schedule
$\{\varepsilon_t\}_{t\ge1}$ with $\varepsilon_t=t^{-1/2}$}

\For{$t=1,2,\dots,T$}{
    Draw $B_t\sim\mathrm{Bernoulli}(1-\varepsilon_t)$\\

    \eIf{$B_t=1$}{
        Sample $\boldsymbol{\theta}_t\sim p_t(\boldsymbol{\theta})$ and construct
        $S_t\gets\mathcal{S}^*(\boldsymbol{\theta}_t)$
        \tcp*{TS-Pareto branch}
    }{
        Set $S_t\gets\mathcal{A}$
        \tcp*{Uniform exploration branch}
    }

    Play $A_t\sim\mathrm{Unif}(S_t)$, observe $\mathbf{Y}_{t,A_t}$ \& Update posterior:
    $p_{t+1}(\boldsymbol{\theta})\propto p_t(\boldsymbol{\theta})\,\mathcal{L}(\mathbf{Y}_{t,A_t}\mid A_t,\boldsymbol{\theta})$.\\

}

\end{algorithm}

We also develop an information-directed sampling (IDS) strategy for the Pareto-violation setting (Algorithm~\ref{alg:IDS}). For each action $a$, define $\Delta_t(a)=1-\mathbb{P}(a\in\mathcal{S}^*\mid\mathcal{H}_t)=\mathbb{P}(a\notin\mathcal{S}^*\mid\mathcal{H}_t)$ as its posterior expected one-step Pareto-violation loss, and $I_t(a)=I(\mathcal{S}^*;\mathbf{Y}_{t,a}\mid\mathcal{H}_t,A_t=a)$ as the information gain about the unknown Pareto set. Algorithm~\ref{alg:IDS} selects a policy $\pi_t^{\mathrm{IDS}}$ minimizing the information ratio $\psi_t(\pi):=\Delta_t(\pi)^2/I_t(\pi)$, where $\Delta_t(\pi)=\mathbb{E}_{a\sim\pi}[\Delta_t(a)]$ and $I_t(\pi)=\mathbb{E}_{a\sim\pi}[I_t(a)]$. IDS then samples $A_t\sim\pi_t^{\mathrm{IDS}}$. Thus, IDS favors sampling distributions that balance low expected Pareto-violation loss against information gain about the unknown Pareto set. Unlike Algorithm~\ref{alg:ts_pareto}, IDS incorporates exploration directly through its objective and requires no separate exploration schedule. Its Bayesian cumulative Pareto-violation regret satisfies $\mathbb{E}\left[\sum_{t=1}^T\mathbbm{1}\{A_t\notin\mathcal{S}^*\}\right]=\mathbb{E}\left[\sum_{t=1}^T\Delta_t(\pi_t^{\mathrm{IDS}})\right]$, since $\Delta_t(\pi_t^{\mathrm{IDS}})$ is exactly the posterior expected one-step violation probability under policy $\pi_t^{\mathrm{IDS}}$.

\begin{restatable}{theorem}{thmTSids}
\label{thm:ts_regret_ids}
Assume that, for some $\Psi>0$, $\psi_t\!\left(\pi_t^{\mathrm{IDS}}\right)\le\Psi$ for all realizable $\mathcal{H}_t$ and $\forall t\in[T]$. Then, for Algorithm~\ref{alg:IDS}, the Bayesian cumulative Pareto-violation regret satisfies $\mathbb{E}\!\left[\sum_{t=1}^T\mathbbm{1}\{A_t\notin\mathcal{S}^*\}\right]=\mathbb{E}\!\left[\sum_{t=1}^T\Delta_t\!\left(\pi_t^{\mathrm{IDS}}\right)\right]\le\sqrt{\Psi\,H(\mathcal{S}^*)\,T}$. Furthermore, under Assumption~\ref{ass:uniform_sep}, $\psi_t\!\left(\pi_t^{\mathrm{IDS}}\right)\le\frac{8}{\kappa}$ for all $t\in[T]$ and $\mathbb{E}\!\left[\sum_{t=1}^T\mathbbm{1}\{A_t\notin\mathcal{S}^*\}\right]\le\sqrt{\frac{8}{\kappa}H(\mathcal{S}^*)\,T}$. Under Assumption~\ref{ass:decay_sep}, $\psi_t\!\left(\pi_t^{\mathrm{IDS}}\right)\le\frac{8}{c}t^\beta$ for all $t\in[T]$, and $\mathbb{E}\!\left[\sum_{t=1}^T\mathbbm{1}\{A_t\notin\mathcal{S}^*\}\right]\le\sqrt{\frac{8}{c}H(\mathcal{S}^*)\,T^{\beta+1}}$, which is sublinear whenever $\beta<1$.
\end{restatable}

Theorem~\ref{thm:ts_regret_ids} gives the Pareto-violation regret guarantee for Algorithm~\ref{alg:IDS}. The proof bounds the IDS information ratio via a two-action comparison distribution combining a low-loss action with an action informative about competing Pareto-set candidates, and then uses mutual-information chain rule and entropy budget. Under Assumption~\ref{ass:decay_sep}, IDS achieves the \(T^{(1+\beta)/2}\) rate and is therefore sublinear for every \(\beta<1\), without requiring a prescribed exploration schedule. By Proposition~\ref{prop_ts_app}, a separation-tuned TS schedule attains the same horizon exponent when the separation profile is known. IDS instead adapts its sampling distribution directly to the current loss-information tradeoff, at the cost of an additional optimization at each round \citep{russo2014learning}. We next show that the separation parameter captures an intrinsic statistical difficulty in the weak-separation regime.

\begin{restatable}{Proposition}{proplb}
\label{prop:lb}
For every $\kappa\in(0,1/8]$, there exists a two-armed, two-objective Bayesian bandit instance for which, at every finite realizable history $\mathcal{H}_t$ with $G_t>0$, the average squared-TV separation satisfies $\kappa_t(\mathcal{H}_t)=\kappa$. Furthermore, for every horizon $T\ge1$, every possibly randomized, adaptive algorithm satisfies $\mathbb{E}\!\left[\sum_{t=1}^T \mathbbm{1}\{A_t\notin\mathcal{S}^*(\boldsymbol{\theta})\}\right]\ge \frac{1}{32}\min\!\left\{T,\frac{1}{\kappa}\right\}$.
\end{restatable}

Proposition~\ref{prop:lb} shows that weak separation creates an intrinsic statistical cost: for every $\kappa\in(0,1/8]$, some instance forces every adaptive randomized algorithm to incur Bayesian Pareto-violation regret at least $\frac{1}{32}\min\{T,1/\kappa\}$. Hence, for $T\le 1/\kappa$ the lower bound is linear in $T$, reaching order $1/\kappa$ at the crossover $T\asymp1/\kappa$. The restriction $\kappa\le1/8$ isolates the weak-separation regime; when $\kappa>1/8$, $1/\kappa$ is bounded by a universal constant. Theorem~\ref{thm:ts_regret_ids} gives Pareto-IDS regret $O(\sqrt{H(\mathcal{S}^*)T/\kappa})$, while Proposition~\ref{prop_ts_app} gives the same $T$--$\kappa$ dependence for tuned Thompson Sampling, using $\varepsilon_t=\min\{\alpha/\sqrt{t},1\}$ with $\alpha=\sqrt{H(\mathcal{S}^*)/\kappa}$. At the crossover $T\asymp1/\kappa$, both upper bounds scale as $O(\sqrt{H(\mathcal{S}^*)}/\kappa)$ versus the lower bound $\Omega(1/\kappa)$, matching the $\kappa^{-1}$ dependence up to the factor $\sqrt{H(\mathcal{S}^*)}$ and universal constants; see Proposition~\ref{prop_ts_app} in Section~\ref{refined_ts_pv}.

\begin{algorithm}[h]
\caption{Pareto-IDS for Minimizing Pareto Violations}
\label{alg:IDS}
\KwIn{Prior $p_1$ over $\boldsymbol{\theta}$}

\For{$t=1,2,\dots,T$}{
    Observe history $\mathcal{H}_t$ and posterior $p_t(\boldsymbol{\theta})$, inducing
    $p_t(s)=\mathbb{P}(\mathcal{S}^*=s\mid\mathcal{H}_t)$ over $\mathcal{S}^*$\\

    Compute $\forall a\in\mathcal{A}$,
    $\Delta_t(a)=1-\mathbb{P}(a\in\mathcal{S}^*\mid\mathcal{H}_t)$
    and
    $I_t(a)=I(\mathcal{S}^*;\mathbf{Y}_{t,a}\mid\mathcal{H}_t,A_t=a)$\\

    Define $\Delta_t(\pi)=\mathbb{E}_{a\sim\pi}[\Delta_t(a)]$ and $I_t(\pi)=\mathbb{E}_{a\sim\pi}[I_t(a)]$ for $\pi\in\Pi(\mathcal{A})$.\\

    Choose $\pi^{\mathrm{IDS}}_t\in\arg\min_{\pi\in\Pi(\mathcal{A})}\frac{\Delta_t(\pi)^2}{I_t(\pi)}$ \tcp*[r]{See Section~\ref{optm_info_ratio}}

    Play $A_t\sim\pi^{\mathrm{IDS}}_t$, observe $\mathbf{Y}_{t,A_t}$, and update the posterior
    $p_{t+1}(\boldsymbol{\theta})\propto p_t(\boldsymbol{\theta})\,\mathcal{L}(\mathbf{Y}_{t,A_t}\mid A_t,\boldsymbol{\theta})$.
}

\end{algorithm}

\section{Learning to Minimize Gap-Based Pareto Regret}
\label{sec:gap_based}
We now turn to learning to minimize gap-based Pareto regret. Unlike the binary Pareto-violation objective in Section~3, $\mathbbm{1}\{a\notin\mathcal{S}^*(\boldsymbol{\theta})\}$, we consider the Pareto gap $\Delta_{\boldsymbol{\theta}}(a):=\sup\{\epsilon\ge0:\exists a'\in\mathcal{A}\text{ such that }\boldsymbol{\mu}_{\boldsymbol{\theta}}(a')\succeq\boldsymbol{\mu}_{\boldsymbol{\theta}}(a)+\epsilon\mathbf{1}\}$, which measures the magnitude of suboptimality and is standard in prior work on multi-objective bandits \citep{drugan2013designing,auer2016pareto,xu2023pareto}. Equivalently, $\Delta_{\boldsymbol{\theta}}(a)=\left(\sup_{b\in\mathcal A}\min_{j\in[d]}\bigl(\mu_{\boldsymbol{\theta},j}(b)-\mu_{\boldsymbol{\theta},j}(a)\bigr)\right)_+$, where $(x)_+:=\max\{x,0\}$ and $\mu_{\boldsymbol{\theta},j}(a)$ is the $j$-th coordinate of $\boldsymbol{\mu}_{\boldsymbol{\theta}}(a)$; see Proposition~\ref{lem:law-equiv}. This creates a key conceptual distinction from Section~3. There, the relevant unknown object is the Pareto-optimal set $\mathcal{S}^*(\boldsymbol{\theta})$, because conditional on $\mathcal S^*(\boldsymbol{\theta})=s$, the violation loss of action $a$ is exactly $\mathbbm{1}\{a\notin s\}$, regardless of which parameter inducing $s$ is the true environment. By contrast, gap-based regret depends on the reward geometry induced by $\boldsymbol{\theta}$ and is not determined by the Pareto set alone. Thus Pareto-set information can be too coarse for magnitude-sensitive regret analysis, and our analysis retains finer parameter-level information about the reward geometry. Importantly, the issue is not merely that the Pareto set fails to determine every numerical Pareto gap: in a single-objective bandit, the identity of the optimal action $A^*$ likewise does not determine every scalar gap. The key distinction is that averaging over parameter values consistent with the same Pareto set can alter the reward geometry and may even change the induced Pareto ordering required to represent gap-based regret itself.

This raises a natural question: why is the single-objective case different, even though $A^*$ likewise does not determine every scalar gap? In information-theoretic analyses of Thompson Sampling \citep{russo2014learning,russo2016information}, the relevant object is the optimal action $A^*:=A^*(\boldsymbol{\theta})$ (with fixed tie-breaking). There, the instantaneous regret of action $a$ is $\Delta_t(a):=\mathbb{E}[R_{t,A^*}-R_{t,a}\mid\mathcal{H}_t]$, which decomposes as $\Delta_t(a)=\sum_{a^*\in\mathcal{A}}\mathbb{P}_t(A^*=a^*)\,\mathbb{E}[\mu_{\boldsymbol{\theta}}(a^*)-\mu_{\boldsymbol{\theta}}(a)\mid A^*=a^*,\mathcal{H}_t]$. For each $a^*\in\mathcal{A}$, let $\Theta_{a^*}:=\{\boldsymbol{\theta}:A^*(\boldsymbol{\theta})=a^*\}$. Since $\mu_{\boldsymbol{\theta}}(a^*)\ge\mu_{\boldsymbol{\theta}}(a)$ for every $\boldsymbol{\theta}\in\Theta_{a^*}$ and all $a\in\mathcal{A}$, conditioning on $A^*=a^*$ preserves these inequalities: $\mathbb{E}[\mu_{\boldsymbol{\theta}}(a^*)\mid A^*=a^*,\mathcal{H}_t]\ge\mathbb{E}[\mu_{\boldsymbol{\theta}}(a)\mid A^*=a^*,\mathcal{H}_t]$. Thus, although scalar gap magnitudes may vary across parameters compatible with $A^*=a^*$, averaging over them preserves the optimality of the fixed comparator $a^*$, and conditional regret remains linear in mean rewards. Hence the $A^*$-level reduction is compatible with scalar information-theoretic analysis: conditioning on $A^*=a^*$ preserves both the comparator and the sign of every regret difference relative to it, while linearity keeps conditional regret aligned with the $A^*$-conditioned predictive model. This is precisely what can fail in the multi-objective setting. The Pareto set $\mathcal{S}^*(\boldsymbol{\theta})$ records which actions are non-dominated, but not the objective-wise tradeoffs responsible for their non-domination. Different parameter values can therefore induce the same Pareto set with different reward geometries, and averaging over them can alter the Pareto relations themselves. Thus conditioning on $\mathcal{S}^*$ does not generally preserve the structure needed for gap-based regret: unlike scalar regret, the Pareto gap is a nonlinear function of the reward geometry, $\Delta_{\boldsymbol{\theta}}(a)=\left(\sup_{b\in\mathcal A}\min_{j\in[d]}(\mu_{\boldsymbol{\theta},j}(b)-\mu_{\boldsymbol{\theta},j}(a))\right)_+$.

A simple example makes this instability concrete. Consider three actions $a_1,a_2,a_3$, two objectives, and two parameter values $\boldsymbol{\theta}_1,\boldsymbol{\theta}_2$, with $\boldsymbol{\mu}_{\boldsymbol{\theta}_1}(a_1)=(1,3)$, $\boldsymbol{\mu}_{\boldsymbol{\theta}_1}(a_2)=(3,1)$, $\boldsymbol{\mu}_{\boldsymbol{\theta}_1}(a_3)=(2.2,2.2)$, while $\boldsymbol{\mu}_{\boldsymbol{\theta}_2}(a_1)=(3,1)$, $\boldsymbol{\mu}_{\boldsymbol{\theta}_2}(a_2)=(1,3)$, and $\boldsymbol{\mu}_{\boldsymbol{\theta}_2}(a_3)=(2.2,2.2)$. In each environment no action dominates another, so $\mathcal{S}^*(\boldsymbol{\theta}_1)=\mathcal{S}^*(\boldsymbol{\theta}_2)=\{a_1,a_2,a_3\}$. Under the equal-weight average $\bar{\boldsymbol{\mu}}(a):=\tfrac12\boldsymbol{\mu}_{\boldsymbol{\theta}_1}(a)+\tfrac12\boldsymbol{\mu}_{\boldsymbol{\theta}_2}(a)$, however, $\bar{\boldsymbol{\mu}}(a_1)=\bar{\boldsymbol{\mu}}(a_2)=(2,2)$ whereas $\bar{\boldsymbol{\mu}}(a_3)=(2.2,2.2)$, so the Pareto set collapses to $\{a_3\}$. Indeed, in the averaged geometry, $a_1$ and $a_2$ have gap $0.2$, although both have zero gap in each original environment. Thus actions that are Pareto-optimal in every constituent environment can become dominated after averaging. Nor is this an isolated equal-weight phenomenon: mixing the environments with weights $\lambda$ and $1-\lambda$ gives $\bar{\boldsymbol{\mu}}_{\lambda}(a_1)=(3-2\lambda,1+2\lambda)$, $\bar{\boldsymbol{\mu}}_{\lambda}(a_2)=(1+2\lambda,3-2\lambda)$, and $\bar{\boldsymbol{\mu}}_{\lambda}(a_3)=(2.2,2.2)$; for every $\lambda\in(0.4,0.6)$, $a_3$ strictly dominates both $a_1$ and $a_2$. Hence an entire interval of mixtures changes the Pareto relations even though both underlying parameters induce exactly the same Pareto set. This example also explains why conditioning on a coarser information target can be problematic. Such coarse conditioning forms a predictive law whose mean averages reward geometries over parameters consistent with that event. In the scalar case, conditioning on $A^*=a^*$ preserves the optimality of the fixed comparator $a^*$ under this averaging. In contrast, conditioning on $\mathcal S^*=s$ can average multi-objective reward geometries in a way that changes their Pareto relations. Thus the problem is not merely that $\mathcal S^*$ does not determine the numerical values of $\Delta_{\boldsymbol{\theta}}(a)$: the Pareto gap of the averaged geometry need not equal the average of the parameter-specific Pareto gaps, because the gap functional is nonlinear in the reward geometry. This blocks a direct $A^*$-level reduction for gap-based Pareto regret and motivates retaining parameter-level information. Accordingly, for each $a\in\mathcal A$, define $I_t(a):=I(\boldsymbol{\theta};\mathbf{Y}_{t,a}\mid\mathcal{H}_t,A_t=a)$, and under the actual sampling distribution $\pi_t(a):=\mathbb{P}(A_t=a\mid\mathcal{H}_t)$, define $I_t:=\sum_{a\in\mathcal A}\pi_t(a)I_t(a)$. Also let $P_{\boldsymbol{\theta}}^a:=\mathcal{L}(\mathbf{Y}_{t,a}\mid\boldsymbol{\theta})$ denote the action-specific reward law. By retaining $\boldsymbol{\theta}$ rather than collapsing parameters that share the same Pareto set, the analysis preserves the reward geometry needed to control the actual Pareto gap, while Lemma~\ref{lem:info_gain_tv_theta} links greater distinguishability among the parameter-conditioned reward laws for action $a$ to greater information gain from playing it.

\begin{restatable}{Lemma}{lemInfoGainTVtheta}
\label{lem:info_gain_tv_theta}
Fix a round \(t\), a realizable history \(\mathcal{H}_t\), and an action
\(a\in\mathcal{A}\). Let \(p_t\) denote the posterior distribution of
\(\boldsymbol{\theta}\) given \(\mathcal{H}_t\), and let
\(\boldsymbol{\theta},\boldsymbol{\theta}'\stackrel{\mathrm{i.i.d.}}{\sim}p_t\).
Under the fixed-action interpretation,
\begin{equation}
I\!\left(
\boldsymbol{\theta};
\mathbf{Y}_{t,a}
\,\middle|\,
\mathcal{H}_t,A_t=a
\right)
\ge
\frac{1}{2}
\mathbb{E}_{
\boldsymbol{\theta},\boldsymbol{\theta}'
\stackrel{\mathrm{i.i.d.}}{\sim}p_t
}
\!\left[
\operatorname{TV}\!\left(
P_{\boldsymbol{\theta}}^a,
P_{\boldsymbol{\theta}'}^a
\right)^2
\right].
\label{eq:info_gain_tv_lower_bound_param}
\end{equation}
\end{restatable}

\begin{algorithm}[h]
\caption{Thompson Sampling for Gap-Based Pareto Regret}
\label{alg:ts_pareto_gap}
\KwIn{Prior \(p_1\) over \(\boldsymbol{\theta}\)}

\For{\(t=1,2,\dots,T\)}{
    Sample \(\boldsymbol{\theta}_t\sim p_t\) and construct
    \(S_t\gets\mathcal{S}^*(\boldsymbol{\theta}_t)\).
    
 Play $A_t\sim\mathrm{Unif}(S_t)$, observe $\mathbf{Y}_{t,A_t}$ \& Update posterior:
    $p_{t+1}(\boldsymbol{\theta})\propto p_t(\boldsymbol{\theta})\,\mathcal{L}(\mathbf{Y}_{t,A_t}\mid A_t,\boldsymbol{\theta})$.\\
}

\end{algorithm}

We propose Thompson Sampling for gap-based Pareto regret in Algorithm~\ref{alg:ts_pareto_gap}. At each round $t$, the algorithm samples $\boldsymbol{\theta}_t\sim p_t$, forms $S_t:=\mathcal{S}^*(\boldsymbol{\theta}_t)$, and selects $A_t\sim\mathrm{Unif}(S_t)$. Unlike in the Pareto-violation setting, the analysis operates at the parameter level and requires no explicit exploration schedule. Regret is controlled through the information budget $I(\boldsymbol{\theta};\mathcal{H}_{T+1})$, where $\mathcal{H}_{T+1}:=(A_1,\mathbf{Y}_{1,A_1},\ldots,A_T,\mathbf{Y}_{T,A_T})$ denotes the observed interaction history; this quantity measures how much the interaction reveals about the underlying environment. Unlike $I(\mathcal{S}^*;\mathcal{H}_{T+1})$ in Theorem~\ref{thm:ts_regret}, it need not admit a finite-entropy bound because $\boldsymbol{\theta}$ may be continuous. Nevertheless, for many standard Bayesian models, $I(\boldsymbol{\theta};\mathcal{H}_{T+1})$ grows only logarithmically in $T$, yielding near-$\sqrt{T}$ regret; model-specific bounds are given later in this section. Dependence on the Pareto-set size is captured by $M:=\sup_{\boldsymbol{\theta}\in\operatorname{supp}(p_1)}|\mathcal{S}^*(\boldsymbol{\theta})|$, the largest Pareto-set size over the prior support. Since $\mathcal{S}^*(\boldsymbol{\theta})\subseteq\mathcal{A}$, we have $M\le|\mathcal{A}|$, with equality possible in the worst case.

\begin{restatable}{theorem}{thmTSgap}
\label{thm:ts_regret_gap}
Under bounded rewards, for Algorithm~\ref{alg:ts_pareto_gap}, the Bayesian cumulative gap-based Pareto regret satisfies \(\mathbb{E}\!\left[\sum_{t=1}^T \Delta_{\boldsymbol{\theta}}(A_t)\right]=O\!\left(\sqrt{|\mathcal{A}|\,M\,\mathrm{IB}_T\,T}\right)\), where \(\mathrm{IB}_T:=I(\boldsymbol{\theta};\mathcal{H}_{T+1})\), \(\mathcal{H}_{T+1}:=(A_1,\mathbf{Y}_{1,A_1},\ldots,A_T,\mathbf{Y}_{T,A_T})\) denotes the observed interaction history up to horizon \(T\), and \(M:=\sup_{\boldsymbol{\theta}\in\operatorname{supp}(p_1)}|\mathcal{S}^*(\boldsymbol{\theta})|\le|\mathcal{A}|\). Since \(M\le|\mathcal{A}|\), the worst-case bound is \(O\!\left(|\mathcal{A}|\sqrt{\mathrm{IB}_T\,T}\right)\).
\end{restatable}

The proof of Theorem~\ref{thm:ts_regret_gap}, in the supplementary material, uses Lemmas~\ref{lem:theta_stability} and~\ref{lem:ts_gap_information}. Lemma~\ref{lem:theta_stability} relates the Pareto gap of an action \(a\in S_t=\mathcal{S}^*(\boldsymbol{\theta}_t)\) to total-variation separation between relevant observation laws by constructing a Pareto-optimal witness under the true parameter \(\boldsymbol{\theta}\). Lemma~\ref{lem:ts_gap_information} then combines this relation with the sampling probabilities and Lemma~\ref{lem:info_gain_tv_theta} to relate one-step expected regret to information gain. The main difficulty is that \(A_t\) is sampled from the random set \(S_t\), and is therefore coupled with the posterior sample \(\boldsymbol{\theta}_t\). Let \(q_t(a):=\mathbb{P}(a\in S_t\mid\mathcal{H}_t)\) and \(\pi_t(a):=\mathbb{P}(A_t=a\mid\mathcal{H}_t)\). Since \(|S_t|\le M\) and \(A_t\) is drawn uniformly from \(S_t\), \(\pi_t(a)\ge q_t(a)/M\), so \(q_t(a)/\pi_t(a)\le M\). Combining Lemma~\ref{lem:theta_stability}, this relation, Cauchy--Schwarz, and Lemma~\ref{lem:info_gain_tv_theta} yields \(\mathbb{E}[\Delta_{\boldsymbol{\theta}}(A_t)\mid\mathcal{H}_t]^2\le 32 R_{\max}^2|\mathcal{A}|M\sum_{a\in\mathcal{A}}\pi_t(a)I(\boldsymbol{\theta};\mathbf{Y}_{t,a}\mid\mathcal{H}_t,A_t=a)\). Summing over \(t\) and applying the mutual-information chain rule gives the stated regret bound. The additional \(\sqrt{M}\) factor arises from the relation \(\pi_t(a)\ge q_t(a)/M\); when \(M=1\), it disappears, leaving the familiar \(\sqrt{|\mathcal{A}|}\) dependence of information-theoretic Thompson Sampling bounds \citep{russo2014learning,russo2016information}.

\begin{restatable}{corollary}{corTSSetQuality}
\label{cor:ts_set_quality}
Under bounded rewards, for Algorithm~\ref{alg:ts_pareto_gap}, let \(S_t:=\mathcal{S}^*(\boldsymbol{\theta}_t)\) denote the sampled Pareto set at round \(t\). Then
\(
\mathbb{E}\!\left[\sum_{t=1}^T\sum_{a\in S_t}\Delta_{\boldsymbol{\theta}}(a)\right]
=
O\!\left(\sqrt{|\mathcal{A}|\,M^3\,\mathrm{IB}_T\,T}\right),
\)
where \(\mathrm{IB}_T:=I(\boldsymbol{\theta};\mathcal{H}_{T+1})\). In particular, since \(M\le|\mathcal{A}|\), the worst-case bound is \(O\!\left(|\mathcal{A}|^2\sqrt{\mathrm{IB}_TT}\right)\).
\end{restatable}

Corollary~\ref{cor:ts_set_quality} further controls the aggregate true Pareto gap of the sampled sets, complementing the action guarantee. Algorithm~\ref{alg:IDS_gap} is the IDS analogue for gap-based Pareto regret, using the same structure as Algorithm~\ref{alg:IDS} but different quantities in the information ratio. In the Pareto-violation setting, the loss depends only on whether an action belongs to \(\mathcal{S}^*\), so IDS trades off the posterior violation probability \(1-\mathbb{P}(a\in\mathcal{S}^*\mid\mathcal{H}_t)\) against information about \(\mathcal{S}^*\). In the gap-based setting, the loss is instead \(\Delta_{\boldsymbol{\theta}}(a)\), which depends on reward geometry encoded by \(\boldsymbol{\theta}\) and is not determined by \(\mathcal{S}^*\) alone. Accordingly, IDS uses the posterior expected gap \(\Delta_t(a):=\mathbb{E}[\Delta_{\boldsymbol{\theta}}(a)\mid\mathcal{H}_t]\) and parameter-level information gain \(I_t(a):=I(\boldsymbol{\theta};\mathbf{Y}_{t,a}\mid\mathcal{H}_t,A_t=a)\). For \(\pi\in\Pi(\mathcal{A})\), define \(\Delta_t(\pi):=\mathbb{E}_{a\sim\pi}[\Delta_t(a)]\), \(I_t(\pi):=\mathbb{E}_{a\sim\pi}[I_t(a)]\), and \(\psi_t(\pi):=\Delta_t(\pi)^2/I_t(\pi)\). Algorithm~\ref{alg:IDS_gap} selects \(\pi_t^{\mathrm{IDS}}\) minimizing \(\psi_t(\pi)\).

\begin{algorithm}[H]
\caption{IDS for Gap-Based Pareto Regret}
\label{alg:IDS_gap}
\KwIn{Prior \(p_1\) over \(\boldsymbol{\theta}\)}

\For{\(t=1,2,\dots,T\)}{
    Observe history \(\mathcal{H}_t\) and posterior \(p_t(\boldsymbol{\theta})\).

    Compute, for each \(a\in\mathcal{A}\),
    \(\Delta_t(a)=\mathbb{E}[\Delta_{\boldsymbol{\theta}}(a)\mid\mathcal{H}_t]\)
    and
    \(I_t(a)=I(\boldsymbol{\theta};\mathbf{Y}_{t,a}\mid\mathcal{H}_t,A_t=a)\).

    Define \(\Delta_t(\pi)=\mathbb{E}_{a\sim\pi}[\Delta_t(a)]\) and
    \(I_t(\pi)=\mathbb{E}_{a\sim\pi}[I_t(a)]\) for \(\pi\in\Pi(\mathcal{A})\).

    Choose
    \(\pi_t^{\mathrm{IDS}}\in\arg\min_{\pi\in\Pi(\mathcal{A})}\Delta_t(\pi)^2/I_t(\pi)\)
    \tcp*[r]{See Section~\ref{optm_info_ratio}}

    Play \(A_t\sim\pi_t^{\mathrm{IDS}}\), observe \(\mathbf{Y}_{t,A_t}\), and update
    \(p_{t+1}(\boldsymbol{\theta})\propto p_t(\boldsymbol{\theta})\,\mathcal{L}(\mathbf{Y}_{t,A_t}\mid A_t,\boldsymbol{\theta})\).
}

\end{algorithm}

\begin{restatable}{theorem}{thmIDSgap}
\label{thm:ids_regret_gap}
Assume that rewards are bounded and that, for some \(\Psi>0\), \(\psi_t(\pi_t^{\mathrm{IDS}})\le\Psi\) for every realizable $\mathcal{H}_t$, \(\forall t\in[T]\). Then Algorithm~\ref{alg:IDS_gap} satisfies
\(
\mathbb{E}\!\left[\sum_{t=1}^T\Delta_{\boldsymbol{\theta}}(A_t)\right]
\le
\sqrt{\Psi\,\mathrm{IB}_T\,T},
\)
where \(\mathrm{IB}_T:=I(\boldsymbol{\theta};\mathcal{H}_{T+1})\). Moreover, there exists a constant \(C>0\), depending only on the fixed reward bound, such that
\(
\psi_t(\pi_t^{\mathrm{IDS}})
\le
C|\mathcal{A}|M
\)
for all \(t\in[T]\). Consequently,
\(
\mathbb{E}\!\left[\sum_{t=1}^T\Delta_{\boldsymbol{\theta}}(A_t)\right]
=
O\!\left(\sqrt{|\mathcal{A}|\,M\,\mathrm{IB}_T\,T}\right)
\). Since \(M\le|\mathcal{A}|\), the worst-case bound is \(O\!\left(|\mathcal{A}|\sqrt{\mathrm{IB}_TT}\right)\).
\end{restatable}

The proof of Theorem~\ref{thm:ids_regret_gap}, given in the supplementary material, converts a uniform information-ratio bound into cumulative regret via Cauchy--Schwarz and the mutual-information chain rule. For an explicit bound, we compare IDS with the one-step Thompson-sampling distribution from Algorithm~\ref{alg:ts_pareto_gap}; since $\pi_t^{\mathrm{IDS}}$ minimizes the information ratio, its ratio is no larger than that of this feasible comparator, whose ratio is $O(|\mathcal{A}|M)$ by the one-step analysis underlying Theorem~\ref{thm:ts_regret_gap}.

\begin{restatable}{Proposition}{propIGBounds}
\label{prop:ig_bounds}
For finite-action Bayesian models, the following bounds hold for \(\mathrm{IB}_T:=I(\boldsymbol{\theta};\mathcal{H}_{T+1})\). For finite parameter space \(\Theta\): \(\mathrm{IB}_T\le H(\boldsymbol{\theta})\le\log|\Theta|\). In the multi-objective categorical reward model with \(d\) objectives and \(K\) categories per objective, suppose that \(\boldsymbol{\theta}\) consists of the categorical probability vectors and that each action--objective pair has an independent Dirichlet prior with parameters \((\alpha_{a,j,1},\ldots,\alpha_{a,j,K})\) satisfying \(\alpha_{a,j,k}\ge1\) for all \(a\in\mathcal{A}\), \(j\in[d]\), and \(k\in[K]\). Then \(\mathrm{IB}_T=O(dK|\mathcal{A}|\log T)\). In the shared-covariance Gaussian reward model of Lemma~\ref{lem:ig_gaussian}, let \(\boldsymbol{\theta}=(\{\mu_a\}_{a\in\mathcal{A}},\Sigma)\) and suppose that
\(\mathbf{Y}_{t,A_t}\mid(\boldsymbol{\theta},\mathcal{H}_t,A_t=a)\sim\mathcal{N}(\mu_a,\Sigma)\).
Suppose \(\Sigma\) has an inverse-Wishart prior and, conditional on \(\Sigma\), the arm means are independent with
\(\mu_a\mid\Sigma\sim\mathcal{N}(m_{a,1},\Sigma/\kappa_{a,1})\) for fixed \(m_{a,1}\in\mathbb{R}^d\) and \(\kappa_{a,1}>0\).
Then \(\mathrm{IB}_T=O((d|\mathcal{A}|+d^2)\log T)\).
\end{restatable}

 Proposition~\ref{prop:ig_bounds} supplies the information-budget bounds
needed in our regret analysis.
For bounded rewards, its logarithmic
budgets yield near-$\sqrt T$ gap-based Pareto regret through
Theorems~\ref{thm:ts_regret_gap} and~\ref{thm:ids_regret_gap}, with
model sizes fixed. For $Y_{t,j}\in[0,R_{\max}]$ and $R_{\max}>0$,
generating $Z_{t,j}\mid Y_{t,j}\sim
\mathrm{Bernoulli}(Y_{t,j}/R_{\max})$ preserves Pareto dominance
and scales every Pareto gap by $1/R_{\max}$. The analysis extends to the induced model when
Proposition~\ref{prop:ig_bounds} applies. For unbounded shared-covariance
Gaussian rewards, Appendix~\ref{app:gaussian_high_probability} conditions
the conjugate prior $p_0$ on the positive-probability event
$E=\{\max_{a,j}|\mu_{a,j}|\le\mu_{\max},\
\max_j\Sigma_{jj}\le\sigma^2\}$. For any fixed
$\delta\in(0,1)$, the thresholds $\mu_{\max},\sigma>0$
can be chosen independently of $T$ so that $p_0(E)\ge\delta$.
The conditioned prior
$p_1=p_0(\cdot\mid E)$ bounds all arm means and
coordinate variances. Conditioning and data processing transfer
the logarithmic budget of Lemma~\ref{lem:ig_gaussian} to the
clipped history, up to the fixed factor $1/p_0(E)\le1/\delta$.
Clipping each coordinate to $[-B_T,B_T]$, where
$B_T=\mu_{\max}+2\sigma\sqrt{\log(2T)}
=O(\sqrt{\log(1+T)})$, gives $o(1)$ cumulative gap bias.
The bounded-reward analysis gives
$O(\sqrt T\log(1+T))$ Bayesian gap regret for clipped TS and IDS
under $p_1$, evaluated using the original Gaussian means.

\begin{restatable}[Joint Sublinear Gap-Based Regret and Pareto-Set Recovery]
  {Proposition}{propTSGapSetRecovery}
\label{prop:ts_gap_set_recovery}
For Algorithm~\ref{alg:ts_pareto_gap} with exact posterior sampling,
assume that every reward coordinate has a finite conditional absolute
first moment for every arm, almost surely under the prior. Let
$e_t=\mathbb P(S_t\ne\mathcal S^*)$ and
$\bar e_T=T^{-1}\sum_{t=1}^T e_t$.
If $\mathrm{IB}_T=o(T)$, then $e_T\le\bar e_T\to0$, and the
posterior probability of the true Pareto set converges to one almost
surely. For the models in Proposition~\ref{prop:ig_bounds}, with
the model and prior fixed, $\bar e_T=O(T^{-1/2})$ in the
finite-parameter case and
$\bar e_T=O((\log(1+T)/T)^{1/4})$ in the categorical and conjugate
Gaussian cases, using raw observations under $p_0$ in the Gaussian
case. For the bounded-reward cases, combining
Theorem~\ref{thm:ts_regret_gap} with
Proposition~\ref{prop:ig_bounds} yields near-$\sqrt T$ Bayesian
gap regret alongside Pareto-set recovery. For shared-covariance
Gaussian rewards, under the conditioned prior
$p_1=p_0(\cdot\mid E)$, which bounds all arm means and coordinate
variances, Algorithm~\ref{alg:ts_pareto_gap} with clipped
observations at the threshold
$B_T=O(\sqrt{\log(1+T)})$ defined above achieves
both $e_T\le\bar e_T=O((\log(1+T)/T)^{1/4})$ and
$O(\sqrt T\log(1+T))$ Bayesian gap regret, with both guarantees
evaluated using the original, unclipped Gaussian means.
\end{restatable}

Since Algorithm~3 selects $A_t\in S_t$, we have $\mathbb{E}[\sum_{t=1}^T\mathbf{1}\{A_t\notin S^\ast\}]\leq T\bar e_T=o(T)$ whenever $\bar e_T\to0$, yielding sublinear Bayesian Pareto-violation regret without Assumptions~3.1--3.2, as a direct implication of Proposition~4.2.
 Fixed-weight scalarization and the stochastic fixed-coordinate UCB of
\citet{xu2023pareto} can control Pareto regret while leaving some
undominated trade-offs unidentified. Recovering the full Pareto- set
on the same run makes these options available when preferences are
specified after learning. The proof of Proposition \ref{prop:ts_gap_set_recovery} in Appendix~\ref{app:ts_gap_set_recovery}
shows that when the true and sampled Pareto sets disagree, some arm
in their union has different reward laws under the two models.
Posterior exchangeability and uniform sampling from the sampled set
let us use Lemma~\ref{lem:info_gain_tv_theta} to relate this discrepancy
to information gained from the arms actually played. Controlling the
prior mass of near ties and using the budgets in
Proposition~\ref{prop:ig_bounds} gives the recovery rates.
The regret bounds follow from Theorem~\ref{thm:ts_regret_gap} and
Appendix~\ref{app:gaussian_high_probability}. 
We also introduce a mixture (Algorithm~\ref{alg:pareto-ts-ids-mixture}) using Pareto TS with probability $\lambda\in(0,1]$ and exact gap-based IDS otherwise. Appendix~\ref{app:pareto-ts-ids-mixture} proves regret bound has no $1/\lambda$ dependence while full Pareto-set recovery holds. Pareto-set error is $O((\lambda T)^{-1/2})$ for finite priors and $O((\log(1+T)/(\lambda T))^{1/4})$ for categorical and truncated Gaussian models. Gap-regret rates remain $O(\sqrt{T})$, $O(\sqrt{T\log(1+T)})$, and $O(\sqrt{T}\log(1+T))$, respectively.

\section{Experiments}

\label{sec:experiments}

\begin{figure*}[h]
\centering
\begin{subfigure}[b]{4.4cm}
\includegraphics[height=3.0cm,width=4.4cm]{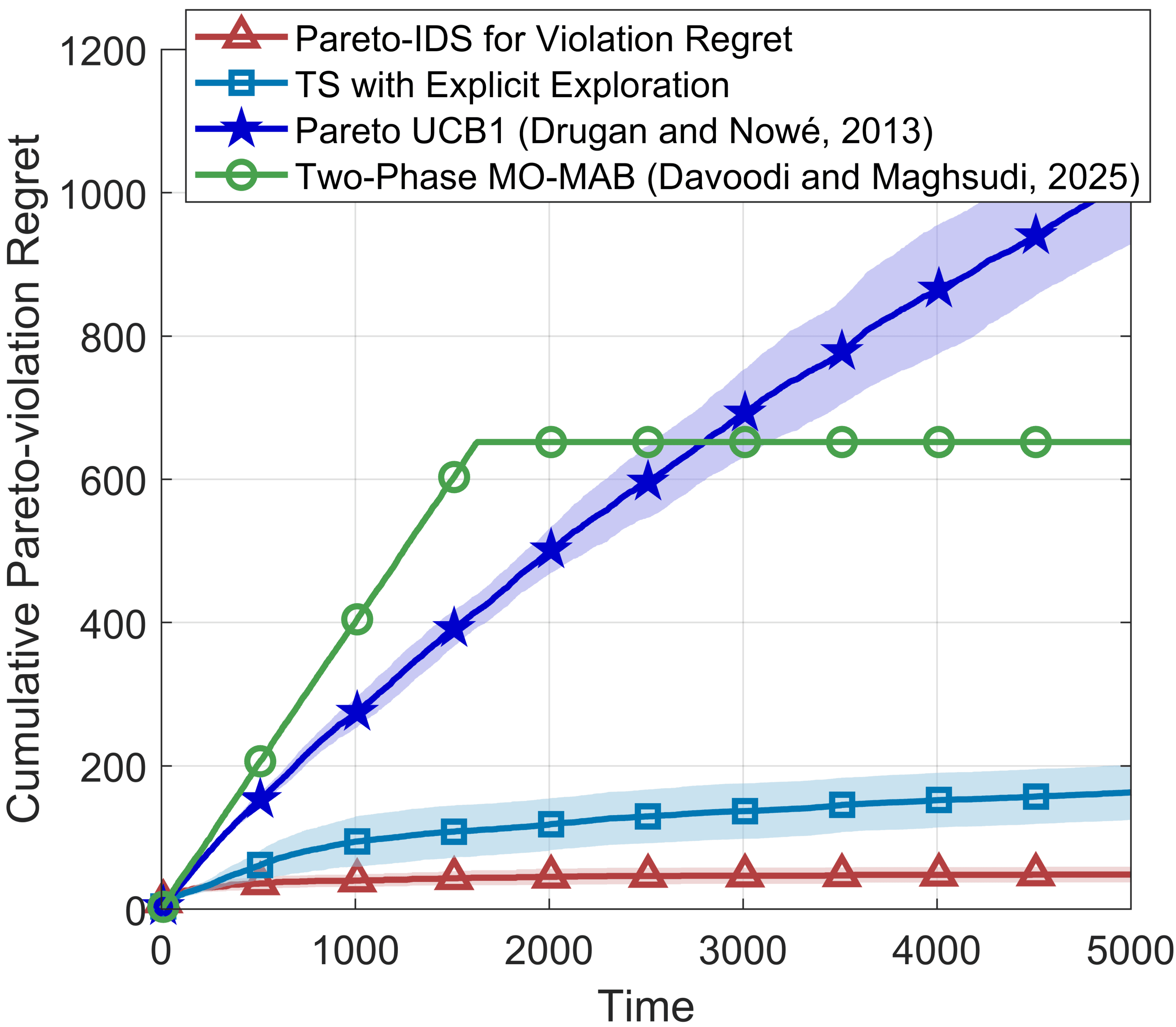}
\subcaption{$|\mathcal A|=10,\;d=4$}
\end{subfigure}
\enspace
\begin{subfigure}[b]{4.4cm}
\includegraphics[height=3.00cm,width=4.4cm]{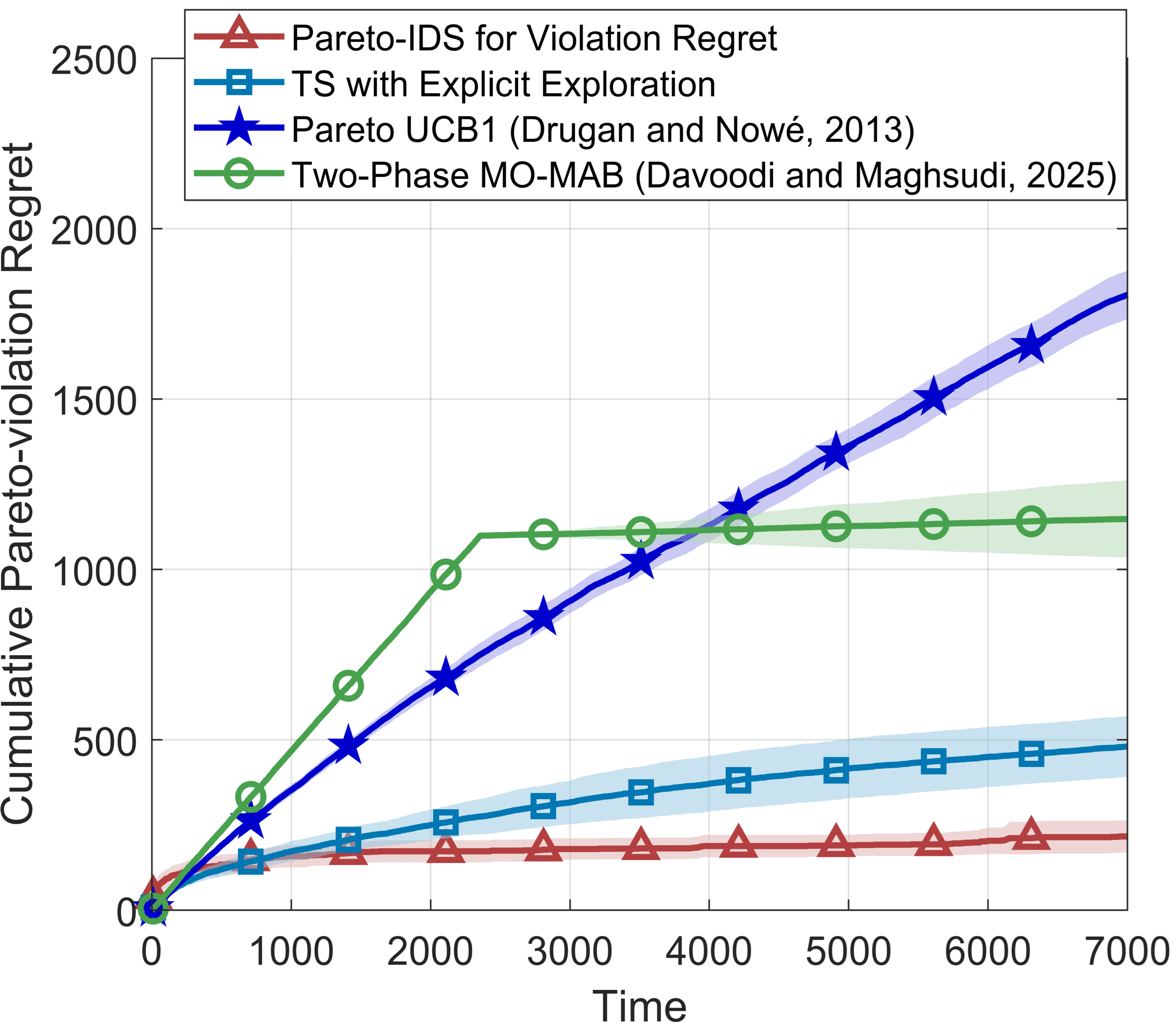}
\subcaption{$|\mathcal A|=15,\;d=5$}
\end{subfigure}
\enspace
\begin{subfigure}[b]{4.4cm}
\includegraphics[height=3.0cm,width=4.4cm]{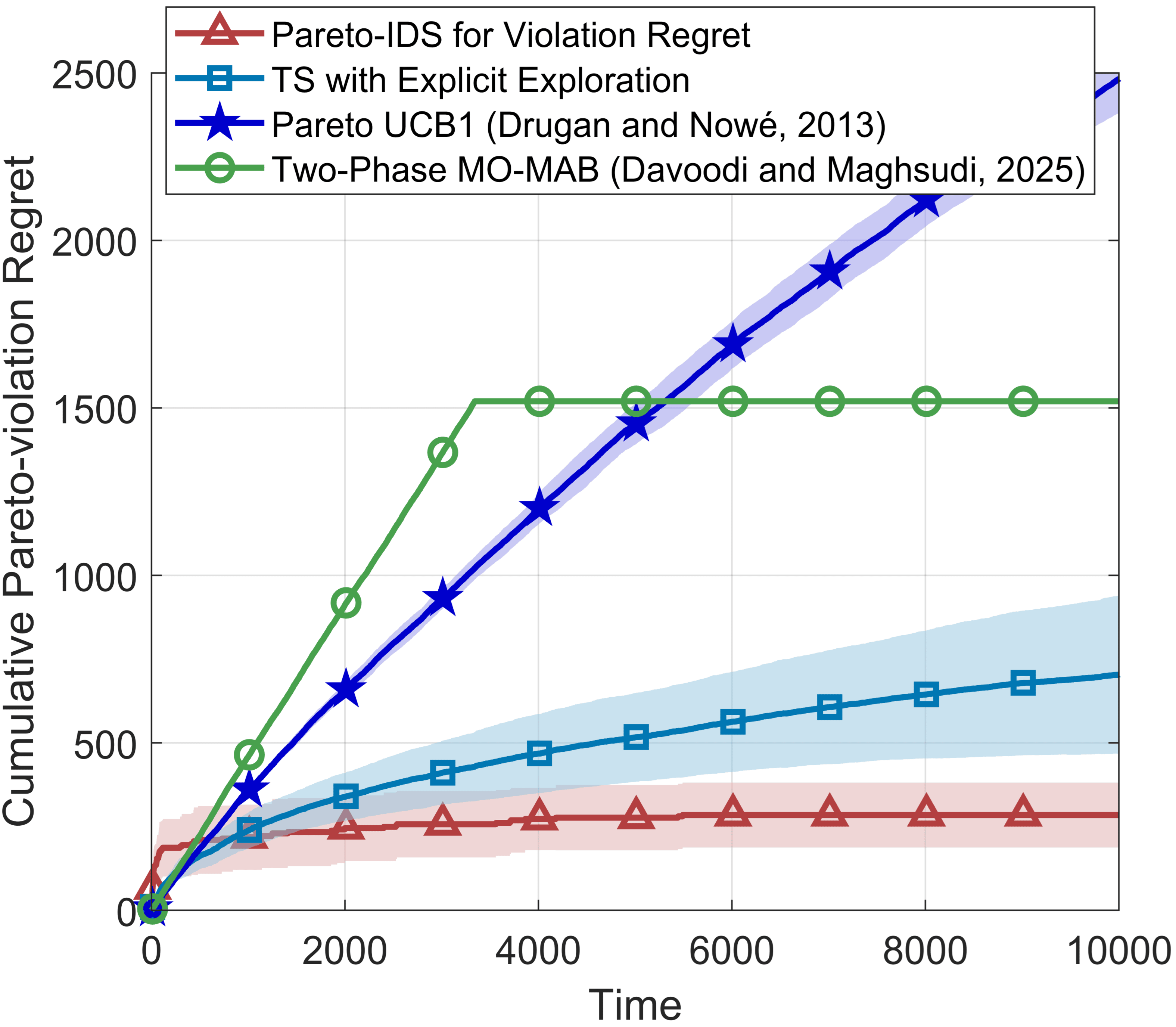}
\subcaption{$|\mathcal A|=20,\;d=6$}
\end{subfigure}

\caption{Cumulative Pareto-violation regret on multi-objective Bernoulli bandit instances.}
\label{pts1}

\end{figure*}

We evaluate the proposed algorithms on synthetic multi-objective Bernoulli bandits under both regret criteria. For each action $a\in\mathcal A$ and objective $j\in[d]$, we sample $\mu_j(a)$ independently and uniformly from $[0.05,0.95]$ and generate $Y_{t,j}(a)\sim\mathrm{Bernoulli}(\mu_j(a))$ independently across objectives. Bernoulli rewards provide a canonical bounded-reward model for our analysis; more general bounded rewards can be handled through the Bernoullization argument discussed earlier, which preserves mean reward vectors up to scaling and hence Pareto dominance. We consider $(|\mathcal A|,d)\in\{(10,4),(15,5),(20,6)\}$. For Pareto-violation regret, we compare TS-Pareto with explicit exploration (Algorithm~\ref{alg:ts_pareto}) and Pareto-IDS (Algorithm~\ref{alg:IDS}) against Pareto-UCB1~\citep{drugan2013designing} and the two-phase explore-then-commit MO-MAB algorithm of~\citet{davoodi2025stochastic}. The latter first explores all arms and then commits to a reduced cover set, with regret $O(T^{2/3}(|\mathcal A|\log T)^{1/3})$ for the exact-cover version and an additional $O(\log|\mathcal A|)$ factor for its polynomial-time approximation. We evaluate all four methods using cumulative Pareto-violation regret, i.e., the number of rounds in which a selected action is not Pareto-optimal under the true mean vectors. For each size, we run $200$ independent trials and report mean cumulative regret with one-standard-deviation bands. Figure~\ref{pts1} shows that Pareto-IDS achieves the lowest regret across all settings, followed by TS-Pareto, while both baselines incur larger regret. This is consistent with IDS directly balancing violation probability against information gain about the unknown Pareto set, whereas TS-Pareto incurs additional cost from explicit exploration to sample informative actions.

\begin{figure*}[h]
\centering
\begin{subfigure}[b]{4.4cm}
\includegraphics[height=3.05cm,width=4.4cm]{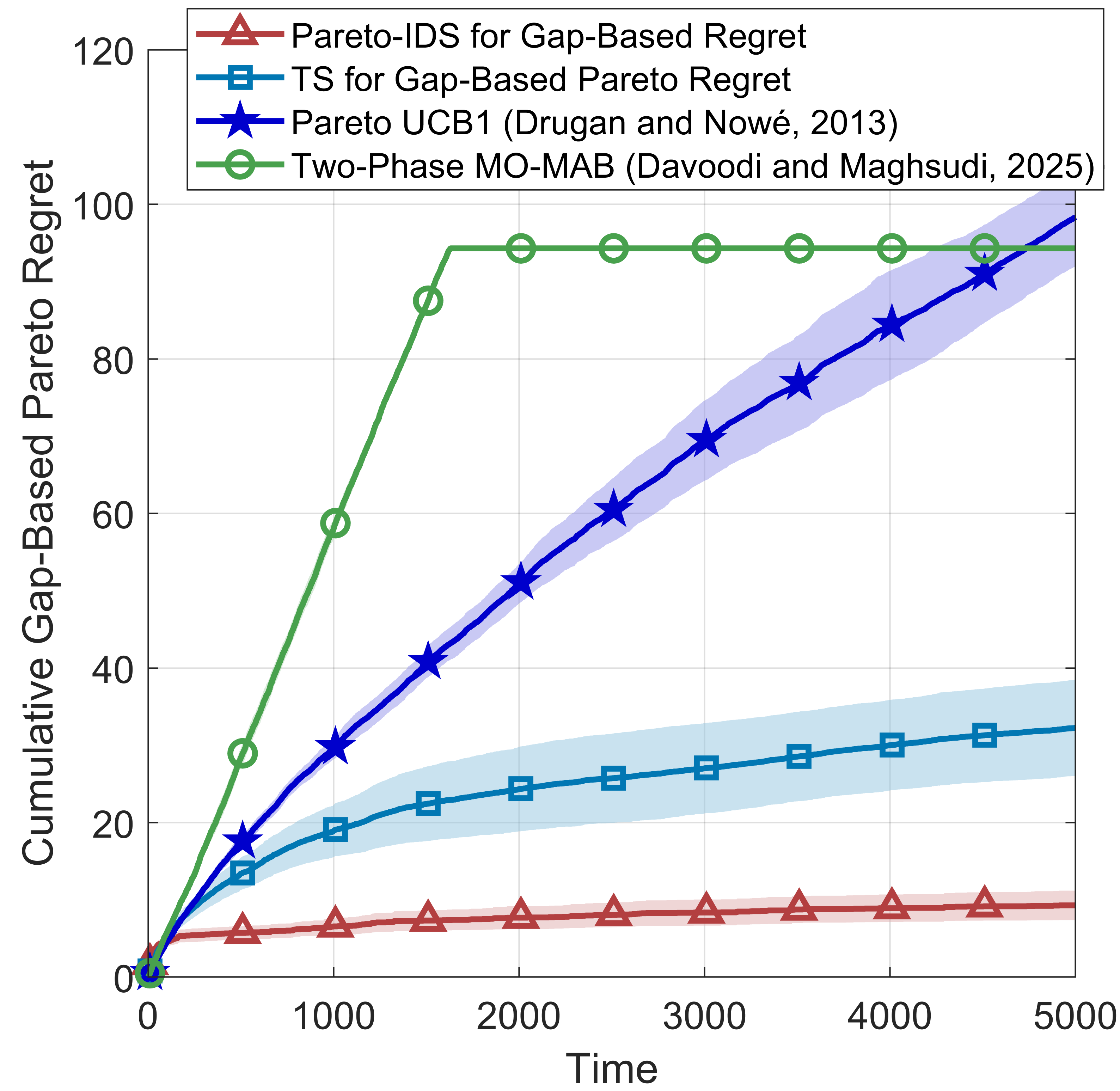}
\subcaption{$|\mathcal A|=10,\;d=4$}
\end{subfigure}
\enspace
\begin{subfigure}[b]{4.4cm}
\includegraphics[height=3.05cm,width=4.4cm]{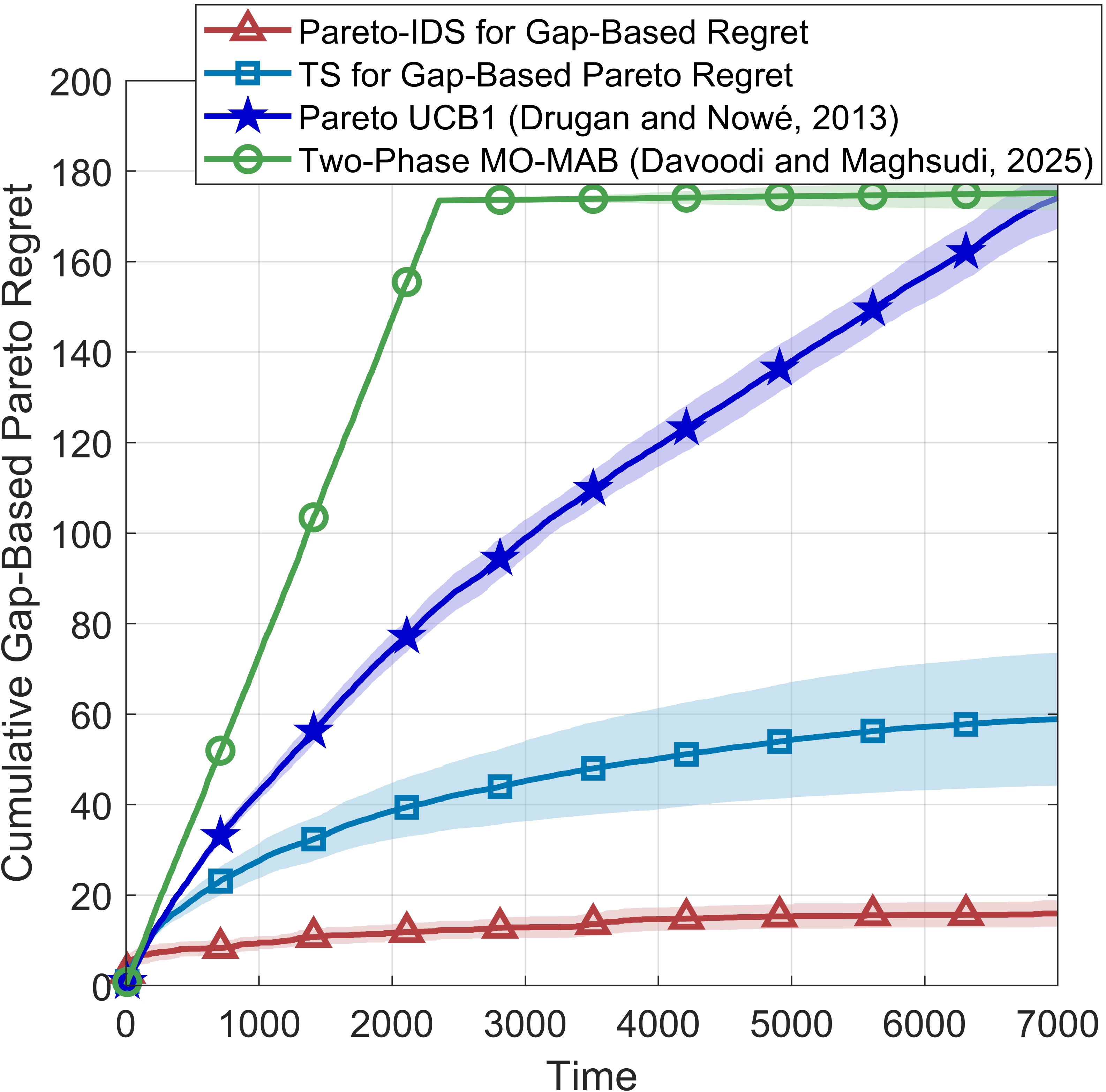}
\subcaption{$|\mathcal A|=15,\;d=5$}
\end{subfigure}
\enspace
\begin{subfigure}[b]{4.4cm}
\includegraphics[height=3.05cm,width=4.4cm]{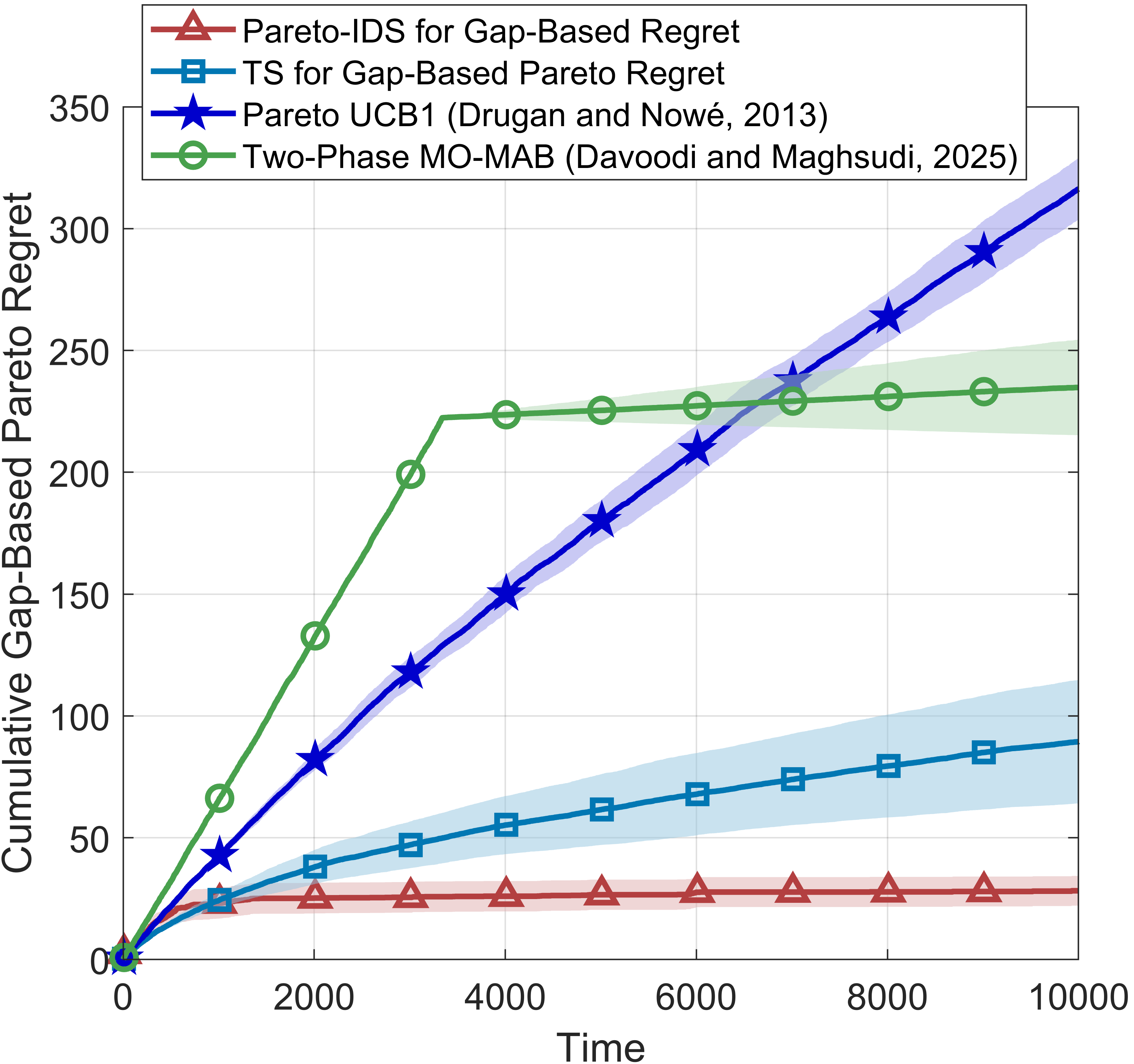}
\subcaption{$|\mathcal A|=20,\;d=6$}
\end{subfigure}

\caption{Cumulative gap-based Pareto regret on multi-objective Bernoulli bandit instances.}
\label{pts2}

\end{figure*}

\begin{figure*}[h]
\centering
\begin{subfigure}[b]{4.4cm}
\includegraphics[height=3.05cm,width=4.4cm]{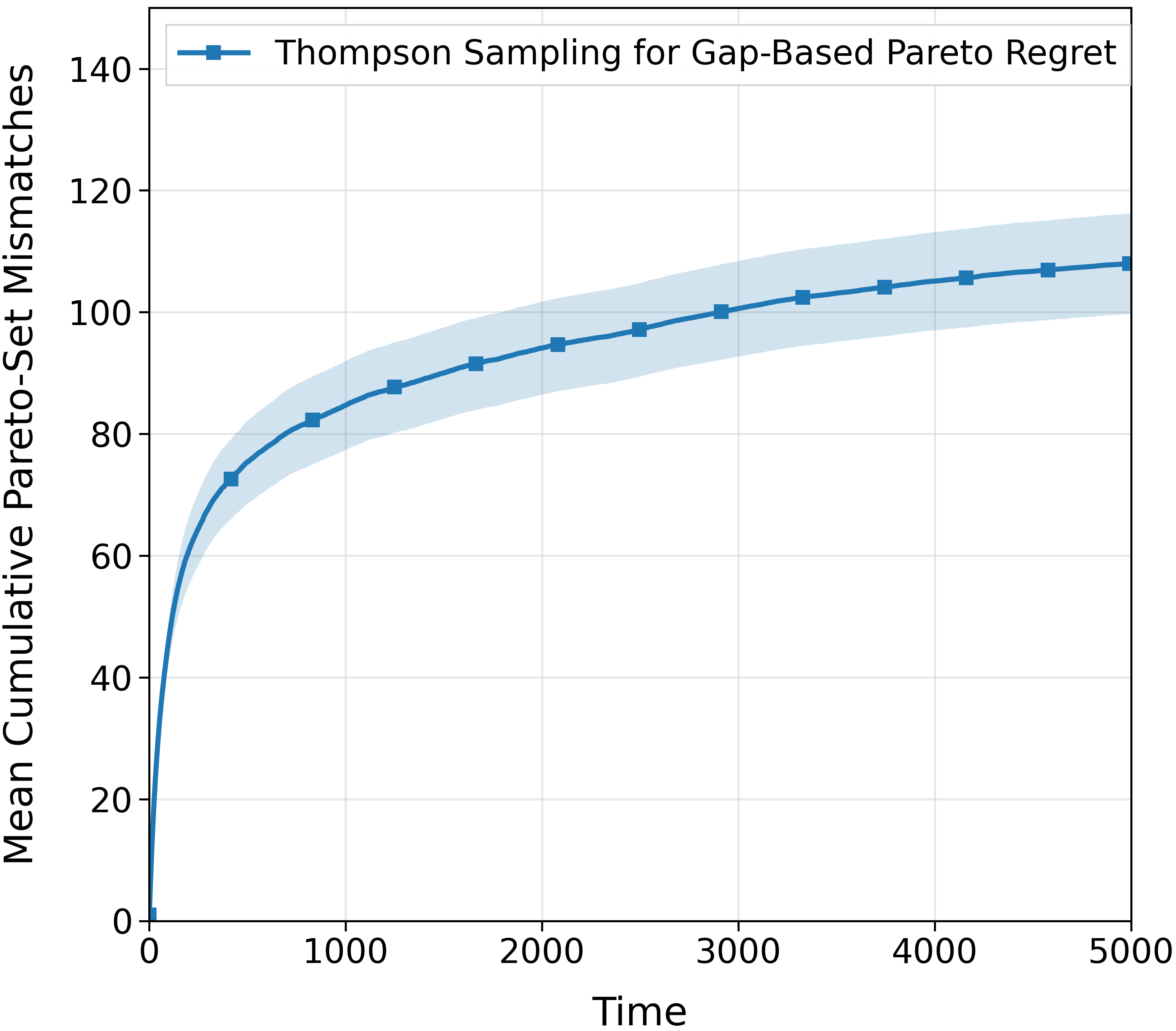}
\subcaption{$|\mathcal A|=5,\;d=2$}
\end{subfigure}
\enspace
\begin{subfigure}[b]{4.4cm}
\includegraphics[height=3.05cm,width=4.4cm]{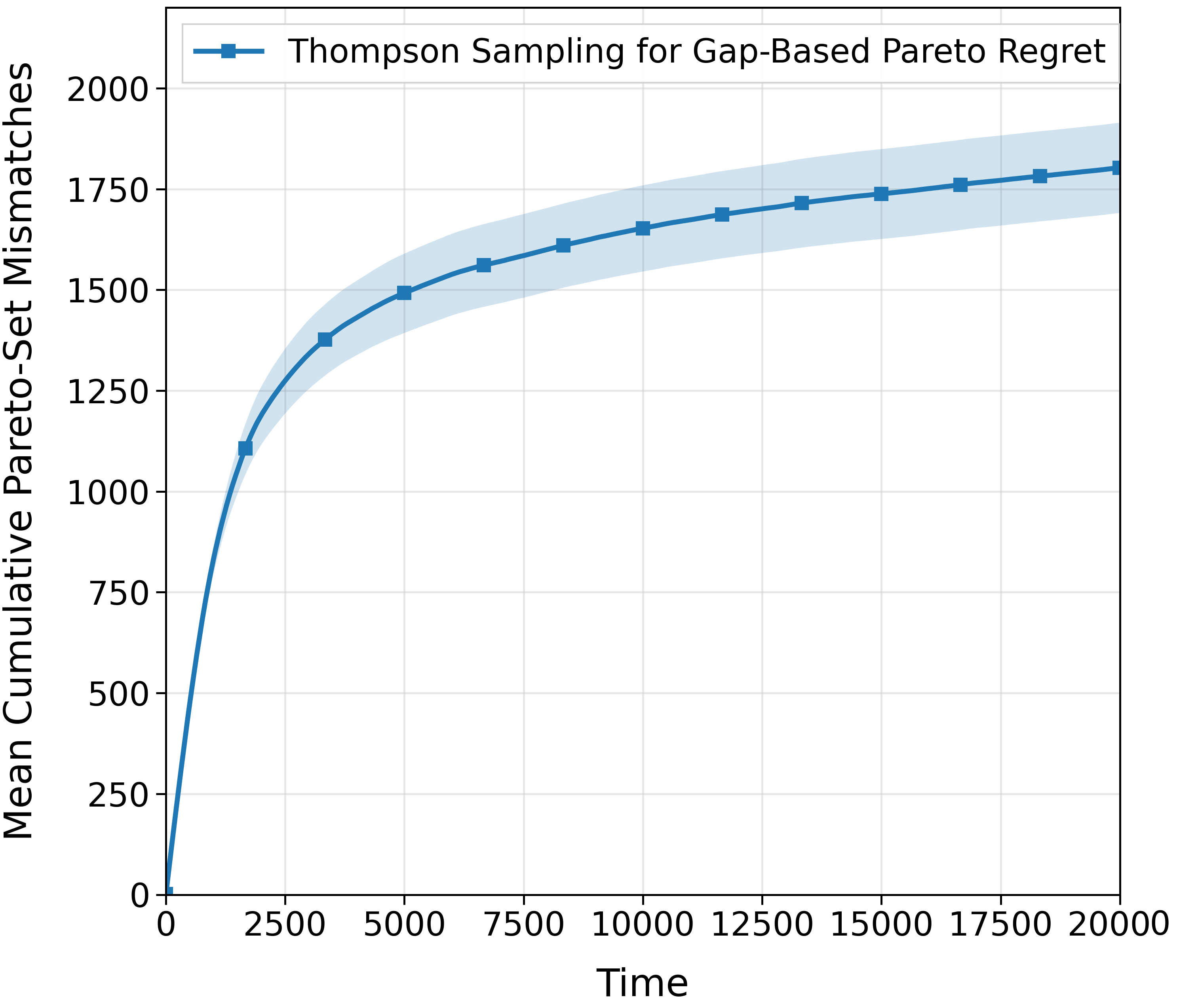}
\subcaption{$|\mathcal A|=10,\;d=4$}
\end{subfigure}
\enspace
\begin{subfigure}[b]{4.4cm}
\includegraphics[height=3.05cm,width=4.4cm]{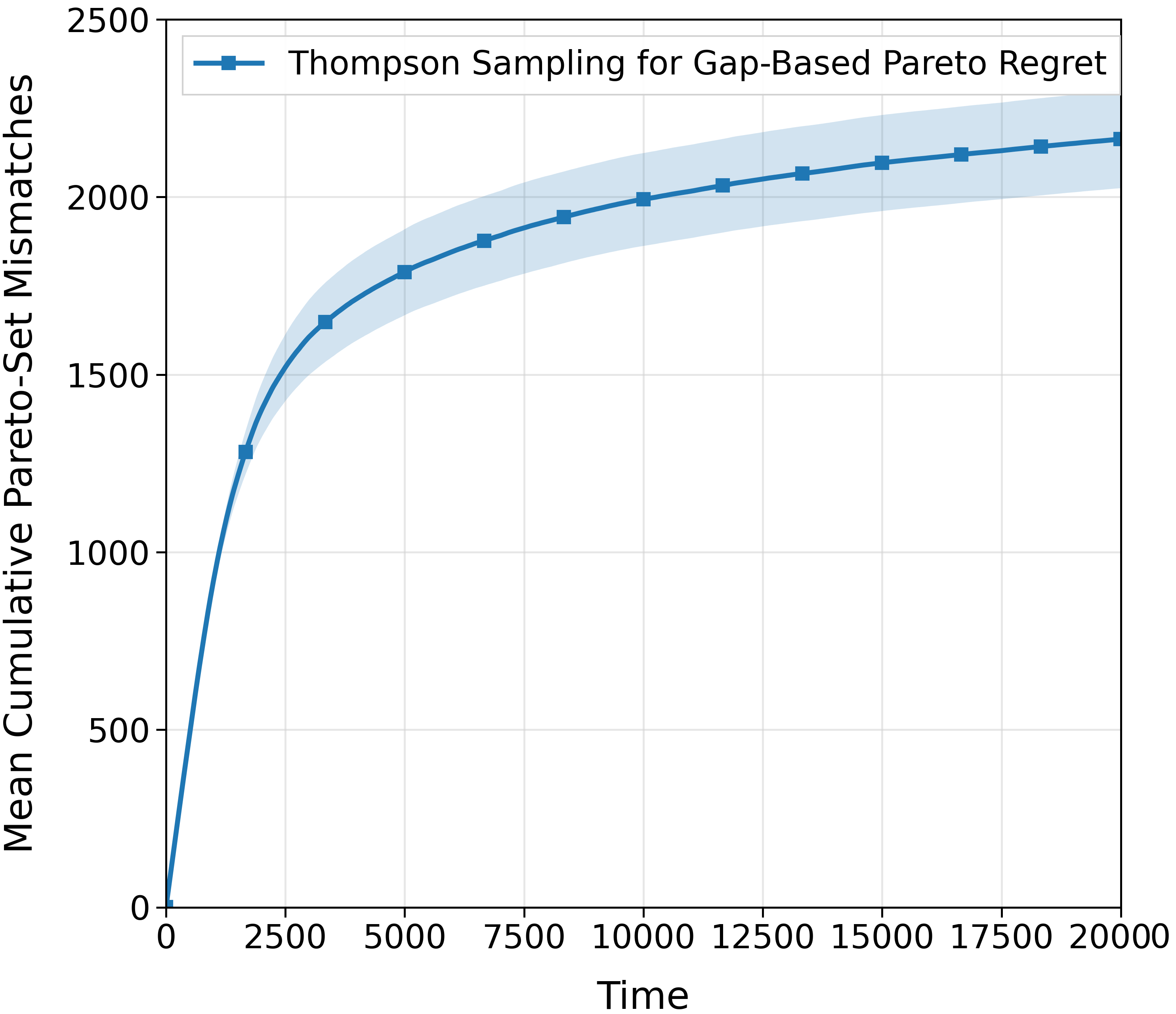}
\subcaption{$|\mathcal A|=15,\;d=5$}
\end{subfigure}

\caption{ Mean cumulative Pareto-Set Mismatches on multi-objective Bernoulli bandit instances}
\label{ptsxx}

\end{figure*}

For gap-based Pareto regret, we use the same environments and problem sizes, with $\Delta(a)$ defined as $\Delta(a) = \left(\sup_{b\in\mathcal A}\min_{j\in[d]}(\mu_j(b)-\mu_j(a))\right)_+$, and compare gap-based TS-Pareto (Algorithm~\ref{alg:ts_pareto_gap}) and Pareto-IDS (Algorithm~\ref{alg:IDS_gap}) against the same baselines. TS-Pareto samples a posterior parameter and plays uniformly from its Pareto set, whereas Pareto-IDS minimizes the information ratio using posterior expected gap and parameter-level information gain. Figure~\ref{pts2} shows the same ordering: Pareto-IDS achieves the lowest regret, followed by TS-Pareto, while Pareto-UCB1 and the two-phase MO-MAB baseline incur larger regret. This supports the Section~\ref{sec:gap_based} principle of balancing expected gap against parameter-level information gain. Continuous-Gaussian experiments in Appendix~\ref{app:gaussian_experiments} show the same ordering, indicating that these empirical conclusions are not specific to Bernoulli rewards.

For Pareto-set recovery, we run gap-based TS-Pareto (Algorithm~\ref{alg:ts_pareto_gap}) for $200$ independent trials per Bernoulli instance. At each round $t$, we sample a posterior parameter, compute its Pareto set $S_t$, play an arm uniformly from $S_t$, and update the posterior using the observed reward vector. Figure~\ref{ptsxx} plots the trial-averaged cumulative mismatches $\sum_{t=1}^{T}\mathbbm{1}\{S_t\ne\mathcal S^\ast\}$, where $\mathcal S^\ast$ is the true Pareto set. In Figure~\ref{ptsxx}(a), arm means are $(1,0)$, $(0,1)$, $(0.4,0.4)$, $(0.1,0.3)$, and $(0.3,0.1)$, giving $\mathcal S^\ast=\{1,2,3\}$. Under any normalized nonnegative linear scalarization, arm~3 scores $0.4$, while an extreme arm scores at least $0.5$; the fixed-coordinate UCB of \citet{xu2023pareto} likewise favors an extreme arm. Moreover, first-coordinate feedback cannot distinguish arm~3's Pareto-optimal mean $(0.4,0.4)$ from $(0.4,0)$, which arm~1 dominates. Using both objectives, TS-Pareto matches the full set, including arm~3, on approximately $98\%$ of the first $5{,}000$ rounds on average. Figures~\ref{ptsxx}(b)--(c) use randomly generated means fixed across trials; mismatches accumulate more slowly after the initial rounds, accounting for approximately $9\%$ and $10\%$ of the first $20{,}000$ rounds, respectively.

\section{Conclusion}

This work provides an information-theoretic analysis of Bayesian multi-objective bandits, demonstrating that different notions of Pareto regret require different levels of information about the unknown environment. For Pareto-violation regret, we introduce a predictive separation measure connecting distinguishability among candidate Pareto sets to information gain and establish sublinear Bayesian regret guarantees for Thompson Sampling and Information-Directed Sampling under uniform and polynomially decaying separation. We also derive a lower bound characterizing the statistical cost of weak separation. For gap-based Pareto regret, we show that Pareto-set information alone may be insufficient, as the magnitude of suboptimality depends on the underlying reward geometry. This motivates parameter-level information-theoretic analyses that yield regret guarantees for both algorithms. We further establish that Pareto Thompson Sampling can simultaneously achieve sublinear gap regret and recover the full Pareto-optimal set, with this joint guarantee extending to a fixed mixture with gap-based IDS. Our experiments support these theoretical findings. Overall, our results reveal a hierarchy of information requirements in multi-objective learning and motivate future work on weaker separation conditions, improved dimensional dependence, and computationally efficient information-directed strategies.

\section{Acknowledgements}
This research has been supported in part by NSF CAREER 2617987, IIS 2617859, Adobe Research and Intuit.
\clearpage

\clearpage

\bibliographystyle{plainnat}
\bibliography{main} 

\section{Technical appendices and supplementary material}

\subsection{Optimizing the Information Ratio}
\label{optm_info_ratio}

At round \(t\), IDS first constructs a sampling distribution
\(\pi_t^{\mathrm{IDS}}\in\Pi(\mathcal{A})\) that minimizes the information ratio and then draws \(A_t\sim\pi_t^{\mathrm{IDS}}\). Specifically,
\(\pi_t^{\mathrm{IDS}}\) is chosen as a solution to
\begin{equation}
\pi_t^{\mathrm{IDS}}
\in
\arg\min_{\pi\in\Pi(\mathcal{A})}
\frac{\left(\pi^\top\vec{\Delta}_t\right)^2}
     {\pi^\top\vec{I}_t}.
\end{equation}
Here, \(\vec{\Delta}_t\in\mathbb{R}^{|\mathcal{A}|}\) and
\(\vec{I}_t\in\mathbb{R}^{|\mathcal{A}|}\) denote the vectors of posterior expected one-step regret and information gain, respectively, with entries \(\Delta_t(a)\) and \(I_t(a)\) for \(a\in\mathcal{A}\). The set
\[
\Pi(\mathcal{A})
=
\left\{
\pi\in\mathbb{R}_+^{|\mathcal{A}|}:
\sum_{a\in\mathcal{A}}\pi(a)=1
\right\}
\]
denotes the probability simplex over the action set. We use the extended-value convention
\[
\frac{x^2}{y}
=
\begin{cases}
x^2/y, & y>0,\\
0, & x=0,\ y=0,\\
+\infty, & x>0,\ y=0,
\end{cases}
\qquad x,y\ge0.
\]
Thus, the information ratio remains well defined even when a candidate sampling distribution yields zero information gain.

\begin{algorithm}[h]
\small
\DontPrintSemicolon
\SetKwFunction{FMain}{IDSDistribution}
\SetKwProg{Fn}{Function}{:}{}

\Fn{\FMain{$\mathcal{A},\vec{\Delta}_t,\vec{I}_t$}}{

    \If{$|\mathcal{A}|=1$}{
        \Return the unique distribution in \(\Pi(\mathcal{A})\)\;
    }

    \If{$\vec I_t=0$}{
        \If{there exists \(a\in\mathcal{A}\) with \(\Delta_t(a)=0\)}{
            \Return the point mass on any such action \(a\)\;
        }
        \Else{
            \Return any distribution in \(\Pi(\mathcal{A})\)\;
        }
    }

    \ForEach{$a,a'\in\mathcal{A}$ with $a\neq a'$}{
        Compute
        \[
        q_{a,a'}
        \in
        \arg\min_{q\in[0,1]}
        \frac{
        \left(q\Delta_t(a)+(1-q)\Delta_t(a')\right)^2
        }{
        qI_t(a)+(1-q)I_t(a')
        },
        \]
        using the extended-value convention above\;
    }

    Select
    \[
    (a^*,a^{**})
    \in
    \arg\min_{a\neq a'}
    \frac{
    \left(
    q_{a,a'}\Delta_t(a)
    +(1-q_{a,a'})\Delta_t(a')
    \right)^2
    }{
    q_{a,a'}I_t(a)
    +(1-q_{a,a'})I_t(a')
    }.
    \]

    Construct \(\pi_t^{\mathrm{IDS}}\in\Pi(\mathcal{A})\) by
    \[
    \pi_t^{\mathrm{IDS}}(a)=
    \begin{cases}
    q_{a^*,a^{**}}, & a=a^*,\\[2mm]
    1-q_{a^*,a^{**}}, & a=a^{**},\\[2mm]
    0, & \text{otherwise}.
    \end{cases}
    \]

    \Return \(\pi_t^{\mathrm{IDS}}\)\;
}

\caption{Computation of the IDS sampling distribution via two-point mixtures}
\label{alg:IDSaction}
\end{algorithm}

When \(\vec I_t\neq0\), Proposition~1 of \citet{russo2014learning} shows that the information-ratio objective is convex on
\(\{\pi\in\Pi(\mathcal A):\pi^\top\vec I_t>0\}\)
and that an optimal solution can be chosen with support on at most two actions. If a zero-regret distribution also has zero information gain, it attains information ratio \(0\) under the convention above and is already optimal. Otherwise, since \(\vec I_t\neq0\), there exists a feasible distribution with positive information gain and hence finite information ratio, so no zero-information distribution with positive regret can be optimal. Accordingly, in the nontrivial case with \(|\mathcal A|\ge2\), Algorithm~\ref{alg:IDSaction} only needs to consider two-point mixtures. It searches over all pairs of distinct actions and, for each pair, solves the resulting one-dimensional convex optimization problem over the mixing weight \(q\in[0,1]\), using the same extended-value convention. The endpoints \(q=0\) and \(q=1\) include singleton distributions. Among these candidate mixtures, it selects one with the smallest information ratio; the resulting distribution defines \(\pi_t^{\mathrm{IDS}}\), from which \(A_t\) is sampled.

The case \(|\mathcal A|=1\) is trivial. If \(\vec I_t=0\), IDS selects any zero-regret action if one exists; otherwise every distribution has infinite information ratio and the information-ratio criterion is uninformative at that history.

\subsection{Proof of Lemma \ref{lem:info_gain_tv}}

\lemInfoGainTV*

\begin{proof}

Fix a round $t$, any realizable history $\mathcal{H}_t$, and an action
$a\in\mathcal{A}$ throughout the proof. Quantities conditioned on
$A_t=a$ are understood under the fixed-action experiment in which
action $a$ is selected at round $t$; whenever
$\mathbb{P}(A_t=a\mid\mathcal{H}_t)>0$, this agrees with ordinary
conditioning on the event $A_t=a$.

Let
\[
p_t(s):=
\mathbb{P}(\mathcal{S}^*=s\mid\mathcal{H}_t)
\]
denote the posterior distribution of the random Pareto set
$\mathcal{S}^*$, and let
\[
\operatorname{supp}(p_t)
:=
\{s\in2^{\mathcal{A}}:p_t(s)>0\}.
\]
Because the policy selects $A_t$ using only $\mathcal{H}_t$ and
independent internal randomization, $A_t$ is conditionally independent
of $\mathcal{S}^*$ given $\mathcal{H}_t$. Hence selecting action $a$ at
round $t$ does not change the posterior distribution of $\mathcal{S}^*$:
\[
\mathbb{P}(\mathcal{S}^*=s
\mid\mathcal{H}_t,A_t=a)
=
\mathbb{P}(\mathcal{S}^*=s
\mid\mathcal{H}_t)
=
p_t(s).
\]

Draw $S,S'\sim p_t$ independently, and let
$B\sim\mathrm{Bernoulli}(1/2)$ be independent of $(S,S')$, all
conditional on $\mathcal{H}_t$ and independently of the policy
randomization at round $t$. Define
\[
\widehat{S}:=
\begin{cases}
S, & B=0,\\
S', & B=1.
\end{cases}
\]
Conditional on
$(\widehat{S},\mathcal{H}_t,A_t=a)$, generate an auxiliary observation
$\widetilde{\mathbf{Y}}$ according to
\[
\widetilde{\mathbf{Y}}
\mid
(\widehat{S}=s,\mathcal{H}_t,A_t=a)
\sim
P_{t,s}^a,
\]
conditionally independently of $(S,S',B)$ given
$(\widehat{S},\mathcal{H}_t,A_t=a)$.

Since $\widehat{S}=S$ when $B=0$ and
$\widehat{S}=S'$ when $B=1$, for every
$s\in\operatorname{supp}(p_t)$,
\[
\{\widehat{S}=s\}
=
\{B=0,S=s\}\cup\{B=1,S'=s\},
\]
where the two events are disjoint. Therefore,
\[
\mathbb{P}(\widehat{S}=s\mid\mathcal{H}_t)
=
\mathbb{P}(B=0,S=s\mid\mathcal{H}_t)
+
\mathbb{P}(B=1,S'=s\mid\mathcal{H}_t).
\]
Since $B$ is independent of $(S,S')$ and $S,S'\sim p_t$,
\[
\mathbb{P}(B=0,S=s\mid\mathcal{H}_t)
=
\mathbb{P}(B=0)\,
\mathbb{P}(S=s\mid\mathcal{H}_t)
=
\tfrac{1}{2}p_t(s),
\]
and similarly,
\[
\mathbb{P}(B=1,S'=s\mid\mathcal{H}_t)
=
\tfrac{1}{2}p_t(s).
\]
Hence,
\[
\mathbb{P}(\widehat{S}=s\mid\mathcal{H}_t)
=
p_t(s).
\]
Moreover, because $(S,S',B)$ are constructed independently of the
policy randomization conditional on $\mathcal{H}_t$,
\[
\mathbb{P}(\widehat{S}=s
\mid\mathcal{H}_t,A_t=a)
=
p_t(s).
\]

Next, by construction of $\widetilde{\mathbf{Y}}$, for any measurable
set $E$ and every $s\in\operatorname{supp}(p_t)$,
\[
\mathbb{P}(\widetilde{\mathbf{Y}}\in E
\mid
\widehat{S}=s,\mathcal{H}_t,A_t=a)
=
P_{t,s}^a(E).
\]
Therefore,
\begin{align*}
&\mathbb{P}(\widehat{S}=s,\widetilde{\mathbf{Y}}\in E
\mid\mathcal{H}_t,A_t=a)\\
&\qquad=
\mathbb{P}(\widehat{S}=s\mid\mathcal{H}_t,A_t=a)
P_{t,s}^a(E)
=
p_t(s)P_{t,s}^a(E).
\end{align*}
By the definitions of $p_t(s)$ and $P_{t,s}^a$, together with the
conditional independence of $A_t$ and $\mathcal{S}^*$ given
$\mathcal{H}_t$, we also have
\[
\mathbb{P}(\mathcal{S}^*=s,\mathbf{Y}_{t,a}\in E
\mid\mathcal{H}_t,A_t=a)
=
p_t(s)P_{t,s}^a(E).
\]
Thus, the conditional joint distributions coincide:
\[
\mathcal{L}(\widehat{S},\widetilde{\mathbf{Y}}
\mid\mathcal{H}_t,A_t=a)
=
\mathcal{L}(\mathcal{S}^*,\mathbf{Y}_{t,a}
\mid\mathcal{H}_t,A_t=a).
\]
Since conditional mutual information depends only on the corresponding
conditional joint distribution,
\begin{equation}
I(\mathcal{S}^*;\mathbf{Y}_{t,a}
\mid\mathcal{H}_t,A_t=a)
=
I(\widehat{S};\widetilde{\mathbf{Y}}
\mid\mathcal{H}_t,A_t=a).
\label{eq:law-equiv}
\end{equation}

Next, by the conditional-independence construction of
$\widetilde{\mathbf{Y}}$,
\[
\widetilde{\mathbf{Y}}
\perp\!\!\!\perp
(S,S',B)
\mid
(\widehat{S},\mathcal{H}_t,A_t=a),
\]
and therefore
\begin{equation}
I(S,S',B;\widetilde{\mathbf{Y}}
\mid
\widehat{S},\mathcal{H}_t,A_t=a)
=
0.
\label{eq:ci-zero}
\end{equation}
Applying the chain rule gives
\begin{align}
I(\widehat{S},S,S',B;\widetilde{\mathbf{Y}}
\mid\mathcal{H}_t,A_t=a)
&=
I(\widehat{S};\widetilde{\mathbf{Y}}
\mid\mathcal{H}_t,A_t=a)
\notag\\
&\quad+
I(S,S',B;\widetilde{\mathbf{Y}}
\mid\widehat{S},\mathcal{H}_t,A_t=a)
\notag\\
&=
I(\widehat{S};\widetilde{\mathbf{Y}}
\mid\mathcal{H}_t,A_t=a),
\label{eq:joint-eq}
\end{align}
where the second equality uses \eqref{eq:ci-zero}.

Since $\widehat{S}$ is a deterministic function of $(S,S',B)$,
adding $\widehat{S}$ does not introduce any additional information
beyond $(S,S',B)$, and hence
\begin{equation}
I(\widehat{S},S,S',B;\widetilde{\mathbf{Y}}
\mid\mathcal{H}_t,A_t=a)
=
I(S,S',B;\widetilde{\mathbf{Y}}
\mid\mathcal{H}_t,A_t=a).
\label{eq:det-nochange}
\end{equation}
Combining \eqref{eq:joint-eq} and \eqref{eq:det-nochange}, we obtain
\begin{equation}
I(\widehat{S};\widetilde{\mathbf{Y}}
\mid\mathcal{H}_t,A_t=a)
=
I(S,S',B;\widetilde{\mathbf{Y}}
\mid\mathcal{H}_t,A_t=a).
\label{eq:mi-hat-eq-sssb}
\end{equation}

Applying the chain rule once more,
\begin{equation}
I(S,S',B;\widetilde{\mathbf{Y}}
\mid\mathcal{H}_t,A_t=a)
=
I(S,S';\widetilde{\mathbf{Y}}
\mid\mathcal{H}_t,A_t=a)
+
I(B;\widetilde{\mathbf{Y}}
\mid S,S',\mathcal{H}_t,A_t=a).
\label{eq:chain2}
\end{equation}
Thus,
\begin{equation}
I(\widehat{S};\widetilde{\mathbf{Y}}
\mid\mathcal{H}_t,A_t=a)
=
I(S,S';\widetilde{\mathbf{Y}}
\mid\mathcal{H}_t,A_t=a)
+
I(B;\widetilde{\mathbf{Y}}
\mid S,S',\mathcal{H}_t,A_t=a).
\label{eq:five}
\end{equation}

Since mutual information is nonnegative, the first term on the
right-hand side of \eqref{eq:five} can be dropped, yielding
\begin{equation}
I(\widehat{S};\widetilde{\mathbf{Y}}
\mid\mathcal{H}_t,A_t=a)
\ge
I(B;\widetilde{\mathbf{Y}}
\mid S,S',\mathcal{H}_t,A_t=a).
\label{eq:drop-first}
\end{equation}

By construction, conditional on $(\mathcal{H}_t,A_t=a)$, the auxiliary
variables $S$ and $S'$ remain independent with common law $p_t$.
Hence, expanding the conditional mutual information over $(S,S')$ gives
\begin{align}
&I(B;\widetilde{\mathbf{Y}}
\mid S,S',\mathcal{H}_t,A_t=a)
\notag\\
&\qquad=
\sum_{s,s'\in\operatorname{supp}(p_t)}
p_t(s)p_t(s')\,
I(B;\widetilde{\mathbf{Y}}
\mid S=s,S'=s',\mathcal{H}_t,A_t=a).
\label{eq:cmi-expansion}
\end{align}
Therefore, from \eqref{eq:drop-first},
\begin{equation}
I(\widehat{S};\widetilde{\mathbf{Y}}
\mid\mathcal{H}_t,A_t=a)
\ge
\sum_{s,s'\in\operatorname{supp}(p_t)}
p_t(s)p_t(s')\,
I(B;\widetilde{\mathbf{Y}}
\mid S=s,S'=s',\mathcal{H}_t,A_t=a).
\label{eq:drop-nonneg-avg}
\end{equation}

Fix $s,s'\in\operatorname{supp}(p_t)$ and define
\[
P:=P_{t,s}^a,
\qquad
Q:=P_{t,s'}^a,
\qquad
M:=\tfrac{1}{2}(P+Q).
\]
Conditional on
$(S=s,S'=s',\mathcal{H}_t,A_t=a)$, the auxiliary observation
$\widetilde{\mathbf{Y}}$ is drawn from $P$ when $B=0$ and from $Q$
when $B=1$. Thus, $\widetilde{\mathbf{Y}}$ has mixture distribution
$M=\tfrac12(P+Q)$, where $B$ indicates whether the observation is drawn
from $P$ or $Q$. It follows that
\begin{equation}
I(B;\widetilde{\mathbf{Y}}
\mid S=s,S'=s',\mathcal{H}_t,A_t=a)
=
\tfrac{1}{2}\mathrm{KL}(P\|M)
+
\tfrac{1}{2}\mathrm{KL}(Q\|M),
\label{eq:ib-js}
\end{equation}
which is the Jensen--Shannon divergence between $P$ and $Q$.

Using natural logarithms, Pinsker's inequality gives, for any
distributions $U,V$,
\[
\mathrm{KL}(U\|V)
\ge
2\,\mathrm{TV}(U,V)^2.
\]
Applying this inequality to both terms in \eqref{eq:ib-js} yields
\[
\tfrac{1}{2}\mathrm{KL}(P\|M)
+
\tfrac{1}{2}\mathrm{KL}(Q\|M)
\ge
\mathrm{TV}(P,M)^2
+
\mathrm{TV}(Q,M)^2.
\]
Since $M=\tfrac{1}{2}(P+Q)$,
\[
\mathrm{TV}(P,M)
=
\mathrm{TV}(Q,M)
=
\tfrac{1}{2}\mathrm{TV}(P,Q).
\]
Therefore,
\begin{equation}
I(B;\widetilde{\mathbf{Y}}
\mid S=s,S'=s',\mathcal{H}_t,A_t=a)
\ge
\frac{1}{2}
\mathrm{TV}\!\left(P_{t,s}^a,P_{t,s'}^a\right)^2.
\label{eq:ib-tv2-pointwise}
\end{equation}

Substituting \eqref{eq:ib-tv2-pointwise} into
\eqref{eq:drop-nonneg-avg}, we obtain
\begin{align}
I(\widehat{S};\widetilde{\mathbf{Y}}
\mid\mathcal{H}_t,A_t=a)
&\ge
\sum_{s,s'\in\operatorname{supp}(p_t)}
p_t(s)p_t(s')\,
I(B;\widetilde{\mathbf{Y}}
\mid S=s,S'=s',\mathcal{H}_t,A_t=a)
\notag\\
&\ge
\frac{1}{2}
\sum_{s,s'\in\operatorname{supp}(p_t)}
p_t(s)p_t(s')\,
\mathrm{TV}\!\left(P_{t,s}^a,P_{t,s'}^a\right)^2
\notag\\
&=
\frac{1}{2}
\mathbb{E}_{S,S'\stackrel{\mathrm{i.i.d.}}{\sim}p_t}
\!\left[
\mathrm{TV}\!\left(P_{t,S}^a,P_{t,S'}^a\right)^2
\right].
\end{align}
Finally, combining this inequality with \eqref{eq:law-equiv}, we conclude
\[
I(\mathcal{S}^*;\mathbf{Y}_{t,a}
\mid\mathcal{H}_t,A_t=a)
\ge
\frac{1}{2}
\mathbb{E}_{S,S'\stackrel{\mathrm{i.i.d.}}{\sim}p_t}
\!\left[
\mathrm{TV}\!\left(P_{t,S}^a,P_{t,S'}^a\right)^2
\right],
\]
which completes the proof.
\end{proof}

\subsection{Proof of Theorem \ref{thm:avgsep-allt-single}}

We restate Theorem~\ref{thm:avgsep-allt-single}, explicitly specifying
the necessary regularity and non-degeneracy conditions, and present a
complete formal proof.

\begin{theorem}
\label{thm:avgsep-allt-single_APP}
Consider a multi-objective bandit model with finite discrete reward
alphabet $\mathcal{Y}\subset\mathbb{R}^d$ and finite action set
$\mathcal{A}$, leading to Dirichlet posterior updates over the parameter
$\boldsymbol{\theta}$. For each $a\in\mathcal A$,
$\boldsymbol{\theta}\in\Theta$, and $y\in\mathcal Y$, write
\[
P^a_{\boldsymbol{\theta}}(y)
:=
\mathbb{P}_{\boldsymbol{\theta}}
(\mathbf{Y}_{t,a}=y)
\]
for the conditional reward probability under action $a$. Let
$q\in\Pi(\mathcal{A})$ have full support, i.e.,
$q(a)>0$ for all $a\in\mathcal{A}$.

Fix a Dirichlet posterior density family
$\{p_\alpha:\alpha\in\mathcal H\subset\mathbb R^m\}$ with
$\mathcal H$ open, where all densities are taken with respect to a fixed
$\sigma$-finite reference measure $\nu$ on $\Theta$. Under the Dirichlet
count-update model, for every round $t$ and every realizable history
$\mathcal H_t$, the posterior parameter can be written as
\[
\alpha_t=\alpha_0+N_t,
\qquad
N_t\in\mathbb N_0^m,
\]
where
\[
\mathbb N_0:=\{0,1,2,\ldots\},
\]
and $\alpha_t\in\mathcal H$.

Let $\mathcal{S}:=2^{\mathcal{A}}$ denote the collection of all subsets
of $\mathcal{A}$, and let $\mathcal{S}^*$ be the random Pareto set
taking values in $\mathcal{S}$. Let $\Theta$ denote the parameter space
of $\boldsymbol{\theta}$, and for each $s\in\mathcal{S}$ define
\[
\Theta_s
:=
\{\boldsymbol{\theta}\in\Theta:
\mathcal{S}^*(\boldsymbol{\theta})=s\},
\]
i.e., the set of parameters for which the Pareto set equals $s$.
Assume that each $\Theta_s$ is measurable. For
$\alpha\in\mathcal H$, define the posterior probability of Pareto-set
candidate $s$ by
\[
p_\alpha(s)
:=
p_\alpha(\Theta_s)
:=
\int_{\Theta_s}
p_\alpha(\boldsymbol{\theta})\,\nu(d\boldsymbol{\theta}).
\]
For each pair $s\neq s'$ with $s,s'\in\mathcal S$, define the
admissible parameter set
\[
\mathcal H_{s,s'}
:=
\{\alpha\in\mathcal H:
p_\alpha(s)>0,\;
p_\alpha(s')>0\}.
\]

Assume that for every $s\neq s'$ with
$\mathcal H_{s,s'}\neq\varnothing$:

\begin{enumerate}
\item\label{nd:analytic-thm}
\textbf{(Analytic parameterization.)}
For each $(a,y)\in\mathcal{A}\times\mathcal{Y}$ and each
$r\in\{s,s'\}$, the maps
\[
\alpha\mapsto
\int_{\Theta_r}
p_\alpha(\boldsymbol{\theta})\,\nu(d\boldsymbol{\theta}),
\qquad
\alpha\mapsto
\int_{\Theta_r}
P^a_{\boldsymbol{\theta}}(y)
p_\alpha(\boldsymbol{\theta})\,\nu(d\boldsymbol{\theta})
\]
are real-analytic on $\mathcal H$.

\item\label{nd:witness-thm}
\textbf{(Witness posterior exists in each component.)}
For every connected component $\mathcal C$ of
$\mathcal H_{s,s'}$, there exist
$\alpha^\star\in\mathcal C$ and $a\in\mathcal A$ such that
\[
P^a_{p_{\alpha^\star},s}
\neq
P^a_{p_{\alpha^\star},s'}.
\]
Here, for $\alpha\in\mathcal H_{s,s'}$, the predictive reward
distribution under action $a$ conditioned on Pareto-set candidate $s$
is
\[
P^a_{p_\alpha,s}(y)
:=
\frac{
\int_{\Theta_s}
P^a_{\boldsymbol{\theta}}(y)
p_\alpha(\boldsymbol{\theta})\,\nu(d\boldsymbol{\theta})
}{
\int_{\Theta_s}
p_\alpha(\boldsymbol{\theta})\,\nu(d\boldsymbol{\theta})
},
\qquad
y\in\mathcal Y.
\]
\end{enumerate}

For a realizable history $\mathcal H_t$, let
\[
p_t(s)
:=
p_{\alpha_t}(s)
=
\mathbb{P}(\mathcal{S}^*=s\mid\mathcal H_t),
\]
and let
\[
S,S'\stackrel{\mathrm{i.i.d.}}{\sim}p_t
\]
conditional on $\mathcal H_t$. Then there exists a Lebesgue measure-zero
set $\mathcal B\subset\mathcal H$ such that, for every
$\alpha_0\in\mathcal H\setminus\mathcal B$, every round $t$, and every
realizable history $\mathcal H_t$,
\begin{equation}
\mathbb{E}\!\left[
\mathbb{E}_{a\sim q}
\left[
\mathrm{TV}\!\left(P^a_{t,S},P^a_{t,S'}\right)^2
\right]
\;\middle|\;
S\neq S',\mathcal H_t
\right]
>0
\label{eq:avgsep-positive-allt-single}
\end{equation}
whenever
\[
\mathbb{P}(S\neq S'\mid\mathcal H_t)>0,
\]
where $P^a_{t,S}:=P^a_{p_{\alpha_t},S}$ and
$\mathrm{TV}$ denotes total variation distance.
\end{theorem}

\begin{proof}
Let $\lambda_m$ denote $m$-dimensional Lebesgue measure on
$\mathbb R^m$. Fix $s\neq s'$ with
$\mathcal H_{s,s'}\neq\varnothing$.

By condition~\ref{nd:analytic-thm}, the map
\[
\alpha\mapsto
p_\alpha(\Theta_r)
=
\int_{\Theta_r}
p_\alpha(\boldsymbol{\theta})\,\nu(d\boldsymbol{\theta})
\]
is real-analytic and hence continuous on $\mathcal H$ for
$r\in\{s,s'\}$. Therefore, the positivity constraints
$p_\alpha(\Theta_s)>0$ and
$p_\alpha(\Theta_{s'})>0$
define an open subset of $\mathcal H$. Since $\mathcal H$ is open in
$\mathbb R^m$, $\mathcal H_{s,s'}$ is open in $\mathbb R^m$.

Fix $(a,y)\in\mathcal A\times\mathcal Y$. For
$\alpha\in\mathcal H_{s,s'}$, define
\[
g_{a,y}(\alpha)
:=
\Big(
\int_{\Theta_s}
P^a_{\boldsymbol{\theta}}(y)
p_\alpha(\boldsymbol{\theta})\,\nu(d\boldsymbol{\theta})
\Big)
\Big(
\int_{\Theta_{s'}}
p_\alpha(\boldsymbol{\theta})\,\nu(d\boldsymbol{\theta})
\Big)
-
\Big(
\int_{\Theta_{s'}}
P^a_{\boldsymbol{\theta}}(y)
p_\alpha(\boldsymbol{\theta})\,\nu(d\boldsymbol{\theta})
\Big)
\Big(
\int_{\Theta_s}
p_\alpha(\boldsymbol{\theta})\,\nu(d\boldsymbol{\theta})
\Big).
\]
By condition~\ref{nd:analytic-thm}, each integral is real-analytic on
$\mathcal H$, and hence $g_{a,y}$ is real-analytic on
$\mathcal H_{s,s'}$.

For $\alpha\in\mathcal H_{s,s'}$, we have
$p_\alpha(\Theta_s)>0$ and
$p_\alpha(\Theta_{s'})>0$, so cross-multiplication is valid.
Moreover, since $\mathcal Y$ is finite,
\[
P^a_{p_\alpha,s}
=
P^a_{p_\alpha,s'}
\iff
P^a_{p_\alpha,s}(y)
=
P^a_{p_\alpha,s'}(y)
\quad\forall y\in\mathcal Y
\iff
g_{a,y}(\alpha)=0
\quad\forall y\in\mathcal Y.
\]
Thus, the no-separation set can be written as
\[
\mathcal E_{s,s'}
:=
\Bigl\{
\alpha\in\mathcal H_{s,s'}:
P^a_{p_\alpha,s}
=
P^a_{p_\alpha,s'}
\ \ \forall a\in\mathcal A
\Bigr\}
=
\bigcap_{(a,y)\in\mathcal A\times\mathcal Y}
\Bigl\{
\alpha\in\mathcal H_{s,s'}:
g_{a,y}(\alpha)=0
\Bigr\}.
\]

Let $\{\mathcal C_j\}_{j\in J}$ denote the connected components of
$\mathcal H_{s,s'}$. Since $\mathcal H_{s,s'}$ is an open subset of
$\mathbb R^m$, each connected component $\mathcal C_j$ is open, and
the collection of connected components is at most countable. Indeed,
every nonempty open component contains a point of $\mathbb Q^m$.
Fix one component $\mathcal C_j$.

By condition~\ref{nd:witness-thm}, there exist
$\alpha_j^\star\in\mathcal C_j$ and $a_j\in\mathcal A$ such that
\[
P^{a_j}_{p_{\alpha_j^\star},s}
\neq
P^{a_j}_{p_{\alpha_j^\star},s'}.
\]
Since $\mathcal Y$ is finite, there exists
$y_j\in\mathcal Y$ with
\[
P^{a_j}_{p_{\alpha_j^\star},s}(y_j)
\neq
P^{a_j}_{p_{\alpha_j^\star},s'}(y_j),
\]
equivalently,
\[
g_{a_j,y_j}(\alpha_j^\star)\neq0.
\]
Hence $g_{a_j,y_j}$ is not identically zero on the connected open set
$\mathcal C_j$.

Let
\[
Z_j
:=
\{
\alpha\in\mathcal C_j:
g_{a_j,y_j}(\alpha)=0
\}.
\]
Because $\mathcal C_j$ is open and connected and
$g_{a_j,y_j}$ is real-analytic and not identically zero on
$\mathcal C_j$, the standard analytic zero-set property implies
\[
\lambda_m(Z_j)=0.
\]
Since
\[
\mathcal E_{s,s'}\cap\mathcal C_j
\subseteq
Z_j,
\]
we conclude
\[
\lambda_m(\mathcal E_{s,s'}\cap\mathcal C_j)=0.
\]
Since $\mathcal H_{s,s'}$ has at most countably many connected
components,
\[
\mathcal E_{s,s'}
=
\bigcup_{j\in J}
\left(
\mathcal E_{s,s'}\cap\mathcal C_j
\right)
\]
is a countable union of measure-zero sets. Therefore,
\[
\lambda_m(\mathcal E_{s,s'})=0.
\]

For each $n\in\mathbb N_0^m$, define
\[
\mathcal B^{(s,s')}_n
:=
\{
\alpha_0\in\mathcal H:
\alpha_0+n\in\mathcal E_{s,s'}
\}.
\]
Equivalently,
\[
\mathcal B^{(s,s')}_n
=
(\mathcal E_{s,s'}-n)\cap\mathcal H.
\]
Translation by $n$ preserves Lebesgue-null sets, and taking an
intersection with $\mathcal H$ preserves nullity. Hence
\[
\lambda_m(\mathcal B^{(s,s')}_n)=0.
\]
Now define
\[
\mathcal B^{(s,s')}
:=
\bigcup_{n\in\mathbb N_0^m}
\mathcal B^{(s,s')}_n.
\]
Since $\mathbb N_0^m$ is countable and each
$\mathcal B^{(s,s')}_n$ has measure zero,
\[
\lambda_m(\mathcal B^{(s,s')})=0.
\]

Finally, define
\[
\mathcal B
:=
\bigcup_{\substack{s\neq s'\\
\mathcal H_{s,s'}\neq\varnothing}}
\mathcal B^{(s,s')}.
\]
Because $\mathcal S=2^{\mathcal A}$ is finite, this is a finite union
of measure-zero sets, hence
\[
\lambda_m(\mathcal B)=0.
\]

Fix $\alpha_0\in\mathcal H\setminus\mathcal B$ and consider any round
$t$ and any realizable history $\mathcal H_t$, with posterior parameter
\[
\alpha_t=\alpha_0+N_t.
\]
Let $s\neq s'$ be any pair with
\[
p_t(s)p_t(s')>0.
\]
Then necessarily
\[
\alpha_t\in\mathcal H_{s,s'},
\]
since
\[
p_{\alpha_t}(\Theta_s)=p_t(s)>0,
\qquad
p_{\alpha_t}(\Theta_{s'})=p_t(s')>0.
\]
In particular, $\mathcal H_{s,s'}\neq\varnothing$, so
$\mathcal B^{(s,s')}\subseteq\mathcal B$. Hence
$\alpha_0\notin\mathcal B^{(s,s')}$. If
$\alpha_t\in\mathcal E_{s,s'}$, then
$\alpha_t=\alpha_0+N_t$ with $N_t\in\mathbb N_0^m$ would imply
\[
\alpha_0
\in
\mathcal B^{(s,s')}_{N_t}
\subseteq
\mathcal B^{(s,s')},
\]
a contradiction. Therefore,
\[
\alpha_t\notin\mathcal E_{s,s'}.
\]
By the definition of $\mathcal E_{s,s'}$, it is therefore not the case
that
\[
P^a_{p_{\alpha_t},s}
=
P^a_{p_{\alpha_t},s'}
\qquad
\forall a\in\mathcal A.
\]
Hence, there exists some $a^\star\in\mathcal A$ such that
\[
P^{a^\star}_{p_{\alpha_t},s}
\neq
P^{a^\star}_{p_{\alpha_t},s'}.
\]
Since total variation distance is zero if and only if the two
probability distributions are equal,
\[
\mathrm{TV}\!\left(
P^{a^\star}_{p_{\alpha_t},s},
P^{a^\star}_{p_{\alpha_t},s'}
\right)
>0.
\]

Since $q$ has full support,
\[
q(a^\star)>0.
\]
Therefore,
\[
\mathbb E_{a\sim q}
\left[
\mathrm{TV}\!\left(
P^a_{p_{\alpha_t},s},
P^a_{p_{\alpha_t},s'}
\right)^2
\right]
\ge
q(a^\star)
\mathrm{TV}\!\left(
P^{a^\star}_{p_{\alpha_t},s},
P^{a^\star}_{p_{\alpha_t},s'}
\right)^2
>0.
\]
Thus, for every pair $s\neq s'$ with
$p_t(s)p_t(s')>0$, the average squared predictive separation is
strictly positive.

For $s,s'\in\operatorname{supp}(p_t)$, define
\[
D_t(s,s')
:=
\mathbb E_{a\sim q}
\left[
\mathrm{TV}\!\left(
P^a_{p_{\alpha_t},s},
P^a_{p_{\alpha_t},s'}
\right)^2
\right],
\]
where
\[
\operatorname{supp}(p_t)
:=
\{s\in\mathcal S:p_t(s)>0\}.
\]
This quantity is well defined because
$p_t(s)>0$ and $p_t(s')>0$. We have shown that
\[
D_t(s,s')>0
\]
whenever
\[
s,s'\in\operatorname{supp}(p_t),
\qquad
s\neq s'.
\]

Let
\[
S,S'\stackrel{\mathrm{i.i.d.}}{\sim}p_t
\]
conditional on $\mathcal H_t$, and assume
\[
G_t
:=
\mathbb P(S\neq S'\mid\mathcal H_t)
>0.
\]
Because $S$ and $S'$ are conditionally independent draws from $p_t$,
\[
G_t
=
\sum_{\substack{s,s'\in\operatorname{supp}(p_t)\\s\neq s'}}
p_t(s)p_t(s').
\]
Furthermore, for
$s,s'\in\operatorname{supp}(p_t)$ with $s\neq s'$,
\begin{align*}
&\mathbb P(
S=s,S'=s'
\mid
S\neq S',\mathcal H_t
)\\
&\qquad=
\frac{
\mathbb P(
S=s,S'=s',S\neq S'
\mid
\mathcal H_t)
}{
\mathbb P(S\neq S'\mid\mathcal H_t)
}\\
&\qquad=
\frac{
\mathbb P(
S=s,S'=s'
\mid
\mathcal H_t)
}{
G_t
}
=
\frac{p_t(s)p_t(s')}{G_t},
\end{align*}
where the second equality uses
$\{S=s,S'=s'\}\subseteq\{S\neq S'\}$ whenever $s\neq s'$.

Therefore,
\begin{align*}
\mathbb E\!\left[
D_t(S,S')
\mid
S\neq S',\mathcal H_t
\right]
&=
\sum_{\substack{s,s'\in\operatorname{supp}(p_t)\\s\neq s'}}
D_t(s,s')\,
\mathbb P(
S=s,S'=s'
\mid
S\neq S',\mathcal H_t
)\\
&=
\frac{1}{G_t}
\sum_{\substack{s,s'\in\operatorname{supp}(p_t)\\s\neq s'}}
p_t(s)p_t(s')D_t(s,s').
\end{align*}
Every term in the final sum is strictly positive. Moreover,
$G_t>0$ implies that at least one distinct pair in
$\operatorname{supp}(p_t)$ exists. Therefore,
\[
\mathbb E\!\left[
D_t(S,S')
\mid
S\neq S',\mathcal H_t
\right]
>0.
\]

Substituting the definition of $D_t(S,S')$, we obtain
\[
\mathbb E\!\left[
\mathbb E_{a\sim q}
\left[
\mathrm{TV}\!\left(
P^a_{p_{\alpha_t},S},
P^a_{p_{\alpha_t},S'}
\right)^2
\right]
\;\middle|\;
S\neq S',\mathcal H_t
\right]
>0,
\]
which is exactly \eqref{eq:avgsep-positive-allt-single}.

Since $\lambda_m(\mathcal B)=0$, the above argument holds for every
Dirichlet prior parameter
$\alpha_0\in\mathcal H\setminus\mathcal B$.
By construction, $\mathcal B$ excludes every initial parameter
$\alpha_0$ for which some count update
$\alpha_0+n$, with $n\in\mathbb N_0^m$, lies in a no-separation set
for any admissible pair $(s,s')$. Since every realizable posterior
parameter has the form
\[
\alpha_t=\alpha_0+N_t,
\qquad
N_t\in\mathbb N_0^m,
\]
the preceding argument applies simultaneously to every round $t$ and
every realizable history $\mathcal H_t$. This completes the proof.
\end{proof}

\subsection{A concrete Beta--Bernoulli model satisfying Theorem~\ref{thm:avgsep-allt-single}}
\label{app:beta-bernoulli-example}

We verify the regularity and witness conditions of
Theorem~\ref{thm:avgsep-allt-single_APP} in a standard two-action,
two-objective Bayesian bandit. Let
\[
\mathcal A=\{a_1,a_2\},
\qquad
\mathbf Y_{t,a_i}
=
(Y_{t,a_i,1},Y_{t,a_i,2})
\in\{0,1\}^2.
\]
Conditional on
\[
\boldsymbol{\theta}
=
(\theta_{11},\theta_{12},\theta_{21},\theta_{22}),
\]
assume the reward coordinates are independent and
\[
Y_{t,a_i,j}\mid\boldsymbol{\theta}
\sim
\mathrm{Bernoulli}(\theta_{ij}),
\qquad
i,j\in\{1,2\}.
\]
Place independent Beta priors
\[
\theta_{ij}
\sim
\mathrm{Beta}(\alpha_{ij},\beta_{ij}),
\qquad
\alpha_{ij},\beta_{ij}>0.
\]
Recall that the Beta distribution is the two-category special case of
the Dirichlet distribution. Thus, each Bernoulli parameter has a
Dirichlet conjugate factor. Collect the hyperparameters into
\[
\alpha
=
(\alpha_{11},\beta_{11},\alpha_{12},\beta_{12},
 \alpha_{21},\beta_{21},\alpha_{22},\beta_{22})
\in\mathcal H,
\qquad
\mathcal H=(0,\infty)^8.
\]
The domain $\mathcal H$ is open and connected. After any realizable
history, the corresponding success and failure counts can be collected
into $N_t\in\mathbb N_0^8$, and conjugacy gives the posterior update
\[
\alpha_t=\alpha_0+N_t,
\]
which is exactly the count-update parameterization required by
Theorem~\ref{thm:avgsep-allt-single_APP}. Here, the reference measure
$\nu$ in Theorem~\ref{thm:avgsep-allt-single_APP} is ordinary Lebesgue
measure on $(0,1)^4$.

The three possible Pareto sets are
\[
s_1=\{a_1\},
\qquad
s_2=\{a_2\},
\qquad
s_{12}=\{a_1,a_2\}.
\]
The interiors of their corresponding parameter regions are, respectively,
\[
\{\theta_{11}>\theta_{21},\ \theta_{12}>\theta_{22}\},
\]
\[
\{\theta_{11}<\theta_{21},\ \theta_{12}<\theta_{22}\},
\]
and
\[
\{\theta_{11}>\theta_{21},\ \theta_{12}<\theta_{22}\}
\cup
\{\theta_{11}<\theta_{21},\ \theta_{12}>\theta_{22}\}.
\]
The omitted tie boundaries have Lebesgue measure zero, and each displayed
region has positive Lebesgue measure. Since every Beta density with
positive parameters is strictly positive on $(0,1)$, it follows that
\[
p_\alpha(s_1)>0,
\qquad
p_\alpha(s_2)>0,
\qquad
p_\alpha(s_{12})>0
\]
for every $\alpha\in\mathcal H$. Hence, for every distinct
$s,s'\in\{s_1,s_2,s_{12}\}$,
\[
\mathcal H_{s,s'}=\mathcal H,
\]
so every admissible set $\mathcal H_{s,s'}$ has a single connected
component.

The joint prior density has the form
\[
p_\alpha(\boldsymbol{\theta})
=
\prod_{i,j\in\{1,2\}}
\frac{
\theta_{ij}^{\alpha_{ij}-1}
(1-\theta_{ij})^{\beta_{ij}-1}
}{
B(\alpha_{ij},\beta_{ij})
}.
\]
We next verify condition~\ref{nd:analytic-thm}. Fix any
$\alpha^0\in\mathcal H$. Because all coordinates of $\alpha^0$ are
strictly positive, there is a sufficiently small complex polydisc
$\mathcal D$ around $\alpha^0$ on which every hyperparameter has
strictly positive real part. The Beta density factors and their
reciprocal normalizing constants admit holomorphic extensions in the
hyperparameters on $\mathcal D$. Moreover, after shrinking
$\mathcal D$ if necessary, there exist constants $C<\infty$ and
$\varepsilon>0$ such that, uniformly over the hyperparameters in
$\mathcal D$,
\[
\left|p_\alpha(\boldsymbol{\theta})\right|
\le
C
\prod_{i,j\in\{1,2\}}
\theta_{ij}^{\varepsilon-1}
(1-\theta_{ij})^{\varepsilon-1},
\]
and the right-hand side is integrable on $(0,1)^4$. Since
$0\le P^a_{\boldsymbol{\theta}}(y)\le1$ for
$y\in\{0,1\}^2$, the same envelope applies to the predictive
numerators. The dominated holomorphic integration theorem therefore
implies that, for every Pareto region $\Theta_s$, action $a$, and
reward $y\in\{0,1\}^2$, the maps
\[
\alpha
\mapsto
\int_{\Theta_s}
p_\alpha(\boldsymbol{\theta})\,d\boldsymbol{\theta},
\qquad
\alpha
\mapsto
\int_{\Theta_s}
P^a_{\boldsymbol{\theta}}(y)
p_\alpha(\boldsymbol{\theta})\,d\boldsymbol{\theta}
\]
are real-analytic on $\mathcal H$. Thus,
condition~\ref{nd:analytic-thm} holds.

It remains to verify the witness condition. Consider the symmetric
prior
\[
\alpha_{ij}=\beta_{ij}=1
\qquad
\text{for all }i,j,
\]
under which the four parameters are independent
$\mathrm{Unif}(0,1)$ random variables. If $U,V$ are independent
$\mathrm{Unif}(0,1)$ variables, then $\mathbb P(U>V)=1/2$ and
\[
\mathbb E[U\mid U>V]
=
\frac{\int_0^1\int_0^u u\,dv\,du}{1/2}
=
\frac{1/3}{1/2}
=
\frac{2}{3}.
\]
By symmetry,
\[
\mathbb E[U\mid U<V]=\frac{1}{3}.
\]
Therefore,
\[
\mathbb E[\theta_{11}\mid\mathcal S^*=s_1]
=
\frac{2}{3},
\qquad
\mathbb E[\theta_{11}\mid\mathcal S^*=s_2]
=
\frac{1}{3}.
\]
Under $s_{12}$, the two trade-off regions have equal probability by
symmetry. In one region $\theta_{11}>\theta_{21}$ and in the other
$\theta_{11}<\theta_{21}$, so
\[
\mathbb E[\theta_{11}\mid\mathcal S^*=s_{12}]
=
\frac12\left(\frac23+\frac13\right)
=
\frac12.
\]
For action $a_1$, the first reward coordinate is Bernoulli conditional
on $\theta_{11}$. After conditioning only on a Pareto-set candidate
$s$, its posterior-predictive marginal is therefore Bernoulli with
success probability $\mathbb E[\theta_{11}\mid\mathcal S^*=s]$.
Hence, the first-coordinate posterior-predictive marginals have distinct
success probabilities $2/3$, $1/3$, and $1/2$. Therefore, the
corresponding joint posterior-predictive reward laws are pairwise
distinct:
\[
P^{a_1}_{p_{\alpha^\star},s_1}
\neq
P^{a_1}_{p_{\alpha^\star},s_2},
\qquad
P^{a_1}_{p_{\alpha^\star},s_1}
\neq
P^{a_1}_{p_{\alpha^\star},s_{12}},
\qquad
P^{a_1}_{p_{\alpha^\star},s_2}
\neq
P^{a_1}_{p_{\alpha^\star},s_{12}},
\]
where $\alpha^\star$ denotes the symmetric hyperparameter vector.
Thus, action $a_1$ is a witness for every pair of competing Pareto-set
candidate at $\alpha^\star$. Because each
$\mathcal H_{s,s'}=\mathcal H$ is connected,
condition~\ref{nd:witness-thm} holds for every pair.

Consequently, this standard Beta--Bernoulli multi-objective bandit
satisfies the regularity and non-degeneracy conditions used in
Theorem~\ref{thm:avgsep-allt-single_APP}.

\subsection{A practical sufficient condition for quantitative separation}
\label{app:mean-separation}

For bounded rewards, Assumptions~\ref{ass:uniform_sep}--\ref{ass:decay_sep}
can be verified through posterior-predictive mean differences, which are
often easier to interpret than total variation distances. All
conditional expectations involving $A_t=a$ use the fixed-action
interpretation from Section~\ref{sec:pareto_violation}.

\begin{Proposition}
\label{prop:mean-separation}
Suppose $R_{\max}>0$ and
$\mathbf{Y}_{t,a}\in[0,R_{\max}]^d$ almost surely for every
$a\in\mathcal A$. Let $(\delta_t)_{t\ge1}$ be a deterministic positive
sequence. Assume that, at every realizable history $\mathcal{H}_t$ with
$G_t>0$, for every pair $s\neq s'$ with $p_t(s)p_t(s')>0$, there exist
an action $a\in\mathcal A$ and coordinate $j\in[d]$ such that
\[
\left|
\mathbb E[Y_{t,a,j}\mid\mathcal H_t,\mathcal S^*=s,A_t=a]
-
\mathbb E[Y_{t,a,j}\mid\mathcal H_t,\mathcal S^*=s',A_t=a]
\right|
\ge \delta_t.
\]
Then
\[
\kappa_t(\mathcal H_t)
\ge
\frac{\delta_t^2}{R_{\max}^2|\mathcal A|}.
\]
Consequently, if $\delta_t\ge\delta>0$, then
Assumption~\ref{ass:uniform_sep} holds with
\[
\kappa=\frac{\delta^2}{R_{\max}^2|\mathcal A|}.
\]
If instead $\delta_t\ge\delta t^{-\gamma}$ for some
$\delta>0$ and $\gamma\ge0$, then
Assumption~\ref{ass:decay_sep} holds with
\[
c=\frac{\delta^2}{R_{\max}^2|\mathcal A|},
\qquad
\beta=2\gamma.
\]
\end{Proposition}

\begin{proof}
Fix a realizable history $\mathcal H_t$ with $G_t>0$ and a pair
$s\neq s'$ with $p_t(s)p_t(s')>0$. By assumption, there exist an action
$a$ and coordinate $j$ for which the corresponding posterior-predictive
means differ by at least $\delta_t$. Let
\[
P=P_{t,s}^a,
\qquad
Q=P_{t,s'}^a.
\]
Since $0\le Y_{t,a,j}\le R_{\max}$, the function
$f(\mathbf y)=y_j/R_{\max}$ takes values in $[0,1]$. By the variational
characterization of total variation,
\[
\mathrm{TV}(P,Q)
\ge
\left|
\mathbb E_P[f]-\mathbb E_Q[f]
\right|
\ge
\frac{\delta_t}{R_{\max}}.
\]
Therefore, for this pair,
\[
\mathbb E_{a'\sim\mathrm{Unif}(\mathcal A)}
\left[
\mathrm{TV}\!\left(P_{t,s}^{a'},P_{t,s'}^{a'}\right)^2
\right]
\ge
\frac{\delta_t^2}{R_{\max}^2|\mathcal A|}.
\]
The same lower bound holds for every posterior-supported pair
$s\neq s'$. Averaging conditional on $S\neq S'$ gives
\[
\kappa_t(\mathcal H_t)
\ge
\frac{\delta_t^2}{R_{\max}^2|\mathcal A|}.
\]
The two stated consequences follow immediately.
\end{proof}

\subsection{Proof of Theorem \ref{thm:ts_regret}}

\thmTS*

\begin{proof}
Recall that the instantaneous Pareto-violation loss is
\[
\ell_t := \mathbbm{1}\{A_t \notin \mathcal{S}^*\},
\]
and that the Bayesian cumulative Pareto-violation regret is
\[
\mathbb{E}\!\left[\sum_{t=1}^T \ell_t\right]
=
\mathbb{E}\left[\sum_{t=1}^T \mathbbm{1}\{A_t \notin \mathcal{S}^*\}\right].
\]
Also recall the disagreement probability
\[
G_t := \mathbb{P}(S \neq S' \mid \mathcal{H}_t),
\]
where \(S,S'\stackrel{\mathrm{i.i.d.}}{\sim}p_t\), and
\(
p_t(s)=\mathbb{P}(\mathcal{S}^*=s\mid\mathcal{H}_t)
\)
denotes the posterior distribution over candidate Pareto sets, with
\(
\mathcal{S}^*:=\mathcal{S}^*(\boldsymbol{\theta})
\)
representing the Pareto set induced by the random parameter
\(\boldsymbol{\theta}\).

Fix versions of the relevant regular conditional distributions as probability
kernels in the history. For every realizable history \(h_t\), define the
history-dependent one-step information gain
\[
I_t(h_t)
:=
I(\mathcal{S}^*;\mathbf{Y}_{t,A_t}
\mid \mathcal{H}_t=h_t,A_t),
\]
equivalently,
\[
I_t(h_t)
=
\sum_{a\in\mathcal{A}}
\mathbb{P}(A_t=a\mid\mathcal{H}_t=h_t)\,
I(\mathcal{S}^*;\mathbf{Y}_{t,a}
\mid\mathcal{H}_t=h_t,A_t=a).
\]
We write \(I_t:=I_t(\mathcal{H}_t)\) for the corresponding
\(\mathcal{H}_t\)-measurable random quantity.

Under Algorithm~\ref{alg:ts_pareto}, we first draw
\[
B_t \sim \mathrm{Bernoulli}(1-\varepsilon_t).
\]
At \(t=1\), we have \(\varepsilon_1=1\), so \(B_1=0\) almost surely and
\[
\mathbb{E}[\ell_1\mid\mathcal{H}_1]\le 1=\varepsilon_1.
\]
Thus, the loss bound established below also holds at \(t=1\). For the
branch-conditioning argument, we may therefore assume \(t\ge2\), for which
\(1-\varepsilon_t>0\).

If \(B_t=0\), then \(S_t=\mathcal{A}\) and
\(A_t\sim\mathrm{Unif}(\mathcal{A})\), and hence trivially
\[
\ell_t\le1.
\]
If \(B_t=1\), then
\(\boldsymbol{\theta}_t\sim p_t(\boldsymbol{\theta})\),
\(S_t=\mathcal{S}^*(\boldsymbol{\theta}_t)\), and
\(A_t\sim\mathrm{Unif}(S_t)\). In this case, if
\(S_t=\mathcal{S}^*\), then \(A_t\in\mathcal{S}^*\) almost surely, and thus
\(\ell_t=0\). Therefore,
\[
\ell_t
\le
\mathbbm{1}\{B_t=0\}
+
\mathbbm{1}\{B_t=1,\;S_t\neq\mathcal{S}^*\}.
\]

Taking conditional expectation given \(\mathcal{H}_t\), for \(t\ge2\) we obtain
\[
\mathbb{E}[\ell_t\mid\mathcal{H}_t]
\le
\varepsilon_t
+
(1-\varepsilon_t)
\mathbb{P}(S_t\neq\mathcal{S}^*
\mid\mathcal{H}_t,B_t=1).
\]

Conditional on \(\mathcal{H}_t\), the branch variable \(B_t\) is
generated using fresh algorithmic randomness independent of the true parameter
\(\boldsymbol{\theta}\). Hence, conditioning additionally on \(B_t=1\) does not
change the posterior distribution of \(\boldsymbol{\theta}\), so
\[
\boldsymbol{\theta}\mid(\mathcal{H}_t,B_t=1)
\sim p_t(\boldsymbol{\theta}).
\]
On this branch, the Thompson sample
\(\boldsymbol{\theta}_t\sim p_t(\boldsymbol{\theta})\) is generated using fresh
randomness independent of \(\boldsymbol{\theta}\) conditional on
\((\mathcal{H}_t,B_t=1)\). Therefore,
\(\boldsymbol{\theta}\) and \(\boldsymbol{\theta}_t\) are conditionally
independent and identically distributed according to
\(p_t(\boldsymbol{\theta})\). Consequently,
\[
\mathcal{S}^*=\mathcal{S}^*(\boldsymbol{\theta})
\quad\text{and}\quad
S_t=\mathcal{S}^*(\boldsymbol{\theta}_t)
\]
are conditionally independent and identically distributed according to the
induced posterior
\[
p_t(s)=\mathbb{P}(\mathcal{S}^*=s\mid\mathcal{H}_t).
\]
Thus,
\[
\mathbb{P}(S_t\neq\mathcal{S}^*
\mid\mathcal{H}_t,B_t=1)
=
\mathbb{P}(S\neq S'\mid\mathcal{H}_t)
=
G_t,
\]
where \(S,S'\stackrel{\mathrm{i.i.d.}}{\sim}p_t\). Together with the already
verified \(t=1\) case, this gives, for every \(t\),
\begin{equation}
\mathbb{E}[\ell_t\mid\mathcal{H}_t]
\le
\varepsilon_t+(1-\varepsilon_t)G_t.
\label{eq:thm32-loss-bound}
\end{equation}

Define \(r_t:=(1-\varepsilon_t)G_t\). Then
\begin{equation}
\mathbb{E}[\ell_t\mid\mathcal{H}_t]
\le
\varepsilon_t+r_t,
\qquad\text{with}\qquad
r_t\le G_t.
\label{eq:thm32-rt-def}
\end{equation}

Fix \(t\), and let
\[
\pi_t(a):=\mathbb{P}(A_t=a\mid\mathcal{H}_t),
\qquad a\in\mathcal{A},
\]
denote the conditional action distribution induced by
Algorithm~\ref{alg:ts_pareto}. Since with probability \(\varepsilon_t\) the
algorithm explores uniformly over \(\mathcal{A}\), we have
\begin{equation}
\pi_t(a)\ge\frac{\varepsilon_t}{|\mathcal{A}|}
\qquad\text{for all }a\in\mathcal{A}.
\label{eq:thm32-pi-lower}
\end{equation}

By the definition of \(I_t\),
\[
I_t
=
\sum_{a\in\mathcal{A}}
\pi_t(a)\,
I(\mathcal{S}^*;\mathbf{Y}_{t,a}
\mid\mathcal{H}_t,A_t=a).
\]
Applying Lemma~\ref{lem:info_gain_tv} to each action \(a\), we obtain
\[
I(\mathcal{S}^*;\mathbf{Y}_{t,a}
\mid\mathcal{H}_t,A_t=a)
\ge
\frac12
\mathbb{E}_{S,S'\stackrel{\mathrm{i.i.d.}}{\sim}p_t}
\!\left[
\mathrm{TV}\!\left(P_{t,S}^a,P_{t,S'}^a\right)^2
\right].
\]
Therefore,
\[
I_t
\ge
\frac12
\sum_{a\in\mathcal{A}}
\pi_t(a)\,
\mathbb{E}_{S,S'\stackrel{\mathrm{i.i.d.}}{\sim}p_t}
\!\left[
\mathrm{TV}\!\left(P_{t,S}^a,P_{t,S'}^a\right)^2
\right].
\]
Using \eqref{eq:thm32-pi-lower} and exchanging the finite sum and expectation,
we obtain
\[
I_t
\ge
\frac{\varepsilon_t}{2|\mathcal{A}|}
\sum_{a\in\mathcal{A}}
\mathbb{E}_{S,S'\stackrel{\mathrm{i.i.d.}}{\sim}p_t}
\!\left[
\mathrm{TV}\!\left(P_{t,S}^a,P_{t,S'}^a\right)^2
\right]
=
\frac{\varepsilon_t}{2}
\mathbb{E}_{S,S'\stackrel{\mathrm{i.i.d.}}{\sim}p_t}
\!\left[
\mathbb{E}_{a\sim\mathrm{Unif}(\mathcal{A})}
\mathrm{TV}\!\left(P_{t,S}^a,P_{t,S'}^a\right)^2
\right].
\]

If \(G_t=0\), then \(r_t=0\), so all subsequent upper bounds on \(r_t\) hold
trivially. We may therefore suppose \(G_t>0\). Since the total variation term
vanishes whenever \(S=S'\), conditioning on the event \(\{S\neq S'\}\) gives
\[
I_t
\ge
\frac{\varepsilon_t}{2}
\mathbb{P}(S\neq S'\mid\mathcal{H}_t)
\mathbb{E}\!\left[
\mathbb{E}_{a\sim\mathrm{Unif}(\mathcal{A})}
\mathrm{TV}\!\left(P_{t,S}^a,P_{t,S'}^a\right)^2
\;\middle|\;
S\neq S',\mathcal{H}_t
\right].
\]
Recalling that
\(\mathbb{P}(S\neq S'\mid\mathcal{H}_t)=G_t\),
this yields
\[
I_t
\ge
\frac{\varepsilon_t}{2}
G_t
\mathbb{E}\!\left[
\mathbb{E}_{a\sim\mathrm{Unif}(\mathcal{A})}
\mathrm{TV}\!\left(P_{t,S}^a,P_{t,S'}^a\right)^2
\;\middle|\;
S\neq S',\mathcal{H}_t
\right].
\]

\medskip
\noindent
\textbf{Case 1:}
By Assumption~\ref{ass:uniform_sep} (Uniform separation),
\[
\mathbb{E}\!\left[
\mathbb{E}_{a\sim\mathrm{Unif}(\mathcal{A})}
\mathrm{TV}\!\left(P_{t,S}^a,P_{t,S'}^a\right)^2
\;\middle|\;
S\neq S',\mathcal{H}_t
\right]
\ge\kappa
\]
whenever \(G_t>0\). Therefore,
\begin{equation}
I_t
\ge
\frac{\varepsilon_t\kappa}{2}G_t.
\label{eq:thm32-info-uniform}
\end{equation}
Since \(r_t\le G_t\) by \eqref{eq:thm32-rt-def},
\[
r_t
\le
\frac{2}{\varepsilon_t\kappa}I_t.
\]
Together with the trivial \(G_t=0\) case, this inequality holds at every
realizable history. With \(\varepsilon_t=t^{-1/2}\),
\begin{equation}
r_t
\le
\frac{2\sqrt{t}}{\kappa}I_t
\le
\frac{2\sqrt{T}}{\kappa}I_t.
\label{eq:thm32-rt-uniform}
\end{equation}

\medskip
\noindent
\textbf{Case 2:}
By Assumption~\ref{ass:decay_sep} (Decaying separation),
\[
\mathbb{E}\!\left[
\mathbb{E}_{a\sim\mathrm{Unif}(\mathcal{A})}
\mathrm{TV}\!\left(P_{t,S}^a,P_{t,S'}^a\right)^2
\;\middle|\;
S\neq S',\mathcal{H}_t
\right]
\ge
c\,t^{-\beta}
\]
whenever \(G_t>0\). Therefore,
\begin{equation}
I_t
\ge
\frac{\varepsilon_t c}{2}
t^{-\beta}G_t.
\label{eq:thm32-info-decay}
\end{equation}
Again using \(r_t\le G_t\),
\[
r_t
\le
\frac{2t^\beta}{\varepsilon_t c}I_t.
\]
Together with the trivial \(G_t=0\) case, this inequality holds at every
realizable history. With \(\varepsilon_t=t^{-1/2}\),
\begin{equation}
r_t
\le
\frac{2}{c}
t^{\beta+\frac12}I_t
\le
\frac{2}{c}
T^{\beta+\frac12}I_t.
\label{eq:thm32-rt-decay}
\end{equation}

We now control the expected cumulative information gain about the unknown
Pareto set \(\mathcal{S}^*\). Since \(\mathcal{H}_1\) is empty and
\[
\mathcal{H}_{t+1}
=
(\mathcal{H}_t,A_t,\mathbf{Y}_{t,A_t}),
\]
the chain rule for mutual information gives
\[
I(\mathcal{S}^*;\mathcal{H}_{T+1})
=
\sum_{t=1}^T
I(\mathcal{S}^*;
A_t,\mathbf{Y}_{t,A_t}
\mid\mathcal{H}_t).
\]
The action \(A_t\) is selected based on \(\mathcal{H}_t\) using only independent
internal randomization of the algorithm. Hence
\[
I(\mathcal{S}^*;A_t\mid\mathcal{H}_t)=0.
\]
Applying the chain rule again,
\[
I(\mathcal{S}^*;
A_t,\mathbf{Y}_{t,A_t}
\mid\mathcal{H}_t)
=
I(\mathcal{S}^*;A_t\mid\mathcal{H}_t)
+
I(\mathcal{S}^*;
\mathbf{Y}_{t,A_t}
\mid\mathcal{H}_t,A_t)
=
\mathbb{E}[I_t],
\]
where the final equality follows by averaging the history-dependent quantity
\(I_t=I_t(\mathcal{H}_t)\) over \(\mathcal{H}_t\). Consequently,
\[
I(\mathcal{S}^*;\mathcal{H}_{T+1})
=
\sum_{t=1}^T\mathbb{E}[I_t]
=
\mathbb{E}\!\left[\sum_{t=1}^T I_t\right].
\]
Therefore,
\begin{equation}
\mathbb{E}\!\left[\sum_{t=1}^T I_t\right]
=
I(\mathcal{S}^*;\mathcal{H}_{T+1})
\le
H(\mathcal{S}^*),
\label{eq:thm32-info-budget}
\end{equation}
where the inequality follows because \(\mathcal{S}^*\) is discrete and mutual
information is upper bounded by entropy.

We now proceed as follows:
\begin{equation}
\mathbb{E}\!\left[\sum_{t=1}^T \ell_t\right]
=
\mathbb{E}\!\left[
\sum_{t=1}^T
\mathbb{E}[\ell_t\mid\mathcal{H}_t]
\right]
\le
\mathbb{E}\!\left[
\sum_{t=1}^T(\varepsilon_t+r_t)
\right]
=
\sum_{t=1}^T\varepsilon_t
+
\mathbb{E}\!\left[
\sum_{t=1}^T r_t
\right].
\label{eq:thm32-master}
\end{equation}
Here, the first equality follows from the tower property of conditional
expectation, and the inequality uses \eqref{eq:thm32-rt-def}.

Since \(\varepsilon_t=t^{-1/2}\),
\begin{equation}
\sum_{t=1}^T\varepsilon_t
=
\sum_{t=1}^T\frac{1}{\sqrt{t}}
\le
2\sqrt{T}.
\label{eq:thm32-eps-sum}
\end{equation}

\medskip
\noindent
\textbf{Under Assumption~\ref{ass:uniform_sep}.}
Using \eqref{eq:thm32-rt-uniform},
\[
r_t
\le
\frac{2\sqrt{T}}{\kappa}I_t.
\]
Summing over \(t=1,\dots,T\) and taking expectations gives
\[
\mathbb{E}\!\left[
\sum_{t=1}^T r_t
\right]
\le
\frac{2\sqrt{T}}{\kappa}
\mathbb{E}\!\left[
\sum_{t=1}^T I_t
\right].
\]
Applying \eqref{eq:thm32-info-budget},
\[
\mathbb{E}\!\left[
\sum_{t=1}^T r_t
\right]
\le
\frac{2H(\mathcal{S}^*)}{\kappa}\sqrt{T}.
\]
Combining this with \eqref{eq:thm32-master} and
\eqref{eq:thm32-eps-sum}, we conclude
\[
\mathbb{E}\!\left[
\sum_{t=1}^T
\mathbbm{1}\{A_t\notin\mathcal{S}^*\}
\right]
\le
\left(
2+\frac{2H(\mathcal{S}^*)}{\kappa}
\right)\sqrt{T}.
\]

\medskip
\noindent
\textbf{Under Assumption~\ref{ass:decay_sep}.}
Using \eqref{eq:thm32-rt-decay},
\[
r_t
\le
\frac{2}{c}
t^{\beta+\frac12}I_t
\le
\frac{2}{c}
T^{\beta+\frac12}I_t,
\]
where the second inequality uses \(t\le T\) and \(\beta\ge0\). Summing over
\(t\) and taking expectations gives
\[
\mathbb{E}\!\left[
\sum_{t=1}^T r_t
\right]
\le
\frac{2T^{\beta+\frac12}}{c}
\mathbb{E}\!\left[
\sum_{t=1}^T I_t
\right].
\]
Applying \eqref{eq:thm32-info-budget},
\[
\mathbb{E}\!\left[
\sum_{t=1}^T r_t
\right]
\le
\frac{2H(\mathcal{S}^*)}{c}
T^{\beta+\frac12}.
\]
Combining this with \eqref{eq:thm32-master} and
\eqref{eq:thm32-eps-sum} yields
\[
\mathbb{E}\!\left[
\sum_{t=1}^T
\mathbbm{1}\{A_t\notin\mathcal{S}^*\}
\right]
\le
2\sqrt{T}
+
\frac{2H(\mathcal{S}^*)}{c}
T^{\beta+\frac12}.
\]

Finally, if \(\beta<\tfrac12\), then
\(\beta+\tfrac12<1\), so both terms on the right-hand side are \(o(T)\),
implying sublinear regret.

This completes the proof.
\end{proof}

\subsection{Proof of Proposition \ref{prop:ts_set_mismatch}}

\propTSMismatch*

\begin{proof}
Recall that under Algorithm~\ref{alg:ts_pareto}, the sampled set \(S_t\) is
defined by
\[
S_t=
\begin{cases}
\mathcal{S}^*(\boldsymbol{\theta}_t), & \text{if } B_t=1,\\
\mathcal{A}, & \text{if } B_t=0,
\end{cases}
\]
where \(B_t\sim\mathrm{Bernoulli}(1-\varepsilon_t)\) and, on the branch
\(B_t=1\), \(\boldsymbol{\theta}_t\sim p_t(\boldsymbol{\theta})\).

Define the instantaneous Pareto-set mismatch loss
\[
m_t:=\mathbbm{1}\{S_t\neq\mathcal{S}^*\}.
\]

We first bound its conditional expectation. At \(t=1\),
\(\varepsilon_1=1\), so \(B_1=0\) almost surely and therefore
\[
\mathbb{E}[m_1\mid\mathcal{H}_1]
\le
1
=
\varepsilon_1.
\]
Thus the bound derived below already holds at \(t=1\). For the
branch-conditioning argument, we may therefore assume \(t\ge2\), for which
\(1-\varepsilon_t>0\).

If \(B_t=0\), then \(S_t=\mathcal{A}\), and hence trivially
\[
m_t\le1.
\]
If \(B_t=1\), then
\[
S_t=\mathcal{S}^*(\boldsymbol{\theta}_t).
\]

Conditional on \(\mathcal{H}_t\), the branch variable \(B_t\) is generated
using fresh algorithmic randomness independent of the true parameter
\(\boldsymbol{\theta}\). Hence conditioning additionally on \(B_t=1\) does
not change the posterior distribution of \(\boldsymbol{\theta}\), so
\[
\boldsymbol{\theta}\mid(\mathcal{H}_t,B_t=1)
\sim p_t(\boldsymbol{\theta}).
\]
On this branch, the Thompson sample
\(\boldsymbol{\theta}_t\sim p_t(\boldsymbol{\theta})\) is generated using
fresh randomness independent of \(\boldsymbol{\theta}\) conditional on
\((\mathcal{H}_t,B_t=1)\). Therefore,
\(\boldsymbol{\theta}\) and \(\boldsymbol{\theta}_t\) are conditionally
independent and identically distributed according to
\(p_t(\boldsymbol{\theta})\). Consequently,
\[
\mathcal{S}^*
=
\mathcal{S}^*(\boldsymbol{\theta})
\quad\text{and}\quad
S_t
=
\mathcal{S}^*(\boldsymbol{\theta}_t)
\]
are conditionally independent and identically distributed according to the
induced posterior
\[
p_t(s)
=
\mathbb{P}(\mathcal{S}^*=s\mid\mathcal{H}_t).
\]
Hence, for \(t\ge2\),
\[
\mathbb{P}(S_t\neq\mathcal{S}^*
\mid\mathcal{H}_t,B_t=1)
=
\mathbb{P}(S\neq S'\mid\mathcal{H}_t)
=
G_t,
\]
where \(S,S'\stackrel{\mathrm{i.i.d.}}{\sim}p_t\).

Moreover,
\[
m_t
=
\mathbbm{1}\{S_t\neq\mathcal{S}^*\}
\le
\mathbbm{1}\{B_t=0\}
+
\mathbbm{1}\{B_t=1,\;S_t\neq\mathcal{S}^*\}.
\]
Taking conditional expectation given \(\mathcal{H}_t\), for \(t\ge2\),
gives
\[
\mathbb{E}[m_t\mid\mathcal{H}_t]
\le
\varepsilon_t
+
(1-\varepsilon_t)
\mathbb{P}(S_t\neq\mathcal{S}^*
\mid\mathcal{H}_t,B_t=1).
\]
Substituting the preceding identity, and combining with the already verified
\(t=1\) case, yields for every \(t\),
\[
\mathbb{E}[m_t\mid\mathcal{H}_t]
\le
\varepsilon_t+(1-\varepsilon_t)G_t.
\]

Define
\[
r_t:=(1-\varepsilon_t)G_t.
\]
Then
\begin{equation}
\mathbb{E}[m_t\mid\mathcal{H}_t]
\le
\varepsilon_t+r_t,
\qquad\text{with}\qquad
r_t\le G_t.
\label{eq:prop-mismatch-rt-def}
\end{equation}

We next relate \(r_t\) to information gain. Fix versions of the relevant
regular conditional distributions as probability kernels in the history.
For every realizable history \(h_t\), define
\[
I_t(h_t)
:=
I(\mathcal{S}^*;\mathbf{Y}_{t,A_t}
\mid\mathcal{H}_t=h_t,A_t),
\]
and write \(I_t:=I_t(\mathcal{H}_t)\). Let
\[
\pi_t(a)
:=
\mathbb{P}(A_t=a\mid\mathcal{H}_t),
\qquad a\in\mathcal{A}.
\]
Since Algorithm~\ref{alg:ts_pareto} explores uniformly over
\(\mathcal{A}\) with probability \(\varepsilon_t\),
\[
\pi_t(a)
\ge
\frac{\varepsilon_t}{|\mathcal{A}|}
\qquad
\text{for all }a\in\mathcal{A}.
\]
By the definition of \(I_t\),
\[
I_t
=
\sum_{a\in\mathcal{A}}
\pi_t(a)\,
I(\mathcal{S}^*;\mathbf{Y}_{t,a}
\mid\mathcal{H}_t,A_t=a).
\]
Applying Lemma~\ref{lem:info_gain_tv} to each action \(a\),
\[
I(\mathcal{S}^*;\mathbf{Y}_{t,a}
\mid\mathcal{H}_t,A_t=a)
\ge
\frac12
\mathbb{E}_{S,S'\stackrel{\mathrm{i.i.d.}}{\sim}p_t}
\!\left[
\mathrm{TV}\!\left(
P_{t,S}^a,P_{t,S'}^a
\right)^2
\right].
\]
Therefore,
\[
I_t
\ge
\frac12
\sum_{a\in\mathcal{A}}
\pi_t(a)\,
\mathbb{E}_{S,S'\stackrel{\mathrm{i.i.d.}}{\sim}p_t}
\!\left[
\mathrm{TV}\!\left(
P_{t,S}^a,P_{t,S'}^a
\right)^2
\right].
\]
Using the lower bound on \(\pi_t(a)\),
\[
I_t
\ge
\frac{\varepsilon_t}{2|\mathcal{A}|}
\sum_{a\in\mathcal{A}}
\mathbb{E}_{S,S'\stackrel{\mathrm{i.i.d.}}{\sim}p_t}
\!\left[
\mathrm{TV}\!\left(
P_{t,S}^a,P_{t,S'}^a
\right)^2
\right],
\]
and hence
\[
I_t
\ge
\frac{\varepsilon_t}{2}
\mathbb{E}_{S,S'\stackrel{\mathrm{i.i.d.}}{\sim}p_t}
\!\left[
\mathbb{E}_{a\sim\mathrm{Unif}(\mathcal{A})}
\mathrm{TV}\!\left(
P_{t,S}^a,P_{t,S'}^a
\right)^2
\right].
\]

If \(G_t=0\), then \(r_t=0\), so all subsequent upper bounds on \(r_t\)
hold trivially. We may therefore assume \(G_t>0\). Since the total
variation term vanishes whenever \(S=S'\), conditioning on
\(\{S\neq S'\}\) gives
\[
I_t
\ge
\frac{\varepsilon_t}{2}
G_t
\mathbb{E}\!\left[
\mathbb{E}_{a\sim\mathrm{Unif}(\mathcal{A})}
\mathrm{TV}\!\left(
P_{t,S}^a,P_{t,S'}^a
\right)^2
\;\middle|\;
S\neq S',\mathcal{H}_t
\right].
\]

Under Assumption~\ref{ass:uniform_sep},
\[
\mathbb{E}\!\left[
\mathbb{E}_{a\sim\mathrm{Unif}(\mathcal{A})}
\mathrm{TV}\!\left(
P_{t,S}^a,P_{t,S'}^a
\right)^2
\;\middle|\;
S\neq S',\mathcal{H}_t
\right]
\ge
\kappa
\]
whenever \(G_t>0\). Consequently,
\[
I_t
\ge
\frac{\varepsilon_t\kappa}{2}G_t.
\]
Using \(r_t\le G_t\) and
\(\varepsilon_t=t^{-1/2}\), we obtain
\begin{equation}
r_t
\le
\frac{2}{\varepsilon_t\kappa}I_t
=
\frac{2\sqrt{t}}{\kappa}I_t
\le
\frac{2\sqrt{T}}{\kappa}I_t.
\label{eq:prop-mismatch-uniform}
\end{equation}
Together with the trivial \(G_t=0\) case, this inequality holds at every
realizable history.

Similarly, under Assumption~\ref{ass:decay_sep},
\[
\mathbb{E}\!\left[
\mathbb{E}_{a\sim\mathrm{Unif}(\mathcal{A})}
\mathrm{TV}\!\left(
P_{t,S}^a,P_{t,S'}^a
\right)^2
\;\middle|\;
S\neq S',\mathcal{H}_t
\right]
\ge
c\,t^{-\beta}
\]
whenever \(G_t>0\). Thus
\[
I_t
\ge
\frac{\varepsilon_t c}{2}
t^{-\beta}G_t.
\]
Using again \(r_t\le G_t\) and
\(\varepsilon_t=t^{-1/2}\), we obtain
\begin{equation}
r_t
\le
\frac{2t^\beta}{\varepsilon_t c}I_t
=
\frac{2}{c}t^{\beta+\frac12}I_t
\le
\frac{2}{c}T^{\beta+\frac12}I_t.
\label{eq:prop-mismatch-decay}
\end{equation}
Again, together with the trivial \(G_t=0\) case, this holds at every
realizable history.

We now control the expected cumulative information gain. Since
\(\mathcal{H}_1\) is empty and
\[
\mathcal{H}_{t+1}
=
(\mathcal{H}_t,A_t,\mathbf{Y}_{t,A_t}),
\]
the chain rule for mutual information gives
\[
I(\mathcal{S}^*;\mathcal{H}_{T+1})
=
\sum_{t=1}^T
I(\mathcal{S}^*;
A_t,\mathbf{Y}_{t,A_t}
\mid\mathcal{H}_t).
\]
Since \(A_t\) is selected based on \(\mathcal{H}_t\) using independent
internal randomization,
\[
I(\mathcal{S}^*;A_t\mid\mathcal{H}_t)=0.
\]
Therefore,
\[
I(\mathcal{S}^*;
A_t,\mathbf{Y}_{t,A_t}
\mid\mathcal{H}_t)
=
I(\mathcal{S}^*;
\mathbf{Y}_{t,A_t}
\mid\mathcal{H}_t,A_t)
=
\mathbb{E}[I_t],
\]
where the final equality follows by averaging the history-dependent quantity
\(I_t=I_t(\mathcal{H}_t)\) over \(\mathcal{H}_t\). Consequently,
\[
I(\mathcal{S}^*;\mathcal{H}_{T+1})
=
\sum_{t=1}^T\mathbb{E}[I_t]
=
\mathbb{E}\!\left[
\sum_{t=1}^T I_t
\right].
\]
Since \(\mathcal{S}^*\) is finite-valued,
\begin{equation}
\mathbb{E}\!\left[
\sum_{t=1}^T I_t
\right]
=
I(\mathcal{S}^*;\mathcal{H}_{T+1})
\le
H(\mathcal{S}^*).
\label{eq:prop-mismatch-info-budget}
\end{equation}

Summing \eqref{eq:prop-mismatch-rt-def} over \(t=1,\dots,T\) and taking
expectations gives
\begin{equation}
\mathbb{E}\!\left[
\sum_{t=1}^T m_t
\right]
\le
\sum_{t=1}^T\varepsilon_t
+
\mathbb{E}\!\left[
\sum_{t=1}^T r_t
\right].
\label{eq:prop-mismatch-master}
\end{equation}
Since \(\varepsilon_t=t^{-1/2}\),
\begin{equation}
\sum_{t=1}^T\varepsilon_t
=
\sum_{t=1}^T\frac{1}{\sqrt{t}}
\le
2\sqrt{T}.
\label{eq:prop-mismatch-eps-sum}
\end{equation}

Under Assumption~\ref{ass:uniform_sep}, using
\eqref{eq:prop-mismatch-uniform} and
\eqref{eq:prop-mismatch-info-budget},
\[
\mathbb{E}\!\left[
\sum_{t=1}^T r_t
\right]
\le
\frac{2\sqrt{T}}{\kappa}
\mathbb{E}\!\left[
\sum_{t=1}^T I_t
\right]
\le
\frac{2H(\mathcal{S}^*)}{\kappa}\sqrt{T}.
\]
Combining this with
\eqref{eq:prop-mismatch-master} and
\eqref{eq:prop-mismatch-eps-sum}, we obtain
\[
\mathbb{E}\!\left[
\sum_{t=1}^T
\mathbbm{1}\{S_t\neq\mathcal{S}^*\}
\right]
\le
\left(
2+\frac{2H(\mathcal{S}^*)}{\kappa}
\right)\sqrt{T}.
\]

Under Assumption~\ref{ass:decay_sep}, using
\eqref{eq:prop-mismatch-decay} and
\eqref{eq:prop-mismatch-info-budget},
\[
\mathbb{E}\!\left[
\sum_{t=1}^T r_t
\right]
\le
\frac{2T^{\beta+\frac12}}{c}
\mathbb{E}\!\left[
\sum_{t=1}^T I_t
\right]
\le
\frac{2H(\mathcal{S}^*)}{c}
T^{\beta+\frac12}.
\]
Combining this with
\eqref{eq:prop-mismatch-master} and
\eqref{eq:prop-mismatch-eps-sum}, we obtain
\[
\mathbb{E}\!\left[
\sum_{t=1}^T
\mathbbm{1}\{S_t\neq\mathcal{S}^*\}
\right]
\le
2\sqrt{T}
+
\frac{2H(\mathcal{S}^*)}{c}
T^{\beta+\frac12}.
\]

Finally, if \(\beta<\tfrac12\), then
\(\beta+\tfrac12<1\), so both terms on the right-hand side are
\(o(T)\), implying sublinear cumulative Pareto-set mismatch.

This completes the proof.
\end{proof}

\subsection{Proof of Theorem \ref{thm:ts_regret_ids}}

\thmTSids*

\begin{proof}
Recall that the instantaneous Pareto-violation loss is
\[
\ell_t:=\mathbbm{1}\{A_t\notin\mathcal{S}^*\}.
\]
For Algorithm~\ref{alg:IDS}, the Bayesian cumulative Pareto-violation regret is
\[
\mathbb{E}\!\left[\sum_{t=1}^T\ell_t\right]
=
\mathbb{E}\!\left[
\sum_{t=1}^T
\mathbbm{1}\{A_t\notin\mathcal{S}^*\}
\right].
\]

Fix versions of the relevant regular conditional distributions as probability
kernels in the history. For every realizable history \(h_t\) and action
\(a\in\mathcal{A}\), define
\[
\Delta_t(a;h_t)
:=
\mathbb{P}
\!\left(
a\notin\mathcal{S}^*
\mid
\mathcal{H}_t=h_t
\right)
\]
and
\[
I_t(a;h_t)
:=
I\!\left(
\mathcal{S}^*;\mathbf{Y}_{t,a}
\mid
\mathcal{H}_t=h_t,A_t=a
\right),
\]
where the latter uses the fixed-action interpretation from Section~3.
Since \(\mathcal{S}^*\in2^{\mathcal{A}}\) and \(\mathcal{A}\) is finite,
\(I_t(a;h_t)\) is finite for every \(a\) and realizable \(h_t\). For
\(\pi\in\Pi(\mathcal{A})\), define
\[
\Delta_t(\pi;h_t)
:=
\sum_{a\in\mathcal{A}}
\pi(a)\Delta_t(a;h_t),
\qquad
I_t(\pi;h_t)
:=
\sum_{a\in\mathcal{A}}
\pi(a)I_t(a;h_t).
\]

For \(x,y\ge0\), define the extended quadratic-over-linear function
\[
\varphi(x,y)
:=
\begin{cases}
x^2/y, & y>0,\\
0, & x=0,\ y=0,\\
+\infty, & x>0,\ y=0.
\end{cases}
\]
Thus the history-wise IDS objective can be written as
\[
\psi_t(\pi;h_t)
:=
\varphi\!\left(
\Delta_t(\pi;h_t),
I_t(\pi;h_t)
\right).
\]

For fixed \(h_t\), both
\(\Delta_t(\pi;h_t)\) and \(I_t(\pi;h_t)\) are affine, hence continuous,
functions of \(\pi\). Moreover, \(\varphi\) is lower semicontinuous on
\(\mathbb{R}_+^2\), so
\(\pi\mapsto\psi_t(\pi;h_t)\) is lower semicontinuous. Since
\(\mathcal{A}\) is finite, \(\Pi(\mathcal{A})\) is a compact metric
simplex. Therefore, for every realizable history \(h_t\), the minimum is
attained and the set
\[
\Gamma_t(h_t)
:=
\operatorname*{arg\,min}_{\pi\in\Pi(\mathcal{A})}
\psi_t(\pi;h_t)
\]
is nonempty and compact.

The history space is standard Borel, and, under the fixed regular
conditional probability kernels above, the maps
\(h_t\mapsto\Delta_t(a;h_t)\) and \(h_t\mapsto I_t(a;h_t)\) are
measurable for every \(a\in\mathcal{A}\). Since \(\mathcal{A}\) is
finite, it follows that
\((h_t,\pi)\mapsto\psi_t(\pi;h_t)\) is jointly measurable and is lower
semicontinuous in \(\pi\). Hence the measurable minimum/selection theorem
for a measurable lower-semicontinuous objective over a compact metric
action space yields a measurable selector
\[
h_t\longmapsto\pi_t^{\mathrm{IDS}}(h_t)
\in\Gamma_t(h_t).
\]
We fix one such selector. Consequently,
\(\pi_t^{\mathrm{IDS}}=\pi_t^{\mathrm{IDS}}(\mathcal{H}_t)\) is
\(\mathcal{H}_t\)-measurable.

We suppress the explicit dependence on \(h_t\) and write
\(\Delta_t(a)\), \(I_t(a)\), \(\Delta_t(\pi)\), and \(I_t(\pi)\) for
the corresponding \(\mathcal{H}_t\)-measurable random quantities evaluated
at \(h_t=\mathcal{H}_t\).

Conditional on \(\mathcal{H}_t\), the distribution
\(\pi_t^{\mathrm{IDS}}\) is fixed, and \(A_t\) is generated from it using
fresh algorithmic randomness independent of the true parameter and hence of
\(\mathcal{S}^*\). Therefore,
\[
A_t\indep\mathcal{S}^*\mid\mathcal{H}_t.
\]
Consequently,
\[
\mathbb{E}[\ell_t\mid\mathcal{H}_t]
=
\sum_{a\in\mathcal{A}}
\pi_t^{\mathrm{IDS}}(a)\,
\mathbb{P}(a\notin\mathcal{S}^*\mid\mathcal{H}_t)
=
\Delta_t(\pi_t^{\mathrm{IDS}}).
\]
Hence,
\begin{equation}
\mathbb{E}\!\left[\sum_{t=1}^T\ell_t\right]
=
\mathbb{E}\!\left[
\sum_{t=1}^T
\Delta_t(\pi_t^{\mathrm{IDS}})
\right].
\label{eq:ids-regret-identity}
\end{equation}

Suppose first that
\[
\psi_t(\pi_t^{\mathrm{IDS}})\le\Psi
\]
at every realizable history and every \(t\in[T]\), for some finite
\(\Psi>0\). Then, at every such history,
\begin{equation}
\Delta_t(\pi_t^{\mathrm{IDS}})^2
\le
\Psi\,I_t(\pi_t^{\mathrm{IDS}}).
\label{eq:ids-ratio-generic}
\end{equation}
Indeed, when \(I_t(\pi_t^{\mathrm{IDS}})>0\), this follows directly from
the definition of the information ratio. When
\(I_t(\pi_t^{\mathrm{IDS}})=0\), finiteness of the ratio implies
\(\Delta_t(\pi_t^{\mathrm{IDS}})=0\), so
\eqref{eq:ids-ratio-generic} also holds.

Taking square roots,
\[
\Delta_t(\pi_t^{\mathrm{IDS}})
\le
\sqrt{\Psi}\,
\sqrt{I_t(\pi_t^{\mathrm{IDS}})}.
\]
Substituting this into \eqref{eq:ids-regret-identity} gives
\[
\mathbb{E}\!\left[\sum_{t=1}^T\ell_t\right]
\le
\sqrt{\Psi}
\sum_{t=1}^T
\mathbb{E}\!\left[
\sqrt{I_t(\pi_t^{\mathrm{IDS}})}
\right].
\]
Since \(x\mapsto\sqrt{x}\) is concave, Jensen's inequality gives
\[
\mathbb{E}\!\left[
\sqrt{I_t(\pi_t^{\mathrm{IDS}})}
\right]
\le
\sqrt{
\mathbb{E}\!\left[
I_t(\pi_t^{\mathrm{IDS}})
\right]
}.
\]
Hence,
\[
\mathbb{E}\!\left[\sum_{t=1}^T\ell_t\right]
\le
\sqrt{\Psi}
\sum_{t=1}^T
\sqrt{
\mathbb{E}\!\left[
I_t(\pi_t^{\mathrm{IDS}})
\right]
}.
\]
Applying Cauchy--Schwarz,
\begin{equation}
\mathbb{E}\!\left[\sum_{t=1}^T\ell_t\right]
\le
\sqrt{
\Psi T
\sum_{t=1}^T
\mathbb{E}\!\left[
I_t(\pi_t^{\mathrm{IDS}})
\right]
}.
\label{eq:ids-before-budget}
\end{equation}

We now control the cumulative expected information gain about
\(\mathcal{S}^*\). Since \(\mathcal{H}_1\) is empty and
\[
\mathcal{H}_{t+1}
=
(\mathcal{H}_t,A_t,\mathbf{Y}_{t,A_t}),
\]
the chain rule for mutual information gives
\[
I(\mathcal{S}^*;\mathcal{H}_{T+1})
=
\sum_{t=1}^T
I\!\left(
\mathcal{S}^*;
A_t,\mathbf{Y}_{t,A_t}
\mid\mathcal{H}_t
\right).
\]
Because \(\pi_t^{\mathrm{IDS}}\) is
\(\mathcal{H}_t\)-measurable and \(A_t\) is sampled from it using fresh
algorithmic randomness independent of \(\mathcal{S}^*\) conditional on
\(\mathcal{H}_t\),
\[
I(\mathcal{S}^*;A_t\mid\mathcal{H}_t)=0.
\]
Therefore,
\[
I\!\left(
\mathcal{S}^*;
A_t,\mathbf{Y}_{t,A_t}
\mid\mathcal{H}_t
\right)
=
I\!\left(
\mathcal{S}^*;
\mathbf{Y}_{t,A_t}
\mid\mathcal{H}_t,A_t
\right).
\]

By the definition of the history-dependent action information quantities
and the law of total expectation,
\[
\begin{aligned}
I\!\left(
\mathcal{S}^*;\mathbf{Y}_{t,A_t}
\mid\mathcal{H}_t,A_t
\right)
&=
\mathbb{E}\!\left[
\sum_{a\in\mathcal{A}}
\mathbb{P}(A_t=a\mid\mathcal{H}_t)\,
I_t(a;\mathcal{H}_t)
\right]\\
&=
\mathbb{E}\!\left[
I_t(\pi_t^{\mathrm{IDS}})
\right].
\end{aligned}
\]
Consequently,
\begin{equation}
\sum_{t=1}^T
\mathbb{E}\!\left[
I_t(\pi_t^{\mathrm{IDS}})
\right]
=
I(\mathcal{S}^*;\mathcal{H}_{T+1})
\le
H(\mathcal{S}^*),
\label{eq:ids-info-budget}
\end{equation}
where the inequality follows because \(\mathcal{S}^*\) is finite-valued and
mutual information is upper bounded by entropy.

Substituting \eqref{eq:ids-info-budget} into
\eqref{eq:ids-before-budget}, we obtain the generic bound
\begin{equation}
\mathbb{E}\!\left[
\sum_{t=1}^T
\mathbbm{1}\{A_t\notin\mathcal{S}^*\}
\right]
\le
\sqrt{
\Psi\,H(\mathcal{S}^*)\,T
}.
\label{eq:ids-generic-bound}
\end{equation}

We now specialize this bound under Assumptions~\ref{ass:uniform_sep}
and~\ref{ass:decay_sep}.

\medskip
\noindent
\textbf{Uniform separation.}
Fix \(t\) and a realizable history \(\mathcal{H}_t\), and write
\[
G_t
:=
\mathbb{P}(S\neq S'\mid\mathcal{H}_t),
\qquad
S,S'\stackrel{\mathrm{i.i.d.}}{\sim}p_t.
\]

If \(G_t=0\), then
\[
G_t
=
1-\sum_s p_t(s)^2
=
0,
\]
so the posterior over \(\mathcal{S}^*\) is concentrated on a single Pareto
set, say \(s_0\), with \(p_t(s_0)=1\). Since \(s_0\) is a realizable Pareto
set, it is nonempty. Choose any \(a_0\in s_0\). Then
\[
\Delta_t(a_0)
=
\mathbb{P}(a_0\notin\mathcal{S}^*\mid\mathcal{H}_t)
=
0.
\]
Moreover, \(\mathcal{S}^*=s_0\) almost surely conditional on
\(\mathcal{H}_t\), and hence
\[
I_t(a_0)
=
I(\mathcal{S}^*;\mathbf{Y}_{t,a_0}
\mid\mathcal{H}_t,A_t=a_0)
=
0.
\]
Thus, under the extended-value convention,
\[
\psi_t(\delta_{a_0})=0.
\]
Since information ratios are nonnegative and
\(\pi_t^{\mathrm{IDS}}\) minimizes the information ratio,
\[
0
\le
\psi_t(\pi_t^{\mathrm{IDS}})
\le
\psi_t(\delta_{a_0})
=
0.
\]
Therefore,
\[
\psi_t(\pi_t^{\mathrm{IDS}})
=
0
\le
\frac{8}{\kappa}.
\]
It remains only to consider the case \(G_t>0\).

For each \(a\in\mathcal{A}\), define
\[
g_t(a)
:=
\mathbb{E}\!\left[
\operatorname{TV}
\!\left(
P_{t,S}^a,P_{t,S'}^a
\right)^2
\;\middle|\;
S\neq S',\mathcal{H}_t
\right].
\]
By Assumption~\ref{ass:uniform_sep},
\[
\mathbb{E}_{a\sim\mathrm{Unif}(\mathcal{A})}
[g_t(a)]
\ge
\kappa.
\]
Choose, using a fixed deterministic tie-breaking rule,
\[
a_t^{\mathrm{inf}}
\in
\arg\max_{a\in\mathcal{A}} g_t(a).
\]
Then
\[
g_t(a_t^{\mathrm{inf}})
\ge
\kappa.
\]

By Lemma~\ref{lem:info_gain_tv},
\[
I_t(a)
\ge
\frac12
\mathbb{E}_{S,S'\stackrel{\mathrm{i.i.d.}}{\sim}p_t}
\!\left[
\operatorname{TV}
\!\left(
P_{t,S}^a,P_{t,S'}^a
\right)^2
\right].
\]
Since the total variation term vanishes when \(S=S'\), and \(G_t>0\),
\[
I_t(a)
\ge
\frac12G_t
\mathbb{E}\!\left[
\operatorname{TV}
\!\left(
P_{t,S}^a,P_{t,S'}^a
\right)^2
\;\middle|\;
S\neq S',\mathcal{H}_t
\right]
=
\frac12G_tg_t(a).
\]
Hence
\begin{equation}
I_t(a_t^{\mathrm{inf}})
\ge
\frac{\kappa}{2}G_t.
\label{eq:ids-informative-uniform}
\end{equation}

Let
\[
s_{\max}\in\arg\max_s p_t(s),
\qquad
p_{\max}:=p_t(s_{\max}),
\]
using a fixed deterministic tie-breaking rule if the maximizer is not unique.
Since \(p_{\max}>0\), \(s_{\max}\) is a posterior-supported Pareto set and
hence is nonempty. Choose \(a_t^{\mathrm{exp}}\in s_{\max}\) using a fixed
deterministic rule. Since \(a_t^{\mathrm{exp}}\) is Pareto-optimal whenever
\(\mathcal{S}^*=s_{\max}\),
\[
\mathbb{P}
\!\left(
a_t^{\mathrm{exp}}\in\mathcal{S}^*
\mid\mathcal{H}_t
\right)
\ge
p_{\max}.
\]
Therefore,
\[
\Delta_t(a_t^{\mathrm{exp}})
\le
1-p_{\max}.
\]
Moreover, since \(p_t(s)\le p_{\max}\) for every \(s\),
\[
\sum_s p_t(s)^2
\le
p_{\max}\sum_s p_t(s)
=
p_{\max}.
\]
Thus
\[
G_t
=
1-\sum_s p_t(s)^2
\ge
1-p_{\max},
\]
and hence
\begin{equation}
\Delta_t(a_t^{\mathrm{exp}})
\le
G_t.
\label{eq:ids-exploit-loss}
\end{equation}

Since \(G_t\) is a probability and we are in the case \(G_t>0\),
\[
0<G_t\le1.
\]
Therefore the comparison distribution
\[
\pi_t^{\mathrm{mix}}
:=
(1-G_t)\delta_{a_t^{\mathrm{exp}}}
+
G_t\delta_{a_t^{\mathrm{inf}}}
\]
is a valid probability distribution on \(\mathcal{A}\), where
\(\delta_a\) denotes the point mass on action \(a\).
Since every Pareto-violation loss is at most one,
\[
\Delta_t(a_t^{\mathrm{inf}})
\le1.
\]

Using \eqref{eq:ids-exploit-loss},
\[
\begin{aligned}
\Delta_t(\pi_t^{\mathrm{mix}})
&=
(1-G_t)\Delta_t(a_t^{\mathrm{exp}})
+
G_t\Delta_t(a_t^{\mathrm{inf}})\\
&\le
(1-G_t)G_t+G_t\\
&=
2G_t-G_t^2\\
&\le
2G_t.
\end{aligned}
\]
Its information gain satisfies, using
\eqref{eq:ids-informative-uniform},
\[
\begin{aligned}
I_t(\pi_t^{\mathrm{mix}})
&=
(1-G_t)I_t(a_t^{\mathrm{exp}})
+
G_tI_t(a_t^{\mathrm{inf}})\\
&\ge
G_tI_t(a_t^{\mathrm{inf}})\\
&\ge
\frac{\kappa}{2}G_t^2.
\end{aligned}
\]
Since \(G_t>0\) and \(\kappa>0\), the denominator above is strictly
positive. Therefore,
\[
\psi_t(\pi_t^{\mathrm{mix}})
\le
\frac{(2G_t)^2}
{(\kappa/2)G_t^2}
=
\frac{8}{\kappa}.
\]
Because \(\pi_t^{\mathrm{IDS}}\) minimizes the information ratio,
\[
\psi_t(\pi_t^{\mathrm{IDS}})
\le
\psi_t(\pi_t^{\mathrm{mix}})
\le
\frac{8}{\kappa}.
\]
Together with the already treated \(G_t=0\) case, this holds at every
realizable history. Taking
\[
\Psi=\frac{8}{\kappa}
\]
in \eqref{eq:ids-generic-bound} yields
\[
\mathbb{E}\!\left[
\sum_{t=1}^T
\mathbbm{1}\{A_t\notin\mathcal{S}^*\}
\right]
\le
\sqrt{
\frac{8}{\kappa}
H(\mathcal{S}^*)T
}.
\]

\medskip
\noindent
\textbf{Decaying separation.}
Fix \(t\) and a realizable history \(\mathcal{H}_t\). If \(G_t=0\), the
same argument as above yields
\[
\psi_t(\pi_t^{\mathrm{IDS}})
=
0
\le
\frac{8}{c}t^\beta.
\]
Suppose therefore that \(G_t>0\), and define
\[
g_t(a)
:=
\mathbb{E}\!\left[
\operatorname{TV}
\!\left(
P_{t,S}^a,P_{t,S'}^a
\right)^2
\;\middle|\;
S\neq S',\mathcal{H}_t
\right].
\]
By Assumption~\ref{ass:decay_sep},
\[
\mathbb{E}_{a\sim\mathrm{Unif}(\mathcal{A})}
[g_t(a)]
\ge
c\,t^{-\beta}.
\]
Choose, using a fixed deterministic tie-breaking rule,
\[
a_t^{\mathrm{inf}}
\in
\arg\max_{a\in\mathcal{A}} g_t(a).
\]
Then
\[
g_t(a_t^{\mathrm{inf}})
\ge
c\,t^{-\beta}.
\]
By Lemma~\ref{lem:info_gain_tv} and the same conditioning argument,
\[
I_t(a_t^{\mathrm{inf}})
\ge
\frac{c}{2}
t^{-\beta}G_t.
\]

Let
\[
s_{\max}\in\arg\max_s p_t(s),
\]
using the same fixed deterministic tie-breaking rule. Since
\(p_t(s_{\max})>0\), \(s_{\max}\) is a posterior-supported Pareto set and
hence is nonempty. Choose \(a_t^{\mathrm{exp}}\in s_{\max}\) using the same
fixed deterministic rule. As proved above,
\[
\Delta_t(a_t^{\mathrm{exp}})
\le
G_t.
\]

Since \(0<G_t\le1\), define the valid comparison distribution
\[
\pi_t^{\mathrm{mix}}
:=
(1-G_t)\delta_{a_t^{\mathrm{exp}}}
+
G_t\delta_{a_t^{\mathrm{inf}}}.
\]
Then
\[
\Delta_t(\pi_t^{\mathrm{mix}})
\le
(1-G_t)G_t+G_t
\le
2G_t,
\]
while
\[
I_t(\pi_t^{\mathrm{mix}})
\ge
G_t I_t(a_t^{\mathrm{inf}})
\ge
\frac{c}{2}
t^{-\beta}G_t^2.
\]
Since \(G_t>0\) and \(c>0\), the denominator is strictly positive.
Therefore,
\[
\psi_t(\pi_t^{\mathrm{mix}})
\le
\frac{(2G_t)^2}
{(c/2)t^{-\beta}G_t^2}
=
\frac{8}{c}t^\beta.
\]
Since IDS minimizes the information ratio,
\[
\psi_t(\pi_t^{\mathrm{IDS}})
\le
\psi_t(\pi_t^{\mathrm{mix}})
\le
\frac{8}{c}t^\beta
\qquad
\text{for every }t\in[T].
\]
Consequently, since \(\beta\ge0\) and \(t\le T\),
\[
\psi_t(\pi_t^{\mathrm{IDS}})
\le
\frac{8}{c}T^\beta
\qquad
\text{for every }t\in[T].
\]
Taking
\[
\Psi=\frac{8}{c}T^\beta
\]
in \eqref{eq:ids-generic-bound} gives
\[
\mathbb{E}\!\left[
\sum_{t=1}^T
\mathbbm{1}\{A_t\notin\mathcal{S}^*\}
\right]
\le
\sqrt{
\frac{8}{c}
H(\mathcal{S}^*)
T^{\beta+1}
}.
\]

Finally, if \(\beta<1\), then
\[
\frac{\beta+1}{2}<1,
\]
so the bound is \(o(T)\), implying sublinear Pareto-violation regret.

This completes the proof.
\end{proof}

\subsection{Proof of Proposition \ref{prop:lb}}
\proplb*

\begin{proof}
We construct a two-armed, two-objective Bayesian bandit for which the average
squared-TV separation is exactly $\kappa$ at every finite realizable history,
and then prove that every possibly randomized and adaptive algorithm incurs
Bayesian cumulative Pareto-violation regret at least
$\frac{1}{32}\min\{T,1/\kappa\}$.

Fix any $\kappa\in(0,1/8]$ and set
\[
\delta:=\sqrt{\frac{\kappa}{2}}.
\]
Then $\delta\in(0,1/4]$ and $\kappa=2\delta^2$. Let
\[
\mathcal{A}=\{a_0,a_1\},
\qquad
\Theta=\{\theta_0,\theta_1\},
\]
with the uniform prior
\[
\mathbb{P}(\boldsymbol{\theta}=\theta_0)
=
\mathbb{P}(\boldsymbol{\theta}=\theta_1)
=
\frac12.
\]

\paragraph{Step 1: Construction and Pareto sets.}
Under both environments, action $a_0$ yields the deterministic reward
\[
\mathbf{Y}_{t,a_0}\equiv(1/2,1/2).
\]
Under $\theta_0$, action $a_1$ yields
\[
\mathbf{Y}_{t,a_1}=(Z_t,Z_t),
\qquad
Z_t\sim\mathrm{Bernoulli}(1/2-\delta),
\]
whereas under $\theta_1$ it yields
\[
\mathbf{Y}_{t,a_1}=(Z_t,Z_t),
\qquad
Z_t\sim\mathrm{Bernoulli}(1/2+\delta).
\]
Observations are independent across rounds conditional on the environment.

Under $\theta_0$,
\[
\boldsymbol{\mu}_{\theta_0}(a_0)=(1/2,1/2),
\qquad
\boldsymbol{\mu}_{\theta_0}(a_1)=(1/2-\delta,1/2-\delta),
\]
so $a_0$ strictly dominates $a_1$ and
\[
\mathcal{S}^*(\theta_0)=\{a_0\}.
\]
Similarly, under $\theta_1$,
\[
\boldsymbol{\mu}_{\theta_1}(a_0)=(1/2,1/2),
\qquad
\boldsymbol{\mu}_{\theta_1}(a_1)=(1/2+\delta,1/2+\delta),
\]
so $a_1$ strictly dominates $a_0$ and
\[
\mathcal{S}^*(\theta_1)=\{a_1\}.
\]
Write
\[
s_0:=\{a_0\},
\qquad
s_1:=\{a_1\}.
\]
Thus $\mathcal{S}^*=s_i$ if and only if
$\boldsymbol{\theta}=\theta_i$.

Because
\[
0<1/2-\delta<1/2+\delta<1,
\]
each outcome in $\{(0,0),(1,1)\}$ from action $a_1$ has strictly positive
probability under both environments, while $a_0$ has the same deterministic
law under both. Since the algorithm uses the same policy kernel at any common
history, it follows inductively that every finite history realizable under one
environment is also realizable under the other. Hence both environments retain
strictly positive posterior probability after every finite realizable history.
In particular,
\[
G_t=\mathbb{P}(S\neq S'\mid\mathcal{H}_t)>0
\]
at every finite realizable history.

\paragraph{Step 2: The separation is exactly $\kappa$.}
Since each Pareto-set candidate uniquely identifies the environment and
future rewards are conditionally independent of the past given the environment,
for every finite realizable history $\mathcal{H}_t$,
\[
P^{a_0}_{t,s_0}=P^{a_0}_{t,s_1},
\]
whereas
\[
P^{a_1}_{t,s_0}
=
\mathcal{L}((Z,Z)),
\quad
Z\sim\mathrm{Bernoulli}(1/2-\delta),
\]
and
\[
P^{a_1}_{t,s_1}
=
\mathcal{L}((Z,Z)),
\quad
Z\sim\mathrm{Bernoulli}(1/2+\delta).
\]

Fix such a history and let
$S,S'\stackrel{\mathrm{i.i.d.}}{\sim}p_t$. Conditional on
$\{S\neq S'\}$, the ordered pair $(S,S')$ is either $(s_0,s_1)$ or
$(s_1,s_0)$. Since total variation is symmetric,
\[
\begin{aligned}
\kappa_t(\mathcal{H}_t)
&=
\mathbb{E}\!\left[
\mathbb{E}_{a\sim\mathrm{Unif}(\mathcal{A})}
\operatorname{TV}\!\left(P^a_{t,S},P^a_{t,S'}\right)^2
\,\middle|\,
S\neq S',\mathcal{H}_t
\right]\\
&=
\frac12
\operatorname{TV}\!\left(
P^{a_0}_{t,s_0},P^{a_0}_{t,s_1}
\right)^2
+
\frac12
\operatorname{TV}\!\left(
P^{a_1}_{t,s_0},P^{a_1}_{t,s_1}
\right)^2.
\end{aligned}
\]
The first term is zero. For $a_1$, the reward takes values only in
$\{(0,0),(1,1)\}$, and therefore
\[
\begin{aligned}
\operatorname{TV}\!\left(
P^{a_1}_{t,s_0},P^{a_1}_{t,s_1}
\right)
&=
\frac12\Bigl(
|(1/2-\delta)-(1/2+\delta)|
+
|(1/2+\delta)-(1/2-\delta)|
\Bigr)\\
&=
2\delta.
\end{aligned}
\]
Consequently,
\[
\kappa_t(\mathcal{H}_t)
=
\frac12(2\delta)^2
=
2\delta^2
=
\kappa.
\]
Thus the first claim of the proposition holds at every finite realizable
history.

\paragraph{Step 3: Express the Bayesian regret through the number of pulls of $a_1$.}
Fix an arbitrary horizon $T\ge1$ and an arbitrary possibly randomized and
adaptive algorithm. For $i\in\{0,1\}$, let $\mathbb{P}_i$ and
$\mathbb{E}_i$ denote probability and expectation under environment
$\theta_i$, and define
\[
N_1(T):=\sum_{t=1}^T\mathbbm{1}\{A_t=a_1\}.
\]
Under $\theta_0$, only $a_0$ is Pareto optimal, so the cumulative violation
loss is $N_1(T)$. Under $\theta_1$, only $a_1$ is Pareto optimal, so the
cumulative violation loss is $T-N_1(T)$. Hence, under the uniform prior,
\begin{equation}
\begin{aligned}
R_T
&:=
\mathbb{E}\!\left[
\sum_{t=1}^T
\mathbbm{1}\{A_t\notin\mathcal{S}^*(\boldsymbol{\theta})\}
\right]\\
&=
\frac12\mathbb{E}_0[N_1(T)]
+
\frac12\bigl(T-\mathbb{E}_1[N_1(T)]\bigr)\\
&=
\frac12\Bigl(
T-
\bigl(\mathbb{E}_1[N_1(T)]
-\mathbb{E}_0[N_1(T)]\bigr)
\Bigr).
\end{aligned}
\label{Rtbar}
\end{equation}

\paragraph{Step 4: Relate the difference in pull counts to trajectory TV.}
Let $\mathbb{P}_i^{\mathrm{traj}}$ be the law under $\theta_i$ of
\[
\mathcal{T}_{T+1}
:=
(A_1,\mathbf{Y}_{1,A_1},\ldots,A_T,\mathbf{Y}_{T,A_T}).
\]
Since
\[
f(\mathcal{T}_{T+1})
:=
\frac{N_1(T)}{T}
\]
takes values in $[0,1]$, the variational characterization of total
variation gives
\[
\left|
\mathbb{E}_1[f]-\mathbb{E}_0[f]
\right|
\le
\operatorname{TV}\!\left(
\mathbb{P}_0^{\mathrm{traj}},
\mathbb{P}_1^{\mathrm{traj}}
\right).
\]
Therefore,
\begin{equation}
\left|
\mathbb{E}_1[N_1(T)]-\mathbb{E}_0[N_1(T)]
\right|
\le
T\,
\operatorname{TV}\!\left(
\mathbb{P}_0^{\mathrm{traj}},
\mathbb{P}_1^{\mathrm{traj}}
\right).
\label{TV_bound}
\end{equation}

\paragraph{Step 5: Upper bound the trajectory KL divergence.}
By the chain rule for KL divergence,
\[
\begin{aligned}
&\operatorname{KL}\!\left(
\mathbb{P}_0^{\mathrm{traj}}
\,\|\, 
\mathbb{P}_1^{\mathrm{traj}}
\right)\\
&\qquad=
\sum_{t=1}^T
\mathbb{E}_0\!\left[
\operatorname{KL}\!\left(
\mathbb{P}_0(A_t,\mathbf{Y}_{t,A_t}\mid\mathcal{H}_t)
\,\|\,
\mathbb{P}_1(A_t,\mathbf{Y}_{t,A_t}\mid\mathcal{H}_t)
\right)
\right].
\end{aligned}
\]
At every history in the common support of the two environments, the algorithm
uses the same conditional action kernel; write it as
$\pi_t(\cdot\mid\mathcal{H}_t)$. Thus
\[
\mathbb{P}_0(A_t=a\mid\mathcal{H}_t)
=
\mathbb{P}_1(A_t=a\mid\mathcal{H}_t)
=
\pi_t(a\mid\mathcal{H}_t).
\]
Consequently, the conditional chain rule for KL gives
\[
\begin{aligned}
&\operatorname{KL}\!\left(
\mathbb{P}_0(A_t,\mathbf{Y}_{t,A_t}\mid\mathcal{H}_t)
\,\|\,
\mathbb{P}_1(A_t,\mathbf{Y}_{t,A_t}\mid\mathcal{H}_t)
\right)\\
&\qquad=
\sum_{a\in\mathcal{A}}
\pi_t(a\mid\mathcal{H}_t)
\operatorname{KL}\!\left(
\mathbb{P}_0(\mathbf{Y}_{t,a}\mid\mathcal{H}_t,A_t=a)
\,\|\,
\mathbb{P}_1(\mathbf{Y}_{t,a}\mid\mathcal{H}_t,A_t=a)
\right).
\end{aligned}
\]
If $a=a_0$, the observation laws are identical, so the corresponding KL
term is zero. If $a=a_1$, conditional independence across rounds implies
that the KL term is
\[
\begin{aligned}
D_\delta
&:=
\left(\frac12-\delta\right)
\log\!\frac{1/2-\delta}{1/2+\delta}
+
\left(\frac12+\delta\right)
\log\!\frac{1/2+\delta}{1/2-\delta}\\
&=
2\delta
\log\!\left(
\frac{1+2\delta}{1-2\delta}
\right).
\end{aligned}
\]
Therefore,
\[
\operatorname{KL}\!\left(
\mathbb{P}_0^{\mathrm{traj}}
\,\|\, 
\mathbb{P}_1^{\mathrm{traj}}
\right)
=
D_\delta\,\mathbb{E}_0[N_1(T)].
\]
For $x\in[0,1)$,
\[
\log\!\left(\frac{1+x}{1-x}\right)
=
\log(1+x)-\log(1-x)
\le
x+\frac{x}{1-x}
\le
\frac{2x}{1-x}.
\]
Since $2\delta\le1/2$, taking $x=2\delta$ gives
\[
D_\delta
\le
\frac{8\delta^2}{1-2\delta}
\le
16\delta^2.
\]
Therefore,
\begin{equation}
\operatorname{KL}\!\left(
\mathbb{P}_0^{\mathrm{traj}}
\,\|\, 
\mathbb{P}_1^{\mathrm{traj}}
\right)
\le
16\delta^2\,\mathbb{E}_0[N_1(T)].
\label{kl_bound}
\end{equation}

From \eqref{Rtbar},
\[
R_T
=
\frac12\mathbb{E}_0[N_1(T)]
+
\frac12\bigl(T-\mathbb{E}_1[N_1(T)]\bigr)
\ge
\frac12\mathbb{E}_0[N_1(T)],
\]
because $N_1(T)\le T$ almost surely. Hence
\[
\mathbb{E}_0[N_1(T)]
\le
2R_T,
\]
and therefore
\begin{equation}
\operatorname{KL}\!\left(
\mathbb{P}_0^{\mathrm{traj}}
\,\|\, 
\mathbb{P}_1^{\mathrm{traj}}
\right)
\le
32\delta^2R_T.
\label{kl_up}
\end{equation}

\paragraph{Step 6: Conclude by a two-case argument.}
If $R_T\ge T/4$, then
\[
R_T
\ge
\frac{T}{4}
\ge
\frac1{32}\min\!\left\{T,\frac1\kappa\right\},
\]
so the desired bound holds.

Now suppose $R_T<T/4$. By \eqref{Rtbar},
\[
\mathbb{E}_1[N_1(T)]-\mathbb{E}_0[N_1(T)]
=
T-2R_T
>
\frac{T}{2}.
\]
Combining this with \eqref{TV_bound} gives
\[
\operatorname{TV}\!\left(
\mathbb{P}_0^{\mathrm{traj}},
\mathbb{P}_1^{\mathrm{traj}}
\right)
>
\frac12.
\]
Pinsker's inequality, with natural logarithms, implies
\[
\operatorname{KL}\!\left(
\mathbb{P}_0^{\mathrm{traj}}
\,\|\, 
\mathbb{P}_1^{\mathrm{traj}}
\right)
\ge
2\,
\operatorname{TV}\!\left(
\mathbb{P}_0^{\mathrm{traj}},
\mathbb{P}_1^{\mathrm{traj}}
\right)^2
>
\frac12.
\]
Together with \eqref{kl_up}, this yields
\[
32\delta^2R_T
>
\frac12,
\]
and hence
\[
R_T
>
\frac{1}{64\delta^2}
=
\frac{1}{32\kappa}.
\]
Since
\[
\min\!\left\{T,\frac1\kappa\right\}
\le
\frac1\kappa,
\]
we obtain
\[
R_T
>
\frac{1}{32\kappa}
\ge
\frac1{32}
\min\!\left\{T,\frac1\kappa\right\}.
\]
Thus the desired lower bound also holds in the second case.

Combining the two cases, for every $T\ge1$ and every possibly randomized
and adaptive algorithm,
\[
R_T
=
\mathbb{E}\!\left[
\sum_{t=1}^T
\mathbbm{1}\{A_t\notin\mathcal{S}^*(\boldsymbol{\theta})\}
\right]
\ge
\frac1{32}
\min\!\left\{
T,\frac1\kappa
\right\}.
\]
Together with
$\kappa_t(\mathcal{H}_t)=\kappa$ at every finite realizable history, this
proves Proposition~\ref{prop:lb}.
\end{proof}

\subsection{Refined Thompson Sampling Guarantees for Pareto-Violation Regret with Separation-Dependent Exploration Schedules}
\label{refined_ts_pv}

The regret bound in Theorem~\ref{thm:ts_regret} uses the generic,
separation-agnostic schedule $\varepsilon_t=t^{-1/2}$. We now show that, when
the separation profile is known, the same sampling rule can be equipped with
a separation-dependent exploration schedule to obtain a sharper guarantee.
Under polynomially decaying separation
$\kappa_t(\mathcal{H}_t)\gtrsim t^{-\beta}$, a schedule matched to $\beta$
yields sublinear regret for every $\beta<1$. Uniform separation is recovered
as the special case $\beta=0$. Thus, the $\beta<1/2$ condition in
Theorem~\ref{thm:ts_regret} is a consequence of the generic exploration
schedule rather than an intrinsic limitation of Thompson Sampling.

\begin{Proposition}
\label{prop_ts_app}
Suppose Assumption~\ref{ass:decay_sep} holds with constants $c>0$ and
$\beta\in[0,1)$. Consider the sampling rule of
Algorithm~\ref{alg:ts_pareto}, but replace its default exploration schedule
by
\[
\varepsilon_t
:=
\min\!\left\{
\alpha\,t^{-\frac{1-\beta}{2}},\,1
\right\},
\qquad
\alpha
:=
\sqrt{\frac{H(\mathcal{S}^*)}{c}}.
\]
Then
\[
\mathbb{E}\!\left[
\sum_{t=1}^T
\mathbbm{1}\{A_t\notin\mathcal{S}^*\}
\right]
\le
4
\sqrt{\frac{H(\mathcal{S}^*)}{c}}\,
T^{\frac{1+\beta}{2}}.
\]
Hence the regret is sublinear for every $\beta<1$.

In particular, under Assumption~\ref{ass:uniform_sep}, taking
$c=\kappa$ and $\beta=0$ gives
\[
\varepsilon_t
=
\min\!\left\{
\frac{\alpha}{\sqrt{t}},\,1
\right\},
\qquad
\alpha
=
\sqrt{\frac{H(\mathcal{S}^*)}{\kappa}},
\]
and
\[
\mathbb{E}\!\left[
\sum_{t=1}^T
\mathbbm{1}\{A_t\notin\mathcal{S}^*\}
\right]
\le
4\sqrt{\frac{T\,H(\mathcal{S}^*)}{\kappa}}.
\]
Therefore, for $\kappa$ in the lower-bound regime of
Proposition~\ref{prop:lb}, when $T\asymp1/\kappa$, the upper bound has the
same $1/\kappa$ dependence as the lower bound, up to universal constants and
the $\sqrt{H(\mathcal{S}^*)}$ factor.
\end{Proposition}

\begin{proof}
Let
\[
\ell_t:=\mathbbm{1}\{A_t\notin\mathcal{S}^*\},
\qquad
G_t:=\mathbb{P}(S\neq S'\mid\mathcal{H}_t),
\qquad
r_t:=(1-\varepsilon_t)G_t,
\]
where $S,S'\stackrel{\mathrm{i.i.d.}}{\sim}p_t$ are independent posterior
draws of $\mathcal{S}^*$.

We first treat the degenerate case $H(\mathcal{S}^*)=0$. Then
$\mathcal{S}^*$ is almost surely constant under the prior and consequently
under every posterior arising from a realizable history. Since $\alpha=0$,
we have $\varepsilon_t=0$ for every $t$. The Thompson Sampling branch
therefore samples a parameter whose Pareto set equals the true deterministic
Pareto set almost surely and selects an action from that set. Hence
$\ell_t=0$ almost surely for every $t$, and the claimed bound is immediate.
Henceforth assume $H(\mathcal{S}^*)>0$, so $\alpha>0$.

Conditional on $\mathcal{H}_t$, the uniform-exploration branch incurs
expected loss at most one. On the Thompson Sampling branch, the sampled
Pareto set and the true Pareto set are independent draws from the same
posterior distribution. If they agree, every action selected from the
sampled set is Pareto optimal under the true environment; if they disagree,
the violation loss is at most one. Hence the TS-branch violation probability
is at most $G_t$, and
\begin{equation}
\mathbb{E}[\ell_t\mid\mathcal{H}_t]
\le
\varepsilon_t+(1-\varepsilon_t)G_t
=
\varepsilon_t+r_t.
\label{eq:tuned-ts-one-step-loss}
\end{equation}

For each $a\in\mathcal{A}$, define
\[
I_t(a)
:=
I(\mathcal{S}^*;\mathbf{Y}_{t,a}
\mid\mathcal{H}_t,A_t=a),
\]
using the fixed-action interpretation, and let
\[
I_t
:=
\sum_{a\in\mathcal{A}}
\mathbb{P}(A_t=a\mid\mathcal{H}_t)\,I_t(a)
\]
denote the history-wise one-step information gain under the actual sampling
distribution. Because the exploration branch chooses uniformly from
$\mathcal{A}$,
\[
\mathbb{P}(A_t=a\mid\mathcal{H}_t)
\ge
\frac{\varepsilon_t}{|\mathcal{A}|}
\qquad
\text{for every }a\in\mathcal{A}.
\]
Therefore,
\[
I_t
\ge
\varepsilon_t\,
\mathbb{E}_{a\sim\mathrm{Unif}(\mathcal{A})}[I_t(a)].
\]
By Lemma~\ref{lem:info_gain_tv} and the definition of
$\kappa_t(\mathcal{H}_t)$,
\[
\mathbb{E}_{a\sim\mathrm{Unif}(\mathcal{A})}[I_t(a)]
\ge
\frac12G_t\kappa_t(\mathcal{H}_t).
\]
Hence, under Assumption~\ref{ass:decay_sep}, whenever $G_t>0$,
\begin{equation}
I_t
\ge
\frac{\varepsilon_t c}{2}t^{-\beta}G_t.
\label{eq:tuned-ts-info-lb}
\end{equation}
If $G_t=0$, then $r_t=0$. If $G_t>0$ and $\varepsilon_t<1$, then
$r_t=(1-\varepsilon_t)G_t\le G_t$, and
\eqref{eq:tuned-ts-info-lb} gives
\begin{equation}
r_t
\le
\frac{2t^\beta}{c\,\varepsilon_t}I_t.
\label{eq:tuned-ts-r-info}
\end{equation}
When $\varepsilon_t=1$, we instead have $r_t=0$ directly.

We will also use the total information budget
\begin{equation}
\mathbb{E}\!\left[\sum_{t=1}^T I_t\right]
=
I(\mathcal{S}^*;\mathcal{H}_{T+1})
\le
H(\mathcal{S}^*).
\label{eq:tuned-ts-info-budget}
\end{equation}
Indeed, the mutual-information chain rule gives
\[
I(\mathcal{S}^*;\mathcal{H}_{T+1})
=
\sum_{t=1}^T
I(\mathcal{S}^*;A_t,\mathbf{Y}_{t,A_t}\mid\mathcal{H}_t).
\]
Conditional on $\mathcal{H}_t$, all posterior samples and branch/action
randomization used by the algorithm are generated using fresh internal
randomness independent of the true parameter and therefore of
$\mathcal{S}^*$. Hence
\[
I(\mathcal{S}^*;A_t\mid\mathcal{H}_t)=0.
\]
The conditional chain rule and the definition of $I_t$ then imply
\[
I(\mathcal{S}^*;A_t,\mathbf{Y}_{t,A_t}\mid\mathcal{H}_t)
=
\mathbb{E}[I_t].
\]
Summing over $t$ yields \eqref{eq:tuned-ts-info-budget}.

Taking expectations in \eqref{eq:tuned-ts-one-step-loss} and summing gives
\begin{equation}
\mathbb{E}\!\left[\sum_{t=1}^T\ell_t\right]
\le
\sum_{t=1}^T\varepsilon_t
+
\mathbb{E}\!\left[\sum_{t=1}^T r_t\right].
\label{eq:tuned-ts-decomp}
\end{equation}

Set
\[
\gamma:=\frac{1-\beta}{2}.
\]
Since $\beta\in[0,1)$, we have $\gamma\in(0,1/2]$ and
\begin{equation}
1-\gamma
=
\beta+\gamma
=
\frac{1+\beta}{2}.
\label{eq:tuned-ts-exponents}
\end{equation}
The schedule is
\[
\varepsilon_t=\min\{\alpha t^{-\gamma},1\}.
\]

We first bound the exploration term. Since
$\varepsilon_t\le\alpha t^{-\gamma}$,
\[
\sum_{t=1}^T\varepsilon_t
\le
\alpha\sum_{t=1}^Tt^{-\gamma}.
\]
For $\gamma\in(0,1)$,
\[
\sum_{t=1}^Tt^{-\gamma}
\le
1+\int_1^T x^{-\gamma}\,dx
=
1+\frac{T^{1-\gamma}-1}{1-\gamma}
\le
\frac{T^{1-\gamma}}{1-\gamma}.
\]
Since
\[
\frac{1}{1-\gamma}
=
\frac{2}{1+\beta}
\le2,
\]
we obtain
\begin{equation}
\sum_{t=1}^T\varepsilon_t
\le
2\alpha T^{\frac{1+\beta}{2}}
=
2\sqrt{\frac{H(\mathcal{S}^*)}{c}}\,
T^{\frac{1+\beta}{2}}.
\label{eq:tuned-ts-exploration}
\end{equation}

For the second term in \eqref{eq:tuned-ts-decomp}, define
\[
\mathcal{U}
:=
\{t\in[T]:\varepsilon_t<1\}.
\]
If $t\notin\mathcal{U}$, then $\varepsilon_t=1$ and hence $r_t=0$.
Therefore,
\[
\sum_{t=1}^T r_t
=
\sum_{t\in\mathcal{U}}r_t.
\]
For every $t\in\mathcal{U}$,
$\varepsilon_t=\alpha t^{-\gamma}$, so
\eqref{eq:tuned-ts-r-info} gives
\[
r_t
\le
\frac{2}{\alpha c}t^{\beta+\gamma}I_t.
\]
Using \eqref{eq:tuned-ts-exponents} and $t\le T$,
\[
r_t
\le
\frac{2}{\alpha c}
T^{\frac{1+\beta}{2}}I_t,
\qquad
t\in\mathcal{U}.
\]
Therefore, by \eqref{eq:tuned-ts-info-budget},
\begin{equation}
\begin{aligned}
\mathbb{E}\!\left[\sum_{t=1}^T r_t\right]
&=
\mathbb{E}\!\left[\sum_{t\in\mathcal{U}}r_t\right]\\
&\le
\frac{2}{\alpha c}
T^{\frac{1+\beta}{2}}
\mathbb{E}\!\left[\sum_{t\in\mathcal{U}}I_t\right]\\
&\le
\frac{2H(\mathcal{S}^*)}{\alpha c}
T^{\frac{1+\beta}{2}}\\
&=
2\sqrt{\frac{H(\mathcal{S}^*)}{c}}\,
T^{\frac{1+\beta}{2}}.
\end{aligned}
\label{eq:tuned-ts-exploitation}
\end{equation}

Combining \eqref{eq:tuned-ts-decomp},
\eqref{eq:tuned-ts-exploration}, and
\eqref{eq:tuned-ts-exploitation} yields
\[
\mathbb{E}\!\left[
\sum_{t=1}^T
\mathbbm{1}\{A_t\notin\mathcal{S}^*\}
\right]
\le
4
\sqrt{\frac{H(\mathcal{S}^*)}{c}}\,
T^{\frac{1+\beta}{2}}.
\]
Since $\beta<1$, the exponent $(1+\beta)/2$ is strictly smaller than one,
so the regret is sublinear.

Finally, under Assumption~\ref{ass:uniform_sep},
Assumption~\ref{ass:decay_sep} holds with $c=\kappa$ and $\beta=0$. Thus
\[
\varepsilon_t
=
\min\!\left\{
\frac{\alpha}{\sqrt{t}},1
\right\},
\qquad
\alpha
=
\sqrt{\frac{H(\mathcal{S}^*)}{\kappa}},
\]
and
\[
\mathbb{E}\!\left[
\sum_{t=1}^T
\mathbbm{1}\{A_t\notin\mathcal{S}^*\}
\right]
\le
4\sqrt{\frac{T\,H(\mathcal{S}^*)}{\kappa}}.
\]
If $T\asymp1/\kappa$, then
\[
\sqrt{\frac{T\,H(\mathcal{S}^*)}{\kappa}}
\asymp
\frac{\sqrt{H(\mathcal{S}^*)}}{\kappa}.
\]
Hence the tuned Thompson Sampling upper bound has the same $1/\kappa$
dependence as the lower bound in Proposition~\ref{prop:lb}, up to universal
constants and the $\sqrt{H(\mathcal{S}^*)}$ factor.
\end{proof}

\subsection{Equivalent Form of the Pareto Gap for Gap-Based Regret}

\begin{Proposition}[Equivalent form of the Pareto gap]
\label{lem:law-equiv}
Let $\mathcal A$ be finite. For any action $a\in\mathcal A$, define
\[
\Delta_{\boldsymbol{\theta}}(a)
:=
\sup\left\{
\varepsilon\ge0:
\exists b\in\mathcal A
\text{ such that }
\boldsymbol{\mu}_{\boldsymbol{\theta}}(b)
\succeq
\boldsymbol{\mu}_{\boldsymbol{\theta}}(a)
+\varepsilon\mathbf 1
\right\}.
\]
Then
\[
\Delta_{\boldsymbol{\theta}}(a)
=
\left(
\sup_{b\in\mathcal A}
\min_{j\in[d]}
\bigl(
\mu_{\boldsymbol{\theta},j}(b)
-
\mu_{\boldsymbol{\theta},j}(a)
\bigr)
\right)_+.
\]
\end{Proposition}

\begin{proof}
Fix $a\in\mathcal A$. For each $b\in\mathcal A$, define
\[
m_b
:=
\min_{j\in[d]}
\bigl(
\mu_{\boldsymbol{\theta},j}(b)
-
\mu_{\boldsymbol{\theta},j}(a)
\bigr).
\]
For any $b\in\mathcal A$ and $\varepsilon\ge0$,
\[
\boldsymbol{\mu}_{\boldsymbol{\theta}}(b)
\succeq
\boldsymbol{\mu}_{\boldsymbol{\theta}}(a)
+\varepsilon\mathbf 1
\]
if and only if
\[
\mu_{\boldsymbol{\theta},j}(b)
-
\mu_{\boldsymbol{\theta},j}(a)
\ge
\varepsilon,
\qquad
\forall j\in[d],
\]
which is equivalent to
\[
\varepsilon
\le
\min_{j\in[d]}
\bigl(
\mu_{\boldsymbol{\theta},j}(b)
-
\mu_{\boldsymbol{\theta},j}(a)
\bigr)
=
m_b.
\]

Let
\[
S_a
:=
\left\{
\varepsilon\ge0:
\exists b\in\mathcal A
\text{ such that }
\boldsymbol{\mu}_{\boldsymbol{\theta}}(b)
\succeq
\boldsymbol{\mu}_{\boldsymbol{\theta}}(a)
+\varepsilon\mathbf 1
\right\}.
\]
The preceding equivalence yields
\[
S_a
=
\left\{
\varepsilon\ge0:
\exists b\in\mathcal A
\text{ such that }
\varepsilon\le m_b
\right\}.
\]

Since $\mathcal A$ is finite,
\[
M
:=
\max_{b\in\mathcal A}m_b
\]
is attained by some $b^*\in\mathcal A$. Moreover, taking $b=a$ gives
\[
m_a
=
\min_{j\in[d]}
\bigl(
\mu_{\boldsymbol{\theta},j}(a)
-
\mu_{\boldsymbol{\theta},j}(a)
\bigr)
=
0,
\]
and therefore $M\ge0$.

For every $\varepsilon\in[0,M]$, choosing $b=b^*$ gives
\[
\varepsilon\le m_{b^*}=M,
\]
so $\varepsilon\in S_a$. Conversely, if $\varepsilon\in S_a$, then there
exists $b\in\mathcal A$ such that
\[
\varepsilon\le m_b\le M.
\]
Hence
\[
S_a=[0,M].
\]
It follows that
\[
\begin{aligned}
\Delta_{\boldsymbol{\theta}}(a)
&=
\sup S_a\\
&=
M\\
&=
\max_{b\in\mathcal A}
\min_{j\in[d]}
\bigl(
\mu_{\boldsymbol{\theta},j}(b)
-
\mu_{\boldsymbol{\theta},j}(a)
\bigr).
\end{aligned}
\]
Because $\mathcal A$ is finite, the maximum equals the corresponding
supremum. Moreover, this quantity is nonnegative because the choice $b=a$
gives value zero. It therefore equals its positive part, and hence
\[
\Delta_{\boldsymbol{\theta}}(a)
=
\left(
\sup_{b\in\mathcal A}
\min_{j\in[d]}
\bigl(
\mu_{\boldsymbol{\theta},j}(b)
-
\mu_{\boldsymbol{\theta},j}(a)
\bigr)
\right)_+.
\]
\end{proof}

\subsection{Proof of Lemma~\ref{lem:info_gain_tv_theta}}

\lemInfoGainTVtheta*

\begin{proof}
Fix a round \(t\), a realizable history \(\mathcal H_t\), and an action
\(a\in\mathcal A\). Fix versions of the relevant regular conditional
distributions throughout. Quantities conditioned on \(A_t=a\) are
understood under the fixed-action experiment in which action \(a\) is
selected at round \(t\). Whenever
\(\mathbb P(A_t=a\mid\mathcal H_t)>0\), this coincides with ordinary
conditioning on the event \(A_t=a\).

Let \(p_t\) denote the posterior law of \(\boldsymbol{\theta}\) given
\(\mathcal H_t\). Since \(A_t\) is selected using only
\(\mathcal H_t\) and fresh algorithmic randomness independent of the
true parameter \(\boldsymbol{\theta}\),
\[
A_t\perp\!\!\!\perp\boldsymbol{\theta}\mid\mathcal H_t.
\]
Hence fixing \(A_t=a\) does not alter the posterior law of
\(\boldsymbol{\theta}\), and under the fixed-action experiment,
\[
\boldsymbol{\theta}\sim p_t.
\]

For each \(\boldsymbol{\vartheta}\in\Theta\), define the
action-specific reward law
\[
P_{\boldsymbol{\vartheta}}^a
:=
\mathcal L\!\left(
\mathbf Y_{t,a}
\mid
\boldsymbol{\theta}=\boldsymbol{\vartheta}
\right).
\]
Because the reward arrays are i.i.d. across rounds conditional on
\(\boldsymbol{\theta}\), the current reward
\(\mathbf Y_{t,a}\) is conditionally independent of the past given
\(\boldsymbol{\theta}\). Moreover, the fresh action-selection
randomization is independent of the current reward conditional on
\((\mathcal H_t,\boldsymbol{\theta})\). Therefore,
\[
\mathcal L\!\left(
\mathbf Y_{t,a}
\mid
\mathcal H_t,A_t=a,
\boldsymbol{\theta}=\boldsymbol{\vartheta}
\right)
=
P_{\boldsymbol{\vartheta}}^a.
\]

Define the posterior-predictive law for action \(a\) by
\[
P_t^a
:=
\int_{\Theta}
P_{\boldsymbol{\vartheta}}^a\,
p_t(d\boldsymbol{\vartheta}).
\]
Since \(\boldsymbol{\theta}\sim p_t\) under the fixed-action law,
marginalizing over \(\boldsymbol{\theta}\) gives, for every measurable
set \(E\),
\[
\mathbb P\!\left(
\mathbf Y_{t,a}\in E
\mid
\mathcal H_t,A_t=a
\right)
=
\int_{\Theta}
P_{\boldsymbol{\vartheta}}^a(E)\,
p_t(d\boldsymbol{\vartheta})
=
P_t^a(E).
\]
Thus \(P_t^a\) is precisely the conditional marginal law of
\(\mathbf Y_{t,a}\) given \(\mathcal H_t\) under the fixed action
\(a\).

By the standard representation of mutual information as the expected
KL divergence between the conditional observation law and its
marginal law,
\begin{align}
I\!\left(
\boldsymbol{\theta};
\mathbf Y_{t,a}
\mid
\mathcal H_t,A_t=a
\right)
&=
\int_{\Theta}
\operatorname{KL}\!\left(
P_{\boldsymbol{\vartheta}}^a
\,\middle\|\,
P_t^a
\right)
p_t(d\boldsymbol{\vartheta})
\nonumber\\
&=
\mathbb E_{\boldsymbol{\theta}\sim p_t}
\!\left[
\operatorname{KL}\!\left(
P_{\boldsymbol{\theta}}^a
\,\middle\|\,
P_t^a
\right)
\right].
\label{eq:param-mi-kl-representation}
\end{align}
This identity is understood in the extended sense if the mutual
information is infinite.

By Pinsker's inequality, using natural logarithms,
\[
\operatorname{KL}\!\left(
P_{\boldsymbol{\theta}}^a
\,\middle\|\,
P_t^a
\right)
\ge
2\,
\operatorname{TV}\!\left(
P_{\boldsymbol{\theta}}^a,
P_t^a
\right)^2.
\]
Taking expectation with respect to
\(\boldsymbol{\theta}\sim p_t\) and using
\eqref{eq:param-mi-kl-representation} yields
\begin{equation}
I\!\left(
\boldsymbol{\theta};
\mathbf Y_{t,a}
\mid
\mathcal H_t,A_t=a
\right)
\ge
2\,
\mathbb E_{\boldsymbol{\theta}\sim p_t}
\!\left[
\operatorname{TV}\!\left(
P_{\boldsymbol{\theta}}^a,
P_t^a
\right)^2
\right].
\label{eq:param-mi-predictive-tv}
\end{equation}

Now let
\(\boldsymbol{\theta},\boldsymbol{\theta}'
\stackrel{\mathrm{i.i.d.}}{\sim}p_t\).
For every realization of
\((\boldsymbol{\theta},\boldsymbol{\theta}')\), the triangle inequality
for total variation gives
\[
\operatorname{TV}\!\left(
P_{\boldsymbol{\theta}}^a,
P_{\boldsymbol{\theta}'}^a
\right)
\le
\operatorname{TV}\!\left(
P_{\boldsymbol{\theta}}^a,
P_t^a
\right)
+
\operatorname{TV}\!\left(
P_{\boldsymbol{\theta}'}^a,
P_t^a
\right).
\]
Using \((x+y)^2\le2x^2+2y^2\), we obtain
\[
\operatorname{TV}\!\left(
P_{\boldsymbol{\theta}}^a,
P_{\boldsymbol{\theta}'}^a
\right)^2
\le
2\operatorname{TV}\!\left(
P_{\boldsymbol{\theta}}^a,
P_t^a
\right)^2
+
2\operatorname{TV}\!\left(
P_{\boldsymbol{\theta}'}^a,
P_t^a
\right)^2.
\]
Taking expectation over the independent posterior draws and using that
both have marginal law \(p_t\) gives
\[
\mathbb E_{
\boldsymbol{\theta},\boldsymbol{\theta}'
\stackrel{\mathrm{i.i.d.}}{\sim}p_t
}
\!\left[
\operatorname{TV}\!\left(
P_{\boldsymbol{\theta}}^a,
P_{\boldsymbol{\theta}'}^a
\right)^2
\right]
\le
4\,
\mathbb E_{\boldsymbol{\theta}\sim p_t}
\!\left[
\operatorname{TV}\!\left(
P_{\boldsymbol{\theta}}^a,
P_t^a
\right)^2
\right].
\]
Equivalently,
\[
\mathbb E_{\boldsymbol{\theta}\sim p_t}
\!\left[
\operatorname{TV}\!\left(
P_{\boldsymbol{\theta}}^a,
P_t^a
\right)^2
\right]
\ge
\frac14
\mathbb E_{
\boldsymbol{\theta},\boldsymbol{\theta}'
\stackrel{\mathrm{i.i.d.}}{\sim}p_t
}
\!\left[
\operatorname{TV}\!\left(
P_{\boldsymbol{\theta}}^a,
P_{\boldsymbol{\theta}'}^a
\right)^2
\right].
\]
Combining this inequality with
\eqref{eq:param-mi-predictive-tv} yields
\[
I\!\left(
\boldsymbol{\theta};
\mathbf Y_{t,a}
\mid
\mathcal H_t,A_t=a
\right)
\ge
\frac12
\mathbb E_{
\boldsymbol{\theta},\boldsymbol{\theta}'
\stackrel{\mathrm{i.i.d.}}{\sim}p_t
}
\!\left[
\operatorname{TV}\!\left(
P_{\boldsymbol{\theta}}^a,
P_{\boldsymbol{\theta}'}^a
\right)^2
\right].
\]
This proves the claim.
\end{proof}

\subsection{Proof of Theorem \ref{thm:ts_regret_gap}}

\thmTSgap*

\begin{Lemma}
\label{lem:theta_stability}
Fix a round \(t\) and condition on \(\mathcal{H}_t\). Let
\(\boldsymbol{\theta},\boldsymbol{\theta}'\stackrel{\mathrm{i.i.d.}}{\sim}p_t\),
where
\(p_t(\boldsymbol{\theta})
=\mathbb{P}(\boldsymbol{\theta}\mid\mathcal{H}_t)\).
Define
\(S:=\mathcal{S}^*(\boldsymbol{\theta})\) and
\(S':=\mathcal{S}^*(\boldsymbol{\theta}')\).
Conditional on \((\boldsymbol{\theta},\boldsymbol{\theta}')\), let
\(A\sim\mathrm{Unif}(S')\).
Suppose that every reward coordinate lies in \([0,1]\). Then
\begin{equation}
\label{eq:theta_stability_new}
\mathbb{E}\!\left[
\Delta_{\boldsymbol{\theta}}(A)
\mid
\mathcal{H}_t
\right]
\le
4
\sum_{a\in\mathcal{A}}
\mathbb{E}\!\left[
\mathbbm{1}\{a\in S'\}
\,
\mathrm{TV}\!\left(
P_{\boldsymbol{\theta}}^a,
P_{\boldsymbol{\theta}'}^a
\right)
\middle|
\mathcal{H}_t
\right].
\end{equation}
\end{Lemma}

\begin{proof}
Fix \(\mathcal{H}_t\) throughout and suppress this conditioning whenever
no ambiguity arises. Fix realizations of
\(\boldsymbol{\theta}\) and \(\boldsymbol{\theta}'\), and let
\(a\in S'=\mathcal{S}^*(\boldsymbol{\theta}')\).

We first construct a Pareto-optimal witness under
\(\boldsymbol{\theta}\). Since \(\mathcal{A}\) is finite, the Pareto gap
admits the finite maximum representation
\[
\Delta_{\boldsymbol{\theta}}(a)
=
\max_{u\in\mathcal{A}}
\left(
\min_{j\in[d]}
\left[
\mu_{\boldsymbol{\theta},j}(u)
-
\mu_{\boldsymbol{\theta},j}(a)
\right]
\right)_{+}.
\]
If \(\Delta_{\boldsymbol{\theta}}(a)>0\), choose, using a fixed
deterministic tie-breaking rule, an action \(u\) attaining this maximum.
Then
\[
\boldsymbol{\mu}_{\boldsymbol{\theta}}(u)
\succeq
\boldsymbol{\mu}_{\boldsymbol{\theta}}(a)
+
\Delta_{\boldsymbol{\theta}}(a)\mathbf{1}.
\]
If \(\Delta_{\boldsymbol{\theta}}(a)=0\), take \(u=a\), for which the
same inequality holds. Thus, in all cases,
\begin{equation}
\label{eq:witness-u-new}
\boldsymbol{\mu}_{\boldsymbol{\theta}}(u)
\succeq
\boldsymbol{\mu}_{\boldsymbol{\theta}}(a)
+
\Delta_{\boldsymbol{\theta}}(a)\mathbf{1}.
\end{equation}

Among the actions that weakly dominate \(u\) under
\(\boldsymbol{\theta}\), choose, again using a fixed deterministic
tie-breaking rule, a \(\succeq\)-maximal action \(b(a)\).
Such an action exists because \(\mathcal{A}\) is finite. If some action
strictly Pareto dominated \(b(a)\), then, since
\(b(a)\succeq u\), that action would also weakly dominate \(u\),
contradicting the maximality of \(b(a)\). Hence
\[
b(a)\in S=\mathcal{S}^*(\boldsymbol{\theta}).
\]
Moreover,
\[
\boldsymbol{\mu}_{\boldsymbol{\theta}}(b(a))
\succeq
\boldsymbol{\mu}_{\boldsymbol{\theta}}(u),
\]
and therefore \eqref{eq:witness-u-new} implies
\begin{equation}
\label{eq:witness-b-new}
\boldsymbol{\mu}_{\boldsymbol{\theta}}(b(a))
\succeq
\boldsymbol{\mu}_{\boldsymbol{\theta}}(a)
+
\Delta_{\boldsymbol{\theta}}(a)\mathbf{1}.
\end{equation}

Since
\(a\in S'=\mathcal{S}^*(\boldsymbol{\theta}')\),
the action \(b(a)\) cannot strictly Pareto dominate \(a\) under
\(\boldsymbol{\theta}'\). Consequently, there exists at least one
coordinate \(i=i(a)\in[d]\) for which
\begin{equation}
\label{eq:nondom-coordinate-new}
\mu_{\boldsymbol{\theta}',i}(b(a))
\le
\mu_{\boldsymbol{\theta}',i}(a).
\end{equation}
At the same coordinate, \eqref{eq:witness-b-new} gives
\begin{equation}
\label{eq:true-gap-coordinate-new}
\mu_{\boldsymbol{\theta},i}(b(a))
\ge
\mu_{\boldsymbol{\theta},i}(a)
+
\Delta_{\boldsymbol{\theta}}(a).
\end{equation}
Combining \eqref{eq:nondom-coordinate-new} and
\eqref{eq:true-gap-coordinate-new},
\begin{align}
\Delta_{\boldsymbol{\theta}}(a)
&\le
\mu_{\boldsymbol{\theta},i}(b(a))
-
\mu_{\boldsymbol{\theta},i}(a)
\nonumber\\
&=
\left(
\mu_{\boldsymbol{\theta},i}(b(a))
-
\mu_{\boldsymbol{\theta}',i}(b(a))
\right)
+
\left(
\mu_{\boldsymbol{\theta}',i}(b(a))
-
\mu_{\boldsymbol{\theta}',i}(a)
\right)
\nonumber\\
&\qquad+
\left(
\mu_{\boldsymbol{\theta}',i}(a)
-
\mu_{\boldsymbol{\theta},i}(a)
\right)
\nonumber\\
&\le
\left|
\mu_{\boldsymbol{\theta},i}(b(a))
-
\mu_{\boldsymbol{\theta}',i}(b(a))
\right|
+
\left|
\mu_{\boldsymbol{\theta}',i}(a)
-
\mu_{\boldsymbol{\theta},i}(a)
\right|.
\label{eq:mean-difference-new}
\end{align}

We next relate these mean differences to total variation. For any two
probability laws \(P,Q\) and any measurable function \(f\) satisfying
\(\|f\|_{\infty}\le1\), the dual characterization of total variation
implies
\[
\left|
\mathbb{E}_P[f]
-
\mathbb{E}_Q[f]
\right|
\le
2\,\mathrm{TV}(P,Q).
\]
Since every reward coordinate lies in \([0,1]\), this applies to each
coordinate projection. Hence
\[
\left|
\mu_{\boldsymbol{\theta},i}(b(a))
-
\mu_{\boldsymbol{\theta}',i}(b(a))
\right|
\le
2\,\mathrm{TV}\!\left(
P_{\boldsymbol{\theta}}^{b(a)},
P_{\boldsymbol{\theta}'}^{b(a)}
\right),
\]
and
\[
\left|
\mu_{\boldsymbol{\theta}',i}(a)
-
\mu_{\boldsymbol{\theta},i}(a)
\right|
\le
2\,\mathrm{TV}\!\left(
P_{\boldsymbol{\theta}}^a,
P_{\boldsymbol{\theta}'}^a
\right).
\]
Substituting these bounds into \eqref{eq:mean-difference-new} gives
\begin{equation}
\label{eq:pointwise-gap-tv-new}
\Delta_{\boldsymbol{\theta}}(a)
\le
2\,\mathrm{TV}\!\left(
P_{\boldsymbol{\theta}}^{b(a)},
P_{\boldsymbol{\theta}'}^{b(a)}
\right)
+
2\,\mathrm{TV}\!\left(
P_{\boldsymbol{\theta}}^a,
P_{\boldsymbol{\theta}'}^a
\right).
\end{equation}

Conditional on
\((\boldsymbol{\theta},\boldsymbol{\theta}',\mathcal{H}_t)\),
the action \(A\) is uniform on \(S'\). Averaging
\eqref{eq:pointwise-gap-tv-new} therefore gives
\begin{align}
&\mathbb{E}\!\left[
\Delta_{\boldsymbol{\theta}}(A)
\mid
\boldsymbol{\theta},
\boldsymbol{\theta}',
\mathcal{H}_t
\right]
\nonumber\\
&\quad\le
\frac{2}{|S'|}
\sum_{a\in S'}
\mathrm{TV}\!\left(
P_{\boldsymbol{\theta}}^{b(a)},
P_{\boldsymbol{\theta}'}^{b(a)}
\right)
+
\frac{2}{|S'|}
\sum_{a\in S'}
\mathrm{TV}\!\left(
P_{\boldsymbol{\theta}}^a,
P_{\boldsymbol{\theta}'}^a
\right).
\label{eq:average-gap-tv-new}
\end{align}

For every \(a\in S'\), the witness \(b(a)\) belongs to \(S\).
Consequently,
\begin{align}
\frac{1}{|S'|}
\sum_{a\in S'}
\mathrm{TV}\!\left(
P_{\boldsymbol{\theta}}^{b(a)},
P_{\boldsymbol{\theta}'}^{b(a)}
\right)
&\le
\max_{u\in S}
\mathrm{TV}\!\left(
P_{\boldsymbol{\theta}}^u,
P_{\boldsymbol{\theta}'}^u
\right)
\nonumber\\
&\le
\sum_{u\in S}
\mathrm{TV}\!\left(
P_{\boldsymbol{\theta}}^u,
P_{\boldsymbol{\theta}'}^u
\right).
\label{eq:witness-average-new}
\end{align}
Substituting \eqref{eq:witness-average-new} into
\eqref{eq:average-gap-tv-new},
\begin{align}
&\mathbb{E}\!\left[
\Delta_{\boldsymbol{\theta}}(A)
\mid
\boldsymbol{\theta},
\boldsymbol{\theta}',
\mathcal{H}_t
\right]
\nonumber\\
&\quad\le
2
\sum_{u\in S}
\mathrm{TV}\!\left(
P_{\boldsymbol{\theta}}^u,
P_{\boldsymbol{\theta}'}^u
\right)
+
\frac{2}{|S'|}
\sum_{a\in S'}
\mathrm{TV}\!\left(
P_{\boldsymbol{\theta}}^a,
P_{\boldsymbol{\theta}'}^a
\right).
\label{eq:before-exchange-new}
\end{align}

We now average over
\(\boldsymbol{\theta},\boldsymbol{\theta}'
\stackrel{\mathrm{i.i.d.}}{\sim}p_t\).
Since the two posterior draws are exchangeable and total variation is
symmetric,
\begin{align}
&\mathbb{E}\!\left[
\sum_{u\in S}
\mathrm{TV}\!\left(
P_{\boldsymbol{\theta}}^u,
P_{\boldsymbol{\theta}'}^u
\right)
\middle|
\mathcal{H}_t
\right]
\nonumber\\
&\quad=
\mathbb{E}\!\left[
\sum_{a\in S'}
\mathrm{TV}\!\left(
P_{\boldsymbol{\theta}}^a,
P_{\boldsymbol{\theta}'}^a
\right)
\middle|
\mathcal{H}_t
\right]
\nonumber\\
&\quad=
\sum_{a\in\mathcal{A}}
\mathbb{E}\!\left[
\mathbbm{1}\{a\in S'\}
\,
\mathrm{TV}\!\left(
P_{\boldsymbol{\theta}}^a,
P_{\boldsymbol{\theta}'}^a
\right)
\middle|
\mathcal{H}_t
\right].
\label{eq:exchangeability-explicit-new}
\end{align}

For the second term in \eqref{eq:before-exchange-new}, the Pareto set
\(S'\) is nonempty because \(\mathcal{A}\) is finite and nonempty.
Therefore \(|S'|\ge1\), and
\[
\frac{\mathbbm{1}\{a\in S'\}}{|S'|}
\le
\mathbbm{1}\{a\in S'\}.
\]
Hence
\begin{align}
&\mathbb{E}\!\left[
\frac{1}{|S'|}
\sum_{a\in S'}
\mathrm{TV}\!\left(
P_{\boldsymbol{\theta}}^a,
P_{\boldsymbol{\theta}'}^a
\right)
\middle|
\mathcal{H}_t
\right]
\nonumber\\
&\quad=
\sum_{a\in\mathcal{A}}
\mathbb{E}\!\left[
\frac{\mathbbm{1}\{a\in S'\}}{|S'|}
\mathrm{TV}\!\left(
P_{\boldsymbol{\theta}}^a,
P_{\boldsymbol{\theta}'}^a
\right)
\middle|
\mathcal{H}_t
\right]
\nonumber\\
&\quad\le
\sum_{a\in\mathcal{A}}
\mathbb{E}\!\left[
\mathbbm{1}\{a\in S'\}
\mathrm{TV}\!\left(
P_{\boldsymbol{\theta}}^a,
P_{\boldsymbol{\theta}'}^a
\right)
\middle|
\mathcal{H}_t
\right].
\label{eq:uniform-term-explicit-new}
\end{align}

Taking expectation in \eqref{eq:before-exchange-new} and applying
\eqref{eq:exchangeability-explicit-new} and
\eqref{eq:uniform-term-explicit-new} yields
\[
\mathbb{E}\!\left[
\Delta_{\boldsymbol{\theta}}(A)
\mid
\mathcal{H}_t
\right]
\le
4
\sum_{a\in\mathcal{A}}
\mathbb{E}\!\left[
\mathbbm{1}\{a\in S'\}
\,
\mathrm{TV}\!\left(
P_{\boldsymbol{\theta}}^a,
P_{\boldsymbol{\theta}'}^a
\right)
\middle|
\mathcal{H}_t
\right].
\]
This proves the lemma \ref{lem:theta_stability}.
\end{proof}

\begin{Lemma}
\label{lem:ts_gap_information}
Fix a round \(t\) and condition on \(\mathcal{H}_t\). Suppose every
reward coordinate lies in \([0,R_{\max}]\) almost surely for some finite
\(R_{\max}\). Let \(\boldsymbol{\theta}\sim p_t\) denote the true posterior
parameter and let
\(\boldsymbol{\theta}_t\sim p_t\) be an independent Thompson sample.
Define
\[
S_t:=\mathcal{S}^*(\boldsymbol{\theta}_t),
\qquad
A_t\sim\mathrm{Unif}(S_t),
\]
and
\[
\pi_t(a)
:=
\mathbb{P}(A_t=a\mid\mathcal{H}_t).
\]
For each realized history, define the fixed-action one-step information
gain
\[
I_t(a)
:=
I\!\left(
\boldsymbol{\theta};
\mathbf{Y}_{t,a}
\mid
\mathcal{H}_t,A_t=a
\right),
\]
where this quantity is understood pointwise in the realized history
\(\mathcal{H}_t\). Define
\[
J_t
:=
\sum_{a\in\mathcal{A}}
\pi_t(a)I_t(a).
\]
Then
\begin{equation}
\label{eq:lemmaA2-final-explicit}
\left(
\mathbb{E}\!\left[
\Delta_{\boldsymbol{\theta}}(A_t)
\mid
\mathcal{H}_t
\right]
\right)^2
\le
32R_{\max}^2|\mathcal{A}|M J_t.
\end{equation}
\end{Lemma}

\begin{proof}
If \(R_{\max}=0\), then every reward coordinate is identically zero and
every Pareto gap is zero, so the result is immediate. Assume
\(R_{\max}>0\).

Apply Lemma~\ref{lem:theta_stability} to the normalized rewards
\[
\overline{\mathbf{Y}}_{t,a}
:=
\frac{\mathbf{Y}_{t,a}}{R_{\max}}.
\]
Every coordinate of \(\overline{\mathbf{Y}}_{t,a}\) lies in \([0,1]\).
Positive rescaling preserves all Pareto-dominance relations, so
\[
\overline{\mathcal{S}}^*(\boldsymbol{\theta})
=
\mathcal{S}^*(\boldsymbol{\theta})
\]
for every parameter value, and the normalized Pareto gap satisfies
\[
\overline{\Delta}_{\boldsymbol{\theta}}(a)
=
\frac{\Delta_{\boldsymbol{\theta}}(a)}{R_{\max}}.
\]
Moreover, \(y\mapsto y/R_{\max}\) is a measurable bijection with
measurable inverse, so total variation is invariant under this common
transformation:
\[
\mathrm{TV}\!\left(
\overline P_{\boldsymbol{\theta}}^a,
\overline P_{\boldsymbol{\theta}_t}^a
\right)
=
\mathrm{TV}\!\left(
P_{\boldsymbol{\theta}}^a,
P_{\boldsymbol{\theta}_t}^a
\right).
\]

Conditional on \(\mathcal{H}_t\),
\(\boldsymbol{\theta}\) and \(\boldsymbol{\theta}_t\) are independent
draws from \(p_t\), and conditional on
\((\boldsymbol{\theta},\boldsymbol{\theta}_t)\),
\(A_t\sim\mathrm{Unif}(S_t)\).
Therefore Lemma~\ref{lem:theta_stability} applies with
\[
(\boldsymbol{\theta},\boldsymbol{\theta}')
=
(\boldsymbol{\theta},\boldsymbol{\theta}_t)
\]
and \(A=A_t\). It gives
\[
\frac{1}{R_{\max}}
\mathbb{E}\!\left[
\Delta_{\boldsymbol{\theta}}(A_t)
\mid
\mathcal{H}_t
\right]
\le
4
\sum_{a\in\mathcal{A}}
\mathbb{E}\!\left[
\mathbbm{1}\{a\in S_t\}
\,
\mathrm{TV}\!\left(
P_{\boldsymbol{\theta}}^a,
P_{\boldsymbol{\theta}_t}^a
\right)
\middle|
\mathcal{H}_t
\right].
\]
Thus
\begin{equation}
\label{eq:A2-start-explicit}
\mathbb{E}\!\left[
\Delta_{\boldsymbol{\theta}}(A_t)
\mid
\mathcal{H}_t
\right]
\le
4R_{\max}
\sum_{a\in\mathcal{A}}
\mathbb{E}\!\left[
\mathbbm{1}\{a\in S_t\}
\,
\mathrm{TV}\!\left(
P_{\boldsymbol{\theta}}^a,
P_{\boldsymbol{\theta}_t}^a
\right)
\middle|
\mathcal{H}_t
\right].
\end{equation}

For each \(a\in\mathcal{A}\), define
\[
q_t(a)
:=
\mathbb{P}(a\in S_t\mid\mathcal{H}_t)
=
\mathbb{E}\!\left[
\mathbbm{1}\{a\in S_t\}
\mid
\mathcal{H}_t
\right].
\]
Since \(A_t\) is drawn uniformly from \(S_t\),
\begin{align}
\pi_t(a)
&=
\mathbb{P}(A_t=a\mid\mathcal{H}_t)
\nonumber\\
&=
\mathbb{E}\!\left[
\mathbb{P}(A_t=a\mid
\boldsymbol{\theta}_t,\mathcal{H}_t)
\middle|
\mathcal{H}_t
\right]
\nonumber\\
&=
\mathbb{E}\!\left[
\frac{\mathbbm{1}\{a\in S_t\}}{|S_t|}
\middle|
\mathcal{H}_t
\right].
\label{eq:pi-explicit-A2}
\end{align}

Recall
\[
M
:=
\sup_{\vartheta\in\operatorname{supp}(p_1)}
|\mathcal{S}^*(\vartheta)|.
\]
Since
\[
\mathbb{P}\!\left(
\boldsymbol{\theta}\in\operatorname{supp}(p_1)
\right)=1,
\]
we also have
\[
\mathbb{P}\!\left(
\boldsymbol{\theta}\in\operatorname{supp}(p_1)
\mid
\mathcal{H}_t
\right)
=
1
\qquad\text{almost surely}.
\]
Since \(\boldsymbol{\theta}_t\) is sampled from this posterior conditional
on \(\mathcal{H}_t\), it follows that
\[
\boldsymbol{\theta}_t\in\operatorname{supp}(p_1)
\qquad\text{almost surely}.
\]
Hence
\[
|S_t|
=
|\mathcal{S}^*(\boldsymbol{\theta}_t)|
\le M
\qquad\text{almost surely}.
\]

Consequently,
\[
\frac{\mathbbm{1}\{a\in S_t\}}{|S_t|}
\ge
\frac{\mathbbm{1}\{a\in S_t\}}{M}.
\]
Taking conditional expectation and using
\eqref{eq:pi-explicit-A2},
\begin{equation}
\label{eq:pi-q-explicit-A2}
\pi_t(a)
\ge
\frac{q_t(a)}{M}.
\end{equation}
Hence, whenever \(q_t(a)>0\),
\(\pi_t(a)>0\) and
\begin{equation}
\label{eq:q-pi-explicit-A2}
\frac{q_t(a)}{\pi_t(a)}
\le
M.
\end{equation}
If \(q_t(a)=0\), then
\(\mathbbm{1}\{a\in S_t\}=0\) almost surely conditional on
\(\mathcal{H}_t\), so its corresponding term below vanishes.

For a fixed action \(a\), conditional Cauchy--Schwarz gives
\begin{align}
&\mathbb{E}\!\left[
\mathbbm{1}\{a\in S_t\}
\,
\mathrm{TV}\!\left(
P_{\boldsymbol{\theta}}^a,
P_{\boldsymbol{\theta}_t}^a
\right)
\middle|
\mathcal{H}_t
\right]
\nonumber\\
&\quad\le
\sqrt{q_t(a)}
\sqrt{
\mathbb{E}\!\left[
\mathrm{TV}\!\left(
P_{\boldsymbol{\theta}}^a,
P_{\boldsymbol{\theta}_t}^a
\right)^2
\middle|
\mathcal{H}_t
\right]
}.
\label{eq:fixed-action-CS-explicit}
\end{align}
For \(q_t(a)>0\), insert \(\pi_t(a)\):
\begin{align}
&\sqrt{q_t(a)}
\sqrt{
\mathbb{E}\!\left[
\mathrm{TV}\!\left(
P_{\boldsymbol{\theta}}^a,
P_{\boldsymbol{\theta}_t}^a
\right)^2
\middle|
\mathcal{H}_t
\right]
}
\nonumber\\
&\quad=
\sqrt{\frac{q_t(a)}{\pi_t(a)}}
\sqrt{
\pi_t(a)
\mathbb{E}\!\left[
\mathrm{TV}\!\left(
P_{\boldsymbol{\theta}}^a,
P_{\boldsymbol{\theta}_t}^a
\right)^2
\middle|
\mathcal{H}_t
\right]
}.
\end{align}
Summing over actions and applying Cauchy--Schwarz gives
\begin{align}
&\left(
\sum_{a\in\mathcal{A}}
\mathbb{E}\!\left[
\mathbbm{1}\{a\in S_t\}
\,
\mathrm{TV}\!\left(
P_{\boldsymbol{\theta}}^a,
P_{\boldsymbol{\theta}_t}^a
\right)
\middle|
\mathcal{H}_t
\right]
\right)^2
\nonumber\\
&\quad\le
\left(
\sum_{\substack{a\in\mathcal{A}\\q_t(a)>0}}
\frac{q_t(a)}{\pi_t(a)}
\right)
\left(
\sum_{a\in\mathcal{A}}
\pi_t(a)
\mathbb{E}\!\left[
\mathrm{TV}\!\left(
P_{\boldsymbol{\theta}}^a,
P_{\boldsymbol{\theta}_t}^a
\right)^2
\middle|
\mathcal{H}_t
\right]
\right).
\label{eq:action-CS-explicit-A2}
\end{align}

By \eqref{eq:q-pi-explicit-A2},
\begin{equation}
\label{eq:first-factor-explicit-A2}
\sum_{\substack{a\in\mathcal{A}\\q_t(a)>0}}
\frac{q_t(a)}{\pi_t(a)}
\le
|\mathcal{A}|M.
\end{equation}

For the second factor, conditional on \(\mathcal{H}_t\),
\(\boldsymbol{\theta}\) and \(\boldsymbol{\theta}_t\) are independent
draws from \(p_t\). Therefore, for every fixed \(a\),
\begin{align}
&\mathbb{E}\!\left[
\mathrm{TV}\!\left(
P_{\boldsymbol{\theta}}^a,
P_{\boldsymbol{\theta}_t}^a
\right)^2
\middle|
\mathcal{H}_t
\right]
\nonumber\\
&\quad=
\mathbb{E}_{\boldsymbol{\theta},\boldsymbol{\theta}'
\stackrel{\mathrm{i.i.d.}}{\sim}p_t}
\!\left[
\mathrm{TV}\!\left(
P_{\boldsymbol{\theta}}^a,
P_{\boldsymbol{\theta}'}^a
\right)^2
\right].
\label{eq:iid-TV-explicit-A2}
\end{align}
Lemma~\ref{lem:info_gain_tv_theta} gives
\[
I_t(a)
\ge
\frac{1}{2}
\mathbb{E}_{\boldsymbol{\theta},\boldsymbol{\theta}'
\stackrel{\mathrm{i.i.d.}}{\sim}p_t}
\!\left[
\mathrm{TV}\!\left(
P_{\boldsymbol{\theta}}^a,
P_{\boldsymbol{\theta}'}^a
\right)^2
\right].
\]
Combining this with \eqref{eq:iid-TV-explicit-A2},
\[
\mathbb{E}\!\left[
\mathrm{TV}\!\left(
P_{\boldsymbol{\theta}}^a,
P_{\boldsymbol{\theta}_t}^a
\right)^2
\middle|
\mathcal{H}_t
\right]
\le
2I_t(a).
\]
Multiplying by \(\pi_t(a)\) and summing gives
\begin{equation}
\label{eq:second-factor-explicit-A2}
\sum_{a\in\mathcal{A}}
\pi_t(a)
\mathbb{E}\!\left[
\mathrm{TV}\!\left(
P_{\boldsymbol{\theta}}^a,
P_{\boldsymbol{\theta}_t}^a
\right)^2
\middle|
\mathcal{H}_t
\right]
\le
2J_t.
\end{equation}

Combining
\eqref{eq:action-CS-explicit-A2},
\eqref{eq:first-factor-explicit-A2},
and
\eqref{eq:second-factor-explicit-A2},
\begin{equation}
\label{eq:coupled-TV-final-A2}
\left(
\sum_{a\in\mathcal{A}}
\mathbb{E}\!\left[
\mathbbm{1}\{a\in S_t\}
\,
\mathrm{TV}\!\left(
P_{\boldsymbol{\theta}}^a,
P_{\boldsymbol{\theta}_t}^a
\right)
\middle|
\mathcal{H}_t
\right]
\right)^2
\le
2|\mathcal{A}|M J_t.
\end{equation}

Squaring \eqref{eq:A2-start-explicit} and applying
\eqref{eq:coupled-TV-final-A2},
\[
\left(
\mathbb{E}\!\left[
\Delta_{\boldsymbol{\theta}}(A_t)
\mid
\mathcal{H}_t
\right]
\right)^2
\le
32R_{\max}^2|\mathcal{A}|M J_t.
\]
This proves the lemma \ref{lem:ts_gap_information}.
\end{proof}

\begin{proof}[Proof of Theorem~\ref{thm:ts_regret_gap}]
Without loss of generality, suppose that the rewards are uniformly bounded
and let \(R_{\max}<\infty\) be such that every reward coordinate lies almost
surely in \([0,R_{\max}]\). Indeed, a fixed coordinate-wise shift is a
bijective transformation of the observations and leaves all reward
differences, Pareto-dominance relations, Pareto gaps, and the mutual
information with the observed history unchanged. The hidden constant in the
bound below may depend on this fixed reward bound.

We prove the nontrivial case
\[
\mathrm{IB}_T
=
I(\boldsymbol{\theta};\mathcal{H}_{T+1})
<
\infty.
\]
If \(\mathrm{IB}_T=\infty\), the explicit upper bound below is understood
in the extended sense and is vacuous.

For each round \(t\), define
\[
r_t
:=
\mathbb{E}\!\left[
\Delta_{\boldsymbol{\theta}}(A_t)
\mid
\mathcal{H}_t
\right].
\]
By Lemma~\ref{lem:ts_gap_information},
\[
r_t^2
\le
32R_{\max}^2|\mathcal{A}|M J_t.
\]
Since \(r_t\ge0\),
\begin{equation}
\label{eq:thm-onestep-root}
r_t
\le
4\sqrt{2}\,R_{\max}
\sqrt{|\mathcal{A}|M J_t}.
\end{equation}

By linearity of expectation and the tower property,
\begin{align}
\mathbb{E}\!\left[
\sum_{t=1}^T
\Delta_{\boldsymbol{\theta}}(A_t)
\right]
&=
\sum_{t=1}^T
\mathbb{E}\!\left[
\Delta_{\boldsymbol{\theta}}(A_t)
\right]
\nonumber\\
&=
\sum_{t=1}^T
\mathbb{E}\!\left[
\mathbb{E}\!\left[
\Delta_{\boldsymbol{\theta}}(A_t)
\mid
\mathcal{H}_t
\right]
\right]
\nonumber\\
&=
\sum_{t=1}^T
\mathbb{E}[r_t].
\label{eq:thm-regret-sum-rt}
\end{align}
Applying \eqref{eq:thm-onestep-root},
\[
\mathbb{E}[r_t]
\le
4\sqrt{2}\,R_{\max}
\sqrt{|\mathcal{A}|M}
\,
\mathbb{E}\!\left[\sqrt{J_t}\right].
\]
Hence
\begin{equation}
\label{eq:thm-before-jensen}
\mathbb{E}\!\left[
\sum_{t=1}^T
\Delta_{\boldsymbol{\theta}}(A_t)
\right]
\le
4\sqrt{2}\,R_{\max}
\sqrt{|\mathcal{A}|M}
\sum_{t=1}^T
\mathbb{E}\!\left[\sqrt{J_t}\right].
\end{equation}

Since \(J_t\ge0\) and \(x\mapsto\sqrt{x}\) is concave,
\[
\mathbb{E}\!\left[\sqrt{J_t}\right]
\le
\sqrt{\mathbb{E}[J_t]}.
\]
Therefore,
\[
\mathbb{E}\!\left[
\sum_{t=1}^T
\Delta_{\boldsymbol{\theta}}(A_t)
\right]
\le
4\sqrt{2}\,R_{\max}
\sqrt{|\mathcal{A}|M}
\sum_{t=1}^T
\sqrt{\mathbb{E}[J_t]}.
\]
Applying Cauchy--Schwarz across time,
\[
\sum_{t=1}^T
\sqrt{\mathbb{E}[J_t]}
\le
\sqrt{
T
\sum_{t=1}^T
\mathbb{E}[J_t]
}.
\]
Thus
\begin{equation}
\label{eq:thm-before-information-budget}
\mathbb{E}\!\left[
\sum_{t=1}^T
\Delta_{\boldsymbol{\theta}}(A_t)
\right]
\le
4\sqrt{2}\,R_{\max}
\sqrt{
|\mathcal{A}|M\,T
\sum_{t=1}^T
\mathbb{E}[J_t]
}.
\end{equation}

It remains to identify the cumulative one-step information gain.
Conditional on \(\mathcal{H}_t\), the true parameter has law \(p_t\),
and Algorithm~\ref{alg:ts_pareto_gap} independently draws
\[
\boldsymbol{\theta}_t\sim p_t.
\]
Hence
\[
\boldsymbol{\theta}
\perp\!\!\!\perp
\boldsymbol{\theta}_t
\mid
\mathcal{H}_t.
\]
The algorithm then forms
\[
S_t
=
\mathcal{S}^*(\boldsymbol{\theta}_t)
\]
and samples
\[
A_t\sim\mathrm{Unif}(S_t)
\]
using fresh internal randomization independent of the true parameter.
Therefore
\[
\boldsymbol{\theta}
\perp\!\!\!\perp
A_t
\mid
\mathcal{H}_t,
\]
and consequently
\begin{equation}
\label{eq:thm-action-zero-mi}
I\!\left(
\boldsymbol{\theta};
A_t
\mid
\mathcal{H}_t
\right)
=
0.
\end{equation}

For each realized history,
\[
J_t
=
\sum_{a\in\mathcal{A}}
\pi_t(a)I_t(a).
\]
For every action with \(\pi_t(a)>0\), the fixed-action interpretation
of \(I_t(a)\) agrees with ordinary conditioning on \(A_t=a\); actions
with \(\pi_t(a)=0\) contribute zero to the sum. Hence the standard
decomposition of conditional mutual information over the discrete
action \(A_t\) gives
\begin{align}
\mathbb{E}[J_t]
&=
\mathbb{E}\!\left[
\sum_{a\in\mathcal{A}}
\pi_t(a)I_t(a)
\right]
\nonumber\\
&=
I\!\left(
\boldsymbol{\theta};
\mathbf{Y}_{t,A_t}
\mid
\mathcal{H}_t,A_t
\right).
\label{eq:thm-J-reward-info}
\end{align}

Applying the mutual-information chain rule,
\begin{align}
I\!\left(
\boldsymbol{\theta};
A_t,\mathbf{Y}_{t,A_t}
\mid
\mathcal{H}_t
\right)
&=
I\!\left(
\boldsymbol{\theta};
A_t
\mid
\mathcal{H}_t
\right)
+
I\!\left(
\boldsymbol{\theta};
\mathbf{Y}_{t,A_t}
\mid
\mathcal{H}_t,A_t
\right)
\nonumber\\
&=
\mathbb{E}[J_t],
\label{eq:thm-round-info-J}
\end{align}
where the final equality follows from
\eqref{eq:thm-action-zero-mi} and
\eqref{eq:thm-J-reward-info}.

Since
\[
\mathcal{H}_{t+1}
=
\left(
\mathcal{H}_t,
A_t,
\mathbf{Y}_{t,A_t}
\right),
\]
and \(\mathcal{H}_1\) is empty, applying the mutual-information chain
rule sequentially over \(t=1,\ldots,T\) gives
\begin{align}
I\!\left(
\boldsymbol{\theta};
\mathcal{H}_{T+1}
\right)
&=
\sum_{t=1}^T
I\!\left(
\boldsymbol{\theta};
A_t,\mathbf{Y}_{t,A_t}
\mid
\mathcal{H}_t
\right)
\nonumber\\
&=
\sum_{t=1}^T
\mathbb{E}[J_t].
\label{eq:thm-information-budget}
\end{align}
By definition,
\[
\mathrm{IB}_T
=
I\!\left(
\boldsymbol{\theta};
\mathcal{H}_{T+1}
\right),
\]
so
\[
\sum_{t=1}^T
\mathbb{E}[J_t]
=
\mathrm{IB}_T.
\]
Substituting into
\eqref{eq:thm-before-information-budget},
\[
\mathbb{E}\!\left[
\sum_{t=1}^T
\Delta_{\boldsymbol{\theta}}(A_t)
\right]
\le
4\sqrt{2}\,R_{\max}
\sqrt{
|\mathcal{A}|M\,\mathrm{IB}_T\,T
}.
\]
Since \(R_{\max}\) is a fixed finite reward bound, its multiplicative
contribution is absorbed into the hidden constant. Therefore,
\[
\mathbb{E}\!\left[
\sum_{t=1}^T
\Delta_{\boldsymbol{\theta}}(A_t)
\right]
=
O\!\left(
\sqrt{
|\mathcal{A}|\,M\,\mathrm{IB}_T\,T
}
\right).
\]

Finally,
\[
M
=
\sup_{\boldsymbol{\theta}\in\operatorname{supp}(p_1)}
|\mathcal{S}^*(\boldsymbol{\theta})|
\le
|\mathcal{A}|.
\]
Hence
\[
\sqrt{|\mathcal{A}|M}
\le
|\mathcal{A}|,
\]
and consequently
\[
\mathbb{E}\!\left[
\sum_{t=1}^T
\Delta_{\boldsymbol{\theta}}(A_t)
\right]
=
O\!\left(
|\mathcal{A}|
\sqrt{\mathrm{IB}_T\,T}
\right).
\]
This completes the proof.
\end{proof}

\subsection{Proof of Corollary \ref{cor:ts_set_quality}}

\corTSSetQuality*

\begin{proof}
Fix a round \(t\). Under Algorithm~\ref{alg:ts_pareto_gap}, after drawing
\(\boldsymbol{\theta}_t\) from the posterior, the algorithm forms
\(S_t=\mathcal{S}^*(\boldsymbol{\theta}_t)\) and samples \(A_t\) uniformly
from \(S_t\). Hence, conditional on
\(\mathcal{H}_t,\boldsymbol{\theta}_t,\boldsymbol{\theta}\),
\[
\mathbb{P}\!\left(
A_t=a
\mid
\mathcal{H}_t,\boldsymbol{\theta}_t,\boldsymbol{\theta}
\right)
=
\frac{1}{|S_t|}
\qquad
\text{for every }a\in S_t.
\]
Therefore,
\[
\mathbb{E}\!\left[
\Delta_{\boldsymbol{\theta}}(A_t)
\mid
\mathcal{H}_t,\boldsymbol{\theta}_t,\boldsymbol{\theta}
\right]
=
\frac{1}{|S_t|}
\sum_{a\in S_t}
\Delta_{\boldsymbol{\theta}}(a),
\]
and thus
\[
\sum_{a\in S_t}
\Delta_{\boldsymbol{\theta}}(a)
=
|S_t|\,
\mathbb{E}\!\left[
\Delta_{\boldsymbol{\theta}}(A_t)
\mid
\mathcal{H}_t,\boldsymbol{\theta}_t,\boldsymbol{\theta}
\right].
\]

Recall that
\[
M
:=
\sup_{\boldsymbol{\vartheta}\in\operatorname{supp}(p_1)}
\left|
\mathcal{S}^*(\boldsymbol{\vartheta})
\right|.
\]
Since
\[
\mathbb{P}\!\left(
\boldsymbol{\theta}\in\operatorname{supp}(p_1)
\right)=1,
\]
we also have
\[
\mathbb{P}\!\left(
\boldsymbol{\theta}\in\operatorname{supp}(p_1)
\mid
\mathcal{H}_t
\right)=1
\qquad\text{almost surely}.
\]
Because \(\boldsymbol{\theta}_t\) is sampled from this posterior conditional
on \(\mathcal{H}_t\), it follows that
\[
\boldsymbol{\theta}_t\in\operatorname{supp}(p_1)
\qquad\text{almost surely},
\]
and hence
\[
|S_t|
=
\left|
\mathcal{S}^*(\boldsymbol{\theta}_t)
\right|
\le
M
\qquad\text{almost surely}.
\]
Since \(\Delta_{\boldsymbol{\theta}}(A_t)\ge0\),
\[
\sum_{a\in S_t}
\Delta_{\boldsymbol{\theta}}(a)
\le
M\,
\mathbb{E}\!\left[
\Delta_{\boldsymbol{\theta}}(A_t)
\mid
\mathcal{H}_t,\boldsymbol{\theta}_t,\boldsymbol{\theta}
\right].
\]

Summing over \(t=1,\ldots,T\), taking expectations, and applying the tower
property term by term gives
\begin{align*}
\mathbb{E}\!\left[
\sum_{t=1}^T
\sum_{a\in S_t}
\Delta_{\boldsymbol{\theta}}(a)
\right]
&\le
M
\sum_{t=1}^T
\mathbb{E}\!\left[
\mathbb{E}\!\left[
\Delta_{\boldsymbol{\theta}}(A_t)
\mid
\mathcal{H}_t,\boldsymbol{\theta}_t,\boldsymbol{\theta}
\right]
\right]
\\
&=
M
\sum_{t=1}^T
\mathbb{E}\!\left[
\Delta_{\boldsymbol{\theta}}(A_t)
\right]
\\
&=
M\,
\mathbb{E}\!\left[
\sum_{t=1}^T
\Delta_{\boldsymbol{\theta}}(A_t)
\right].
\end{align*}

By Theorem~\ref{thm:ts_regret_gap},
\[
\mathbb{E}\!\left[
\sum_{t=1}^T
\Delta_{\boldsymbol{\theta}}(A_t)
\right]
=
O\!\left(
\sqrt{
|\mathcal{A}|\,M\,\mathrm{IB}_T\,T
}
\right).
\]
Therefore,
\begin{align*}
\mathbb{E}\!\left[
\sum_{t=1}^T
\sum_{a\in S_t}
\Delta_{\boldsymbol{\theta}}(a)
\right]
&=
O\!\left(
M
\sqrt{
|\mathcal{A}|\,M\,\mathrm{IB}_T\,T
}
\right)
\\
&=
O\!\left(
\sqrt{
|\mathcal{A}|\,M^3\,\mathrm{IB}_T\,T
}
\right).
\end{align*}

Finally, since \(M\le|\mathcal{A}|\),
\[
\sqrt{
|\mathcal{A}|\,M^3\,\mathrm{IB}_T\,T
}
\le
\sqrt{
|\mathcal{A}|^4\,\mathrm{IB}_T\,T
}
=
|\mathcal{A}|^2
\sqrt{
\mathrm{IB}_T\,T
}.
\]
Hence, in the worst case,
\[
\mathbb{E}\!\left[
\sum_{t=1}^T
\sum_{a\in S_t}
\Delta_{\boldsymbol{\theta}}(a)
\right]
=
O\!\left(
|\mathcal{A}|^2
\sqrt{
\mathrm{IB}_T\,T
}
\right).
\]
This proves the corollary.
\end{proof}

\subsection{Proof of Theorem~\ref{thm:ids_regret_gap}}

\thmIDSgap*

\begin{proof}
Fix a round \(t\) and condition on the realized history \(\mathcal{H}_t\). Recall that
\[
\Delta_t(a)
:=
\mathbb{E}\!\left[
\Delta_{\boldsymbol{\theta}}(a)
\mid
\mathcal{H}_t
\right]
\]
and
\[
I_t(a)
:=
I\!\left(
\boldsymbol{\theta};
\mathbf{Y}_{t,a}
\mid
\mathcal{H}_t,A_t=a
\right),
\]
where \(I_t(a)\) is understood pointwise for the realized history
\(\mathcal{H}_t\) under the fixed-action interpretation. For any
\(\pi\in\Pi(\mathcal{A})\), define
\[
\Delta_t(\pi)
:=
\sum_{a\in\mathcal{A}}
\pi(a)\Delta_t(a),
\qquad
I_t(\pi)
:=
\sum_{a\in\mathcal{A}}
\pi(a)I_t(a),
\]
and
\[
\psi_t(\pi)
:=
\frac{\Delta_t(\pi)^2}{I_t(\pi)}.
\]
We use the standard extended-value convention that
\(\psi_t(\pi)=0\) if
\(\Delta_t(\pi)=I_t(\pi)=0\), and
\(\psi_t(\pi)=+\infty\) if
\(I_t(\pi)=0\) while \(\Delta_t(\pi)>0\).

We first establish the generic information-ratio bound. Fix an
\(\mathcal{H}_t\)-measurable minimizer
\(\pi_t^{\mathrm{IDS}}\), using a fixed deterministic tie-breaking rule
when the minimizer is not unique. Conditional on \(\mathcal{H}_t\),
Algorithm~\ref{alg:IDS_gap} samples
\(A_t\sim\pi_t^{\mathrm{IDS}}\) using fresh internal randomness.
Hence
\[
\boldsymbol{\theta}
\perp\!\!\!\perp
A_t
\mid
\mathcal{H}_t.
\]
Therefore,
\begin{align}
\mathbb{E}\!\left[
\Delta_{\boldsymbol{\theta}}(A_t)
\mid
\mathcal{H}_t
\right]
&=
\sum_{a\in\mathcal{A}}
\mathbb{P}(A_t=a\mid\mathcal{H}_t)
\mathbb{E}\!\left[
\Delta_{\boldsymbol{\theta}}(a)
\mid
\mathcal{H}_t,A_t=a
\right]
\nonumber\\
&=
\sum_{a\in\mathcal{A}}
\pi_t^{\mathrm{IDS}}(a)
\mathbb{E}\!\left[
\Delta_{\boldsymbol{\theta}}(a)
\mid
\mathcal{H}_t
\right]
\nonumber\\
&=
\sum_{a\in\mathcal{A}}
\pi_t^{\mathrm{IDS}}(a)\Delta_t(a)
\nonumber\\
&=
\Delta_t(\pi_t^{\mathrm{IDS}}).
\label{eq:ids-cond-gap-final}
\end{align}

Under the assumed bound
\[
\psi_t(\pi_t^{\mathrm{IDS}})
\le
\Psi
<
\infty,
\]
if \(I_t(\pi_t^{\mathrm{IDS}})=0\), then necessarily
\[
\Delta_t(\pi_t^{\mathrm{IDS}})=0,
\]
since otherwise the extended-value convention would give
\(\psi_t(\pi_t^{\mathrm{IDS}})=+\infty\).
If \(I_t(\pi_t^{\mathrm{IDS}})>0\), then
\[
\Delta_t(\pi_t^{\mathrm{IDS}})^2
=
\psi_t(\pi_t^{\mathrm{IDS}})
I_t(\pi_t^{\mathrm{IDS}})
\le
\Psi I_t(\pi_t^{\mathrm{IDS}}).
\]
Hence, in either case,
\[
\Delta_t(\pi_t^{\mathrm{IDS}})
\le
\sqrt{
\Psi\,I_t(\pi_t^{\mathrm{IDS}})
}.
\]

Using \eqref{eq:ids-cond-gap-final}, linearity of expectation, and the
tower property,
\begin{align}
\mathbb{E}\!\left[
\sum_{t=1}^T
\Delta_{\boldsymbol{\theta}}(A_t)
\right]
&=
\sum_{t=1}^T
\mathbb{E}\!\left[
\mathbb{E}\!\left[
\Delta_{\boldsymbol{\theta}}(A_t)
\mid
\mathcal{H}_t
\right]
\right]
\nonumber\\
&=
\sum_{t=1}^T
\mathbb{E}\!\left[
\Delta_t(\pi_t^{\mathrm{IDS}})
\right]
\nonumber\\
&\le
\sqrt{\Psi}
\sum_{t=1}^T
\mathbb{E}\!\left[
\sqrt{
I_t(\pi_t^{\mathrm{IDS}})
}
\right].
\label{eq:ids-before-jensen-final}
\end{align}

Since \(x\mapsto\sqrt{x}\) is concave, Jensen's inequality gives
\[
\mathbb{E}\!\left[
\sqrt{
I_t(\pi_t^{\mathrm{IDS}})
}
\right]
\le
\sqrt{
\mathbb{E}\!\left[
I_t(\pi_t^{\mathrm{IDS}})
\right]
}.
\]
Hence,
\[
\mathbb{E}\!\left[
\sum_{t=1}^T
\Delta_{\boldsymbol{\theta}}(A_t)
\right]
\le
\sqrt{\Psi}
\sum_{t=1}^T
\sqrt{
\mathbb{E}\!\left[
I_t(\pi_t^{\mathrm{IDS}})
\right]
}.
\]
Applying Cauchy--Schwarz across time,
\begin{equation}
\label{eq:ids-generic-before-budget-final}
\mathbb{E}\!\left[
\sum_{t=1}^T
\Delta_{\boldsymbol{\theta}}(A_t)
\right]
\le
\sqrt{
\Psi T
\sum_{t=1}^T
\mathbb{E}\!\left[
I_t(\pi_t^{\mathrm{IDS}})
\right]
}.
\end{equation}

We now identify the information budget. Since \(A_t\) is generated from
\(\mathcal{H}_t\) and fresh internal randomness,
\[
\boldsymbol{\theta}
\perp\!\!\!\perp
A_t
\mid
\mathcal{H}_t,
\]
and therefore
\begin{equation}
\label{eq:ids-zero-action-info-final}
I\!\left(
\boldsymbol{\theta};
A_t
\mid
\mathcal{H}_t
\right)
=
0.
\end{equation}
For every action with
\(\pi_t^{\mathrm{IDS}}(a)>0\), the fixed-action interpretation of
\(I_t(a)\) agrees with ordinary conditioning on \(A_t=a\); actions with
zero selection probability contribute zero. Therefore, averaging over
the selected action and then over the random history gives
\begin{equation}
\label{eq:ids-info-average-final}
\mathbb{E}\!\left[
I_t(\pi_t^{\mathrm{IDS}})
\right]
=
I\!\left(
\boldsymbol{\theta};
\mathbf{Y}_{t,A_t}
\mid
\mathcal{H}_t,A_t
\right).
\end{equation}
Thus, by the conditional mutual-information chain rule and
\eqref{eq:ids-zero-action-info-final},
\begin{align}
I\!\left(
\boldsymbol{\theta};
A_t,\mathbf{Y}_{t,A_t}
\mid
\mathcal{H}_t
\right)
&=
I\!\left(
\boldsymbol{\theta};
A_t
\mid
\mathcal{H}_t
\right)
+
I\!\left(
\boldsymbol{\theta};
\mathbf{Y}_{t,A_t}
\mid
\mathcal{H}_t,A_t
\right)
\nonumber\\
&=
\mathbb{E}\!\left[
I_t(\pi_t^{\mathrm{IDS}})
\right].
\label{eq:ids-one-step-mi-final}
\end{align}

Since
\[
\mathcal{H}_{T+1}
=
\left(
A_1,\mathbf{Y}_{1,A_1},
\ldots,
A_T,\mathbf{Y}_{T,A_T}
\right),
\]
the chain rule for mutual information yields
\begin{align}
\sum_{t=1}^T
\mathbb{E}\!\left[
I_t(\pi_t^{\mathrm{IDS}})
\right]
&=
\sum_{t=1}^T
I\!\left(
\boldsymbol{\theta};
A_t,\mathbf{Y}_{t,A_t}
\mid
\mathcal{H}_t
\right)
\nonumber\\
&=
I\!\left(
\boldsymbol{\theta};
\mathcal{H}_{T+1}
\right)
\nonumber\\
&=
\mathrm{IB}_T.
\label{eq:ids-budget-final}
\end{align}
Combining \eqref{eq:ids-generic-before-budget-final} and
\eqref{eq:ids-budget-final} gives
\[
\mathbb{E}\!\left[
\sum_{t=1}^T
\Delta_{\boldsymbol{\theta}}(A_t)
\right]
\le
\sqrt{
\Psi\,\mathrm{IB}_T\,T
}.
\]
This proves the first claim.

We next establish the uniform information-ratio bound. Fix a round \(t\)
and a realized history \(\mathcal{H}_t\). Let
\(\pi_t^{\mathrm{TS}}\) denote the action distribution obtained by
applying one step of Algorithm~\ref{alg:ts_pareto_gap} to the current
posterior \(p_t\): independently draw
\(\boldsymbol{\theta}_t\sim p_t\), form
\[
S_t
=
\mathcal{S}^*(\boldsymbol{\theta}_t),
\]
and sample
\[
A_t^{\mathrm{TS}}
\sim
\mathrm{Unif}(S_t).
\]
Thus,
\[
\pi_t^{\mathrm{TS}}(a)
=
\mathbb{P}\!\left(
A_t^{\mathrm{TS}}=a
\mid
\mathcal{H}_t
\right).
\]

Lemma~\ref{lem:ts_gap_information} is a pointwise statement conditional
on the current realized history and posterior. Therefore, it applies to
this Thompson-sampling comparator even though the history
\(\mathcal{H}_t\) was generated by Algorithm~\ref{alg:IDS_gap}.
Moreover, the quantities \(I_t(a)\) are fixed-action information gains
determined by the current posterior and observation kernels, so the same
\(I_t(a)\) are used for both the IDS policy and this comparator.
Let the fixed bound on the reward coordinates be \([0,R_{\max}]\).
Then Lemma~\ref{lem:ts_gap_information} gives
\begin{equation}
\label{eq:ids-ts-comparator-final}
\left(
\mathbb{E}\!\left[
\Delta_{\boldsymbol{\theta}}(A_t^{\mathrm{TS}})
\mid
\mathcal{H}_t
\right]
\right)^2
\le
32R_{\max}^2
|\mathcal{A}|M
\sum_{a\in\mathcal{A}}
\pi_t^{\mathrm{TS}}(a)I_t(a).
\end{equation}

By definition,
\begin{align}
\Delta_t(\pi_t^{\mathrm{TS}})
&=
\sum_{a\in\mathcal{A}}
\pi_t^{\mathrm{TS}}(a)\Delta_t(a)
\nonumber\\
&=
\sum_{a\in\mathcal{A}}
\pi_t^{\mathrm{TS}}(a)
\mathbb{E}\!\left[
\Delta_{\boldsymbol{\theta}}(a)
\mid
\mathcal{H}_t
\right]
\nonumber\\
&=
\mathbb{E}\!\left[
\Delta_{\boldsymbol{\theta}}(A_t^{\mathrm{TS}})
\mid
\mathcal{H}_t
\right].
\label{eq:ids-ts-gap-final}
\end{align}
Also,
\[
I_t(\pi_t^{\mathrm{TS}})
=
\sum_{a\in\mathcal{A}}
\pi_t^{\mathrm{TS}}(a)I_t(a).
\]
Therefore, \eqref{eq:ids-ts-comparator-final} becomes
\begin{equation}
\label{eq:ids-ts-info-ratio-pre-final}
\Delta_t(\pi_t^{\mathrm{TS}})^2
\le
32R_{\max}^2
|\mathcal{A}|M
I_t(\pi_t^{\mathrm{TS}}).
\end{equation}

If \(I_t(\pi_t^{\mathrm{TS}})>0\), dividing by
\(I_t(\pi_t^{\mathrm{TS}})\) gives
\[
\psi_t(\pi_t^{\mathrm{TS}})
\le
32R_{\max}^2
|\mathcal{A}|M.
\]
If \(I_t(\pi_t^{\mathrm{TS}})=0\), then
\eqref{eq:ids-ts-info-ratio-pre-final} implies
\[
\Delta_t(\pi_t^{\mathrm{TS}})=0,
\]
and hence, by the extended-value convention,
\[
\psi_t(\pi_t^{\mathrm{TS}})
=
0.
\]
Thus, in all cases,
\begin{equation}
\label{eq:ids-ts-ratio-final}
\psi_t(\pi_t^{\mathrm{TS}})
\le
32R_{\max}^2
|\mathcal{A}|M.
\end{equation}

Since \(\pi_t^{\mathrm{IDS}}\) minimizes \(\psi_t(\pi)\) over all
\(\pi\in\Pi(\mathcal{A})\), and
\(\pi_t^{\mathrm{TS}}\in\Pi(\mathcal{A})\) is feasible,
\[
\psi_t(\pi_t^{\mathrm{IDS}})
\le
\psi_t(\pi_t^{\mathrm{TS}}).
\]
Combining this with \eqref{eq:ids-ts-ratio-final},
\[
\psi_t(\pi_t^{\mathrm{IDS}})
\le
32R_{\max}^2
|\mathcal{A}|M.
\]
Therefore, there exists a constant
\[
C
=
32R_{\max}^2,
\]
depending only on the fixed reward bound, such that
\[
\psi_t(\pi_t^{\mathrm{IDS}})
\le
C|\mathcal{A}|M
\]
for every \(t\in[T]\). This proves the second claim.

Substituting \(\Psi=C|\mathcal{A}|M\) into the generic information-ratio
bound gives
\[
\mathbb{E}\!\left[
\sum_{t=1}^T
\Delta_{\boldsymbol{\theta}}(A_t)
\right]
\le
\sqrt{
C|\mathcal{A}|M\,\mathrm{IB}_T\,T
}
=
O\!\left(
\sqrt{
|\mathcal{A}|M\,\mathrm{IB}_T\,T
}
\right),
\]
where the hidden constant depends only on the fixed reward bound.

Finally, since \(M\le|\mathcal{A}|\),
\[
\sqrt{|\mathcal{A}|M}
\le
|\mathcal{A}|,
\]
and therefore
\[
\mathbb{E}\!\left[
\sum_{t=1}^T
\Delta_{\boldsymbol{\theta}}(A_t)
\right]
=
O\!\left(
|\mathcal{A}|
\sqrt{\mathrm{IB}_T\,T}
\right)
\]
in the worst case. This completes the proof.
\end{proof}

\subsection{Proof of Proposition \ref{prop:ig_bounds}}

\propIGBounds*

\begin{proof}
The three bounds follow from Lemmas~\ref{lem:ig_finite_parameter},
\ref{lem:ig_categorical}, and~\ref{lem:ig_gaussian}, respectively.
\end{proof}

\begin{Lemma}[Information budget for finite parameter spaces]
\label{lem:ig_finite_parameter}
Suppose that the unknown model parameter \(\boldsymbol{\theta}\) takes values in a finite parameter space \(\Theta\). Then, for every \(T\ge 1\),
\[
\mathrm{IB}_T := I(\boldsymbol{\theta};\mathcal H_{T+1})
\le
H(\boldsymbol{\theta})
\le
\log|\Theta|.
\]
\end{Lemma}

\begin{proof}
The result follows from the entropy bound on mutual information. Since \(\boldsymbol{\theta}\) is finite-valued,
\[
I(\boldsymbol{\theta};\mathcal H_{T+1})
=
H(\boldsymbol{\theta})
-
H(\boldsymbol{\theta}\mid \mathcal H_{T+1})
\le
H(\boldsymbol{\theta}),
\]
because \(H(\boldsymbol{\theta}\mid \mathcal H_{T+1})\ge 0\). Also, since \(\boldsymbol{\theta}\) is supported on \(\Theta\), we have
\[
H(\boldsymbol{\theta})\le \log|\Theta|.
\]
Therefore,
\[
\mathrm{IB}_T
=
I(\boldsymbol{\theta};\mathcal H_{T+1})
\le
H(\boldsymbol{\theta})
\le
\log|\Theta|.
\]
\end{proof}

\begin{Lemma}[Logarithmic information growth for the multi-objective categorical reward model]
\label{lem:ig_categorical}
Fix a finite action set \(\mathcal A\), an objective dimension \(d \in \mathbb N\), and a category size \(K \in \mathbb N\). For each action \(a \in \mathcal A\) and coordinate \(j \in [d]:=\{1,\dots,d\}\), let
\[
p_{a,j}=(p_{a,j,1},\dots,p_{a,j,K}) \in \Delta^{K-1},
\qquad
\Delta^{K-1}:=\Bigl\{x \in [0,1]^K : \sum_{k=1}^K x_k = 1\Bigr\}.
\]
Let
\(
\boldsymbol{\theta}:=(p_{a,j})_{a \in \mathcal A,\; j \in [d]}.
\) At round \(t\), an action \(A_t \in \mathcal A\) is selected, and the observed outcome is
\(
Y_t=(Y_{t,1},\dots,Y_{t,d}) \in [K]^d
\). Assume that conditional on \((\boldsymbol{\theta},A_t=a,\mathcal H_t)\), the coordinates are independent and satisfy
\(
Y_{t,j} \mid (\boldsymbol{\theta},A_t=a,\mathcal H_t) \sim \mathrm{Cat}(p_{a,j}), \forall 
 j \in [d].
\) Assume independent Dirichlet priors across \((a,j)\):
\[
p_{a,j} \sim \mathrm{Dir}(\alpha_{a,j}),
\qquad
\alpha_{a,j}=(\alpha_{a,j,1},\dots,\alpha_{a,j,K}) \in (0,\infty)^K,
\]
and 
\(
\alpha_{a,j,0}:=\sum_{k=1}^K \alpha_{a,j,k}
\).  Also assume that
\(
\alpha_{a,j,k} \ge 1
\text{ for all } a \in \mathcal A,\ j \in [d],\ k \in [K].
\)
Let
\(
\mathcal H_t := (A_1,Y_1,\dots,A_{t-1},Y_{t-1})
\) denote the history before round \(t\), and define the information budget
\(
\mathrm{IB}_T:=I(\boldsymbol{\theta};\mathcal H_{T+1})
\). Assume that the action-selection rule uses only \(\mathcal H_t\) and fresh randomness independent of \(\boldsymbol{\theta}\).  Then
\[
\mathrm{IB}_T \le d(K+1)|\mathcal A|(1+\log T)
=
O(dK|\mathcal A|\log T).
\]
\end{Lemma}

\begin{proof}
We proceed in four steps.

\medskip
\noindent
\textbf{Step 1: Posterior update.}

For each \(a\in\mathcal A\), \(j\in[d]\), and \(k\in[K]\), define
\[
N_{a,j,k}(t)
:=
\sum_{s=1}^t
\mathbbm{1}\{A_s=a,\;Y_{s,j}=k\},
\qquad
N_a(t)
:=
\sum_{s=1}^t
\mathbbm{1}\{A_s=a\}.
\]

Fix a time \(t\ge1\). We derive the posterior of \(p_{a,j}\) given the history \(\mathcal H_t\). Because the prior factorizes independently across pairs \((a,j)\) and the observation model factorizes across coordinates conditional on \((\boldsymbol{\theta},A_s,\mathcal H_s)\), it is enough to consider a fixed pair \((a,j)\). Moreover, since the policy selects \(A_s\) using only \(\mathcal H_s\) and fresh randomness independent of \(\boldsymbol{\theta}\), the policy factors do not contribute any \(\boldsymbol{\theta}\)-dependent terms to the posterior kernel.

Condition on the realized history \(\mathcal H_t\). Among the first \(t-1\) rounds, the observations relevant to \(p_{a,j}\) are exactly those rounds \(s\le t-1\) for which \(A_s=a\), in which case the coordinate \(Y_{s,j}\) is drawn from \(\mathrm{Cat}(p_{a,j})\). Hence the likelihood contribution of these observations is
\[
L_{a,j}(p_{a,j};\mathcal H_t)
=
\prod_{k=1}^K
p_{a,j,k}^{\,N_{a,j,k}(t-1)}.
\]

The prior density of \(p_{a,j}\) is Dirichlet:
\[
p_{a,j}
\sim
\mathrm{Dir}(\alpha_{a,j}),
\]
so, up to normalization,
\[
\pi_{a,j}(p_{a,j})
\propto
\prod_{k=1}^K
p_{a,j,k}^{\,\alpha_{a,j,k}-1}.
\]
Multiplying prior and likelihood, the posterior density is proportional to
\[
\pi_{a,j}(p_{a,j}\mid\mathcal H_t)
\propto
\prod_{k=1}^K
p_{a,j,k}^{\,\alpha_{a,j,k}+N_{a,j,k}(t-1)-1}.
\]
This is exactly the kernel of a Dirichlet distribution with parameter vector
\[
\beta_{a,j}(t)
=
\bigl(
\beta_{a,j,1}(t),
\dots,
\beta_{a,j,K}(t)
\bigr),
\qquad
\beta_{a,j,k}(t)
:=
\alpha_{a,j,k}
+
N_{a,j,k}(t-1).
\]
Therefore,
\[
p_{a,j}
\mid
\mathcal H_t
\sim
\mathrm{Dir}(\beta_{a,j}(t)).
\]
Its total concentration is
\[
\beta_{a,j,0}(t)
=
\sum_{k=1}^K
\beta_{a,j,k}(t)
=
\alpha_{a,j,0}
+
\sum_{k=1}^K
N_{a,j,k}(t-1).
\]
Since whenever action \(a\) is played, exactly one category is observed in coordinate \(j\),
\[
\sum_{k=1}^K
N_{a,j,k}(t-1)
=
N_a(t-1).
\]
Hence,
\[
\beta_{a,j,0}(t)
=
\alpha_{a,j,0}
+
N_a(t-1).
\]

\medskip
\noindent
\textbf{Step 2: Information decomposition.}

By definition,
\[
\mathrm{IB}_T
=
I(\boldsymbol{\theta};\mathcal H_{T+1}),
\qquad
\mathcal H_{T+1}
=
(A_1,Y_1,\dots,A_T,Y_T).
\]
Applying the chain rule for mutual information,
\[
I(\boldsymbol{\theta};\mathcal H_{T+1})
=
\sum_{t=1}^T
I(
\boldsymbol{\theta};
(A_t,Y_t)
\mid
\mathcal H_t
).
\]
Applying the conditional chain rule again,
\[
I(
\boldsymbol{\theta};
(A_t,Y_t)
\mid
\mathcal H_t
)
=
I(
\boldsymbol{\theta};
A_t
\mid
\mathcal H_t
)
+
I(
\boldsymbol{\theta};
Y_t
\mid
\mathcal H_t,A_t
).
\]
By assumption, \(A_t\) is generated using only \(\mathcal H_t\) and fresh randomness independent of \(\boldsymbol{\theta}\). Hence,
\[
I(
\boldsymbol{\theta};
A_t
\mid
\mathcal H_t
)
=
0.
\]
Therefore,
\[
\mathrm{IB}_T
=
\sum_{t=1}^T
I(
\boldsymbol{\theta};
Y_t
\mid
\mathcal H_t,A_t
).
\]

\medskip
\noindent
\textbf{Step 3: Reduction to the played action and one-step mutual-information bound.}

Fix \(t\). By the chain rule for mutual information,
\[
I(
\boldsymbol{\theta};
Y_t
\mid
\mathcal H_t,A_t
)
=
\sum_{j=1}^d
I(
\boldsymbol{\theta};
Y_{t,j}
\mid
\mathcal H_t,A_t,Y_{t,1:j-1}
).
\]
For each \(j\),
\[
I(
\boldsymbol{\theta};
Y_{t,j}
\mid
\mathcal H_t,A_t,Y_{t,1:j-1}
)
\le
I(
\boldsymbol{\theta};
Y_{t,j}
\mid
\mathcal H_t,A_t
).
\]
Indeed,
\begin{align*}
&I(
\boldsymbol{\theta};
Y_{t,j}
\mid
\mathcal H_t,A_t,Y_{t,1:j-1}
)\\
&=
H(
Y_{t,j}
\mid
\mathcal H_t,A_t,Y_{t,1:j-1}
)
-
H(
Y_{t,j}
\mid
\boldsymbol{\theta},\mathcal H_t,A_t,Y_{t,1:j-1}
).
\end{align*}
The first entropy is at most
\[
H(
Y_{t,j}
\mid
\mathcal H_t,A_t
),
\]
since conditioning reduces entropy. The second entropy equals
\[
H(
Y_{t,j}
\mid
\boldsymbol{\theta},\mathcal H_t,A_t
),
\]
since the coordinates are conditionally independent given \((\boldsymbol{\theta},A_t,\mathcal H_t)\). Hence,
\[
I(
\boldsymbol{\theta};
Y_t
\mid
\mathcal H_t,A_t
)
\le
\sum_{j=1}^d
I(
\boldsymbol{\theta};
Y_{t,j}
\mid
\mathcal H_t,A_t
).
\]

Now fix \(j\in[d]\) and condition on \(\{A_t=a\}\). Because \(Y_{t,j}\) depends only on the block \(p_{a,j}\) and is conditionally independent of all other components of \(\boldsymbol{\theta}\) given \((p_{a,j},\mathcal H_t,A_t=a)\), we have
\[
I(
\boldsymbol{\theta};
Y_{t,j}
\mid
\mathcal H_t,A_t=a
)
=
I(
p_{a,j};
Y_{t,j}
\mid
\mathcal H_t,A_t=a
).
\]
Moreover, since \(A_t\) is conditionally independent of \(\boldsymbol{\theta}\) given \(\mathcal H_t\), conditioning additionally on \(A_t=a\) does not alter the posterior. Thus,
\[
p_{a,j}
\mid
\mathcal H_t,A_t=a
\sim
\mathrm{Dir}(\beta_{a,j}(t)).
\]

We now derive the one-step Dirichlet--categorical information bound. Let
\[
p
\sim
\mathrm{Dir}(\beta),
\]
where
\[
\beta
=
(\beta_1,\ldots,\beta_K),
\qquad
\beta_k\ge1,
\qquad
\beta_0
:=
\sum_{k=1}^K\beta_k.
\]
Given \(p\), let
\[
Y\mid p
\sim
\mathrm{Cat}(p).
\]
The posterior predictive distribution of \(Y\) is categorical with probabilities
\[
\bar p_k
:=
\mathbb E[p_k]
=
\frac{\beta_k}{\beta_0},
\qquad
k\in[K].
\]
Therefore,
\[
I(p;Y)
=
\mathbb E_p
\left[
\mathrm{KL}
\left(
\mathrm{Cat}(p)
\,\Vert\,
\mathrm{Cat}(\bar p)
\right)
\right]
=
\mathbb E_p
\left[
\sum_{k=1}^K
p_k
\log\frac{p_k}{\bar p_k}
\right].
\]
Expanding the expectation gives
\[
I(p;Y)
=
\sum_{k=1}^K
\mathbb E[p_k\log p_k]
-
\sum_{k=1}^K
\mathbb E[p_k]\log\bar p_k.
\]
For a Dirichlet random vector,
\[
\mathbb E[p_k]
=
\frac{\beta_k}{\beta_0}
\]
and
\[
\mathbb E[p_k\log p_k]
=
\frac{\beta_k}{\beta_0}
\left(
\psi(\beta_k+1)
-
\psi(\beta_0+1)
\right),
\]
where \(\psi\) is the digamma function. Hence,
\[
I(p;Y)
=
\sum_{k=1}^K
\frac{\beta_k}{\beta_0}
\left(
\psi(\beta_k+1)
-
\psi(\beta_0+1)
-
\log\frac{\beta_k}{\beta_0}
\right).
\]

We now upper bound the term inside the parentheses. Using the standard digamma bounds
\[
\psi(x)\le\log x
\qquad\text{and}\qquad
\psi(x)\ge\log x-\frac1x,
\qquad
x\ge1,
\]
we get
\[
\psi(\beta_k+1)
-
\psi(\beta_0+1)
\le
\log(\beta_k+1)
-
\log(\beta_0+1)
+
\frac1{\beta_0+1}.
\]
Therefore,
\begin{align*}
&\psi(\beta_k+1)
-
\psi(\beta_0+1)
-
\log\frac{\beta_k}{\beta_0}\\
&\le
\log\frac{\beta_k+1}{\beta_k}
-
\log\frac{\beta_0+1}{\beta_0}
+
\frac1{\beta_0+1}\\
&=
\log\left(1+\frac1{\beta_k}\right)
-
\log\left(1+\frac1{\beta_0}\right)
+
\frac1{\beta_0+1}.
\end{align*}
Since
\[
-\log\left(1+\frac1{\beta_0}\right)
\le0,
\]
dropping this negative term gives
\[
\psi(\beta_k+1)
-
\psi(\beta_0+1)
-
\log\frac{\beta_k}{\beta_0}
\le
\log\left(1+\frac1{\beta_k}\right)
+
\frac1{\beta_0+1}.
\]
Using
\[
\log(1+u)\le u,
\qquad
u\ge0,
\]
we obtain
\[
\psi(\beta_k+1)
-
\psi(\beta_0+1)
-
\log\frac{\beta_k}{\beta_0}
\le
\frac1{\beta_k}
+
\frac1{\beta_0+1}.
\]
Substituting this into the exact expression for \(I(p;Y)\), we get
\[
I(p;Y)
\le
\sum_{k=1}^K
\frac{\beta_k}{\beta_0}
\left(
\frac1{\beta_k}
+
\frac1{\beta_0+1}
\right).
\]
Now,
\[
\sum_{k=1}^K
\frac{\beta_k}{\beta_0}
\frac1{\beta_k}
=
\frac K{\beta_0},
\]
and
\[
\sum_{k=1}^K
\frac{\beta_k}{\beta_0}
\frac1{\beta_0+1}
=
\frac1{\beta_0+1}.
\]
Therefore,
\[
I(p;Y)
\le
\frac K{\beta_0}
+
\frac1{\beta_0+1}
\le
\frac{K+1}{\beta_0}.
\]
Thus, if
\[
p\sim\mathrm{Dir}(\beta),
\qquad
\beta_k\ge1,
\qquad
Y\mid p\sim\mathrm{Cat}(p),
\]
then
\[
I(p;Y)
\le
\frac{K+1}{\beta_0}.
\]

Now fix a realized history \(\mathcal H_t=h_t\). Applying the one-step bound conditionally on \((\mathcal H_t=h_t,A_t=a)\), we take
\[
p=p_{a,j},
\qquad
\beta=\beta_{a,j}(t).
\]
Since
\[
\beta_{a,j,k}(t)
=
\alpha_{a,j,k}
+
N_{a,j,k}(t-1)
\ge1,
\]
the assumptions of the one-step bound hold. Moreover,
\[
\beta_{a,j,0}(t)
=
\alpha_{a,j,0}
+
N_a(t-1).
\]
Hence,
\[
I(
p_{a,j};
Y_{t,j}
\mid
\mathcal H_t=h_t,A_t=a
)
\le
\frac{K+1}{
\alpha_{a,j,0}+N_a(t-1)
}.
\]
Since
\[
\alpha_{a,j,0}
=
\sum_{k=1}^K
\alpha_{a,j,k}
\ge
K
\ge1,
\]
we obtain
\[
I(
p_{a,j};
Y_{t,j}
\mid
\mathcal H_t=h_t,A_t=a
)
\le
\frac{K+1}{
1+N_a(t-1)
}.
\]

We now remove the conditioning on the realized action. Averaging over the conditional distribution of \(A_t\) given this history,
\begin{align*}
&I(
\boldsymbol{\theta};
Y_{t,j}
\mid
\mathcal H_t=h_t,A_t
)\\
&=
\sum_{a\in\mathcal A}
\Pr(A_t=a\mid\mathcal H_t=h_t)
I(
\boldsymbol{\theta};
Y_{t,j}
\mid
\mathcal H_t=h_t,A_t=a
).
\end{align*}
Therefore,
\[
I(
\boldsymbol{\theta};
Y_{t,j}
\mid
\mathcal H_t=h_t,A_t
)
\le
(K+1)
\sum_{a\in\mathcal A}
\Pr(A_t=a\mid\mathcal H_t=h_t)
\frac1{1+N_a(t-1)}.
\]
Using the earlier coordinate-wise decomposition,
\[
I(
\boldsymbol{\theta};
Y_t
\mid
\mathcal H_t=h_t,A_t
)
\le
d(K+1)
\sum_{a\in\mathcal A}
\Pr(A_t=a\mid\mathcal H_t=h_t)
\frac1{1+N_a(t-1)}.
\]
Finally, averaging over \(\mathcal H_t\) gives
\[
I(
\boldsymbol{\theta};
Y_t
\mid
\mathcal H_t,A_t
)
\le
d(K+1)
\sum_{a\in\mathcal A}
\mathbb E\!\left[
\Pr(A_t=a\mid\mathcal H_t)
\frac1{1+N_a(t-1)}
\right].
\]

\medskip
\noindent
\textbf{Step 4: Summation over time via harmonic counts.}

Combining the previous steps gives
\[
\mathrm{IB}_T
\le
d(K+1)
\sum_{t=1}^T
\sum_{a\in\mathcal A}
\mathbb E
\left[
\Pr(A_t=a\mid\mathcal H_t)
\frac1{1+N_a(t-1)}
\right].
\]
For fixed \(a\in\mathcal A\) and \(t\), define
\[
f_t(\mathcal H_t)
:=
\frac1{1+N_a(t-1)}.
\]
This quantity is \(\mathcal H_t\)-measurable because \(N_a(t-1)\) depends only on the actions taken before round \(t\), all of which are contained in \(\mathcal H_t\). Hence, by the tower property,
\begin{align*}
&\mathbb E
\left[
\mathbbm{1}\{A_t=a\}
f_t(\mathcal H_t)
\right]\\
&=
\mathbb E
\left[
\mathbb E
\left[
\mathbbm{1}\{A_t=a\}
f_t(\mathcal H_t)
\mid
\mathcal H_t
\right]
\right]\\
&=
\mathbb E
\left[
f_t(\mathcal H_t)
\mathbb E
\left[
\mathbbm{1}\{A_t=a\}
\mid
\mathcal H_t
\right]
\right]\\
&=
\mathbb E
\left[
\Pr(A_t=a\mid\mathcal H_t)
\frac1{1+N_a(t-1)}
\right].
\end{align*}
Hence,
\[
\mathrm{IB}_T
\le
d(K+1)
\sum_{a\in\mathcal A}
\mathbb E
\left[
\sum_{t=1}^T
\mathbbm{1}\{A_t=a\}
\frac1{1+N_a(t-1)}
\right].
\]

It remains to bound the inner sum deterministically. Fix an action \(a\). Let
\[
\tau_1^a
<
\tau_2^a
<
\cdots
<
\tau_{N_a(T)}^a
\]
be the times at which action \(a\) is played up to time \(T\). On the \(m\)-th play of action \(a\), that is, at time \(t=\tau_m^a\), the action has already been played \(m-1\) times, so
\[
N_a(\tau_m^a-1)
=
m-1.
\]
Therefore,
\begin{align*}
\sum_{t=1}^T
\mathbbm{1}\{A_t=a\}
\frac1{1+N_a(t-1)}
&=
\sum_{m=1}^{N_a(T)}
\frac1{1+N_a(\tau_m^a-1)}\\
&=
\sum_{m=1}^{N_a(T)}
\frac1m.
\end{align*}
If \(N_a(T)=0\), the sum is empty and equals zero. Otherwise,
\[
\sum_{m=1}^{N_a(T)}
\frac1m
\le
1+\log N_a(T)
\le
1+\log T.
\]
Thus, for every action \(a\),
\[
\sum_{t=1}^T
\mathbbm{1}\{A_t=a\}
\frac1{1+N_a(t-1)}
\le
1+\log T.
\]
Substituting this deterministic bound and summing over \(a\in\mathcal A\) yields
\[
\mathrm{IB}_T
\le
d(K+1)
|\mathcal A|
(1+\log T).
\]
Hence,
\[
\mathrm{IB}_T
=
O(dK|\mathcal A|\log T),
\]
as claimed.
\end{proof}

\begin{Lemma}[Logarithmic information growth for finite-action Gaussian reward models]
\label{lem:ig_gaussian}
Fix a finite action set \(\mathcal A\) and reward dimension \(d\in\mathbb N\). Consider the finite-action Gaussian reward model
\[
Y_t\mid(\boldsymbol{\theta},A_t=a,\mathcal H_t)\sim \mathcal N(\mu_a,\Sigma),
\qquad
\boldsymbol{\theta}:=(\{\mu_a\}_{a\in\mathcal A},\Sigma),
\]
where \(\mu_a\in\mathbb R^d\) is an unknown action-specific mean and \(\Sigma\) is an unknown \(d\times d\) symmetric positive definite shared covariance matrix. Suppose \(\Sigma\sim\mathrm{IW}(\Psi_1,\nu_1)\), where \(\Psi_1\) is a \(d\times d\) symmetric positive definite matrix, and, conditional on \(\Sigma\), the means are independent with
\[
\mu_a\mid \Sigma \sim \mathcal N\!\left(m_{a,1},\frac{1}{\kappa_{a,1}}\Sigma\right),
\qquad a\in\mathcal A,
\]
where \(m_{a,1}\in\mathbb R^d\) and \(\kappa_{a,1}>0\). Let \(\mathcal H_t=(A_1,Y_1,\ldots,A_{t-1},Y_{t-1})\), and assume the action-selection rule uses only \(\mathcal H_t\) and fresh randomness independent of \(\boldsymbol{\theta}\), so that \(I(\boldsymbol{\theta};A_t\mid\mathcal H_t)=0\) for each \(t\). Then, for \(\mathrm{IB}_T:=I(\boldsymbol{\theta};\mathcal H_{T+1})\),
\[
\mathrm{IB}_T
=
O\!\left(d|\mathcal A|\log T+d^2\log T\right).
\]
In particular, for fixed \(d\) and \(|\mathcal A|\), we have \(\mathrm{IB}_T=O(\log T)\).
\end{Lemma}

\begin{proof}
Before proving Lemma~\ref{lem:ig_gaussian}, we record an auxiliary bound controlling the amount of information contained in one Gaussian observation about an inverse-Wishart covariance matrix. This lemma will be applied in Step~5, conditionally on the history \(\mathcal H_t\) and the selected action \(A_t=a\), with
\[
Z
=
Y_t-m_{a,t},
\qquad
c
=
1+\frac1{\kappa_{a,t}},
\qquad
\Sigma\mid\mathcal H_t
\sim
\mathrm{IW}(\Psi_t,\nu_t).
\]

\begin{Lemma}[Covariance-information bound under inverse-Wishart mixing]
\label{lem:IW_cov_info_bound}
Fix \(d\in\mathbb N\), and define
\[
\widetilde C_d
:=
\frac{d(3d+1)}4.
\]
Suppose
\[
\Sigma
\sim
\mathrm{IW}(\Psi,\nu),
\qquad
Z\mid\Sigma
\sim
\mathcal N(0,c\Sigma),
\]
where \(\Psi\) is a \(d\times d\) symmetric positive definite matrix, \(c>0\), and \(\nu>d-1\). Then \(I(\Sigma;Z)<\infty\). Moreover, if \(\nu>d+3\), then
\[
I(\Sigma;Z)
\le
\frac{\widetilde C_d}{\nu-d}.
\]
\end{Lemma}

\begin{proof}
Because mutual information is invariant under invertible deterministic transformations, it suffices to treat the canonical case
\[
\Sigma
\sim
\mathrm{IW}(I_d,\nu),
\qquad
Z\mid\Sigma
\sim
\mathcal N(0,\Sigma).
\]
Indeed, define
\[
\widetilde\Sigma
:=
\Psi^{-1/2}
\Sigma
\Psi^{-1/2},
\qquad
\widetilde Z
:=
c^{-1/2}
\Psi^{-1/2}
Z.
\]
Then
\[
I(\Sigma;Z)
=
I(
\widetilde\Sigma;
\widetilde Z
),
\]
while
\[
\widetilde\Sigma
\sim
\mathrm{IW}(I_d,\nu),
\qquad
\widetilde Z
\mid
\widetilde\Sigma
\sim
\mathcal N(
0,\widetilde\Sigma
).
\]

Hence assume from now on that
\[
\Sigma
\sim
\mathrm{IW}(I_d,\nu),
\qquad
Z\mid\Sigma
\sim
\mathcal N(0,\Sigma).
\]
Set
\[
x
:=
\frac{\nu-d+1}{2},
\qquad
\delta
:=
\frac d2.
\]
Since \(\nu>d-1\), we have \(x>0\), and
\[
x+\delta
=
\frac{\nu+1}{2},
\qquad
\frac\nu2
=
x+\frac{d-1}{2}.
\]

We use
\[
I(\Sigma;Z)
=
h(Z)
-
\mathbb E[
h(Z\mid\Sigma)
].
\]

First,
\[
h(
Z\mid\Sigma=\sigma
)
=
\frac d2
\log(2\pi e)
+
\frac12\log|\sigma|.
\]
Since
\[
\Sigma^{-1}
\sim
\mathrm{Wishart}_d(\nu,I_d),
\]
the standard Wishart identity gives
\[
\mathbb E[
\log|\Sigma|
]
=
-
\psi_d
\left(
\frac\nu2
\right)
-
d\log2,
\]
where
\[
\psi_d(a)
:=
\sum_{i=1}^d
\psi
\left(
a-\frac{i-1}{2}
\right).
\]
Therefore,
\[
\mathbb E[
h(Z\mid\Sigma)
]
=
\frac d2
\log(\pi e)
-
\frac12
\psi_d
\left(
\frac\nu2
\right).
\]

Second, the marginal distribution of \(Z\) is multivariate Student-\(t\) with
\[
r
:=
\nu-d+1
=
2x
\]
degrees of freedom, location \(0\), and scale matrix \(I_d/r\). The entropy of a \(d\)-variate Student-\(t\) distribution with \(r\) degrees of freedom and scale matrix \(\Sigma_t\) is
\begin{align*}
h(Z)
&=
\log\Gamma
\left(
\frac r2
\right)
-
\log\Gamma
\left(
\frac{r+d}{2}
\right)
+
\frac d2
\log(r\pi)\\
&\quad+
\frac12
\log|\Sigma_t|
+
\frac{r+d}{2}
\left[
\psi
\left(
\frac{r+d}{2}
\right)
-
\psi
\left(
\frac r2
\right)
\right].
\end{align*}
Here,
\[
\Sigma_t
=
\frac{I_d}{r},
\]
so
\[
\frac d2
\log(r\pi)
+
\frac12
\log|I_d/r|
=
\frac d2
\log\pi.
\]
Thus,
\[
h(Z)
=
\log\Gamma(x)
-
\log\Gamma(x+\delta)
+
\frac d2\log\pi
+
(x+\delta)
\bigl(
\psi(x+\delta)-\psi(x)
\bigr).
\]

Combining the two entropy formulas gives
\begin{align*}
I(\Sigma;Z)
&=
\log\Gamma(x)
-
\log\Gamma(x+\delta)
-
\delta\\
&\quad+
(x+\delta)
\bigl(
\psi(x+\delta)-\psi(x)
\bigr)
+
\frac12
\sum_{j=0}^{d-1}
\psi
\left(
x+\frac j2
\right).
\end{align*}
Since \(x>0\), all Gamma and digamma terms in the preceding display are finite. Therefore,
\[
I(\Sigma;Z)<\infty.
\]
For later use, denote this finite quantity by
\[
G_d(\nu)
:=
\log\Gamma(x)
-
\log\Gamma(x+\delta)
-
\delta
+
(x+\delta)
\bigl(
\psi(x+\delta)-\psi(x)
\bigr)
+
\frac12
\sum_{j=0}^{d-1}
\psi
\left(
x+\frac j2
\right),
\qquad
x=\frac{\nu-d+1}{2}.
\]
We now establish the sharper bound under the additional condition \(\nu>d+3\). Under this condition, \(x>2\). Adding and subtracting \(\delta\psi(x)\), we obtain
\begin{align*}
I(\Sigma;Z)
&=
\Bigl[
\log\Gamma(x)
-
\log\Gamma(x+\delta)
+
\delta\psi(x)
\Bigr]
-\delta\\
&\quad+
(x+\delta)
\bigl(
\psi(x+\delta)-\psi(x)
\bigr)\\
&\quad+
\frac12
\sum_{j=0}^{d-1}
\left[
\psi
\left(
x+\frac j2
\right)
-
\psi(x)
\right].
\end{align*}

By Taylor's theorem,
\[
\log\Gamma(x+\delta)
=
\log\Gamma(x)
+
\delta\psi(x)
+
\frac{\delta^2}{2}
\psi'(\xi_1)
\]
for some
\[
\xi_1
\in
(x,x+\delta).
\]
Since
\[
\psi'(\xi_1)>0,
\]
we have
\[
\log\Gamma(x)
-
\log\Gamma(x+\delta)
+
\delta\psi(x)
=
-
\frac{\delta^2}{2}
\psi'(\xi_1)
\le0.
\]
Therefore,
\begin{align*}
I(\Sigma;Z)
&\le
-\delta
+
(x+\delta)
\bigl(
\psi(x+\delta)-\psi(x)
\bigr)\\
&\quad+
\frac12
\sum_{j=0}^{d-1}
\left[
\psi
\left(
x+\frac j2
\right)
-
\psi(x)
\right].
\end{align*}

Since the trigamma function \(\psi'\) is positive and decreasing on \((0,\infty)\),
\[
\psi(x+\delta)-\psi(x)
\le
\delta\psi'(x),
\]
and, for every \(j=0,\ldots,d-1\),
\[
\psi
\left(
x+\frac j2
\right)
-
\psi(x)
=
\int_x^{x+j/2}
\psi'(u)\,du
\le
\frac j2
\psi'(x).
\]
Hence,
\[
I(\Sigma;Z)
\le
-\delta
+
(x+\delta)\delta\psi'(x)
+
\frac12
\sum_{j=0}^{d-1}
\frac j2
\psi'(x).
\]
Since
\[
\sum_{j=0}^{d-1}
j
=
\frac{d(d-1)}2,
\]
this becomes
\[
I(\Sigma;Z)
\le
-\delta
+
(x+\delta)\delta\psi'(x)
+
\frac{d(d-1)}8
\psi'(x).
\]
Using the standard trigamma bound
\[
\psi'(x)
\le
\frac1{x-\frac12},
\qquad
x>\frac12,
\]
we obtain
\[
I(\Sigma;Z)
\le
-\delta
+
\frac{
(x+\delta)\delta
}{
x-\frac12
}
+
\frac{
d(d-1)
}{
8(x-\frac12)
}.
\]
Now,
\[
-\delta
+
\frac{
(x+\delta)\delta
}{
x-\frac12
}
=
\delta
\left(
\frac{x+\delta}{x-\frac12}
-
1
\right)
=
\delta
\frac{
\delta+\frac12
}{
x-\frac12
}.
\]
Therefore,
\[
I(\Sigma;Z)
\le
\frac{
\delta(\delta+\frac12)
}{
x-\frac12
}
+
\frac{
d(d-1)
}{
8(x-\frac12)
}.
\]
Since
\[
\delta
=
\frac d2,
\]
we have
\[
\delta
\left(
\delta+\frac12
\right)
=
\frac{d(d+1)}4.
\]
Thus,
\begin{align*}
I(\Sigma;Z)
&\le
\frac{
d(d+1)
}{
4(x-\frac12)
}
+
\frac{
d(d-1)
}{
8(x-\frac12)
}\\
&=
\frac{
d(3d+1)
}{
8(x-\frac12)
}.
\end{align*}
Finally,
\[
x-\frac12
=
\frac{\nu-d}{2},
\]
and hence
\[
I(\Sigma;Z)
\le
\frac{
d(3d+1)
}{
4(\nu-d)
}
=
\frac{
\widetilde C_d
}{
\nu-d
}.
\]
This proves the lemma \ref{lem:IW_cov_info_bound}.
\end{proof}

We now return to the proof of Lemma~\ref{lem:ig_gaussian}. We first describe the posterior updates for the Gaussian reward model with unknown action-specific means and an unknown shared covariance matrix.

\medskip
\noindent
\textbf{Step 1: Posterior update.}

Because the policy selects each \(A_t\) using only \(\mathcal H_t\) and fresh randomness independent of \(\boldsymbol{\theta}\), the joint distribution factors as
\[
p(
\boldsymbol{\theta},
\mathcal H_{T+1}
)
=
p(\boldsymbol{\theta})
\prod_{t=1}^T
p(
A_t\mid\mathcal H_t
)
p(
Y_t
\mid
\boldsymbol{\theta},
A_t,
\mathcal H_t
).
\]
Since
\[
p(
A_t\mid\mathcal H_t
)
\]
does not depend on \(\boldsymbol{\theta}\), these policy factors do not affect the posterior kernel. Therefore, for a realized history, the observation likelihood can be regrouped according to the actions selected.

For each \(a\in\mathcal A\), let
\[
\mathcal I_a(T)
:=
\{
t\in[T]:
A_t=a
\}
\]
be the set of times at which action \(a\) is played, and let
\[
N_a(T)
:=
|\mathcal I_a(T)|.
\]
If \(N_a(T)>0\), define
\[
\bar Y_a(T)
:=
\frac1{N_a(T)}
\sum_{t\in\mathcal I_a(T)}
Y_t
\]
and
\[
W_a(T)
:=
\sum_{t\in\mathcal I_a(T)}
\bigl(
Y_t-\bar Y_a(T)
\bigr)
\bigl(
Y_t-\bar Y_a(T)
\bigr)^\top.
\]
If \(N_a(T)=0\), set
\[
W_a(T)
:=
0,
\]
and choose any value for \(\bar Y_a(T)\); its value is immaterial because all terms involving it below are multiplied by \(N_a(T)\).

For the realized observations associated with action \(a\), the likelihood contribution is
\[
L_a(\mu_a,\Sigma)
=
(2\pi)^{-dN_a(T)/2}
|\Sigma|^{-N_a(T)/2}
\exp
\left(
-\frac12
\sum_{t\in\mathcal I_a(T)}
(Y_t-\mu_a)^\top
\Sigma^{-1}
(Y_t-\mu_a)
\right).
\]
Using the standard quadratic decomposition,
\begin{align*}
&\sum_{t\in\mathcal I_a(T)}
(Y_t-\mu_a)
(Y_t-\mu_a)^\top\\
&=
W_a(T)
+
N_a(T)
\bigl(
\bar Y_a(T)-\mu_a
\bigr)
\bigl(
\bar Y_a(T)-\mu_a
\bigr)^\top,
\end{align*}
we may rewrite the likelihood as
\begin{align*}
L_a(\mu_a,\Sigma)
&\propto
|\Sigma|^{-N_a(T)/2}\\
&\quad\times
\exp
\left(
-\frac12
\operatorname{tr}
\bigl(
\Sigma^{-1}W_a(T)
\bigr)
-
\frac{N_a(T)}2
(\bar Y_a(T)-\mu_a)^\top
\Sigma^{-1}
(\bar Y_a(T)-\mu_a)
\right).
\end{align*}

Now multiply this by the conditional Gaussian prior
\[
\mu_a
\mid
\Sigma
\sim
\mathcal N
\left(
m_{a,1},
\frac1{\kappa_{a,1}}
\Sigma
\right),
\]
whose density is
\[
p(\mu_a\mid\Sigma)
=
(2\pi)^{-d/2}
\kappa_{a,1}^{d/2}
|\Sigma|^{-1/2}
\exp
\left(
-\frac{\kappa_{a,1}}2
(\mu_a-m_{a,1})^\top
\Sigma^{-1}
(\mu_a-m_{a,1})
\right).
\]
Combining the quadratic terms in \(\mu_a\), we obtain
\begin{align*}
&N_a(T)
(\bar Y_a(T)-\mu_a)^\top
\Sigma^{-1}
(\bar Y_a(T)-\mu_a)\\
&\quad+
\kappa_{a,1}
(\mu_a-m_{a,1})^\top
\Sigma^{-1}
(\mu_a-m_{a,1})\\
&=
\bigl(
\kappa_{a,1}+N_a(T)
\bigr)
(\mu_a-m_{a,T+1})^\top
\Sigma^{-1}
(\mu_a-m_{a,T+1})\\
&\quad+
\frac{
\kappa_{a,1}N_a(T)
}{
\kappa_{a,1}+N_a(T)
}
(\bar Y_a(T)-m_{a,1})^\top
\Sigma^{-1}
(\bar Y_a(T)-m_{a,1}),
\end{align*}
where
\[
m_{a,T+1}
=
\frac{
\kappa_{a,1}m_{a,1}
+
N_a(T)\bar Y_a(T)
}{
\kappa_{a,1}
+
N_a(T)
}.
\]
Hence,
\[
\mu_a
\mid
(
\Sigma,
\mathcal H_{T+1}
)
\sim
\mathcal N
\left(
m_{a,T+1},
\frac1{\kappa_{a,T+1}}
\Sigma
\right),
\qquad
\kappa_{a,T+1}
:=
\kappa_{a,1}
+
N_a(T).
\]

Integrating out \(\mu_a\), the Gaussian integral contributes the factor
\[
(2\pi)^{d/2}
\kappa_{a,T+1}^{-d/2}
|\Sigma|^{1/2},
\]
which cancels the additional
\[
|\Sigma|^{-1/2}
\]
arising from the Gaussian prior density. Therefore, after marginalizing \(\mu_a\), the contribution of action \(a\) to the kernel in \(\Sigma\) is
\begin{align*}
&|\Sigma|^{-N_a(T)/2}\\
&\quad\times
\exp
\left(
-\frac12
\operatorname{tr}
\left(
\Sigma^{-1}
\left[
W_a(T)
+
\frac{
\kappa_{a,1}N_a(T)
}{
\kappa_{a,1}+N_a(T)
}
(\bar Y_a(T)-m_{a,1})
(\bar Y_a(T)-m_{a,1})^\top
\right]
\right)
\right).
\end{align*}

The prior on \(\Sigma\) is inverse-Wishart:
\[
\Sigma
\sim
\mathrm{IW}(\Psi_1,\nu_1),
\]
whose density is proportional to
\[
|\Sigma|^{-(\nu_1+d+1)/2}
\exp
\left(
-\frac12
\operatorname{tr}
(
\Psi_1\Sigma^{-1}
)
\right).
\]
Multiplying over all actions \(a\in\mathcal A\), the posterior density in \(\Sigma\) is proportional to
\[
|\Sigma|^{-(\nu_1+d+1+T)/2}
\exp
\left(
-\frac12
\operatorname{tr}
\bigl(
\Psi_{T+1}
\Sigma^{-1}
\bigr)
\right),
\]
where
\[
\nu_{T+1}
:=
\nu_1+T
\]
and
\[
\Psi_{T+1}
=
\Psi_1
+
\sum_{a\in\mathcal A}
\left[
W_a(T)
+
\frac{
\kappa_{a,1}N_a(T)
}{
\kappa_{a,1}+N_a(T)
}
\bigl(
\bar Y_a(T)-m_{a,1}
\bigr)
\bigl(
\bar Y_a(T)-m_{a,1}
\bigr)^\top
\right].
\]
This is exactly the kernel of an inverse-Wishart distribution, so
\[
\Sigma
\mid
\mathcal H_{T+1}
\sim
\mathrm{IW}
(
\Psi_{T+1},
\nu_{T+1}
).
\]

Applying the same update to the first \(t-1\) observations gives
\[
\Sigma
\mid
\mathcal H_t
\sim
\mathrm{IW}
(
\Psi_t,
\nu_t
),
\qquad
\nu_t
=
\nu_1+t-1,
\]
and
\[
\mu_a
\mid
(
\mathcal H_t,
\Sigma
)
\sim
\mathcal N
\left(
m_{a,t},
\frac1{\kappa_{a,t}}
\Sigma
\right),
\qquad
\kappa_{a,t}
=
\kappa_{a,1}
+
N_a(t-1).
\]

\medskip
\noindent
\textbf{Step 2: Chain rule for the information budget.}

By definition,
\[
\mathrm{IB}_T
=
I(
\boldsymbol{\theta};
\mathcal H_{T+1}
),
\qquad
\mathcal H_{T+1}
=
(A_1,Y_1,\dots,A_T,Y_T).
\]
Applying the chain rule for mutual information,
\[
I(
\boldsymbol{\theta};
\mathcal H_{T+1}
)
=
\sum_{t=1}^T
I
\bigl(
\boldsymbol{\theta};
(A_t,Y_t)
\mid
\mathcal H_t
\bigr).
\]
For each \(t\),
\[
I
\bigl(
\boldsymbol{\theta};
(A_t,Y_t)
\mid
\mathcal H_t
\bigr)
=
I(
\boldsymbol{\theta};
A_t
\mid
\mathcal H_t
)
+
I(
\boldsymbol{\theta};
Y_t
\mid
\mathcal H_t,A_t
).
\]
By assumption,
\[
I(
\boldsymbol{\theta};
A_t
\mid
\mathcal H_t
)
=
0.
\]
Therefore,
\[
\mathrm{IB}_T
=
\sum_{t=1}^T
I(
\boldsymbol{\theta};
Y_t
\mid
\mathcal H_t,A_t
).
\]

\medskip
\noindent
\textbf{Step 3: Per-round decomposition into mean and covariance information.}

For each round \(t\), since
\[
\boldsymbol{\theta}
=
(\mu,\Sigma),
\qquad
\mu
:=
(\mu_a)_{a\in\mathcal A},
\]
the chain rule gives
\[
I(
\boldsymbol{\theta};
Y_t
\mid
\mathcal H_t,A_t
)
=
I(
\Sigma;
Y_t
\mid
\mathcal H_t,A_t
)
+
I(
\mu;
Y_t
\mid
\mathcal H_t,A_t,\Sigma
).
\]
Summing over \(t\) and using Step~2,
\[
\mathrm{IB}_T
=
\sum_{t=1}^T
I(
\mu;
Y_t
\mid
\mathcal H_t,A_t,\Sigma
)
+
\sum_{t=1}^T
I(
\Sigma;
Y_t
\mid
\mathcal H_t,A_t
).
\]

\medskip
\noindent
\textbf{Step 4: Bound the mean-information term.}

Fix a round \(t\) and condition on
\[
(
\mathcal H_t=h_t,
A_t=a,
\Sigma=\sigma
).
\]
By conjugacy,
\[
\mu_a
\mid
(
\mathcal H_t=h_t,
\Sigma=\sigma
)
\sim
\mathcal N
\left(
m_{a,t},
\frac1{\kappa_{a,t}}
\sigma
\right),
\]
where
\[
\kappa_{a,t}
=
\kappa_{a,1}
+
N_a(t-1).
\]
Since \(A_t\) is conditionally independent of \(\boldsymbol{\theta}\) given \(\mathcal H_t\), conditioning additionally on \(A_t=a\) does not change this posterior.

Moreover,
\[
Y_t
\mid
(
\mu_a,
\mathcal H_t=h_t,
A_t=a,
\Sigma=\sigma
)
\sim
\mathcal N(
\mu_a,
\sigma
).
\]
Conditional on
\[
(
\mathcal H_t=h_t,
A_t=a,
\Sigma=\sigma
),
\]
only \(\mu_a\) enters the observation model, so
\[
\mu_{-a}
:=
(\mu_b)_{b\neq a}
\]
drops out by conditional independence. Therefore,
\[
I(
\mu;
Y_t
\mid
\mathcal H_t=h_t,
A_t=a,
\Sigma=\sigma
)
=
I(
\mu_a;
Y_t
\mid
\mathcal H_t=h_t,
A_t=a,
\Sigma=\sigma
).
\]
Thus, conditional on these quantities, this is a Gaussian linear channel with prior covariance
\[
\frac{\sigma}{\kappa_{a,t}}
\]
and noise covariance \(\sigma\). Therefore,
\begin{align*}
&I(
\mu;
Y_t
\mid
\mathcal H_t=h_t,
A_t=a,
\Sigma=\sigma
)\\
&=
\frac12
\log\det
\left(
I_d
+
\frac1{\kappa_{a,t}}
I_d
\right)\\
&=
\frac d2
\log
\left(
1+\frac1{\kappa_{a,t}}
\right).
\end{align*}
Averaging over
\[
(
\mathcal H_t,
A_t,
\Sigma
)
\]
gives
\[
I(
\mu;
Y_t
\mid
\mathcal H_t,A_t,\Sigma
)
=
\frac d2
\mathbb E
\left[
\log
\left(
1+\frac1{\kappa_{A_t,t}}
\right)
\right].
\]
Summing over \(t\),
\[
\sum_{t=1}^T
I(
\mu;
Y_t
\mid
\mathcal H_t,A_t,\Sigma
)
=
\frac d2
\mathbb E
\left[
\sum_{t=1}^T
\log
\left(
1+\frac1{\kappa_{A_t,t}}
\right)
\right].
\]

Now group the sum by action. For a fixed action \(a\), every time action \(a\) is played the value of \(\kappa_{a,t}\) runs through
\[
\kappa_{a,1},
\quad
\kappa_{a,1}+1,
\quad
\dots,
\quad
\kappa_{a,1}+N_a(T)-1.
\]
Therefore, pathwise,
\begin{align*}
\sum_{t:\,A_t=a}
\log
\left(
1+\frac1{\kappa_{A_t,t}}
\right)
&=
\sum_{j=0}^{N_a(T)-1}
\log
\left(
1+\frac1{\kappa_{a,1}+j}
\right)\\
&=
\sum_{j=0}^{N_a(T)-1}
\log
\left(
\frac{
\kappa_{a,1}+j+1
}{
\kappa_{a,1}+j
}
\right)\\
&=
\log
\left(
\frac{
\kappa_{a,1}+N_a(T)
}{
\kappa_{a,1}
}
\right)\\
&=
\log
\left(
1+\frac{N_a(T)}{\kappa_{a,1}}
\right).
\end{align*}
Summing over actions yields the pathwise identity
\[
\sum_{t=1}^T
\log
\left(
1+\frac1{\kappa_{A_t,t}}
\right)
=
\sum_{a\in\mathcal A}
\log
\left(
1+\frac{N_a(T)}{\kappa_{a,1}}
\right).
\]
Taking expectation gives
\[
\sum_{t=1}^T
I(
\mu;
Y_t
\mid
\mathcal H_t,A_t,\Sigma
)
=
\frac d2
\mathbb E
\left[
\sum_{a\in\mathcal A}
\log
\left(
1+\frac{N_a(T)}{\kappa_{a,1}}
\right)
\right].
\]

\medskip
\noindent
\textbf{Step 5: Bound the covariance-information term.}

Fix \(t\), a realized history \(\mathcal H_t=h_t\), and an action \(a\in\mathcal A\). By Step~1 applied at horizon \(t-1\),
\[
\Sigma
\mid
\mathcal H_t=h_t
\sim
\mathrm{IW}(\Psi_t,\nu_t),
\qquad
\nu_t
=
\nu_1+t-1.
\]
Because \(A_t\) is conditionally independent of \(\boldsymbol{\theta}\) given \(\mathcal H_t\), conditioning additionally on \(A_t=a\) does not change this posterior:
\[
\Sigma
\mid
(
\mathcal H_t=h_t,
A_t=a
)
\sim
\mathrm{IW}(\Psi_t,\nu_t).
\]

Also, by conjugacy,
\[
\mu_a
\mid
(
\mathcal H_t=h_t,
\Sigma
)
\sim
\mathcal N
\left(
m_{a,t},
\frac1{\kappa_{a,t}}
\Sigma
\right).
\]
Since
\[
Y_t
\mid
(
\mu_a,
\mathcal H_t=h_t,
A_t=a,
\Sigma
)
\sim
\mathcal N(
\mu_a,
\Sigma
),
\]
integrating out \(\mu_a\) gives
\[
Y_t
\mid
(
\mathcal H_t=h_t,
A_t=a,
\Sigma
)
\sim
\mathcal N
\left(
m_{a,t},
\left(
1+\frac1{\kappa_{a,t}}
\right)
\Sigma
\right).
\]
Because \(m_{a,t}\) is fixed conditional on \(\mathcal H_t=h_t\), translation invariance of mutual information yields
\[
I(
\Sigma;
Y_t
\mid
\mathcal H_t=h_t,A_t=a
)
=
I
\left(
\Sigma;
Y_t-m_{a,t}
\mid
\mathcal H_t=h_t,A_t=a
\right).
\]
Conditional on \((\mathcal H_t=h_t,A_t=a)\), the pair therefore has the form
\[
\Sigma
\sim
\mathrm{IW}(\Psi_t,\nu_t),
\qquad
Z\mid\Sigma
\sim
\mathcal N(0,c_t\Sigma),
\qquad
c_t
:=
1+\frac1{\kappa_{a,t}}.
\]

Because \(\mathrm{IW}(\Psi_1,\nu_1)\) is a proper inverse-Wishart prior, \(\nu_1>d-1\). Define
\[
t_0
:=
\min\left\{
t\ge1:
\nu_1+t-1>d+3
\right\}.
\]
Then \(t_0\) is finite (indeed, \(t_0\le5\)). For \(t<t_0\), the first part of Lemma~\ref{lem:IW_cov_info_bound} gives, pointwise in \((h_t,a)\),
\[
I(
\Sigma;
Y_t
\mid
\mathcal H_t=h_t,A_t=a
)
=
G_d(\nu_t)
<
\infty.
\]
For \(t\ge t_0\), the second part of Lemma~\ref{lem:IW_cov_info_bound} gives
\[
I(
\Sigma;
Y_t
\mid
\mathcal H_t=h_t,A_t=a
)
\le
\frac{\widetilde C_d}{\nu_t-d}.
\]
Both bounds are independent of the realized history \(h_t\) and action \(a\). Hence, defining the finite prior-dependent constant
\[
C_{\mathrm{early}}
:=
\sum_{t=1}^{t_0-1}
G_d(\nu_t),
\]
we obtain, after averaging over \((\mathcal H_t,A_t)\) and summing over time,
\[
\sum_{t=1}^T
I(
\Sigma;
Y_t
\mid
\mathcal H_t,A_t
)
\le
C_{\mathrm{early}}
+
\widetilde C_d
\sum_{t=t_0}^T
\frac1{\nu_1+t-d-1},
\]
where the second sum is understood to be zero when \(T<t_0\).

Let
\[
b_0
:=
\nu_1+t_0-d-1
>
3.
\]
If \(T\ge t_0\), then
\begin{align*}
\sum_{t=t_0}^T
\frac1{\nu_1+t-d-1}
&=
\sum_{s=0}^{T-t_0}
\frac1{b_0+s}\\
&\le
\frac1{b_0}
+
\int_{b_0}^{b_0+T-t_0}
\frac{dx}{x}\\
&=
\frac1{b_0}
+
\log
\left(
1+\frac{T-t_0}{b_0}
\right)\\
&\le
\frac1{b_0}
+
\log(1+T).
\end{align*}
Thus, for every \(T\ge1\),
\[
\sum_{t=1}^T
I(
\Sigma;
Y_t
\mid
\mathcal H_t,A_t
)
\le
C_{\mathrm{early}}
+
\widetilde C_d
\left(
\frac1{b_0}
+
\log(1+T)
\right).
\]

\medskip
\noindent
\textbf{Step 6: Combine the bounds and derive the final rate.}

Combining Steps~3--5 gives
\begin{align*}
\mathrm{IB}_T
&\le
\frac d2
\mathbb E
\left[
\sum_{a\in\mathcal A}
\log
\left(
1+\frac{N_a(T)}{\kappa_{a,1}}
\right)
\right]\\
&\quad+
C_{\mathrm{early}}
+
\widetilde C_d
\left(
\frac1{b_0}
+
\log(1+T)
\right).
\end{align*}

We now bound the mean contribution. For every \(a\in\mathcal A\),
\[
0
\le
N_a(T)
\le
T.
\]
Since \(\kappa_{a,1}>0\) is fixed by the prior,
\[
\log
\left(
1+\frac{N_a(T)}{\kappa_{a,1}}
\right)
\le
\log
\left(
1+\frac{T}{\kappa_{a,1}}
\right).
\]
Moreover,
\begin{align*}
\log
\left(
1+\frac{T}{\kappa_{a,1}}
\right)
&\le
\log
\left(
(1+T)
\left(
1+\frac1{\kappa_{a,1}}
\right)
\right)\\
&=
\log(1+T)
+
\log
\left(
1+\frac1{\kappa_{a,1}}
\right).
\end{align*}
Therefore,
\[
\sum_{a\in\mathcal A}
\log
\left(
1+\frac{N_a(T)}{\kappa_{a,1}}
\right)
\le
|\mathcal A|
\log(1+T)
+
\sum_{a\in\mathcal A}
\log
\left(
1+\frac1{\kappa_{a,1}}
\right).
\]
Taking expectations preserves the same bound. Hence,
\begin{align*}
\mathrm{IB}_T
&\le
\frac d2
|\mathcal A|
\log(1+T)
+
\frac d2
\sum_{a\in\mathcal A}
\log
\left(
1+\frac1{\kappa_{a,1}}
\right)\\
&\quad+
\widetilde C_d
\log(1+T)
+
C_{\mathrm{early}}
+
\frac{\widetilde C_d}{b_0}.
\end{align*}
Equivalently,
\[
\mathrm{IB}_T
\le
\left(
\frac d2
|\mathcal A|
+
\widetilde C_d
\right)
\log(1+T)
+
C_{\mathrm{prior}},
\]
where
\[
C_{\mathrm{prior}}
:=
\frac d2
\sum_{a\in\mathcal A}
\log
\left(
1+\frac1{\kappa_{a,1}}
\right)
+
C_{\mathrm{early}}
+
\frac{\widetilde C_d}{b_0}.
\]
The quantity \(C_{\mathrm{prior}}\) is finite and depends only on the fixed prior hyperparameters and model dimension, not on \(T\). Finally, since
\[
\widetilde C_d
=
\frac{d(3d+1)}4
=
O(d^2),
\]
we obtain
\[
\mathrm{IB}_T
=
O
\left(
d|\mathcal A|
\log(1+T)
+
d^2
\log(1+T)
\right).
\]
Since
\[
\log(1+T)
=
O(\log T),
\]
this is equivalently
\[
\mathrm{IB}_T
=
O
\left(
d|\mathcal A|
\log T
+
d^2
\log T
\right),
\]
as claimed.
\end{proof}

\subsection{Gaussian Rewards: Clipping and Tail Localization}
\label{app:gaussian_high_probability}

We bound the gap-based regret of the clipped versions of
Algorithms~\ref{alg:ts_pareto_gap} and~\ref{alg:IDS_gap}.
The learner uses the same prior \(p_1\) from which the true
parameter is drawn. Conditional on \(\boldsymbol{\theta}\), the
reward arrays are i.i.d.\ across rounds, with arm laws
\(P_{\boldsymbol{\theta}}^a\). Regret is evaluated at the original mean
rewards, so the cumulative loss is
\(R_T:=\sum_{t=1}^T\Delta_{\boldsymbol{\theta}}(A_t)\).
The proof uses uniform bounds on the conditional means and
centered reward tails, together with an information budget for the
raw rewards along the actions of the clipped policy.
For the conditioned shared-covariance Gaussian model below, the
same clipped Thompson Sampling run also recovers the Pareto set
of the original means. We establish the properties needed for
that recovery here and give the error bound in
Appendix~\ref{app:ts_gap_set_recovery}.

\paragraph{Uniform mean and tail conditions.}
Let \(\mu_{\max}>0\) and \(\sigma>0\) be deterministic constants
known to the learner and fixed independently of \(T\). They will
determine the clipping level. Assume that, for \(p_1\)-almost every
\(\boldsymbol{\theta}\), all \(a\in\mathcal A\), \(j\in[d]\), and
\(\lambda\in\mathbb R\),
\begin{align}
 |\mu_{\boldsymbol{\theta},j}(a)|&\le\mu_{\max}, \notag\\
 \mathbb E_{\boldsymbol{\theta}}\!\left[
   \exp\!\bigl(\lambda(Y_{t,a,j}-\mu_{\boldsymbol{\theta},j}(a))\bigr)
 \right]
 &\le \exp(\sigma^2\lambda^2/2).
 \label{eq:gaussian-clipping-uniform-conditions}
\end{align}
For the shared-covariance Gaussian model, the centered moment
generating function is \(\exp(\lambda^2\Sigma_{jj}/2)\). Thus
\eqref{eq:gaussian-clipping-uniform-conditions} holds when
\(\max_{a,j}|\mu_{a,j}|\le\mu_{\max}\) and
\(\max_j\Sigma_{jj}\le\sigma^2\) with prior probability one.
The unconditioned normal--inverse-Wishart prior gives unbounded
support to the arm means and diagonal covariance entries.
Lemma~\ref{lem:ig_gaussian} establishes a logarithmic information
budget for that prior. Below we construct a conditioned Gaussian
prior with both a logarithmic budget and the uniform bounds in
\eqref{eq:gaussian-clipping-uniform-conditions}.

\paragraph{Clipping bias in the Pareto gap.}
Fix \(B>\mu_{\max}\) for the run. Apply the following clipping
map to each coordinate and define its mean reward by
\[
 c_B(y):=\max\{-B,\min\{y,B\}\},
 \qquad
 \widetilde{\mu}_{\boldsymbol{\theta},j}(a)
 :=\mathbb E_{\boldsymbol{\theta}}[c_B(Y_{t,a,j})].
\]
Set \(u_B:=B-\mu_{\max}>0\). Fix a parameter satisfying
\eqref{eq:gaussian-clipping-uniform-conditions} and write
\(X=Y_{t,a,j}-\mu_{\boldsymbol{\theta},j}(a)\). For \(u>0\),
Markov's inequality with \(\lambda=u/\sigma^2\) gives
\[
 \mathbb P_{\boldsymbol{\theta}}(X>u)
 \le e^{-\lambda u}
       \mathbb E_{\boldsymbol{\theta}}[e^{\lambda X}]
 \le \exp(-u^2/(2\sigma^2)).
\]
Applying the same inequality to \(-X\) and adding the two bounds
gives
\(\mathbb P_{\boldsymbol{\theta}}(|X|>u)
  \le 2\exp(-u^2/(2\sigma^2))\).
Since \(|y-c_B(y)|=(|y|-B)_+\) and
\(|Y_{t,a,j}|\le |X|+\mu_{\max}\), we have
\begin{align*}
 |\mu_{\boldsymbol{\theta},j}(a)
      -\widetilde{\mu}_{\boldsymbol{\theta},j}(a)|
 &\le \mathbb E_{\boldsymbol{\theta}}[(|Y_{t,a,j}|-B)_+]\\
 &\le \mathbb E_{\boldsymbol{\theta}}[(|X|-u_B)_+]\\
 &\le 2\int_{u_B}^\infty \exp(-u^2/(2\sigma^2))\,du\\
 &\le \frac{2}{u_B}\int_{u_B}^\infty
        u\exp(-u^2/(2\sigma^2))\,du
 =:\varepsilon_B.
\end{align*}
The third line uses
\(\mathbb E[(Z-u_B)_+]=\int_{u_B}^{\infty}\mathbb P(Z>u)\,du\)
with \(Z=|X|\). The last line uses \(u/u_B\ge1\) on that
integration range. Evaluating the last integral gives
\[
 \varepsilon_B
 =\frac{2\sigma^2}{B-\mu_{\max}}
   \exp\!\left(-\frac{(B-\mu_{\max})^2}{2\sigma^2}\right).
\]
The Pareto gap of the clipped means is
\[
 \widetilde{\Delta}_{\boldsymbol{\theta},B}(a)
 :=\left[
 \max_{b\in\mathcal A}\min_{j\in[d]}
   \bigl(\widetilde{\mu}_{\boldsymbol{\theta},j}(b)
        -\widetilde{\mu}_{\boldsymbol{\theta},j}(a)\bigr)
 \right]_+ .
\]
For each \(b\in\mathcal A\) and \(j\in[d]\), the original
coordinate-wise mean difference
\(\mu_{\boldsymbol{\theta},j}(b)
  -\mu_{\boldsymbol{\theta},j}(a)\)
and its clipped counterpart differ by at most
\(2\varepsilon_B\), since each mean changes by at most
\(\varepsilon_B\). For each \(b\), taking the minimum over
coordinates preserves this bound: if two collections of
values differ elementwise by at most \(2\varepsilon_B\),
their minima differ by at most the same amount. The same
reasoning applies when taking the maximum over \(b\).
Finally, the positive-part operation satisfies
\(|[x]_+-[y]_+|\le |x-y|\). Hence, with prior
probability one,
\begin{equation}
 |\Delta_{\boldsymbol{\theta}}(a)
   -\widetilde{\Delta}_{\boldsymbol{\theta},B}(a)|
 \le 2\varepsilon_B
 \qquad\text{for every }a\in\mathcal A .
 \label{eq:gaussian-clipping-gap-bias}
\end{equation}

\paragraph{Regret and information under the clipped policy.}
Apply \(c_B\) coordinatewise to reward vectors. The learner
observes \(c_B(\mathbf Y_{t,A_t})\), updates the exact posterior
under \(p_1\) using the clipped observation law, and uses the
clipped means for its Pareto sets and gaps. The clipped law for
arm \(a\) is the pushforward
\[
 \widetilde P_{\boldsymbol{\theta},B}^a(D)
 :=P_{\boldsymbol{\theta}}^a(c_B^{-1}(D))
\]
for each measurable reward event \(D\). This kernel includes any
mass introduced at the clipping boundaries. For IDS, the
same kernel defines the per-arm information gain.
Write \(\widetilde{\mathcal H}_{T+1}\) for the history of
actions and clipped observations and
\(\widetilde{\mathrm{IB}}_T
 :=I_{p_1}(\boldsymbol{\theta};\widetilde{\mathcal H}_{T+1})\).

Generate the raw and clipped histories on the same run of the
clipped policy. The raw history
\(\mathcal H_{T+1}^{Y}
 =(A_1,\mathbf Y_{1,A_1},\ldots,A_T,\mathbf Y_{T,A_T})\)
records the original rewards along the chosen actions. Let
\(\mathcal C_B\) clip the reward entries of a history and retain
its action entries. Thus
\(\widetilde{\mathcal H}_{T+1}
=\mathcal C_B(\mathcal H_{T+1}^{Y})\).
The clipped policy is admissible in the raw experiment because
it can apply \(\mathcal C_B\) to each raw history prefix and then
use fresh internal randomness. By data processing,
\begin{equation}
 \widetilde{\mathrm{IB}}_T
 \le I_{p_1}(\boldsymbol{\theta};\mathcal H_{T+1}^{Y}).
 \label{eq:gaussian-clipping-ib-dpi}
\end{equation}
Suppose a finite \(\mathcal I_T\) bounds the raw information
budget under \(p_1\) for every admissible policy. This also gives
finite fixed-action information gains at almost every clipped
history. Indeed, for any round \(t\le T\) and arm \(a\), consider
the policy that uses the original rule through round \(t-1\),
forces \(a\) at round \(t\), and uses any admissible rule thereafter.
Its pre-round history has the same law as under the original policy.
The chain rule bounds the
expected information gain from this forced pull by the modified
policy's clipped-history budget, which is at most \(\mathcal I_T\)
by \eqref{eq:gaussian-clipping-ib-dpi}. The fixed-action gain is
nonnegative, so its finite expectation implies finiteness almost
surely. There are finitely many rounds and arms, so this holds
jointly for all the candidate gains used by IDS.

Let \(M_B\le|\mathcal A|\) be the maximum clipped Pareto-set size
over the prior support. The affine map \(y\mapsto(y+B)/(2B)\)
sends each clipped coordinate to \([0,1]\). Its inverse on the
transformed range is \(z\mapsto2Bz-B\), so the two histories
determine each other and induce the same posterior. Under a fixed
action the reward and its transform also determine each other, so
their parameter-level information gains are equal by data processing
in both directions. The transform applies the same positive affine
change to every arm mean, preserving the clipped Pareto sets and
dividing each gap by \(2B\). Thus for each action distribution the
IDS ratio is multiplied by \((2B)^{-2}\), with the same
zero-information conventions as in Algorithm~\ref{alg:IDS_gap}.
The ratio minimizers are therefore the same. Starting with the
common prior, use the same random draws and tie breaking at
corresponding histories. Induction over the rounds gives the
same actions in the clipped and normalized runs, for either
algorithm.
Let \(r_t\) be the conditional expected gap in the normalized
model and \(J_t\) the conditional information gain of the policy's
action distribution. For TS, Lemma~\ref{lem:ts_gap_information}
gives \(r_t^2\le32|\mathcal A|M_B J_t\). For IDS, the TS
distribution at the same posterior is a feasible comparator with
ratio at most \(32|\mathcal A|M_B\). The IDS ratio is no greater,
so the same inequality holds. If \(J_t=0\), finiteness of the
ratio forces \(r_t=0\) under the zero-information convention.
Since the action uses only the clipped past and fresh randomness,
it contributes zero conditional information about
\(\boldsymbol{\theta}\) given that past. The chain rule then
gives \(\sum_{t=1}^T\mathbb E_{p_1}[J_t]
=\widetilde{\mathrm{IB}}_T\), using the information equivalence
of the clipped and normalized observations. Thus the unit-range
argument of Theorems~\ref{thm:ts_regret_gap} and~\ref{thm:ids_regret_gap},
followed by rescaling, gives
\begin{align*}
 \mathbb E_{p_1}\!\left[\sum_{t=1}^T
         \widetilde{\Delta}_{\boldsymbol{\theta},B}(A_t)\right]
 &=2B\sum_{t=1}^T\mathbb E_{p_1}[r_t]\\
 &\le 2B\sqrt{32|\mathcal A|M_B}
        \sum_{t=1}^T\mathbb E_{p_1}[\sqrt{J_t}]\\
 &\le C B\sqrt{|\mathcal A|M_B T\,\widetilde{\mathrm{IB}}_T},
\end{align*}
where \(C=8\sqrt2\). The last inequality uses Jensen's inequality
and Cauchy--Schwarz over the rounds.

\eqref{eq:gaussian-clipping-gap-bias} holds simultaneously for
all arms, so it applies to each \(A_t\) in this run. Summing,
taking expectations, and applying
\eqref{eq:gaussian-clipping-ib-dpi} gives
\begin{align}
 \mathbb E_{p_1}[R_T]
 &\le
 \mathbb E_{p_1}\!\left[\sum_{t=1}^T
        \widetilde{\Delta}_{\boldsymbol{\theta},B}(A_t)\right]
        +2T\varepsilon_B \notag\\
 &\le C B\sqrt{|\mathcal A|M_B T\,\widetilde{\mathrm{IB}}_T}
        +2T\varepsilon_B \notag\\
 &\le C B\sqrt{|\mathcal A|M_B T\,\mathcal I_T}
        +2T\varepsilon_B .
 \label{eq:gaussian-clipping-regret-transfer}
\end{align}
For horizon \(T\ge1\), take
\(B_T:=\mu_{\max}+2\sigma\sqrt{\log(2T)}\).
Use this level throughout the run. Substituting
\(u_{B_T}=2\sigma\sqrt{\log(2T)}\) gives
\(\exp(-u_{B_T}^2/(2\sigma^2))=(2T)^{-2}\) and
\[
 2T\varepsilon_{B_T}
 =\frac{\sigma}{2T\sqrt{\log(2T)}}.
\]
If \(\mathcal I_T=O(\log(1+T))\), then
\(B_T=O(\sqrt{\log(1+T)})\) and \(M_{B_T}\le|\mathcal A|\)
give \(O(\sqrt T\log(1+T))\) expected cumulative gap regret
for fixed \(|\mathcal A|,d,\mu_{\max},\sigma\) and prior.

\paragraph{A Gaussian prior satisfying both conditions.}
Let \(p_0\) be the unconditioned conjugate prior of
Lemma~\ref{lem:ig_gaussian}. In this model the reward of
arm \(a\) given the parameter has law \(\mathcal N(\mu_a,\Sigma)\),
with \(\Sigma\) shared across arms. We have
\(\Sigma\sim\mathrm{IW}(\Psi_1,\nu_1)\) with
\(\Psi_1\succ0\) and \(\nu_1>d-1\). Given \(\Sigma\),
the means are independent with
\(\mu_a\mid\Sigma\sim\mathcal N(m_{a,1},\Sigma/\kappa_{a,1})\),
where \(\kappa_{a,1}>0\). For fixed
\(\mu_{\max},\sigma>0\), define the parameter event
\[
 E:=\left\{
   \max_{a\in\mathcal A,\,j\in[d]}|\mu_{a,j}|\le\mu_{\max},
   \quad \max_{j\in[d]}\Sigma_{jj}\le\sigma^2
 \right\}.
\]
To prove \(p_0(E)>0\), let \(\Sigma_\circ=(\sigma^2/2)I_d\)
and consider the open ball of symmetric matrices
\[
 \mathcal U
 :=\{\Sigma=\Sigma^\top:
     \|\Sigma-\Sigma_\circ\|_{\mathrm{op}}<\sigma^2/4\}.
\]
For every unit vector \(x\) and every \(\Sigma\in\mathcal U\),
\[
 \bigl|x^\top\Sigma x-\sigma^2/2\bigr|
 \le\|\Sigma-\Sigma_\circ\|_{\mathrm{op}}
 <\sigma^2/4 .
\]
Thus \(x^\top\Sigma x>\sigma^2/4\), so \(\mathcal U\) consists
of positive-definite matrices. Taking \(x\) to be each coordinate
vector also gives
\(\Sigma_{jj}<3\sigma^2/4<\sigma^2\). The proper inverse-Wishart
prior has a positive density on the positive-definite cone, so
\(\mathcal U\) has positive prior probability. For each
\(\Sigma\in\mathcal U\), the conditional joint density of the arm
means is a product of nondegenerate Gaussian densities. It is
positive on \((-\mu_{\max},\mu_{\max})^{d|\mathcal A|}\), an open
box with positive volume. Integrating the resulting positive
conditional probability over \(\Sigma\in\mathcal U\) proves
\(p_0(E)>0\). Moreover, for any prescribed
\(\delta\in(0,1)\), the finite thresholds \(\mu_{\max}\) and
\(\sigma\) can be chosen, independently of \(T\), so that
\(p_0(E)\ge\delta\). Indeed, as both thresholds increase to
infinity, the events \(E\) increase to the full parameter space
up to a \(p_0\)-null set, since the proper prior assigns finite
means and covariance entries almost surely. Continuity of
probability from below gives \(p_0(E)\uparrow1\). Thus the
conditioning factor \(1/p_0(E)\le1/\delta\) can be controlled by
fixed thresholds; their magnitudes still enter the clipping level
and the finite-horizon regret constant.

Take the environment prior and the learner's prior to be
\(p_1=p_0(\,\cdot\mid E)\). Then \(p_1(E)=1\). For a fixed
parameter in \(E\), the reward kernel remains Gaussian with
that parameter's means and covariance. The bound on the means
and the centered Gaussian moment generating function verify
\eqref{eq:gaussian-clipping-uniform-conditions}.

To bound the raw information budget under \(p_1\), fix any
admissible policy. Write \(\pi_t(da_t\mid h_t^Y)\) for its
measurable action kernel, defined on all raw histories by
extending it arbitrarily where a posterior version was
unspecified. Use these same kernels in the experiments with
environment priors \(p_0\) and \(p_1\). In particular, for
clipped TS or IDS the kernel first forms \(\mathcal C_B(h_t^Y)\)
and uses the learner's \(p_1\) posterior in both experiments.
Under either environment prior \(p\), the joint law is
\[
 p(d\boldsymbol{\theta})
 \prod_{t=1}^T
 \pi_t(da_t\mid h_t^Y)\,
 P_{\boldsymbol{\theta}}^{a_t}(d\mathbf y_t).
\]
Here \(h_t^Y=(a_1,\mathbf y_1,\ldots,a_{t-1},\mathbf y_{t-1})\).
Conditioning this expression under \(p=p_0\) on \(E\) multiplies
the prior factor by \(\mathbbm{1}_E(\boldsymbol{\theta})/p_0(E)\).
The action and reward kernels stay the same. The conditional
joint law of the parameter and raw history is therefore the joint
law of the \(p_1\) experiment.

Let \(U=\mathbbm{1}\{\boldsymbol{\theta}\in E\}\). Since \(U\)
is a function of \(\boldsymbol{\theta}\), the chain rule gives
\(I_{p_0}(\boldsymbol{\theta};\mathcal H_{T+1}^Y)
  =I_{p_0}(U,\boldsymbol{\theta};\mathcal H_{T+1}^Y)\).
Expanding the right-hand side in the other order yields
\begin{align}
 I_{p_0}(\boldsymbol{\theta};\mathcal H_{T+1}^{Y})
 &= I_{p_0}(U;\mathcal H_{T+1}^{Y})
  + I_{p_0}(\boldsymbol{\theta};\mathcal H_{T+1}^{Y}\mid U)
  \notag\\
 &\ge p_0(E)\,
   I_{p_0}(\boldsymbol{\theta};\mathcal H_{T+1}^{Y}\mid U=1)
  \notag\\
 &= p_0(E)\,I_{p_1}(\boldsymbol{\theta};\mathcal H_{T+1}^{Y}).
 \label{eq:gaussian-clipping-conditioned-budget}
\end{align}
Indeed, the conditional mutual information given \(U\) is the
average over its two values, and the terms omitted in the
inequality are nonnegative. The final equality uses the
conditional joint law just identified.

The calculation proving Lemma~\ref{lem:ig_gaussian} supplies a
finite constant \(C_{\mathrm{prior}}\), depending only on the
fixed prior and dimension, such that the following deterministic
bound holds under \(p_0\) for every admissible policy and \(T\ge1\).
Set \(\widetilde C_d=d(3d+1)/4\) and
\[
 \mathcal I_T^0:=
 \left(\frac d2|\mathcal A|+\widetilde C_d\right)\log(1+T)
 +C_{\mathrm{prior}}.
\]
Then \(I_{p_0}(\boldsymbol{\theta};\mathcal H_{T+1}^Y)
\le\mathcal I_T^0\). In the proof of
Lemma~\ref{lem:ig_gaussian}, the information about the means
conditional on the covariance is controlled by the play counts
\(N_a(T)\le T\). The information about the covariance is bounded
at each round uniformly over the realized histories and actions.
Consequently the bound applies to all the fixed policy kernels
above, including the policy using \(B=B_T\) for the given horizon.
Dividing \eqref{eq:gaussian-clipping-conditioned-budget} by
\(p_0(E)>0\) gives the uniform bound
\(\mathcal I_T=\mathcal I_T^0/p_0(E)\) under \(p_1\).
Together with \eqref{eq:gaussian-clipping-ib-dpi}, this yields
\begin{equation}
 \widetilde{\mathrm{IB}}_T
 \le I_{p_1}(\boldsymbol{\theta};\mathcal H_{T+1}^{Y})
 \le \frac{\mathcal I_T^0}{p_0(E)}
 =O(\log(1+T)).
 \label{eq:gaussian-clipping-budget-final}
\end{equation}
The event \(E\) and its probability are fixed as \(T\) varies,
and the bound applies to the actual clipped policy for each \(T\).

\paragraph{Sublinear Gaussian gap regret.}
Applying \eqref{eq:gaussian-clipping-regret-transfer} with
\(B=B_T\) and the budget just proved gives
\begin{equation}
 \mathbb E_{p_1}[R_T]
 \le C B_T
      \sqrt{\frac{|\mathcal A|M_{B_T} T\,\mathcal I_T^0}{p_0(E)}}
      +\frac{\sigma}{2T\sqrt{\log(2T)}}.
 \label{eq:gaussian-clipping-regret-final}
\end{equation}
Since \(M_{B_T}\le|\mathcal A|\), the first term of
\eqref{eq:gaussian-clipping-regret-final} is
\(O(\sqrt T\log(1+T))\) at fixed model and prior. One factor
of \(\sqrt{\log(1+T)}\) comes from \(B_T\), and the other comes
from the square root of the information budget. The bias term
tends to zero. Thus the original-mean gap regret of clipped TS
and IDS under \(p_1\) satisfies
\(\mathbb E_{p_1}[R_T]=O(\sqrt T\log(1+T))\). Its per-round
average is \(O(\log(1+T)/\sqrt T)\to0\).

\paragraph{Pareto sets under shared-covariance Gaussian clipping.}
Fix a parameter \(\boldsymbol{\vartheta}\in E\), a coordinate \(j\),
and a level \(B>\mu_{\max}\). Write
\(\sigma_j(\boldsymbol{\vartheta})=\sqrt{\Sigma_{jj}}>0\) and
let \(Z\sim\mathcal N(0,1)\). The function
\[
 g_{j,B}(m)
 :=\mathbb E[c_B(m+\sigma_j(\boldsymbol{\vartheta})Z)]
\]
is the same for every arm at this parameter, since the covariance
is shared. For \(m<m'\), use the same \(Z\) in both arguments.
Monotonicity of \(c_B\) gives
\(c_B(m+\sigma_j(\boldsymbol{\vartheta})Z)
 \le c_B(m'+\sigma_j(\boldsymbol{\vartheta})Z)\) pointwise.
The inequality is strict on
\(\{-B<m+\sigma_j(\boldsymbol{\vartheta})Z<B\}\), since the
larger argument either stays in the clipping interval or maps
to \(B\), and in both cases its clipped value exceeds the first.
This event has positive probability
under the nondegenerate normal law. Taking expectations shows
that \(g_{j,B}\) is strictly increasing.

The clipped mean of arm \(a\) in coordinate \(j\) is
\(g_{j,B}(\mu_{\boldsymbol{\vartheta},j}(a))\). Therefore, for
every pair of arms, the coordinatewise mean comparisons, including
equalities, are preserved. Let
\(\widetilde{\mathcal S}_{B}^*(\boldsymbol{\vartheta})\) be the
Pareto set of the clipped means. Then
\begin{equation}
 \widetilde{\mathcal S}_{B}^*(\boldsymbol{\vartheta})
 =\mathcal S^*(\boldsymbol{\vartheta})
 \qquad (\boldsymbol{\vartheta}\in E).
 \label{eq:gaussian-clipping-same-set}
\end{equation}
For a coordinate \(j\) and threshold \(h\in(-B,B)\), the event
\(c_B(y_j)\le h\) holds exactly when \(y_j\le h\). Hence, with
\(D_{j,h}:=\{\mathbf y:y_j\le h\}\), the clipped law satisfies
\begin{equation}
 \widetilde P_{\boldsymbol{\vartheta},B}^a(D_{j,h})
 =P_{\boldsymbol{\vartheta}}^a(D_{j,h}).
 \label{eq:gaussian-clipping-threshold-event}
\end{equation}
The thresholds in Step~8 of Appendix~\ref{app:ts_gap_set_recovery}
are coordinate means of candidate models in \(E\), so they lie
strictly inside \([-B_T,B_T]\). This preserves the differences
in event probabilities used there to distinguish competing
Pareto sets. Step~8 combines these differences with
\eqref{eq:gaussian-clipping-same-set} and
\eqref{eq:gaussian-clipping-budget-final} to bound set error
on the same clipped TS run as \eqref{eq:gaussian-clipping-regret-final}.

\subsection{Proof of Proposition~\ref{prop:ts_gap_set_recovery}}
\label{app:ts_gap_set_recovery}

Corollary~\ref{cor:ts_set_quality} bounds the aggregate gap of the
sampled Pareto sets. A dominated arm can have zero Pareto gap when a
coordinate is tied. Exact set recovery therefore calls for a comparison
that detects changes in Pareto membership, including such ties.
We compare the reward laws of arms in the union of the true and
sampled sets. This gives the following quantitative version of
Proposition~\ref{prop:ts_gap_set_recovery}.

\paragraph{Detailed statement of Proposition~\ref{prop:ts_gap_set_recovery}.}
Run Algorithm~\ref{alg:ts_pareto_gap} in the Bayesian model with a
nonempty finite action set \(\mathcal A\). Conditional on the true
parameter \(\boldsymbol{\theta}\sim p_1\), the reward arrays are
i.i.d.\ across rounds, and \(P_{\boldsymbol{\vartheta}}^a\) is the
measurable reward-vector kernel for arm \(a\) at parameter
\(\boldsymbol{\vartheta}\). Write \(\mathcal H_t\) for the observed
action--reward history before round \(t\) and
\(p_t=\mathcal L(\boldsymbol{\theta}\mid\mathcal H_t)\) for the
exact posterior. At round \(t\), the algorithm draws
\(\boldsymbol{\theta}_t\sim p_t\), sets
\(S_t=\mathcal S^*(\boldsymbol{\theta}_t)\), and plays an arm
uniformly from \(S_t\). The fresh random seeds used in each
round are independent of the environment and of earlier seeds.

Let \(\Theta_{\mathrm{int}}\) be the set of
\(\boldsymbol{\vartheta}\) for which
\(\mathbb E_{\boldsymbol{\vartheta}}[|Y_{t,a,j}|]<\infty\) for all
\(a\in\mathcal A\) and \(j\in[d]\), where \(Y_{t,a,j}\) is the
\(j\)-th coordinate of \(\mathbf Y_{t,a}\). Assume
\(p_1(\Theta_{\mathrm{int}})=1\). Let \(M\) be a deterministic
bound with \(1\le M\le|\mathcal A|\) and
\(|\mathcal S^*|\le M\) almost surely under \(p_1\), where
\(\mathcal S^*=\mathcal S^*(\boldsymbol{\theta})\).
For \(s\subseteq\mathcal A\), write
\(p_t(s):=\mathbb P(\mathcal S^*=s\mid\mathcal H_t)\) for the
posterior mass of the candidate set \(s\).
Define the mismatch probabilities and information budget by
\[
 e_t:=\mathbb P(S_t\ne\mathcal S^*),
 \qquad
 \bar e_T:=\frac1T\sum_{t=1}^T e_t,
 \qquad
 \mathrm{IB}_T:=I(\boldsymbol{\theta};\mathcal H_{T+1}).
\]
These quantities are under the joint Bayesian law of a single run
and its prefixes. The unconditional error probability \(e_t\) is
nonincreasing in \(t\), so \(e_T\le\bar e_T\) for every \(T\ge1\).

\emph{Finite-horizon recovery bound.}
For a given horizon \(T\), fix a deterministic measurable set
\(\mathcal G\subseteq\Theta_{\mathrm{int}}\) and \(\rho>0\).
Suppose that whenever
\(\boldsymbol{\vartheta},\boldsymbol{\vartheta}'\in\mathcal G\)
have different Pareto sets, some arm
\(a\in\mathcal S^*(\boldsymbol{\vartheta})\cup
        \mathcal S^*(\boldsymbol{\vartheta}')\) satisfies
\(\operatorname{TV}(P_{\boldsymbol{\vartheta}}^a,
                    P_{\boldsymbol{\vartheta}'}^a)\ge\rho\).
Then
\[
 e_T\le \bar e_T
 \le 2p_1(\mathcal G^c)
      +\frac{2}{\rho}\sqrt{\frac{2|\mathcal A|M\,\mathrm{IB}_T}{T}}.
\]

\emph{Consistency.}
If \(\mathrm{IB}_T=o(T)\), then \(\bar e_T\to0\) and
\[
 \mathbb P(S_T=\mathcal S^*)\longrightarrow1,
 \qquad
 p_T(\mathcal S^*)\longrightarrow1 \quad\text{almost surely},
 \qquad
 H(\mathcal S^*\mid\mathcal H_T)\longrightarrow0 .
\]
The almost-sure convergence concerns the posterior mass
\(p_T(\mathcal S^*)\) at the realized true set. The exact-match
probability concerns the sampled set \(S_T\).

\emph{Model-specific recovery rates.}
For the models of Proposition~\ref{prop:ig_bounds}, the following
upper bounds hold with the model and prior fixed.
\begin{itemize}
 \item For a finite parameter space, \(\bar e_T=O(T^{-1/2})\).
 \item For the categorical model of Lemma~\ref{lem:ig_categorical}
 with conditionally independent coordinates and independent
 Dirichlet priors satisfying \(\alpha_{a,j,k}\ge1\),
 \(\bar e_T=O((\log(1+T)/T)^{1/4})\).
 \item For the shared-covariance Gaussian model of
 Lemma~\ref{lem:ig_gaussian}, use its unconditioned conjugate
 prior \(p_0\) and the original Gaussian observations. Then
 \(\bar e_T=O((\log(1+T)/T)^{1/4})\).
\end{itemize}

\emph{Gap regret on the same run.}
If the rewards are uniformly bounded, the Bayesian cumulative
gap regret satisfies
\[
 \mathbb E\!\left[\sum_{t=1}^T
                 \Delta_{\boldsymbol{\theta}}(A_t)\right]
 = O\!\left(\sqrt{|\mathcal A|M T\,\mathrm{IB}_T}\right),
\]
where the implied constant may depend on the reward bound.
At fixed model sizes and reward bound, \(\mathrm{IB}_T=o(T)\)
makes this regret sublinear and gives the consistency conclusions
on the same run.
At fixed model sizes and prior, Proposition~\ref{prop:ig_bounds}
gives \(\mathrm{IB}_T=O(1)\) for the finite-parameter case and
\(\mathrm{IB}_T=O(\log(1+T))\) for the categorical case.
For bounded rewards, these budgets give gap-regret bounds of
\(O(\sqrt T)\) and \(O(\sqrt{T\log(1+T)})\), respectively.
Both budgets satisfy the consistency condition
\(\mathrm{IB}_T=o(T)\), so each bound holds on the same run
as the corresponding recovery guarantee.

\emph{Gaussian gap regret and recovery on the same clipped run.}
Use the conditioned prior \(p_1=p_0(\,\cdot\mid E)\) and the
clipping level \(B_T\) from Appendix~\ref{app:gaussian_high_probability}.
For each horizon \(T\), run clipped
Algorithm~\ref{alg:ts_pareto_gap} with its exact posterior and
that level throughout the run. Denote the sampled set at
round \(t\le T\) by \(\widetilde S_t^{(T)}\), and let
\(\widetilde e_t^{(T)}
 :=\mathbb P_{p_1}(\widetilde S_t^{(T)}\ne\mathcal S^*)\).
Here \(\mathcal S^*\) is the Pareto set of the original Gaussian
means. The clipped history has information budget
\(\widetilde{\mathrm{IB}}_T=O(\log(1+T))\). For fixed model and
prior, the same run satisfies
\begin{align*}
 \widetilde e_T^{(T)}
 \le \frac1T\sum_{t=1}^T\widetilde e_t^{(T)}
 &=O\!\left(\left(\frac{\log(1+T)}{T}\right)^{1/4}\right),\\
 \mathbb E_{p_1}[R_T]&=O(\sqrt T\log(1+T)),
\end{align*}
where \(R_T=\sum_{t=1}^T\Delta_{\boldsymbol{\theta}}(A_t)\)
uses the original Gaussian means. The round-\(T\) posterior
mass of the true set also converges to one in probability
for these horizon-dependent runs.
In the clipped clause of Proposition~\ref{prop:ts_gap_set_recovery},
\(e_T\) and \(\bar e_T\) refer to \(\widetilde e_T^{(T)}\) and
\(T^{-1}\sum_{t=1}^T\widetilde e_t^{(T)}\), respectively.

\begin{proof}
Let \(\mathcal F_t:=\sigma(\mathcal H_t)\) and write
\(\mathbb{E}_t[\cdot]:=\mathbb{E}[\cdot\mid\mathcal F_t]\).
Recall that \(P_{\boldsymbol{\vartheta}}^a\) denotes the
reward-vector law under parameter \(\boldsymbol{\vartheta}\) and
action \(a\). As in Lemma~\ref{lem:info_gain_tv_theta}, define
\[
 \operatorname{TV}(P,Q):=\sup_E|P(E)-Q(E)|,
\]
where \(E\) ranges over measurable reward events.
We consider a single run of Algorithm~\ref{alg:ts_pareto_gap}
and its prefixes. In particular, \(\mathcal H_1\) is empty and
\(\mathcal H_{t+1}=(\mathcal H_t,A_t,\mathbf Y_{t,A_t})\).

For each \(a,j\), the absolute first moment is the increasing
limit of the measurable integrals of \(\min\{|y_j|,n\}\)
against \(P_{\boldsymbol{\vartheta}}^a\). Thus
\(\Theta_{\mathrm{int}}\) is measurable. The integrals of the
positive and negative parts of \(y_j\) are measurable by the
same argument, and their difference gives a measurable
coordinate mean on \(\Theta_{\mathrm{int}}\). Since Pareto
membership is determined by finitely many comparisons of these
means, the set map \(\mathcal S^*(\cdot)\) is measurable there.
Extend the map by a fixed singleton on the prior-null
complement of \(\Theta_{\mathrm{int}}\).

\paragraph{Step 1. Posterior sampling probabilities and the information budget.}
By definition, \(\boldsymbol{\theta}\) has law \(p_t\)
conditional on \(\mathcal F_t\). Given that history, the sampling
rule draws \(\boldsymbol{\theta}_t\) from \(p_t\) using randomness
independent of the true parameter and the environment's rewards.
Thus for measurable parameter sets \(D,E\), conditioning first on
\((\mathcal F_t,\boldsymbol{\theta})\) gives
\[
 \mathbb{P}(\boldsymbol{\theta}\in D,\boldsymbol{\theta}_t\in E
             \mid\mathcal F_t)
 = \mathbb{E}_t[\mathbbm{1}\{\boldsymbol{\theta}\in D\}\,p_t(E)]
 = p_t(D)p_t(E).
\]
In particular, the two parameters are conditionally independent
with the same law, and for each \(a\in\mathcal A\),
\[
 q_t(a):=\mathbb{P}(a\in S_t\mid\mathcal H_t)
        =\mathbb{P}(a\in\mathcal S^*\mid\mathcal H_t).
\]
The events \(\boldsymbol{\theta}\in\Theta_{\mathrm{int}}\) and
\(|\mathcal S^*|\le M\) have prior probability one. By the tower
property they have conditional probability one at almost every
history. The equality in conditional law gives
\(\boldsymbol{\theta}_t\in\Theta_{\mathrm{int}}\) and
\(|S_t|\le M\) almost surely as well. Each Pareto set is nonempty
because the dominance order in a finite nonempty action set has a
maximal element.
Define \(\pi_t(a):=\mathbb P(A_t=a\mid\mathcal F_t)\).
Given \((\mathcal H_t,S_t)\), the algorithm plays \(a\) with
probability \(\mathbbm{1}\{a\in S_t\}/|S_t|\).
Taking conditional expectation over \(S_t\) gives
\begin{equation}
 \frac{q_t(a)}{M}
 \le \pi_t(a)
 =\mathbb{E}_t\!\left[\frac{\mathbbm{1}\{a\in S_t\}}{|S_t|}\right]
 \le q_t(a).
 \label{eq:ts-gap-recovery-play-prob}
\end{equation}
The inequalities follow from \(1\le |S_t|\le M\). In particular,
\(\pi_t(a)=0\) if and only if \(q_t(a)=0\).

At a realized history, let
\(\overline P_t^a:=\int P_{\boldsymbol{\vartheta}}^a
\,p_t(d\boldsymbol{\vartheta})\) be the posterior predictive
reward law for arm \(a\). The fixed-action mutual information
in Lemma~\ref{lem:info_gain_tv_theta} is
\[
 I_t(a):=\int
 D_{\mathrm{KL}}(P_{\boldsymbol{\vartheta}}^a\Vert
                 \overline P_t^a)\,p_t(d\boldsymbol{\vartheta}).
\]
It is computed from the current posterior and the reward kernel of
arm \(a\). The information gain of the policy at this history is
\(I_t:=\sum_{a:\pi_t(a)>0}\pi_t(a)I_t(a)\).
Since \(A_t\) is generated from \(\mathcal H_t\) and fresh policy
randomness, it is conditionally independent of
\(\boldsymbol{\theta}\) given \(\mathcal H_t\). The information
chain rule, starting from the empty history \(\mathcal H_1\),
and conditioning on the realized history and action give
\begin{align}
 \mathrm{IB}_T
 &=\sum_{t=1}^T I(\boldsymbol{\theta};A_t,\mathbf Y_{t,A_t}
                      \mid\mathcal H_t) \notag\\
 &=\sum_{t=1}^T I(\boldsymbol{\theta};\mathbf Y_{t,A_t}
                      \mid\mathcal H_t,A_t)
 =\sum_{t=1}^T\mathbb{E}[I_t].
 \label{eq:ts-gap-recovery-budget}
\end{align}
For the last equality, fix a history at which \(\pi_t(a)>0\).
Conditional on this history and \(A_t=a\), the parameter law is
still \(p_t\) because the action uses fresh randomization. Given
the parameter, the current reward array is independent of the past
and of that randomization. Thus the joint law of the parameter and
the selected reward is
\(p_t(d\boldsymbol{\vartheta})
  P_{\boldsymbol{\vartheta}}^a(d\mathbf y)\), which is exactly the
forced-play law defining \(I_t(a)\). Averaging over \(A_t\) gives
\(I_t\) at that history. Averaging over histories gives
\(I(\boldsymbol{\theta};\mathbf Y_{t,A_t}\mid\mathcal H_t,A_t)
=\mathbb{E}[I_t]\).

\paragraph{Step 2. A change of Pareto set gives a changed comparison within the union.}
Consider parameters
\(\boldsymbol{\vartheta},\boldsymbol{\vartheta}'\in\Theta_{\mathrm{int}}\)
whose Pareto sets differ. By exchanging their roles if needed,
choose
\[
 a\in\mathcal S^*(\boldsymbol{\vartheta})
        \setminus\mathcal S^*(\boldsymbol{\vartheta}').
\]
Under \(\boldsymbol{\vartheta}'\), the arm \(a\) is dominated.
Starting from \(a\), repeatedly choose a dominating arm under
\(\boldsymbol{\vartheta}'\). Each new mean vector is at least
as large in every coordinate and strictly larger in at least
one coordinate. By transitivity of these comparisons, each
later arm dominates every earlier arm, so all arms on the
chain are distinct. Since \(\mathcal A\) is finite, the chain
ends at an undominated arm
\(b\in\mathcal S^*(\boldsymbol{\vartheta}')\) that dominates
\(a\) under \(\boldsymbol{\vartheta}'\).

For each coordinate \(j\), define the mean-reward difference
\[
 D_{a,b,j}(\boldsymbol{\xi})
 :=\mu_{\boldsymbol{\xi},j}(b)-\mu_{\boldsymbol{\xi},j}(a).
\]
Because \(b\) dominates \(a\) under
\(\boldsymbol{\vartheta}'\), these differences are nonnegative
for all coordinates and strictly positive for at least one.
Since \(a\) is Pareto-optimal under
\(\boldsymbol{\vartheta}\), either every difference is zero
or at least one is negative; otherwise, \(b\) would dominate
\(a\) there as well. In the first case, choose a coordinate
\(j\) with \(D_{a,b,j}(\boldsymbol{\vartheta}')>0\).
In the second case, choose a coordinate \(j\) with
\(D_{a,b,j}(\boldsymbol{\vartheta})<0\).
For the selected coordinate, we have
\(D_{a,b,j}(\boldsymbol{\vartheta})\le0\le
D_{a,b,j}(\boldsymbol{\vartheta}')\), with at least
one strict inequality. Consequently,
\[
 |D_{a,b,j}(\boldsymbol{\vartheta})
    -D_{a,b,j}(\boldsymbol{\vartheta}')|
 = |D_{a,b,j}(\boldsymbol{\vartheta})|
   +|D_{a,b,j}(\boldsymbol{\vartheta}')|.
\]
If at least one of these two differences has magnitude at least
\(r>0\), this sum is at least \(r\). Expanding the two
mean-reward differences and applying the triangle inequality
bounds the left-hand side by
\[
 |\mu_{\boldsymbol{\vartheta},j}(a)
   -\mu_{\boldsymbol{\vartheta}',j}(a)|
 +|\mu_{\boldsymbol{\vartheta},j}(b)
   -\mu_{\boldsymbol{\vartheta}',j}(b)|.
\]
Consequently, at least one arm \(u\in\{a,b\}\) satisfies
\begin{equation}
 |\mu_{\boldsymbol{\vartheta},j}(u)
    -\mu_{\boldsymbol{\vartheta}',j}(u)|\ge r/2,
 \qquad
 u\in\mathcal S^*(\boldsymbol{\vartheta})
       \cup\mathcal S^*(\boldsymbol{\vartheta}').
 \label{eq:ts-gap-recovery-witness}
\end{equation}
This includes comparisons that change from a tie to a strict
inequality. At least one of the two differences is nonzero, so
one can choose \(r>0\) for this pair. Since equal reward laws
have equal coordinate means, \eqref{eq:ts-gap-recovery-witness}
gives an arm in the union whose reward laws differ.

We will use the following bound for a measurable function \(f\)
with values in \([L,U]\) and probability laws \(P,Q\).
Since \(f(y)-L=\int_L^U\mathbbm{1}\{f(y)>v\}\,dv\),
integrating under each law and applying Fubini gives the first
inequality below. The definition of total variation gives the second.
\begin{align}
 |\mathbb{E}_P[f]-\mathbb{E}_Q[f]|
 &\le\int_L^U |P(f>v)-Q(f>v)|\,dv \notag\\
 &\le (U-L)\operatorname{TV}(P,Q).
 \label{eq:ts-gap-recovery-bounded-test}
\end{align}

\paragraph{Step 3. Bound the mismatch probability on a set of parameters.}
We work at a horizon with finite \(\mathrm{IB}_T\), so \(I_t\)
is finite almost surely for \(t\le T\) by
\eqref{eq:ts-gap-recovery-budget}. Then \(I_t(a)\) is finite
whenever \(\pi_t(a)>0\), since \(0\le\pi_t(a)I_t(a)\le I_t\).
When \(\mathrm{IB}_T=\infty\), the bound holds in the extended
sense. Fix \(\mathcal G\) and \(\rho\) as in the detailed statement.
For each pair in \(\mathcal G\) with different Pareto sets, the assumption
provides an arm \(u\) in their union satisfying
\begin{equation}
 \operatorname{TV}(P_{\boldsymbol{\vartheta}}^u,
                    P_{\boldsymbol{\vartheta}'}^u)\ge\rho .
 \label{eq:ts-gap-recovery-good-set}
\end{equation}
Write \(\eta_{\mathcal G}:=p_1(\mathcal G^c)\), and for the
true and sampled parameters at round \(t\) define
\[
 V_t(a):=\operatorname{TV}(P_{\boldsymbol{\theta}}^a,
                           P_{\boldsymbol{\theta}_t}^a).
\]
These are measurable random variables. In fact, on \(\mathbb R^d\)
one can take the supremum defining total variation over the
countable algebra generated by rational boxes. For any fixed \(P,Q\),
every Borel event can be approximated in \((P+Q)\)-measure by
this algebra. The class of approximable events contains the algebra
and is closed under complements and countable unions. For a countable
union, first approximate the union by a finite subunion using
finiteness of \(P+Q\), then approximate its finitely many terms.
The approximable events thus form a sigma algebra containing the
rational-box algebra, which generates the Borel sets. For events
\(E,F\),
\(\bigl|(P(E)-Q(E))-(P(F)-Q(F))\bigr|
 \le (P+Q)(E\mathbin{\triangle}F)\).
Thus the supremum over the countable algebra gives the same total
variation. Each term in that supremum is measurable by
measurability of the kernels. Almost surely,
\begin{align}
 \mathbbm{1}\{S_t\ne\mathcal S^*\}
 &\le
 \mathbbm{1}\{\boldsymbol{\theta}\notin\mathcal G\}
 +\mathbbm{1}\{\boldsymbol{\theta}_t\notin\mathcal G\}
 +\frac{1}{\rho}\sum_{a\in\mathcal S^*\cup S_t}V_t(a).
 \label{eq:ts-gap-recovery-pathwise}
\end{align}
Indeed, if either parameter is outside \(\mathcal G\), the
corresponding indicator is one. If both are in \(\mathcal G\) and
the sets differ, \eqref{eq:ts-gap-recovery-good-set} makes the
sum at least \(\rho\). All terms are nonnegative when the sets agree.

The pair \((\boldsymbol{\theta},\boldsymbol{\theta}_t)\) is
conditionally exchangeable by Step~1. Interchanging the parameters
interchanges \(\mathcal S^*\) and \(S_t\), and each \(V_t(a)\)
is symmetric in the two parameters. Thus
\begin{align}
 \mathbb{E}_t\!\left[\sum_{a\in\mathcal S^*\cup S_t}V_t(a)\right]
 &\le
 \sum_{a\in\mathcal A}\mathbb{E}_t\!\left[
   \bigl(\mathbbm{1}\{a\in\mathcal S^*\}
        +\mathbbm{1}\{a\in S_t\}\bigr)V_t(a)\right] \notag\\
 &=2\sum_{a\in\mathcal A}\mathbb{E}_t[
                  \mathbbm{1}\{a\in S_t\}V_t(a)] \notag\\
 &\le 2\sum_{a:\pi_t(a)>0}
             \sqrt{q_t(a)\,\mathbb{E}_t[V_t(a)^2]} \notag\\
 &\le 2\sqrt{2}\sum_{a:\pi_t(a)>0}\sqrt{q_t(a) I_t(a)} .
 \label{eq:ts-gap-recovery-exchange}
\end{align}
The first inequality bounds the indicator of the union by the
sum of the two indicators. In the third line we used
Cauchy--Schwarz and the identity
\(\mathbb{E}_t[\mathbbm{1}\{a\in S_t\}^2]=q_t(a)\).
The last line uses \eqref{eq:info_gain_tv_lower_bound_param}
from Lemma~\ref{lem:info_gain_tv_theta}, which gives
\(\mathbb{E}_t[V_t(a)^2]\le2 I_t(a)\) because the conditional
joint law of the two parameters is \(p_t\otimes p_t\).
Arms with \(\pi_t(a)=0\)
contribute zero because \(q_t(a)=0\) by
\eqref{eq:ts-gap-recovery-play-prob}.

For each term with \(\pi_t(a)>0\), write
\(\sqrt{q_t(a)I_t(a)}
=\sqrt{q_t(a)/\pi_t(a)}\,\sqrt{\pi_t(a)I_t(a)}\).
A second application of Cauchy--Schwarz gives
\begin{align}
 \sum_{a:\pi_t(a)>0}\sqrt{q_t(a)I_t(a)}
 &\le
 \sqrt{\sum_{a:\pi_t(a)>0}\frac{q_t(a)}{\pi_t(a)}}
 \sqrt{\sum_{a:\pi_t(a)>0}\pi_t(a)I_t(a)}
 \le\sqrt{|\mathcal A| M I_t},
 \label{eq:ts-gap-recovery-action-cs}
\end{align}
where the last inequality uses \(q_t(a)/\pi_t(a)\le M\).

Taking unconditional expectations in
\eqref{eq:ts-gap-recovery-pathwise}, the first indicator has mean
\(\eta_{\mathcal G}\). The second indicator has the same mean,
since
\[
 \mathbb{P}(\boldsymbol{\theta}_t\notin\mathcal G)
 =\mathbb{E}\!\left[\mathbb{P}
        (\boldsymbol{\theta}_t\notin\mathcal G\mid\mathcal H_t)\right]
 =\mathbb{E}\!\left[\mathbb{P}
        (\boldsymbol{\theta}\notin\mathcal G\mid\mathcal H_t)\right]
 =\eta_{\mathcal G}.
\]
Consequently, \eqref{eq:ts-gap-recovery-exchange} and
\eqref{eq:ts-gap-recovery-action-cs} yield
\(e_t\le 2\eta_{\mathcal G}
 +(2/\rho)\sqrt{2|\mathcal A|M}\,\mathbb{E}[\sqrt{I_t}]\).
Using Jensen's inequality, Cauchy--Schwarz over rounds, and
\eqref{eq:ts-gap-recovery-budget}, respectively, gives
\[
 \frac{1}{T}\sum_{t=1}^T\mathbb{E}[\sqrt{I_t}]
 \le \frac{1}{T}\sum_{t=1}^T\sqrt{\mathbb{E}[I_t]}
 \le \sqrt{\frac{1}{T}\sum_{t=1}^T\mathbb{E}[I_t]}
 =\sqrt{\frac{\mathrm{IB}_T}{T}}.
\]
Since \(\bar e_T=T^{-1}\sum_{t=1}^T e_t\), we obtain
\begin{equation}
 \bar e_T
 \le
 2\eta_{\mathcal G}
 +\frac{2}{\rho}\sqrt{\frac{2|\mathcal A|M\,\mathrm{IB}_T}{T}}.
 \label{eq:ts-gap-recovery-master}
\end{equation}

\paragraph{Step 4. Pass from the average error to the last-round error.}
The true set \(\mathcal S^*\) is determined by the environment,
and its conditional law given \(\mathcal F_t\) is \(p_t(\cdot)\).
Step~1 shows that \(S_t\) has the same conditional law and is
conditionally independent of \(\mathcal S^*\). Thus
\[
 \mathbb P(S_t=\mathcal S^*\mid\mathcal F_t)
 =\sum_{s\subseteq\mathcal A}
   \mathbb P(S_t=s,\mathcal S^*=s\mid\mathcal F_t)
 =\sum_{s\subseteq\mathcal A}p_t(s)^2 .
\]
The posterior mass at the realized true set can be written as
\[
 p_t(\mathcal S^*)
 =\sum_{s\subseteq\mathcal A}\mathbbm{1}\{\mathcal S^*=s\}\,p_t(s).
\]
For each fixed \(s\), \(p_t(s)\) is \(\mathcal F_t\)-measurable
and \(\mathbb E_t[\mathbbm{1}\{\mathcal S^*=s\}]=p_t(s)\).
Hence the conditional expectation of this last sum is also
\(\sum_{s\subseteq\mathcal A}p_t(s)^2\). Taking expectations of
both conditional identities gives
\begin{equation}
 1-e_t=\mathbb E\!\left[\sum_{s\subseteq\mathcal A}p_t(s)^2\right]
      =\mathbb E[p_t(\mathcal S^*)].
 \label{eq:ts-gap-recovery-error-posterior}
\end{equation}

For each fixed \(s\), the histories generate nested sigma fields
\(\mathcal F_t\subseteq\mathcal F_{t+1}\). By the tower property,
\[
 \mathbb E_t[p_{t+1}(s)]
 =\mathbb E_t\!\left[
      \mathbb E[\mathbbm{1}\{\mathcal S^*=s\}\mid\mathcal F_{t+1}]
   \right]
 =p_t(s).
\]
Thus \((p_t(s))_{t\ge1}\) is a martingale bounded in \([0,1]\).
Conditional Jensen gives
\(\mathbb E_t[p_{t+1}(s)^2]\ge p_t(s)^2\).
Taking expectations and summing over \(s\) shows that
\(\mathbb E[\sum_s p_t(s)^2]\) is nondecreasing in \(t\).
By \eqref{eq:ts-gap-recovery-error-posterior},
\(e_{t+1}\le e_t\), so \(e_T\le\bar e_T\).

\paragraph{Step 5. Consistency under a sublinear information budget.}
Fix a target error \(\delta\in(0,1)\) and put \(\eta=\delta/4\).
For a parameter \(\boldsymbol{\vartheta}\) with finite conditional
reward means, let
\[
 m(\boldsymbol{\vartheta})
 :=
 \min_{\substack{a,b\in\mathcal A,\ a\ne b,\ j\in[d]\\
          D_{a,b,j}(\boldsymbol{\vartheta})\ne0}}
 |D_{a,b,j}(\boldsymbol{\vartheta})|.
\]
The minimum of an empty collection is \(+\infty\). Equivalently,
assign \(+\infty\) to each zero mean difference before taking the
minimum over all pairs and coordinates. This gives a measurable
function, since the mean differences are measurable and the minimum is
over a finite collection. Every value of \(m\) is positive,
including \(+\infty\) when all mean differences are zero.
The events \(\{m(\boldsymbol{\theta})<1/n\}\), \(n\ge1\),
decrease to a null event because \(m(\boldsymbol{\theta})>0\)
almost surely. Continuity of probability from above yields
\(\mathbb{P}(m(\boldsymbol{\theta})<\epsilon)\to0\) as
\(\epsilon\downarrow0\).
Choose a fixed \(\epsilon>0\) so that this prior probability is
at most \(\eta/2\).

For \(B>0\), set \(f_B(y):=\max\{-B,\min\{B,y\}\}\) and define
\[
 \mu^{(B)}_{\boldsymbol{\vartheta},j}(a)
 :=\mathbb{E}_{\boldsymbol{\vartheta}}[f_B(Y_{t,a,j})],
 \qquad
 L_B(\boldsymbol{\vartheta})
 :=\max_{a\in\mathcal A,\,j\in[d]}
 \bigl|\mu^{(B)}_{\boldsymbol{\vartheta},j}(a)
          -\mu_{\boldsymbol{\vartheta},j}(a)\bigr|.
\]
For each arm and coordinate,
\[
 \bigl|\mu^{(B)}_{\boldsymbol{\vartheta},j}(a)
          -\mu_{\boldsymbol{\vartheta},j}(a)\bigr|
 \le
 \mathbb{E}_{\boldsymbol{\vartheta}}\!\left[
     |Y_{t,a,j}|\mathbbm{1}\{|Y_{t,a,j}|>B\}\right].
\]
This follows from \(|f_B(y)-y|\le |y|\mathbbm{1}\{|y|>B\}\).
For each \(\boldsymbol{\vartheta}\in\Theta_{\mathrm{int}}\),
the right-hand side tends to zero as \(B\to\infty\) by dominated
convergence under \(P_{\boldsymbol{\vartheta}}^a\), using
\(|Y_{t,a,j}|\) as the integrable dominating variable.
The maximum is over finitely many terms, so
\(L_B(\boldsymbol{\vartheta})\to0\). The clipped means are
measurable integrals against the reward kernels, so \(L_B\)
is measurable. Since \(p_1(\Theta_{\mathrm{int}})=1\), bounded
convergence applied to the indicator of
\(\{L_B(\boldsymbol{\theta})>\epsilon/8\}\) gives
\[
 \mathbb P\bigl(L_B(\boldsymbol{\theta})>\epsilon/8\bigr)
 \longrightarrow0 \qquad\text{as }B\to\infty.
\]
Choose a fixed \(B\ge\max\{1,\epsilon\}\) so that this probability
is at most \(\eta/2\). Define
\[
 \mathcal G:=
 \left\{\boldsymbol{\vartheta}\in\Theta_{\mathrm{int}}:
   m(\boldsymbol{\vartheta})\ge\epsilon,\quad
   L_B(\boldsymbol{\vartheta})\le\epsilon/8\right\}.
\]
Since \(p_1(\Theta_{\mathrm{int}})=1\), the union bound gives
\(\eta_{\mathcal G}\le\eta/2+\eta/2=\eta\).

For two parameters in \(\mathcal G\) with different Pareto sets,
Step~2 identifies a comparison for which at least one of the two
mean differences is nonzero. By the definition of \(\mathcal G\), that
mean difference has magnitude at least \(\epsilon\). It follows from
\eqref{eq:ts-gap-recovery-witness} that there is an arm \(u\)
in the union whose coordinate mean differs by at least
\(\epsilon/2\) between the two parameters. Its two means after
applying \(f_B\) therefore differ by at least
\(\epsilon/2-2(\epsilon/8)=\epsilon/4\), by the triangle
inequality. Applying \eqref{eq:ts-gap-recovery-bounded-test}
with \(f(\mathbf y)=f_B(y_j)\) gives
\[
 \operatorname{TV}(P_{\boldsymbol{\vartheta}}^u,
                     P_{\boldsymbol{\vartheta}'}^u)
 \ge \frac{\epsilon}{8B}.
\]
Thus \eqref{eq:ts-gap-recovery-master} holds with
\(\rho_\eta:=\epsilon/(8B)>0\). The function \(f_B\) is a
bounded test of the original observation law, and the algorithm
uses the original observations. Both \(\epsilon\) and \(B\)
were chosen for the fixed \(\eta\), so \(\rho_\eta\) is fixed
as \(T\) grows. Steps~3--4 now give
\[
 0\le e_T\le\bar e_T\le
 \frac{\delta}{2}+
 \frac{2}{\rho_\eta}\sqrt{\frac{2|\mathcal A|M\,\mathrm{IB}_T}{T}} .
\]
If \(\mathrm{IB}_T=o(T)\), then there is a \(T_\delta\) such
that \(\mathrm{IB}_T/T\le
\delta^2\rho_\eta^2/(32|\mathcal A|M)\) for all \(T\ge T_\delta\).
At those horizons the second term is at most \(\delta/2\),
so \(e_T\le\bar e_T\le\delta\).
The horizon \(T_\delta\) may grow as \(\rho_\eta\) decreases for
smaller target errors. Since the argument applies to every
\(\delta\in(0,1)\), both \(e_T\) and \(\bar e_T\) converge to zero.

\paragraph{Step 6. Rates for the models in Proposition~\ref{prop:ig_bounds}.}
The exceptional-set probability in
\eqref{eq:ts-gap-recovery-master} is evaluated under \(p_1\).
This uses the fact from Step~3 that the true parameter and the
posterior draw have the same marginal law after averaging over
the history.

\medskip\noindent\emph{Finite parameter space.}
Let \(\Theta_0:=\{\boldsymbol{\vartheta}\in\Theta:
p_1(\{\boldsymbol{\vartheta}\})>0\}\). By
\(p_1(\Theta_{\mathrm{int}})=1\), each point of \(\Theta_0\)
belongs to \(\Theta_{\mathrm{int}}\). If all parameters in
\(\Theta_0\) have the same Pareto set, the true and sampled
sets equal it almost surely at every round, giving \(e_t=0\).
Otherwise, for any two parameters with different Pareto sets,
Step~2 gives an arm in their union whose mean differs between
the parameters. Equal reward laws have equal means, so the
total-variation distance of its reward laws is positive. Define
\[
 \rho_{\mathrm{fin}}
 :=
 \min_{\substack{\boldsymbol{\vartheta},\boldsymbol{\vartheta}'
                   \in\Theta_0,\ a\in\mathcal A\\
       \operatorname{TV}(P_{\boldsymbol{\vartheta}}^a,
                       P_{\boldsymbol{\vartheta}'}^a)>0}}
       \operatorname{TV}(P_{\boldsymbol{\vartheta}}^a,
                          P_{\boldsymbol{\vartheta}'}^a).
\]
This minimum is over a finite nonempty collection of positive
numbers, so \(\rho_{\mathrm{fin}}>0\).
Taking \(\mathcal G=\Theta_0\) gives
\(\eta_{\mathcal G}=0\) and \(\rho=\rho_{\mathrm{fin}}\).
By \eqref{eq:ts-gap-recovery-master} and
Lemma~\ref{lem:ig_finite_parameter},
\[
 e_T\le\bar e_T
 \le \frac{2}{\rho_{\mathrm{fin}}}
     \sqrt{\frac{2|\mathcal A|M\log|\Theta|}{T}} .
\]
This is \(O(T^{-1/2})\) for a fixed model. The argument uses
positive distances between distinct reward laws and permits ties
between coordinate means.

\medskip\noindent\emph{Categorical model.}
Use the reward levels \([K]\) and the conditionally independent
coordinate laws in Lemma~\ref{lem:ig_categorical}. If \(K=1\),
all mean reward vectors coincide and the Pareto set is always
\(\mathcal A\). Assume \(K\ge2\), and for
\(0<\epsilon\le1\) put a parameter in
\(\mathcal G_{\epsilon}^{\mathrm{cat}}\) whenever every nonzero
mean difference \(D_{a,b,j}(\boldsymbol{\vartheta})\) has magnitude at
least \(\epsilon\).

To bound the prior mass of its complement, fix \(a,j\).
Write \(x_1,\ldots,x_{K-1}\) for the first \(K-1\) entries of
the vector \(p_{a,j}\), and
\(x_K=1-\sum_{k=1}^{K-1}x_k\). On the simplex, the Dirichlet
density in the first \(K-1\) entries is a finite constant times
\(\prod_{k=1}^{K}x_k^{\alpha_{a,j,k}-1}\).
Since \(\alpha_{a,j,k}\ge1\) and each \(x_k\in[0,1]\), this
density is bounded by a finite constant \(C_{a,j}\).
The coordinate mean is
\[
 \mu_{\boldsymbol{\theta},j}(a)
   =\sum_{k=1}^{K} k x_k
   =K+\sum_{k=1}^{K-1}(k-K)x_k .
\]
For fixed \(x_2,\ldots,x_{K-1}\), this mean is an affine
function of \(x_1\) with slope \(1-K\). Therefore, for any
\(z\in\mathbb R\), the condition
\(|\mu_{\boldsymbol{\theta},j}(a)-z|<\epsilon\) restricts
\(x_1\) to an interval of length at most \(2\epsilon/(K-1)\).
The other entries range in a subset of \([0,1]^{K-2}\), which
has volume at most one. For \(K=2\) there are no other entries
to integrate. Integrating the density over these slices gives
\[
 \mathbb{P}\bigl(|\mu_{\boldsymbol{\theta},j}(a)-z|<\epsilon\bigr)
 \le \frac{2C_{a,j}}{K-1}\,\epsilon .
\]
For \(b\ne a\), the two coordinate means are independent under
the prior, because their categorical probability vectors have
independent Dirichlet priors. Conditioning on
\(\mu_{\boldsymbol{\theta},j}(b)\) and using the preceding
bound, uniformly in \(z\), gives the same bound for
\(\mathbb{P}(|D_{a,b,j}(\boldsymbol{\theta})|<\epsilon)\).
The complement of \(\mathcal G_\epsilon^{\mathrm{cat}}\) is the
union of the events \(0<|D_{a,b,j}(\boldsymbol{\theta})|<\epsilon\)
over unordered pairs \(\{a,b\}\) and coordinates \(j\). Each event
has probability at most \(2\epsilon\max_{a,j}C_{a,j}/(K-1)\).
There are \(d\binom{|\mathcal A|}{2}\) such events, so the union
bound yields
\[
 p_1\bigl((\mathcal G_\epsilon^{\mathrm{cat}})^c\bigr)
 \le C_{\mathrm{cat}}\epsilon,
 \qquad
 C_{\mathrm{cat}}
 :=\frac{2d\binom{|\mathcal A|}{2}}{K-1}
          \max_{a,j}C_{a,j}.
\]

For two parameters in \(\mathcal G_\epsilon^{\mathrm{cat}}\)
with different sets, \eqref{eq:ts-gap-recovery-witness} gives a
coordinate mean difference of at least \(\epsilon/2\) on an arm
in their union. The coordinate reward has range \([1,K]\), so
\eqref{eq:ts-gap-recovery-bounded-test} gives a total-variation
distance at least \(\epsilon/[2(K-1)]\).
Substituting this value of \(\rho\) and the bound on
\(\eta_{\mathcal G}\) into \eqref{eq:ts-gap-recovery-master}
gives
\[
 e_T\le\bar e_T\le
 2C_{\mathrm{cat}}\epsilon
 +\frac{4(K-1)}{\epsilon}
       \sqrt{\frac{2|\mathcal A|M\,\mathrm{IB}_T}{T}} .
\]
For fixed \(|\mathcal A|,d,K\) and prior hyperparameters,
Lemma~\ref{lem:ig_categorical} gives
\(\mathrm{IB}_T\le C_{\mathrm{IB,cat}}\log(1+T)\) for some
finite constant \(C_{\mathrm{IB,cat}}\) and all \(T\ge1\).
Take \(\epsilon_T=(\log(1+T)/T)^{1/4}\), which lies in
\((0,1]\) because \(0<\log(1+T)\le T\). Since
\(\epsilon_T^2=\sqrt{\log(1+T)/T}\), substitution gives
\[
 \bar e_T\le
 \left(2C_{\mathrm{cat}}
       +4(K-1)\sqrt{2|\mathcal A|M C_{\mathrm{IB,cat}}}\right)
 \epsilon_T.
\]
This proves the categorical rate. Independent Bernoulli coordinates
with independent \(\mathrm{Beta}(\alpha_{a,j,2},\alpha_{a,j,1})\)
priors on their success probabilities give the two-category model
after adding one to each reward coordinate. The stated rate applies
when both shape parameters are at least one. The bijection sends
success to category \(2\). It preserves the likelihood as a
function of the parameter and preserves all coordinate mean
differences. The posterior and Pareto sets therefore agree at
corresponding histories. Using the same policy randomization
gives the same actions in the two models, and the bijection of
observed histories preserves the information budget.

\medskip\noindent\emph{Shared-covariance Gaussian model with raw observations.}
Write \(p_0\) for the unconditioned conjugate prior of
Lemma~\ref{lem:ig_gaussian}, as in
Appendix~\ref{app:gaussian_high_probability}. For this raw run,
take the prior in Steps~1--5 to be \(p_0\) and use the original
Gaussian observations.
In particular, \(\Sigma\sim\mathrm{IW}(\Psi_1,\nu_1)\) is
proper with \(\Psi_1\) positive definite and \(\nu_1>d-1\).
Given \(\Sigma\) the means are independent with
\(\mu_a\mid\Sigma\sim\mathcal N(m_{a,1},\Sigma/\kappa_{a,1})\)
for \(\kappa_{a,1}>0\). For a parameter
\(\boldsymbol{\vartheta}\), write
\(\sigma_j(\boldsymbol{\vartheta}):=\sqrt{\Sigma_{jj}}\) for
its shared conditional standard deviation in coordinate \(j\).
It is positive and finite under the proper prior. Define the
standardized mean difference
\[
 Z_{a,b,j}(\boldsymbol{\vartheta})
 :=\frac{D_{a,b,j}(\boldsymbol{\vartheta})}
         {\sigma_j(\boldsymbol{\vartheta})}.
\]
For \(0<\epsilon\le1\), let \(\mathcal G_\epsilon^{\mathrm{G}}\)
consist of the parameters for which each nonzero standardized
mean difference has magnitude at least \(\epsilon\).

Fix \(a\ne b\) and \(j\). Conditional on \(\Sigma\),
the two arm means in coordinate \(j\) are independent normals.
Their difference is normal with mean
\((m_{b,1}-m_{a,1})_j\) and variance
\(\Sigma_{jj}(\kappa_{a,1}^{-1}+\kappa_{b,1}^{-1})\).
Dividing by \(\sqrt{\Sigma_{jj}}\) gives
\[
 Z_{a,b,j}(\boldsymbol{\theta})\mid\Sigma
 \sim\mathcal N\!\left(
    \frac{(m_{b,1}-m_{a,1})_j}{\sqrt{\Sigma_{jj}}},
    v_{a,b}\right),
 \qquad
 v_{a,b}:=\kappa_{a,1}^{-1}+\kappa_{b,1}^{-1}>0.
\]
The normal density is bounded by \((2\pi v_{a,b})^{-1/2}\),
independently of \(\Sigma\) and of the conditional mean.
Integrating over \((-\epsilon,\epsilon)\) and averaging over
\(\Sigma\) gives
\[
 p_0\bigl(|Z_{a,b,j}(\boldsymbol{\theta})|<\epsilon\bigr)
 \le \frac{2\epsilon}{\sqrt{2\pi v_{a,b}}}.
\]
A union bound over arm pairs and coordinates gives
\begin{equation}
 p_0\bigl((\mathcal G_\epsilon^{\mathrm{G}})^c\bigr)
 \le C_{\mathrm{G}}\epsilon,
 \qquad
 C_{\mathrm{G}}
 :=2d\sum_{\{a,b\}\subseteq\mathcal A,\ a\ne b}
       (2\pi v_{a,b})^{-1/2}.
 \label{eq:ts-gap-recovery-gaussian-prior}
\end{equation}
where the sum is over unordered arm pairs. The constant is
finite because each \(v_{a,b}>0\) and the sum is finite.

Now consider two parameters in \(\mathcal G_\epsilon^{\mathrm{G}}\)
with different Pareto sets. Step~2 gives arms \(a,b\) and a
coordinate \(j\) for which the mean differences lie on opposite
sides of zero (allowing one to equal zero), with at least one nonzero. Denote one parameter with a nonzero
mean difference by \(\boldsymbol{\xi}\) and the other by
\(\boldsymbol{\xi}'\). By membership in
\(\mathcal G_\epsilon^{\mathrm{G}}\),
\(|D_{a,b,j}(\boldsymbol{\xi})|
 \ge\epsilon\sigma_j(\boldsymbol{\xi})\).
Both arms have the same reward standard deviation in this
coordinate under \(\boldsymbol{\xi}\) because its covariance
is shared across arms. Applying
\eqref{eq:ts-gap-recovery-witness} with
\(r=\epsilon\sigma_j(\boldsymbol{\xi})\) gives an arm \(u\) in
the union satisfying
\[
 |\mu_{\boldsymbol{\xi},j}(u)-\mu_{\boldsymbol{\xi}',j}(u)|
 \ge \frac{\epsilon}{2}\sigma_j(\boldsymbol{\xi}).
\]
The two parameters may have different covariance matrices.
Use the event
\(D:=\{\mathbf y:y_j\le\mu_{\boldsymbol{\xi}',j}(u)\}\).
Its probability under \(\boldsymbol{\xi}'\) is exactly \(1/2\)
because the threshold is the mean of that nondegenerate normal
coordinate. Under \(\boldsymbol{\xi}\),
\[
 P_{\boldsymbol{\xi}}^u(D)
 =\Phi\!\left(
   \frac{\mu_{\boldsymbol{\xi}',j}(u)
         -\mu_{\boldsymbol{\xi},j}(u)}
        {\sigma_j(\boldsymbol{\xi})}\right),
\]
where \(\Phi\) is the standard normal cumulative distribution
function. Denote its argument by \(z\). Then \(|z|\ge\epsilon/2\),
and symmetry gives \(|\Phi(z)-1/2|=\Phi(|z|)-1/2\).
By monotonicity, the absolute difference between the two event
probabilities is at least \(\Phi(\epsilon/2)-1/2\). If \(\phi\) is the
standard normal density, then
\[
 \Phi(\epsilon/2)-\frac12
 =\int_0^{\epsilon/2}\phi(x)\,dx
 \ge \frac{\epsilon}{2}\,\phi(1/2),
\]
where we used \(0<\epsilon\le1\) and the fact that
\(\phi\) decreases on \([0,\infty)\). In particular, with
\(c_0:=\phi(1/2)/2>0\), the event \(D\) satisfies
\begin{equation}
 |P_{\boldsymbol{\xi}}^u(D)-P_{\boldsymbol{\xi}'}^u(D)|
 \ge c_0\epsilon.
 \label{eq:ts-gap-recovery-gaussian-witness}
\end{equation}
The total-variation distance is at least this difference, so
\eqref{eq:ts-gap-recovery-good-set} holds with \(\rho=c_0\epsilon\).

Substituting this value into
\eqref{eq:ts-gap-recovery-master} gives
\[
 e_T\le\bar e_T\le
 2C_{\mathrm{G}}\epsilon+
 \frac{2}{c_0\epsilon}
    \sqrt{\frac{2|\mathcal A|M\,\mathrm{IB}_T}{T}} .
\]
With the model and prior fixed, Lemma~\ref{lem:ig_gaussian}
gives \(\mathrm{IB}_T\le C_{\mathrm{IB,G}}\log(1+T)\) for
some finite \(C_{\mathrm{IB,G}}\) and all \(T\ge1\).
Substituting \(\epsilon=\epsilon_T=(\log(1+T)/T)^{1/4}\)
and using \(\sqrt{\log(1+T)/T}=\epsilon_T^2\) gives
\[
 \bar e_T\le
 \left(2C_{\mathrm{G}}
       +\frac{2}{c_0}\sqrt{2|\mathcal A|M C_{\mathrm{IB,G}}}\right)
 \epsilon_T.
\]
The recovery bound uses the original Gaussian reward laws.
The conditional density bound is uniform in \(\Sigma\), so
the argument applies to the proper inverse-Wishart priors
in Lemma~\ref{lem:ig_gaussian}, including when the prior
mean of \(\operatorname{tr}(\Sigma)\) is infinite.

\paragraph{Step 7. Almost-sure posterior concentration and set entropy.}
Under \(\mathrm{IB}_T=o(T)\), Step~5 gives \(e_t\to0\).
By \eqref{eq:ts-gap-recovery-error-posterior}, the mean
posterior mass of the true set therefore satisfies
\[
 \mathbb E[p_t(\mathcal S^*)]=1-e_t\longrightarrow1.
\]
For every fixed \(s\subseteq\mathcal A\), Step~4 shows that
\(p_t(s)\) is a bounded martingale. By the martingale convergence
theorem, there is a limit \(p_\infty(s)\in[0,1]\) such that
\(p_t(s)\to p_\infty(s)\) almost surely. The intersection of
these events over the finitely many sets \(s\) has probability
one. On that event, the finite sum for \(p_t(\mathcal S^*)\)
in Step~4 converges term by term to
\[
 p_\infty(\mathcal S^*)
 :=\sum_{s\subseteq\mathcal A}
       \mathbbm{1}\{\mathcal S^*=s\}\,p_\infty(s).
\]
This gives an almost-sure limit for the mass at the realized
true set, whichever set the environment induces.

Since \(0\le p_t(\mathcal S^*)\le1\), bounded convergence gives
\[
 \mathbb E[p_\infty(\mathcal S^*)]
 =\lim_{t\to\infty}\mathbb E[p_t(\mathcal S^*)]=1.
\]
Exactly one indicator in the definition of
\(p_\infty(\mathcal S^*)\) equals one, so
\(1-p_\infty(\mathcal S^*)\) is nonnegative. Its expectation
is zero, which implies that it is zero almost surely. Thus
\(p_t(\mathcal S^*)\to1\) almost surely.

Let \(h_t\) be the entropy of the posterior at the realized
history,
\[
 h_t:=-\sum_{s\subseteq\mathcal A}p_t(s)\log p_t(s),
 \qquad 0\log0=0.
\]
On the event of almost-sure concentration, each \(s\ne\mathcal S^*\)
satisfies \(0\le p_t(s)\le1-p_t(\mathcal S^*)\to0\), while
\(p_t(\mathcal S^*)\to1\). The function \(x\mapsto-x\log x\)
is continuous on \([0,1]\) with value zero at both endpoints.
Each term of the finite entropy sum therefore tends to zero,
so \(h_t\to0\) almost surely. Since there are at most
\(2^{|\mathcal A|}\) candidate sets, the entropy is bounded by
\(0\le h_t\le|\mathcal A|\log2\). The conditional entropy is
the expected posterior entropy, so bounded convergence gives
\[
 H(\mathcal S^*\mid\mathcal H_t)
 =\mathbb E[h_t]\longrightarrow0.
\]
Since \(\mathcal S^*\) is finite-valued, the entropy identity
for mutual information then gives
\[
 I(\mathcal S^*;\mathcal H_t)
 =H(\mathcal S^*)-H(\mathcal S^*\mid\mathcal H_t)
 \longrightarrow H(\mathcal S^*).
\]

The set \(\mathcal S^*\) is a deterministic function of
\(\boldsymbol{\theta}\), so
\(I(\mathcal S^*;\mathcal H_{T+1}\mid\boldsymbol{\theta})=0\).
Applying the chain rule in the two orders gives
\begin{align*}
 \mathrm{IB}_T
 &=I((\boldsymbol{\theta},\mathcal S^*);\mathcal H_{T+1})\\
 &=I(\mathcal S^*;\mathcal H_{T+1})
   + I(\boldsymbol{\theta};\mathcal H_{T+1}\mid\mathcal S^*).
\end{align*}
The first term tends to \(H(\mathcal S^*)\) as \(T\to\infty\),
so the history resolves the prior uncertainty about Pareto
membership. The second term measures what the history reveals
about the model conditional on the Pareto set.

\paragraph{Step 8. Gap regret and recovery on the same run.}
\emph{Uniformly bounded rewards.}
For a common bounded reward interval \([L,U]\), subtracting \(L\)
gives rewards in \([0,U-L]\) and preserves all
coordinate-wise mean differences.
This translation is invertible, so the translated and original
histories give the same posterior and information budget. Under
the same randomization, the preserved Pareto sets give the same
actions. Thus Theorem~\ref{thm:ts_regret_gap} uses the same posterior,
actions, and information budget as the recovery proof.
Its proof uses \(M\) through the almost-sure bound
\(|S_t|\le M\) at each round, which Step~1 verifies for the
deterministic \(M\) in the statement. The information budget in
that theorem is therefore exactly \(\mathrm{IB}_T\) in
\eqref{eq:ts-gap-recovery-budget}.
Dividing the regret bound by \(T\) gives
\(O(\sqrt{|\mathcal A|M\,\mathrm{IB}_T/T})\). At fixed model sizes
and reward bound, this tends to zero when \(\mathrm{IB}_T=o(T)\),
which is also the condition used in Steps~5 and~7 for set recovery
and almost-sure posterior concentration.

Substitute the budgets from Proposition~\ref{prop:ig_bounds}
into the regret bound \(O(\sqrt{|\mathcal A|M T\,\mathrm{IB}_T})\).
For bounded rewards on a finite parameter space,
\(\mathrm{IB}_T\le\log|\Theta|\) gives \(O(\sqrt T)\) at fixed
model size. For the categorical model, the budget is
\(O(\log(1+T))\) at fixed model sizes and prior, giving
\(O(\sqrt{T\log(1+T)})\).
These two budget bounds also verify \(\mathrm{IB}_T=o(T)\)
for Steps~5 and~7. Step~6 uses the same budgets to obtain the
corresponding set-recovery rates.

\medskip\noindent\emph{Gaussian rewards with clipping.}
Fix the horizon \(T\). Use the conditioned prior
\(p_1=p_0(\,\cdot\mid E)\) and the clipped observation kernels
\(\widetilde P_{\boldsymbol{\vartheta},B_T}^a\) of
Appendix~\ref{app:gaussian_high_probability}. 
The thresholds defining $E$ are fixed independently of $T$
and may be chosen so that $p_0(E)\ge\delta_E$ for any
prescribed $\delta_E\in(0,1)$. The clipped TS policy
is Algorithm~\ref{alg:ts_pareto_gap} for these kernels, with
the exact posterior given the clipped history. Its observations
are bounded by \(B_T\) within this run, so the conditional moment
assumption holds. Since \(p_1(E)=1\),
the true parameter and every posterior draw belong to \(E\)
almost surely. By \eqref{eq:gaussian-clipping-same-set},
their clipped and original Pareto sets coincide. In particular,
the true set for the clipped model is \(\mathcal S^*\), and the
bound \(M\) on its size applies to the clipped sets as well.

To apply the finite-horizon recovery bound to this model, take
\(\mathcal G=E\cap\mathcal G_\epsilon^{\mathrm{G}}\) for
\(0<\epsilon\le1\). By \eqref{eq:ts-gap-recovery-gaussian-prior},
\[
 p_1(\mathcal G^c)
 =\frac{p_0(E\cap(\mathcal G_\epsilon^{\mathrm{G}})^c)}{p_0(E)}
 \le\frac{C_{\mathrm{G}}\epsilon}{p_0(E)}.
\]
Two parameters in \(\mathcal G\) with different clipped Pareto
sets have different original Pareto sets by
\eqref{eq:gaussian-clipping-same-set}. Step~6 therefore gives
an arm \(u\) in their union and a coordinate \(j\) with the
witness event \(D=\{\mathbf y:y_j\le h\}\) from
\eqref{eq:ts-gap-recovery-gaussian-witness}, where
\(h=\mu_{\boldsymbol{\xi}',j}(u)\). Since
\(|h|\le\mu_{\max}<B_T\), \eqref{eq:gaussian-clipping-threshold-event}
gives the same probability of \(D\) before and after clipping,
under each of the two parameters. It follows that
\[
 \operatorname{TV}(\widetilde P_{\boldsymbol{\xi},B_T}^u,
                    \widetilde P_{\boldsymbol{\xi}',B_T}^u)
 \ge |P_{\boldsymbol{\xi}}^u(D)-P_{\boldsymbol{\xi}'}^u(D)|
 \ge c_0\epsilon .
\]
Thus the recovery bound uses \(\rho=c_0\epsilon\) for the
clipped observation laws, with a constant independent of \(B_T\).

Apply Steps~3 and~4 to the prefixes of this fixed horizon-\(T\)
run. Substituting \eqref{eq:gaussian-clipping-budget-final} into
\eqref{eq:ts-gap-recovery-master} yields
\[
 \widetilde e_T^{(T)}
 \le \frac1T\sum_{t=1}^T\widetilde e_t^{(T)}
 \le \frac{2C_{\mathrm{G}}\epsilon}{p_0(E)}
      +\frac{2}{c_0\epsilon}
       \sqrt{\frac{2|\mathcal A|M\,\mathcal I_T^0}{T p_0(E)}} .
\]
The budget here is for the clipped history observed by the learner.
At fixed model and prior, take \(C_{\mathrm{IB},0}<\infty\) such that
\(\mathcal I_T^0\le C_{\mathrm{IB},0}\log(1+T)\) for all \(T\ge1\).
Choose \(\epsilon_T=(\log(1+T)/T)^{1/4}\in(0,1]\). Since
\(\sqrt{\log(1+T)/T}=\epsilon_T^2\), substitution gives
\[
 \widetilde e_T^{(T)}
 \le \frac1T\sum_{t=1}^T\widetilde e_t^{(T)}
 \le \left(
       \frac{2C_{\mathrm G}}{p_0(E)}
       +\frac{2}{c_0}
        \sqrt{\frac{2|\mathcal A|M C_{\mathrm{IB},0}}{p_0(E)}}
     \right)\epsilon_T .
\]
The last-round inequality compares prefixes within each run,
allowing the level \(B_T\) to vary across horizons.

Write \(\widetilde p_T^{(T)}(\mathcal S^*)\) for the posterior
mass assigned to the realized true set before round \(T\).
By \eqref{eq:ts-gap-recovery-error-posterior} applied to the
clipped history,
\(\mathbb E_{p_1}[1-\widetilde p_T^{(T)}(\mathcal S^*)]
=\widetilde e_T^{(T)}\). For each fixed \(\delta>0\), Markov's
inequality gives
\[
 \mathbb P_{p_1}\!\left(
   1-\widetilde p_T^{(T)}(\mathcal S^*)>\delta\right)
 \le \frac{\widetilde e_T^{(T)}}{\delta}\longrightarrow0.
\]

Finally, \eqref{eq:gaussian-clipping-regret-final} uses these
same actions, prior, and clipping level. It gives
\(\mathbb E_{p_1}[R_T]=O(\sqrt T\log(1+T))=o(T)\) for the
original-mean gaps. This proves recovery of the original Pareto
set and sublinear gap regret on the same clipped TS run.
\end{proof}

\subsection{A fixed mixture of Pareto Thompson Sampling and gap-based IDS}
\label{app:pareto-ts-ids-mixture}

We show that a fixed mixture of Pareto Thompson Sampling
(Algorithm~\ref{alg:ts_pareto_gap}) and exact gap-based IDS
(Algorithm~\ref{alg:IDS_gap}) controls gap regret and recovers the
entire Pareto-optimal set. Both policies satisfy a common information-ratio
bound at the same history. A positive Thompson Sampling probability
supplies the sampling coverage used in the recovery proof.

Let $\mathcal A$ be a nonempty finite action set, $A:=|\mathcal A|$, and
$M:=\operatorname*{ess\,sup}_{\vartheta\sim p_1}|S^*(\vartheta)|\le A$.
Conditional on $\theta\sim p_1$, the reward arrays are i.i.d. across
rounds. Let $P_\vartheta^a$ be the measurable reward-vector kernel
of arm $a$ at parameter $\vartheta$,
$H_t=(A_s,Y_{s,A_s})_{s<t}$ the observed history, and
$p_t=\mathcal L(\theta\mid H_t)$ the exact posterior.
Assume $p_1(\Theta_{\mathrm{int}})=1$, where $\Theta_{\mathrm{int}}$
contains the parameters having finite conditional absolute first moments
in every reward coordinate of every arm. The means, Pareto sets, and gaps
are then well defined. We further assume that the posterior expected
gaps below are finite whenever IDS is computed. A sufficient condition
is $\mathbb E_{p_1}[\max_{a,j}|\mu_{\theta,j}(a)|]<\infty$.
At each realized history $H_t$, let
$\bar P_t^a=\int P_\vartheta^a\,p_t(d\vartheta)$ be the posterior
predictive reward law and put
\begin{equation}
  \Delta_t(a)=\mathbb E[\Delta_\theta(a)\mid H_t],\qquad
  I_t(a)=\int D_{\mathrm{KL}}(P_\vartheta^a\Vert\bar P_t^a)
                  \,p_t(d\vartheta).
  \label{eq:mix-arm-quantities}
\end{equation}
This $I_t(a)$ is the fixed-action information gain and is defined
even if the executed policy assigns zero probability to arm $a$.
Whenever that probability is positive it equals
$I(\theta;Y_{t,a}\mid H_t,A_t=a)$. For $\pi\in\Pi(\mathcal A)$, define
\begin{equation}
 r_t(\pi)=\mathbb E_{a\sim\pi}[\Delta_t(a)],\quad
 J_t(\pi)=\sum_{a:\pi(a)>0}\pi(a)I_t(a),\quad
 \psi_t(\pi)=\frac{r_t(\pi)^2}{J_t(\pi)},
 \label{eq:mix-ratio}
\end{equation}
using $0/0=0$, $x/0=+\infty$ for $x>0$, and $x/(+\infty)=0$
for finite $x\ge0$. Fix $\lambda\in(0,1]$. Define the TS
distribution and, for $\lambda<1$, the exact IDS distribution by
\begin{equation}
 \pi_t^{\mathrm{TS}}(a)
  :=\mathbb E_{\vartheta\sim p_t}\left[
     \frac{\mathbf 1\{a\in S^*(\vartheta)\}}{|S^*(\vartheta)|}\right],
 \qquad
 \pi_t^{\mathrm{IDS}}\in\operatorname*{arg\,min}_{\pi\in\Pi(\mathcal A)}
 \psi_t(\pi),
 \label{eq:mix-constituents}
\end{equation}
The executed action distribution is
\begin{equation}
 \pi_t^{\mathrm{mix}}=
 \begin{cases}
 \lambda\pi_t^{\mathrm{TS}}+(1-\lambda)\pi_t^{\mathrm{IDS}},
    &0<\lambda<1,\\
 \pi_t^{\mathrm{TS}},&\lambda=1.
 \end{cases}
 \label{eq:mix-action-dist}
\end{equation}
For $k\in\{\mathrm{TS},\mathrm{IDS},\mathrm{mix}\}$, write
$r_t^k=r_t(\pi_t^k)$ and $J_t^k=J_t(\pi_t^k)$ for each
defined distribution at the current mixed history.
Fix an ordering of the arms for the IDS minimization.
If some $I_t(a)=+\infty$, the point mass on the first such arm has
ratio zero and is an exact minimizer. Otherwise the ratio is lower
semicontinuous on the compact simplex. It is continuous where
$J_t>0$, diverges on approach to a point with $J_t=0<r_t$, and
has its smallest possible value at $r_t=J_t=0$. A minimizer can be
computed by the two-point search in Appendix~\ref{optm_info_ratio}.
Evaluate the endpoints and the admissible interior candidates
described in Appendix~\ref{app:ids_estimation_section4} and take the
first minimizing candidate in a fixed order. This gives a measurable
choice of the exact IDS distribution from the history.

All sampling uses fresh randomness independent of the environment.
The set $S_t$ is drawn from the pre-action posterior on every round.
It determines the set from which TS samples its arm on TS rounds.
On IDS rounds it serves as the set estimate for the recovery guarantee.

For the recovery guarantee, a measurable good set
$\mathcal G\subseteq\Theta_{\mathrm{int}}$ and $\rho>0$ satisfy the
\emph{union-of-sets witness condition} when
\begin{equation}
 \begin{gathered}
 S^*(\vartheta)\ne S^*(\vartheta'),\quad
 \vartheta,\vartheta'\in\mathcal G
 \quad\Longrightarrow\quad
 \exists u\in S^*(\vartheta)\cup S^*(\vartheta'):\ \\
 \operatorname{TV}(P_\vartheta^u,P_{\vartheta'}^u)\ge\rho.
 \end{gathered}
 \label{eq:mix-witness}
\end{equation}

\begin{algorithm}[h]
\small
\DontPrintSemicolon
\caption{Fixed mixture of Pareto TS and exact gap-based IDS}
\label{alg:pareto-ts-ids-mixture}
\KwIn{Prior $p_1$ over $\theta$, fixed $\lambda\in(0,1]$, horizon $T$}
\For{$t=1,\ldots,T$}{
  Form the exact posterior $p_t=\mathcal L(\theta\mid H_t)$\;
  Draw $\theta_t\sim p_t$ using fresh randomness and set
  $S_t=S^*(\theta_t)$\;
  Independently draw $B_t\sim\operatorname{Bernoulli}(\lambda)$\;
  \eIf{$B_t=1$}{
    Draw $A_t\sim\operatorname{Unif}(S_t)$ \tcp*{Pareto TS}
  }{
    Compute $\{\Delta_t(a),I_t(a)\}_{a\in\mathcal A}$ and the exact
    minimizer $\pi_t^{\mathrm{IDS}}$ of \eqref{eq:mix-ratio}\;
    Draw $A_t\sim\pi_t^{\mathrm{IDS}}$ \tcp*{Exact gap IDS}
  }
  Observe $Y_{t,A_t}$ and update the posterior
  $p_{t+1}=\mathcal L(\theta\mid H_{t+1})$ using the observation kernel\;
}
\end{algorithm}

\begin{theorem}[Joint gap regret and exact Pareto-set recovery for the mixture]
\label{thm:ts_ids_mixture_joint}
Run Algorithm~\ref{alg:pareto-ts-ids-mixture} with a fixed
$\lambda\in(0,1]$ and exact posterior updates. Define
\begin{equation}
 \mathrm{IB}_T:=I(\theta;H_{T+1}),\quad
 e_t:=\Pr(S_t\ne S^*(\theta)),\quad
 \bar e_T:=\frac1T\sum_{t=1}^T e_t.
 \label{eq:mix-error-def}
\end{equation}
\textup{(i) Gap regret.} If all reward coordinates take values in a
common deterministic interval of width $L<\infty$, then
\begin{equation}
 R_T^{\mathrm{gap}}
 :=\mathbb E\Bigl[\sum_{t=1}^T\Delta_\theta(A_t)\Bigr]
 \le\sqrt{32}\,L\sqrt{AMT\,\mathrm{IB}_T}.
 \label{eq:mix-gap-theorem}
\end{equation}
The prefactor is independent of $\lambda$, and $\mathrm{IB}_T$
is evaluated along the executed mixed policy.

\textup{(ii) Exact-set recovery.} Without any reward-boundedness
assumption, for every $\mathcal G,\rho$ satisfying
\eqref{eq:mix-witness},
\begin{equation}
 e_T\le\bar e_T\le 2p_1(\mathcal G^c)
 +\frac{2}{\rho}\sqrt{\frac{2AM\,\mathrm{IB}_T}{\lambda T}}.
 \label{eq:mix-recovery-theorem}
\end{equation}
For each $\eta>0$ there exist $\mathcal G_\eta$ and
$\rho_\eta>0$ satisfying \eqref{eq:mix-witness} and
$p_1(\mathcal G_\eta^c)\le\eta$. Consequently, if
$\mathrm{IB}_T=o(T)$ along the prefixes of a single run of a fixed
observation model and mixed policy, then $e_T,\bar e_T\to0$, and the
posterior mass $p_t(S^*(\theta))$ converges to one almost surely.

\textup{(iii) Model-specific guarantees.} Fix the underlying reward
model and prior as $T$ varies. The following are simultaneous guarantees
on the \emph{same mixed run}. The constants hidden in $O(\cdot)$
depend only on the fixed model and prior.
For the Gaussian row, use the horizon-$T$ clipped experiment with
$B_T=\mu_{\max}+2\sigma\sqrt{\log(2T)}$ and the conditioned prior
$p_1=p_0(\cdot\mid E)$ of
Appendix~\ref{app:gaussian_high_probability}. Regret and membership
are measured using the \emph{original} Gaussian means. For the
categorical row, use the model of Lemma~\ref{lem:ig_categorical} with
$K\ge2$, conditionally independent reward coordinates given
$(\theta,A_t,H_t)$, and independent Dirichlet priors across
arm--objective pairs with $\alpha_{a,j,k}\ge1$. When $K=1$, every
arm is Pareto optimal and both regret and set error vanish.

\par\medskip\noindent
\begin{minipage}{\linewidth}
\centering\normalfont\footnotesize
\refstepcounter{table}\label{tab:mix-joint-guarantees}
\setlength{\tabcolsep}{3pt}
\renewcommand{\arraystretch}{1.24}
\begin{tabular}{p{0.23\linewidth}p{0.36\linewidth}p{0.32\linewidth}}
\hline
Model & Bayesian gap regret & Exact-set error $e_T$ \\
\hline
\raggedright Finite prior, bounded rewards
 & $\sqrt{32}\,L\sqrt{AMT H(\theta)}$
 & $\displaystyle\frac{2}{\rho_{\rm fin}}
    \sqrt{\frac{2AMH(\theta)}{\lambda T}}$ \\
\raggedright Product categorical, independent Dirichlet priors
 & $\sqrt{32}\,LA\sqrt{Md(K+1)T(1+\log T)}$
 & $\displaystyle O\!\left(
  \left[\frac{\log(1+T)}{\lambda T}\right]^{1/4}\right)$ \\
\raggedright Conditioned Gaussian, $B_T$ clipping
 & $O(\sqrt T\log(1+T))$ (original means)
 & $\displaystyle O\!\left(
  \left[\frac{\log(1+T)}{\lambda T}\right]^{1/4}\right)$ \\
\hline
\end{tabular}
\par\smallskip
\textit{Table~\thetable: Model-specific bounds forming part of
Theorem~\thetheorem.}
\end{minipage}
\par\medskip
The set-error bounds also apply to $\bar e_T$.
In the finite-prior row $\rho_{\rm fin}>0$ is the minimum positive TV
distance between the laws of a common arm under two supported
parameters. If all supported parameters induce the same Pareto set,
$e_t=0$ for every $t$. The Gaussian recovery rate compares prefixes
within each horizon-$T$ clipped run, with $B_T$ fixed throughout that run.
\end{theorem}

\begin{proof}
We verify the general inequalities, last-round statement,
consistency, and every model-specific table entry in turn.

\paragraph{Step 1: Posterior sampling, conditional independence, and the
information chain rule.}
Fix an arbitrary realized history $H_t=h$ and write $p_t$ for the
posterior there. The true parameter $\theta$ has law $p_t$, whereas
the auxiliary $\theta_t$ is generated using independent fresh randomness
from the same $p_t$. Thus, for measurable $D,D'$, explicitly,
\begin{equation}
 \Pr(\theta\in D,\theta_t\in D'\mid H_t)
 =p_t(D)p_t(D').
 \label{eq:mix-post-indep}
\end{equation}
The action is generated from $(H_t,\theta_t,B_t)$ and fresh independent
randomness. Thus $A_t\perp\!\!\!\perp\theta\mid H_t$. For any
$\pi$ selected from $H_t$ and fresh randomness, the conditional
joint law of the true parameter and the reward upon playing $a$
is $p_t(d\vartheta)P_\vartheta^a(dy)$. Its mutual information
is precisely $I_t(a)$. More explicitly, the chain rule yields
\begin{align}
 I(\theta;H_{T+1})
 &=\sum_{t=1}^T I(\theta;A_t,Y_{t,A_t}\mid H_t)\\
 &=\sum_{t=1}^T I(\theta;Y_{t,A_t}\mid H_t,A_t)\\
 &=\sum_{t=1}^T\mathbb E\left[
     \sum_{a:\pi_t^{\mathrm{mix}}(a)>0}
     \pi_t^{\mathrm{mix}}(a)I_t(a)\right]
 =\sum_{t=1}^T\mathbb E[J_t^{\mathrm{mix}}].
 \label{eq:mix-chain-rule}
\end{align}
The first equality expands the entire adaptive history, the second
uses $I(\theta;A_t\mid H_t)=0$, and the third conditions on the
selected action.

\paragraph{Step 2: A common pointwise information-ratio constant.}
For this step assume rewards lie in a common interval of width $L$.
If $L=0$, all gaps and regret are zero, so assume $L>0$. Translation
and division by $L$ take every coordinate to $[0,1]$, preserve the
Pareto sets, and scale every gap by $1/L$. Their invertibility preserves
mutual information and TV. We recall the witness argument behind
Lemmas~\ref{lem:theta_stability} and~\ref{lem:ts_gap_information}
to identify the constant used here.

Draw independent $\theta,\vartheta\sim p_t$, put
$S=S^*(\theta)$, $S'=S^*(\vartheta)$, and let
$\widetilde A\sim\operatorname{Unif}(S')$. If
$\Delta_\theta(a)>0$ for $a\in S'$, select an arm $u$ achieving
$\Delta_\theta(a)$. Among the arms weakly dominating $u$ under
$\theta$, choose a Pareto-maximal arm $b$. Any arm dominating $b$
would also weakly dominate $u$, so $b\in S$. Then
$\mu_\theta(b)\succeq\mu_\theta(a)+\Delta_\theta(a)\mathbf1$.
The Pareto optimality of $a$ under $\vartheta$ gives a coordinate
$j$ with
$\mu_{\vartheta,j}(b)\le\mu_{\vartheta,j}(a)$.
Consequently,
\begin{equation}
 \Delta_\theta(a)
 \le |\mu_{\theta,j}(b)-\mu_{\vartheta,j}(b)|
   +|\mu_{\theta,j}(a)-\mu_{\vartheta,j}(a)|.
 \label{eq:mix-witness-gap}
\end{equation}
If $\Delta_\theta(a)=0$, the bound holds with any $b\in S$ and $j$.
Set $V_t(u)=\operatorname{TV}(P_\theta^u,P_\vartheta^u)$. Since a
reward coordinate has range width $L$,
$|\mu_{\theta,j}(u)-\mu_{\vartheta,j}(u)|
\le L V_t(u)$. Using the conservative constant $2L$ and
averaging \eqref{eq:mix-witness-gap} over $a\in S'$ gives
\[
 \mathbb E[\Delta_\theta(\widetilde A)\mid\theta,\vartheta,H_t]
 \le 2L\left(\sum_{a\in S'} V_t(a)+\sum_{b\in S} V_t(b)\right).
\]
Indeed, $|S'|\ge1$, and each witness term $V_t(b)$ is bounded by
the sum over $S$. Taking the conditional expectation given $H_t$
and exchanging $\theta,\vartheta$ gives
\begin{equation}
 r_t^{\mathrm{TS}}
 \le4L\sum_{a\in\mathcal A}
   \mathbb E[\mathbf1\{a\in S'\}V_t(a)\mid H_t].
 \label{eq:mix-ts-tv}
\end{equation}
Let $q_t(a):=\Pr(a\in S'\mid H_t)$. Conditional Cauchy--Schwarz
and Lemma~\ref{lem:info_gain_tv_theta} give, for each $a$ with
$q_t(a)>0$,
\begin{align}
 \mathbb E[\mathbf1\{a\in S'\}V_t(a)\mid H_t]
 &\le\sqrt{q_t(a)\mathbb E[V_t(a)^2\mid H_t]}\\
 &\le\sqrt{2q_t(a)I_t(a)}.
 \label{eq:mix-ts-arm-tv}
\end{align}
Moreover, the definition of the TS distribution and $1\le|S'|\le M$
show that
\begin{equation}
 \pi_t^{\mathrm{TS}}(a)
 =\mathbb E[\mathbf1\{a\in S'\}/|S'|\mid H_t]
 \ge q_t(a)/M.
 \label{eq:mix-ts-coverage}
\end{equation}
Arms with $q_t(a)=0$ contribute zero to \eqref{eq:mix-ts-tv}.
For all $q_t(a)>0$, $q_t(a)/\pi_t^{\mathrm{TS}}(a)\le M$.
By Cauchy--Schwarz across these arms,
\begin{align}
 \sum_{a:q_t(a)>0}\sqrt{q_t(a)I_t(a)}
 &=\sum_{a:q_t(a)>0}
  \sqrt{\frac{q_t(a)}{\pi_t^{\mathrm{TS}}(a)}}
  \sqrt{\pi_t^{\mathrm{TS}}(a)I_t(a)}\\
 &\le\sqrt{AM\,J_t^{\mathrm{TS}}}.
 \label{eq:mix-ts-sum-tv}
\end{align}
Combining \eqref{eq:mix-ts-tv}--\eqref{eq:mix-ts-sum-tv}
gives the pointwise comparator guarantee
\begin{equation}
 (r_t^{\mathrm{TS}})^2\le32L^2 AM\,J_t^{\mathrm{TS}}.
 \label{eq:mix-ts-ratio-bound}
\end{equation}
For $0<\lambda<1$, exact gap IDS minimizes $\psi_t$ over the simplex
containing $\pi_t^{\mathrm{TS}}$. If $0<J_t^{\mathrm{TS}}<\infty$, its minimal ratio
is at most $32L^2 AM$. If $J_t^{\mathrm{TS}}=0$, then
\eqref{eq:mix-ts-ratio-bound} forces $r_t^{\mathrm{TS}}=0$,
so the TS comparator has ratio zero. If
$J_t^{\mathrm{TS}}=+\infty$, its ratio is also zero because
$r_t^{\mathrm{TS}}$ is finite under bounded rewards.
In either zero-ratio case an exact IDS minimizer has
$r_t^{\mathrm{IDS}}=0$ whenever $J_t^{\mathrm{IDS}}<\infty$.
For $J_t^{\mathrm{IDS}}=+\infty$ the following inequality
holds in the extended sense. Hence in all cases,
\begin{equation}
 (r_t^{\mathrm{IDS}})^2\le32L^2 AM\,J_t^{\mathrm{IDS}}.
 \label{eq:mix-ids-ratio-bound}
\end{equation}
When $\lambda=1$, the mixed-policy bound is
\eqref{eq:mix-ts-ratio-bound} in square-root form. When
$J_t^{\mathrm{mix}}=+\infty$, the bound holds in the extended sense.
For $0<\lambda<1$ and finite $J_t^{\mathrm{mix}}$, both branch gains
are finite. Linearity in the action distribution and concavity of the
square root then yield
\begin{align}
 r_t^{\mathrm{mix}}
 &=\lambda r_t^{\mathrm{TS}}+(1-\lambda)r_t^{\mathrm{IDS}}\\
 &\le\sqrt{32}\,L\sqrt{AM}
 \bigl(\lambda\sqrt{J_t^{\mathrm{TS}}}
       +(1-\lambda)\sqrt{J_t^{\mathrm{IDS}}}\bigr)\\
 &\le\sqrt{32}\,L\sqrt{AM\bigl(
  \lambda J_t^{\mathrm{TS}}+(1-\lambda)J_t^{\mathrm{IDS}}\bigr)}\\
 &=\sqrt{32}\,L\sqrt{AM\,J_t^{\mathrm{mix}}}.
 \label{eq:mix-common-ratio}
\end{align}
\paragraph{Step 3: Cumulative regret.}
Conditioning first on $H_t$ and the action chosen by the mixture gives
$\mathbb E[\Delta_\theta(A_t)\mid H_t]=r_t^{\mathrm{mix}}$.
Using \eqref{eq:mix-common-ratio}, Jensen's inequality, the
Cauchy--Schwarz inequality across the $T$ rounds, and
\eqref{eq:mix-chain-rule} in that order yields
\begin{align}
 R_T^{\mathrm{gap}}
 &=\sum_{t=1}^T\mathbb E[r_t^{\mathrm{mix}}]\\
 &\le\sqrt{32}\,L\sqrt{AM}\sum_{t=1}^T
    \mathbb E\sqrt{J_t^{\mathrm{mix}}}\\
 &\le\sqrt{32}\,L\sqrt{AM}\sum_{t=1}^T
   \sqrt{\mathbb E[J_t^{\mathrm{mix}}]}\\
 &\le\sqrt{32}\,L\sqrt{AMT\sum_{t=1}^T
    \mathbb E[J_t^{\mathrm{mix}}]}\\
 &=\sqrt{32}\,L\sqrt{AMT\,\mathrm{IB}_T}.
 \label{eq:mix-regret-proved}
\end{align}
This proves part~(i).

\paragraph{Step 4: Mixed-action sampling coverage.}
Return to arbitrary integrable reward laws. From
\eqref{eq:mix-post-indep}, the two sets $S_t$ and $S^*(\theta)$
have the same conditional membership probabilities
\begin{equation}
 q_t(a)=\Pr(a\in S_t\mid H_t)
       =\Pr(a\in S^*(\theta)\mid H_t).
 \label{eq:mix-q}
\end{equation}
The almost-sure definition of $M$ gives $1\le|S_t|\le M$.
By \eqref{eq:mix-action-dist},
\begin{align}
 \pi_t^{\mathrm{mix}}(a)
 &\ge\lambda\pi_t^{\mathrm{TS}}(a)
 =\lambda\mathbb E[\mathbf1\{a\in S_t\}/|S_t|\mid H_t]\\
 &\ge(\lambda/M)q_t(a).
 \label{eq:mix-coverage}
\end{align}
Thus, if $q_t(a)>0$, then $\pi_t^{\mathrm{mix}}(a)>0$ and
$q_t(a)/\pi_t^{\mathrm{mix}}(a)\le M/\lambda$. Conversely,
$q_t(a)=0$ eliminates the arm from all expressions multiplied by
$\mathbf1\{a\in S_t\}$.

\paragraph{Step 5: A changed Pareto set admits a witness in its union.}
For $\vartheta,\vartheta'\in\Theta_{\mathrm{int}}$ with
$S^*(\vartheta)\ne S^*(\vartheta')$, exchange them if needed
and choose $a\in S^*(\vartheta)\setminus S^*(\vartheta')$.
Because $a$ is dominated under $\vartheta'$, follow a finite
chain of strict dominance to a $b\in S^*(\vartheta')$ that
dominates $a$ there. For the coordinate-wise mean differences
$D_{a,b,j}(\xi):=\mu_{\xi,j}(b)-\mu_{\xi,j}(a)$, all
$D_{a,b,j}(\vartheta')\ge0$ and at least one is positive.
Because $a\in S^*(\vartheta)$, either some
$D_{a,b,j}(\vartheta)<0$ or all are zero. In the first case choose a
coordinate $j$ with $D_{a,b,j}(\vartheta)<0$ and
$D_{a,b,j}(\vartheta')\ge0$. In the second case choose a $j$ with
$D_{a,b,j}(\vartheta')>0$. Then
\begin{equation}
 |D_{a,b,j}(\vartheta)-D_{a,b,j}(\vartheta')|
 =|D_{a,b,j}(\vartheta)|+|D_{a,b,j}(\vartheta')|>0.
 \label{eq:mix-contrast-sign}
\end{equation}
If either nonzero mean difference has magnitude at least $\epsilon$,
the triangle inequality shows that
\begin{align}
 \epsilon
 &\le |D_{a,b,j}(\vartheta)-D_{a,b,j}(\vartheta')|\\
 &\le|\mu_{\vartheta,j}(a)-\mu_{\vartheta',j}(a)|
    +|\mu_{\vartheta,j}(b)-\mu_{\vartheta',j}(b)|.
 \label{eq:mix-contrast-mean}
\end{align}
Therefore at least one $u\in\{a,b\}\subseteq
S^*(\vartheta)\cup S^*(\vartheta')$ has a coordinate-wise mean
difference of at least $\epsilon/2$. In particular, the reward
laws of that arm cannot coincide. This proves the union-witness
property whenever a nonzero coordinate-wise mean difference
is uniformly bounded away from zero or the reward laws admit
a suitable distinguishing event.

\paragraph{Step 6: Conditional mismatch and information gain.}
Fix a good set $\mathcal G$ and $\rho$ satisfying
\eqref{eq:mix-witness}. Let
$V_t(a)=\operatorname{TV}(P_\theta^a,P_{\theta_t}^a)$.
For every realization, the following indicator bound holds:
\begin{equation}
 \mathbf1\{S_t\ne S^*(\theta)\}
 \le\mathbf1\{\theta\notin\mathcal G\}
     +\mathbf1\{\theta_t\notin\mathcal G\}
     +\rho^{-1}\sum_{a\in S^*(\theta)\cup S_t}V_t(a).
 \label{eq:mix-mismatch-indicator}
\end{equation}
Indeed, if either parameter lies outside $\mathcal G$, the
right side is at least one. If both lie inside and the sets
differ, the witness condition makes the sum at least $\rho$.
Otherwise the left side is zero. Now use the elementary bound
$\mathbf1\{a\in S^*(\theta)\cup S_t\}
\le\mathbf1\{a\in S^*(\theta)\}+\mathbf1\{a\in S_t\}$.
Conditional exchangeability of $(\theta,\theta_t)$ and the
symmetry of TV make the expected two sums equal. Thus
\begin{align}
 &\mathbb E\left[\sum_{a\in S^*(\theta)\cup S_t}V_t(a)
             \Bigm|H_t\right]\\
 &\quad\le2\sum_a
    \mathbb E[\mathbf1\{a\in S_t\}V_t(a)\mid H_t]\\
 &\quad\le2\sum_{a:q_t(a)>0}
    \sqrt{q_t(a)\mathbb E[V_t(a)^2\mid H_t]}\\
 &\quad\le2\sqrt2\sum_{a:q_t(a)>0}\sqrt{q_t(a)I_t(a)}\\
 &\quad=2\sqrt2\sum_{a:q_t(a)>0}
   \sqrt{\frac{q_t(a)}{\pi_t^{\mathrm{mix}}(a)}}
   \sqrt{\pi_t^{\mathrm{mix}}(a)I_t(a)}\\
 &\quad\le2\sqrt{\frac{2AM}{\lambda}
                 J_t^{\mathrm{mix}}}.
 \label{eq:mix-mismatch-information}
\end{align}
The second inequality uses conditional Cauchy--Schwarz and omits
the zero terms with $q_t(a)=0$. The third uses
Lemma~\ref{lem:info_gain_tv_theta}, which gives
$\mathbb E[V_t(a)^2\mid H_t]\le2I_t(a)$. The final inequality uses
Cauchy--Schwarz across at most $A$ arms and \eqref{eq:mix-coverage}.
If $J_t^{\mathrm{mix}}=+\infty$, the final bound holds in the
extended sense.

Taking expectations in \eqref{eq:mix-mismatch-indicator}, the
true parameter has marginal $p_1$ by definition. The
posterior-drawn parameter also has marginal $p_1$, since
$\mathbb E[p_t(D)]=\Pr(\theta\in D)=p_1(D)$ for every
measurable $D$. Hence the two exceptional indicators together
have expectation $2p_1(\mathcal G^c)$. By
\eqref{eq:mix-mismatch-information} and Jensen,
\begin{equation}
 e_t\le2p_1(\mathcal G^c)
 +\frac2\rho\sqrt{\frac{2AM}{\lambda}}
  \sqrt{\mathbb E[J_t^{\mathrm{mix}}]}.
 \label{eq:mix-each-t}
\end{equation}
Average this inequality over $t=1,\ldots,T$ and apply
Cauchy--Schwarz and \eqref{eq:mix-chain-rule}:
\begin{align}
 \bar e_T
 &\le2p_1(\mathcal G^c)
 +\frac2\rho\sqrt{\frac{2AM}{\lambda}}
      \frac1T\sum_{t=1}^T
        \sqrt{\mathbb E[J_t^{\mathrm{mix}}]}\\
 &\le2p_1(\mathcal G^c)
 +\frac2\rho\sqrt{\frac{2AM}{\lambda T}
             \sum_{t=1}^T\mathbb E[J_t^{\mathrm{mix}}]}\\
 &=2p_1(\mathcal G^c)
 +\frac2\rho\sqrt{\frac{2AM\,\mathrm{IB}_T}{\lambda T}}.
 \label{eq:mix-average-e}
\end{align}

\paragraph{Step 7: From average to last-round mismatch.}
Let $p_t(s)=\Pr(S^*(\theta)=s\mid H_t)$, for each
$s\subseteq\mathcal A$. Given $H_t$, the true and
posterior-drawn sets are independent with the same distribution,
so
\begin{equation}
 \Pr(S_t=S^*(\theta)\mid H_t)=\sum_{s\subseteq\mathcal A}p_t(s)^2,
 \quad e_t=1-\mathbb E\left[\sum_s p_t(s)^2\right].
 \label{eq:mix-mismatch-post}
\end{equation}
For every fixed $s$, $p_t(s)=\mathbb E[
\mathbf1\{S^*(\theta)=s\}\mid H_t]$ is a bounded martingale
in the increasing history filtration. Conditional Jensen gives
$\mathbb E[p_{t+1}(s)^2\mid H_t]\ge p_t(s)^2$.
Summing over the finitely many $s$ and taking expectations
shows $e_{t+1}\le e_t$. Therefore
$e_T\le T^{-1}\sum_{t=1}^T e_t=\bar e_T$, and
\eqref{eq:mix-average-e} proves part~(ii)'s finite-horizon
inequality.

\paragraph{Step 8: Good sets of arbitrarily high prior probability.}
We now prove the asserted existence of $\mathcal G_\eta$,
using only finite absolute first moments. For each parameter
$\vartheta\in\Theta_{\mathrm{int}}$, let
\begin{equation}
 m(\vartheta)
 :=\min\left\{|D_{a,b,j}(\vartheta)|:
    a\ne b,\ j\in[d],\ D_{a,b,j}(\vartheta)\ne0\right\},
 \label{eq:mix-margin}
\end{equation}
with $\min\varnothing=+\infty$. If there are any nonzero
mean differences, their finite collection has a strictly positive
minimum; otherwise $m(\vartheta)=+\infty$. In either case,
$m(\vartheta)>0$. For any $\eta>0$, continuity of
probabilities from above allows a fixed $\epsilon>0$ such
that $p_1\{m<\epsilon\}\le\eta/2$.
For $C>0$, put $f_C(x)=\max\{-C,\min\{C,x\}\}$ and
\begin{equation}
 b_C(\vartheta):=\max_{a,j}\left|
     \mathbb E_\vartheta[f_C(Y_{t,a,j})]
                   -\mu_{\vartheta,j}(a)\right|.
 \label{eq:mix-clipping-bias-g}
\end{equation}
Since $|f_C(x)-x|\le|x|\mathbf1\{|x|>C\}$ and every
reward coordinate has finite absolute first moment,
dominated convergence gives $b_C(\vartheta)\to0$ for
all $\vartheta\in\Theta_{\mathrm{int}}$. The mean integrals are
measurable in $\vartheta$, so the finite maximum $b_C$ is
measurable. Bounded convergence applied to
$\mathbf1\{b_C>\epsilon/8\}$ gives
$p_1\{b_C>\epsilon/8\}\to0$ as $C\to\infty$.
Choose a fixed $C>0$ with $p_1\{b_C>\epsilon/8\}\le\eta/2$. Define
\begin{equation}
 \mathcal G_\eta
 =\{\vartheta\in\Theta_{\mathrm{int}}:
      m(\vartheta)\ge\epsilon,
      b_C(\vartheta)\le\epsilon/8\}.
 \label{eq:mix-good-set}
\end{equation}
The union bound gives $p_1(\mathcal G_\eta^c)\le\eta$.
For $\vartheta,\vartheta'\in\mathcal G_\eta$ with
different Pareto sets, Step~5 yields an arm in their union
with one original-coordinate mean difference at least
$\epsilon/2$. The two truncated means differ by at least
$\epsilon/2-2(\epsilon/8)=\epsilon/4$. Because the bounded
test $f_C(Y_j)$ takes values in $[-C,C]$,
\begin{equation}
 \left|\mathbb E_\vartheta f_C(Y_{t,u,j})
       -\mathbb E_{\vartheta'}f_C(Y_{t,u,j})\right|
 \le2C\operatorname{TV}(P_\vartheta^u,P_{\vartheta'}^u).
 \label{eq:mix-g-tv}
\end{equation}
Therefore the witness condition holds with
$\rho_\eta=\epsilon/(8C)>0$, independent of $T$.

For a fixed $\lambda>0$, apply \eqref{eq:mix-recovery-theorem}
with this good set. If $\mathrm{IB}_T=o(T)$, then
$\limsup_T\bar e_T\le2\eta$. Since $\eta>0$ is arbitrary,
$\bar e_T\to0$, and hence $e_T\to0$. Finally,
\begin{equation}
 \mathbb E[p_t(S^*(\theta))]
 =\mathbb E\left[\sum_s p_t(s)^2\right]=1-e_t\to1.
 \label{eq:mix-true-postmass}
\end{equation}
The bounded martingale convergence theorem gives, for each
of finitely many $s$, $p_t(s)\to p_\infty(s)$ almost surely.
Thus $p_t(S^*(\theta))\to p_\infty(S^*(\theta))$ almost surely.
Bounded convergence and \eqref{eq:mix-true-postmass} imply
$\mathbb E[p_\infty(S^*(\theta))]=1$. Since this random
variable is at most one, it equals one almost surely.
This completes the consistency claim for a fixed infinite-run
model and policy.

\paragraph{Step 9: Finite-prior table entry.}
Suppose $p_1$ is supported on a finite set $\Theta_0$ of
positive-mass parameters. Then
$\mathrm{IB}_T\le H(\theta)\le\log|\Theta_0|$ by
$I(\theta;H_{T+1})=H(\theta)-H(\theta\mid H_{T+1})$.
Substituting this budget into \eqref{eq:mix-gap-theorem}
gives the finite-prior regret entry.
If all members of $\Theta_0$ induce the same Pareto set,
$e_t=0$ for every $t$. Otherwise, by Step~5 every supported
pair with different Pareto sets has an arm with strictly
positive TV distance. The finite collection of positive
TV distances between supported action laws has a strictly
positive minimum $\rho_{\mathrm{fin}}$. Taking
$\mathcal G=\Theta_0$ gives $p_1(\mathcal G^c)=0$,
and hence from \eqref{eq:mix-recovery-theorem},
\begin{equation}
 e_T\le\bar e_T\le \frac2{\rho_{\mathrm{fin}}}
          \sqrt{\frac{2AM\,H(\theta)}{\lambda T}}.
 \label{eq:mix-finite-rec}
\end{equation}
This gives the finite-prior recovery entry.

\paragraph{Step 10: Independent Dirichlet categorical table entry.}
Use the product categorical model of Lemma~\ref{lem:ig_categorical}
with $K\ge2$ reward levels $1,\ldots,K$ and $\alpha_{a,j,k}\ge1$.
By that lemma, every admissible adaptive action rule, including
this mixture, satisfies
\begin{equation}
 \mathrm{IB}_T\le Ad(K+1)(1+\log T).
 \label{eq:mix-cat-ib}
\end{equation}
All rewards lie in $[1,K]$, of width $L=K-1$.
Substitution of \eqref{eq:mix-cat-ib}
into \eqref{eq:mix-gap-theorem} gives
\begin{equation}
 R_T^{\mathrm{gap}}
 \le\sqrt{32}\,L A\sqrt{Md(K+1)T(1+\log T)},
 \label{eq:mix-cat-reg}
\end{equation}
the table's explicit categorical regret bound.

For recovery, define $\mathcal G_\epsilon^{\rm cat}$ as
the parameters at which each nonzero coordinate-wise mean difference
$D_{a,b,j}$ has absolute value at least $\epsilon$.
The Dirichlet density on each probability simplex is bounded
because all concentration parameters are at least one.
Writing the coordinate mean as
$\mu_{\theta,j}(a)=K-\sum_{k=1}^{K-1}(K-k)p_{a,j,k}$,
the coefficient of $p_{a,j,1}$ is $-(K-1)\ne0$.
For each fixed choice of the other simplex coordinates, an
interval of width $2\epsilon$ in the mean restricts
$p_{a,j,1}$ to a length at most $2\epsilon/(K-1)$.
Integrating the bounded prior density over the remaining
compact simplex gives a constant $C_{a,j}<\infty$ such that
\begin{equation}
 \sup_z p_1\bigl(|\mu_{\theta,j}(a)-z|<\epsilon\bigr)
 \le \frac{2C_{a,j}}{K-1}\epsilon.
 \label{eq:mix-cat-anticonc}
\end{equation}
For two distinct arms $a,b$, the coordinate means are independent
under the prior. Condition on the mean of $b$ and apply
\eqref{eq:mix-cat-anticonc}. A union bound over the finitely
many arm pairs and coordinates gives
\begin{equation}
 p_1((\mathcal G_\epsilon^{\rm cat})^c)
 \le C_{\mathrm{cat}}\epsilon
 \quad(0<\epsilon\le1).
 \label{eq:mix-cat-goodprob}
\end{equation}
for a fixed finite $C_{\mathrm{cat}}$. If two such good
parameters have different Pareto sets, Step~5 provides an arm in
the union with a coordinate mean difference at least $\epsilon/2$.
For any two laws on $[1,K]$, the expectation difference of
that coordinate is at most $(K-1)$ times their TV distance.
Therefore the witness threshold is
\begin{equation}
 \rho_\epsilon=\frac{\epsilon}{2(K-1)}.
 \label{eq:mix-cat-rho}
\end{equation}
Inserting \eqref{eq:mix-cat-ib}, \eqref{eq:mix-cat-goodprob},
and \eqref{eq:mix-cat-rho} into
\eqref{eq:mix-recovery-theorem} yields
\begin{equation}
 e_T\le\bar e_T\le2C_{\rm cat}\epsilon
 +\frac{4(K-1)}{\epsilon}
  \sqrt{\frac{2AM\,Ad(K+1)(1+\log T)}{\lambda T}}.
 \label{eq:mix-cat-eps-bound}
\end{equation}
The constants in front of $\epsilon$ and
$\epsilon^{-1}\sqrt{\log(1+T)/(\lambda T)}$
are fixed by the model. If
$x_T:=\log(1+T)/(\lambda T)\le1$, choose
$\epsilon=x_T^{1/4}$. Then both terms are
$O(x_T^{1/4})$ (using $1+\log T\le C\log(1+T)$
for $T\ge1$). If $x_T>1$, use $\bar e_T\le1\le x_T^{1/4}$.
Thus the categorical recovery rate in the table holds for every
$T$ and every fixed
$\lambda\in(0,1]$, with constants independent of $T$ and
$\lambda$. When $K=1$, all reward means coincide and
$S_t=S^*(\theta)=\mathcal A$ for every $t$, and the regrets
and mismatch probabilities are zero.

\paragraph{Step 11: Conditioned shared-covariance Gaussian entry.}
Let $p_0$ be the normal--inverse-Wishart prior of
Lemma~\ref{lem:ig_gaussian}. The covariance $\Sigma$ is shared
across arms, and the arm means are independent given $\Sigma$ with
$\mu_a\mid\Sigma\sim\mathcal N(m_{a,1},\Sigma/\kappa_{a,1})$.
The reward law is $Y_{t,a}\mid(\mu_a,\Sigma)\sim\mathcal N(\mu_a,\Sigma)$.
Fix $\mu_{\max},\sigma>0$ and define
\begin{equation}
 E=\bigl\{\max_{a,j}|\mu_{a,j}|\le\mu_{\max},\quad
            \max_j\Sigma_{jj}\le\sigma^2\bigr\},
 \qquad p_1=p_0(\cdot\mid E).
 \label{eq:mix-gaussian-event}
\end{equation}
Appendix~\ref{app:gaussian_high_probability} shows that $p_0(E)>0$.
For each horizon $T$, choose
$B_T=\mu_{\max}+2\sigma\sqrt{\log(2T)}$ once, apply
$c_{B_T}(y)=\max\{-B_T,\min\{y,B_T\}\}$ coordinatewise,
and run Algorithm~\ref{alg:pareto-ts-ids-mixture} on the clipped
observations. Use their pushforward laws, including any boundary
mass, for the exact posterior updates, the clipped-mean gaps, and
the IDS information gains. Write $\widetilde P_{\vartheta,B_T}^a$
for the clipped observation law, $\widetilde{\mathrm{IB}}_T$
for the information budget along this clipped mixed run, and
$\widetilde\Delta_{\vartheta,B_T}$ for its mean-based gaps.

First, we verify both Pareto membership and information control.
For each fixed parameter $\vartheta\in E$ and coordinate $j$,
all arms have the same conditional standard deviation
$\sigma_j(\vartheta)=\sqrt{\Sigma_{jj}}>0$. For $B>0$, the map
\begin{equation}
 g_{\vartheta,j,B}(m)
 :=\mathbb E[c_B(m+\sigma_j(\vartheta)Z)],
 \qquad Z\sim N(0,1),
 \label{eq:mix-gaussian-monotonic}
\end{equation}
is strictly increasing in $m$: for $m'<m$, monotonicity of
$c_B$ gives a pointwise inequality, strict on a positive-
probability event on which the lower argument lies inside
$(-B,B)$. This same strictly increasing map applies to
every arm in coordinate $j$. Consequently all coordinate
mean comparisons, including ties, are preserved and
\begin{equation}
 S^*_{\rm clipped}(\vartheta)=S^*_{\rm original}(\vartheta),
 \qquad\vartheta\in E.
 \label{eq:mix-gaussian-pareto}
\end{equation}
Let $H^Y_{T+1}=(A_t,Y_{t,A_t})_{t=1}^T$ record the raw observations
along the actions of the clipped policy. The policy is admissible
in the raw experiment because it can clip the raw history and use
its clipped-posterior computation. For this same action kernel,
conditioning the $p_0$ experiment on $E$ gives the $p_1$
experiment. Because $\mathbf1_E$ is a function of $\theta$,
the chain rule and nonnegativity give
\begin{equation}
 I_{p_0}(\theta;H^Y_{T+1})
 \ge p_0(E) I_{p_1}(\theta;H^Y_{T+1}).
 \label{eq:mix-gauss-conditioned-ig}
\end{equation}
The observed clipped history is a measurable function of the
raw history with the same actions, so data processing and
Lemma~\ref{lem:ig_gaussian} (valid for all such adaptive policies) give
\begin{equation}
 \widetilde{\mathrm{IB}}_T
 \le I_{p_1}(\theta;H^Y_{T+1})
 \le\frac{I_{p_0}(\theta;H^Y_{T+1})}{p_0(E)}
 \le C_{\rm G}\log(1+T),
 \label{eq:mix-gauss-ib}
\end{equation}
where $C_{\rm G}<\infty$ depends only on the fixed model and prior.

Second, the clipped rewards lie in a common interval of width
$2B_T$, so part~(i), applied to the clipped experiment and
using \eqref{eq:mix-gaussian-pareto}, gives
\begin{equation}
 \mathbb E\sum_{t=1}^T
   \widetilde\Delta_{\theta,B_T}(A_t)
 \le\sqrt{32}(2B_T)\sqrt{AMT\,
                       \widetilde{\mathrm{IB}}_T}.
 \label{eq:mix-gauss-clipped-gap}
\end{equation}
For every coordinate the centered conditional Gaussian tail
is $\sigma$-sub-Gaussian under $E$. Standard one-sided Chernoff
bounds on $Y-\mu$ and $-(Y-\mu)$, integrated beyond
$u_B=B-\mu_{\max}>0$, yield the uniform mean-clipping bias
\begin{equation}
 \max_{a,j}|\mathbb E_\vartheta[c_B(Y_{t,a,j})]
                     -\mu_{\vartheta,j}(a)|
 \le \epsilon_B,
 \quad \epsilon_B:=\frac{2\sigma^2}{B-\mu_{\max}}
    \exp\!\left(-\frac{(B-\mu_{\max})^2}{2\sigma^2}\right).
 \label{eq:mix-gaussian-bias}
\end{equation}
For each pair of arms $a,b$ and coordinate $j$, clipping
changes the mean difference
$\mu_{\vartheta,j}(b)-\mu_{\vartheta,j}(a)$ by at most
$2\epsilon_B$, because each of the two means changes by at most
$\epsilon_B$. Taking the minimum over coordinates, the maximum
over comparator arms, and the positive part cannot increase this
uniform error. Consequently,
$\Delta_\vartheta(a)\le
\widetilde\Delta_{\vartheta,B}(a)+2\epsilon_B$.
Write $R_T^{\mathrm{gap,orig}}=\mathbb E[\sum_{t=1}^T
\Delta_\theta(A_t)]$ for the Bayesian regret of the original-mean
gaps on this same mixed run. Then
\begin{align}
 R_T^{\mathrm{gap,orig}}
 &\le\sqrt{32}(2B_T)\sqrt{AMT\,
                  \widetilde{\mathrm{IB}}_T}+2T\epsilon_{B_T}\\
 &\le\sqrt{32}(2B_T)\sqrt{AMT\,C_{\rm G}\log(1+T)}
    +\frac{\sigma}{2T\sqrt{\log(2T)}}\\
 &=O\!\left(\sqrt T\log(1+T)\right),
 \label{eq:mix-gaussian-orig-regret}
\end{align}
for fixed $A,M,\mu_{\max},\sigma$, and prior. The second
line substitutes $B_T-\mu_{\max}=2\sigma\sqrt{\log(2T)}$
into \eqref{eq:mix-gaussian-bias}.

Third, we establish the recovery witness directly for the
clipped laws. For $a\ne b$ and $j$, define the
standardized mean difference
$Z_{a,b,j}(\vartheta):=
[\mu_{\vartheta,j}(b)-\mu_{\vartheta,j}(a)]/
\sqrt{\Sigma_{jj}}$.
Let $\mathcal G_\epsilon^{\rm G}$ contain the parameters
for which every nonzero standardized mean difference has absolute value
at least $\epsilon$, for $0<\epsilon\le1$. Under $p_0$,
conditional on $\Sigma$, $Z_{a,b,j}$ is Gaussian with
variance $v_{a,b}=\kappa_{a,1}^{-1}+\kappa_{b,1}^{-1}>0$
(its conditional mean may depend on $\Sigma$). Its density
is bounded by $(2\pi v_{a,b})^{-1/2}$ uniformly in that mean.
Integrating over $(-\epsilon,\epsilon)$, averaging over
$\Sigma$, and taking a finite union bound yields
\begin{equation}
 p_0((\mathcal G_\epsilon^{\rm G})^c)
 \le C_{\rm near}\epsilon,
 \qquad
 C_{\rm near}:=2d\sum_{\{a,b\}:a\ne b}
       (2\pi v_{a,b})^{-1/2}<\infty.
 \label{eq:mix-gaussian-prior-ties}
\end{equation}
Set $\mathcal G=E\cap\mathcal G_\epsilon^{\rm G}$.
Its complement under $p_1$ has mass at most
$C_{\rm near}\epsilon/p_0(E)$.
For two parameters $\vartheta,\vartheta'$ in this set
with different Pareto sets, Step~5 gives a sign-changing
coordinate-wise mean difference that is nonpositive at one
parameter and nonnegative at the other, with at least one value
nonzero. A nonzero value has absolute value at least
$\epsilon\sigma_j(\xi)$ for the corresponding parameter
$\xi\in\{\vartheta,\vartheta'\}$. By
\eqref{eq:mix-contrast-mean}, an arm $u$ in the union has
\begin{equation}
 |\mu_{\xi,j}(u)-\mu_{\xi',j}(u)|
 \ge \tfrac\epsilon2\sigma_j(\xi),
 \label{eq:mix-gauss-mean-witness}
\end{equation}
where $\xi'$ denotes the other parameter.
Take the measurable reward event
$D=\{y:y_j\le\mu_{\xi',j}(u)\}$. Under $\xi'$, its
probability is $1/2$. Under $\xi$, it is $\Phi(z)$ with
$z=(\mu_{\xi',j}(u)-\mu_{\xi,j}(u))/\sigma_j(\xi)$, so
$|z|\ge\epsilon/2$. Consequently
\begin{equation}
 |P_\xi^u(D)-P_{\xi'}^u(D)|
 \ge\Phi(\epsilon/2)-\tfrac12
 \ge c_0\epsilon,
 \qquad c_0:=\tfrac12\varphi(1/2)>0,
 \label{eq:mix-gauss-event-witness}
\end{equation}
using $0<\epsilon\le1$ and monotonicity of the standard
normal density $\varphi$ on $[0,1/2]$.
The threshold $\mu_{\xi',j}(u)$ lies in
$[-\mu_{\max},\mu_{\max}]\subset(-B_T,B_T)$, so
$\{c_{B_T}(y_j)\le\mu_{\xi',j}(u)\}$ is exactly $D$.
Thus the same probability difference survives clipping and
\begin{equation}
 \operatorname{TV}(\widetilde P_{\xi,B_T}^u,
                       \widetilde P_{\xi',B_T}^u)
 \ge c_0\epsilon.
 \label{eq:mix-gauss-clipped-tv}
\end{equation}
Applying part~(ii) to \emph{this horizon-$T$ clipped
experiment}, whose original and clipped Pareto sets coincide
by \eqref{eq:mix-gaussian-pareto}, gives
\begin{equation}
 e_T^{(T)}\le\bar e_T^{(T)}
 \le\frac{2C_{\rm near}}{p_0(E)}\epsilon
   +\frac{2}{c_0\epsilon}
     \sqrt{\frac{2AM\,C_{\rm G}\log(1+T)}{\lambda T}}.
 \label{eq:mix-gauss-rec-eps}
\end{equation}
If $x_T=\log(1+T)/(\lambda T)\le1$, select
$\epsilon=x_T^{1/4}$, which makes the two terms
$O(x_T^{1/4})$. If $x_T>1$, use
$e_T^{(T)}\le\bar e_T^{(T)}\le1\le x_T^{1/4}$.
This proves the Gaussian recovery rate in the theorem's table
for the exact Pareto set of the original Gaussian means.
Within each fixed horizon-$T$ clipped run, monotonicity
$e_T^{(T)}\le\bar e_T^{(T)}$ follows from Step~7.
Together with \eqref{eq:mix-gaussian-orig-regret},
this proves the joint guarantee for that \emph{same} clipped
mixed run and finishes the proof of all parts of the theorem.
\end{proof}

\subsection{Computation of the IDS Information Ratios}
\label{app:ids_estimation_section4}

We describe how the IDS quantities used for Pareto-violation and
gap-based regret are computed in the experiments. The two formulations
share the same information-ratio optimization, but differ in both the
one-step loss and the information target. We use
\(N_{\mathrm{MC}}=100\) posterior samples per round for quantities that
depend nonlinearly on the unknown Pareto geometry. For gap-based IDS,
the parameter-level information gain can be estimated using a general
posterior-sampling procedure; in the Beta--Bernoulli model and in a
Gaussian model with known observation covariance, it also admits useful
closed-form expressions.

\paragraph{Pareto-violation IDS.}
For Algorithm~\ref{alg:IDS}, the one-step loss and information gain are
\[
\Delta_t^{\mathrm{viol}}(a)
:=
1-\mathbb{P}
\!\left(
a\in\mathcal{S}^*
\mid
\mathcal H_t
\right),
\qquad
I_t^{\mathrm{viol}}(a)
:=
I\!\left(
\mathcal{S}^*;
\mathbf Y_{t,a}
\mid
\mathcal H_t,A_t=a
\right).
\]
At round \(t\), we draw
\[
\boldsymbol{\theta}^{(1)},\ldots,
\boldsymbol{\theta}^{(N_{\mathrm{MC}})}
\stackrel{\mathrm{i.i.d.}}{\sim}
p_t(\boldsymbol{\theta}),
\]
and compute the sampled Pareto sets
\[
S^{(m)}
:=
\mathcal S^*
\!\left(
\boldsymbol{\theta}^{(m)}
\right).
\]
The posterior violation probability is approximated by
\[
\widehat{\Delta}_t^{\mathrm{viol}}(a)
=
1-
\frac{1}{N_{\mathrm{MC}}}
\sum_{m=1}^{N_{\mathrm{MC}}}
\mathbbm{1}\{a\in S^{(m)}\}.
\]

For the Bernoulli experiments, let
\(\mathcal Y=\{0,1\}^d\). Conditional on posterior sample
\(\boldsymbol{\theta}^{(m)}\), the probability of
\(\mathbf y\in\mathcal Y\) under action \(a\) is
\[
p_m^a(\mathbf y)
:=
\prod_{j=1}^{d}
\left(
\theta_{a,j}^{(m)}
\right)^{y_j}
\left(
1-\theta_{a,j}^{(m)}
\right)^{1-y_j}.
\]
Let \(\mathcal U_t\) denote the distinct Pareto sets among
\(S^{(1)},\ldots,S^{(N_{\mathrm{MC}})}\), and define
\[
\widehat p_t(s)
=
\frac{1}{N_{\mathrm{MC}}}
\sum_{m=1}^{N_{\mathrm{MC}}}
\mathbbm{1}\{S^{(m)}=s\},
\]
\[
\widehat p_t^a(\mathbf y\mid s)
=
\frac{
\sum_{m:S^{(m)}=s}
p_m^a(\mathbf y)
}{
\sum_{m=1}^{N_{\mathrm{MC}}}
\mathbbm{1}\{S^{(m)}=s\}
},
\]
and
\[
\widehat p_t^a(\mathbf y)
=
\frac{1}{N_{\mathrm{MC}}}
\sum_{m=1}^{N_{\mathrm{MC}}}
p_m^a(\mathbf y).
\]
The Pareto-set information gain is approximated by
\[
\widehat I_t^{\mathrm{viol}}(a)
=
\sum_{s\in\mathcal U_t}
\widehat p_t(s)
\sum_{\mathbf y\in\mathcal Y}
\widehat p_t^a(\mathbf y\mid s)
\log
\frac{
\widehat p_t^a(\mathbf y\mid s)
}{
\widehat p_t^a(\mathbf y)
}.
\]
Thus, although the conditional Bernoulli reward probabilities are
available explicitly, the information gain about
\(\mathcal S^*\) is evaluated through posterior samples. This is
necessary because the Pareto set is a nonlinear function of the complete
reward geometry, so Beta--Bernoulli conjugacy does not in general yield
a comparable closed-form expression for
\(I(\mathcal S^*;\mathbf Y_{t,a}\mid\mathcal H_t,A_t=a)\).

The same posterior-sampling construction extends to continuous
observation models. For Gaussian rewards, let
\(p_m^a(\mathbf y)\) denote the Gaussian observation density under
posterior sample \(\boldsymbol{\theta}^{(m)}\), and define
\[
\widehat p_t^a(\mathbf y)
=
\frac{1}{N_{\mathrm{MC}}}
\sum_{m=1}^{N_{\mathrm{MC}}}
p_m^a(\mathbf y)
\]
and, for each \(s\in\mathcal U_t\),
\[
\widehat p_t^a(\mathbf y\mid s)
=
\frac{
\sum_{m:S^{(m)}=s}
p_m^a(\mathbf y)
}{
N_s
},
\qquad
N_s
:=
\sum_{m=1}^{N_{\mathrm{MC}}}
\mathbbm{1}\{S^{(m)}=s\}.
\]
The remaining continuous expectation may be approximated by drawing
\[
\mathbf Y_{t,a}^{(m,r)}
\sim
p_m^a,
\qquad
r=1,\ldots,R,
\]
which gives the plug-in Monte Carlo approximation
\[
\widehat I_t^{\mathrm{viol}}(a)
=
\sum_{s\in\mathcal U_t}
\widehat p_t(s)
\frac{1}{N_sR}
\sum_{m:S^{(m)}=s}
\sum_{r=1}^{R}
\log
\frac{
\widehat p_t^a(
\mathbf Y_{t,a}^{(m,r)}
\mid s)
}{
\widehat p_t^a(
\mathbf Y_{t,a}^{(m,r)})
}.
\]
Hence Pareto-violation IDS can be implemented through posterior
sampling for both discrete and continuous reward models while retaining
\(\mathcal S^*\) as the information target.

\paragraph{Gap-based IDS.}
For Algorithm~\ref{alg:IDS_gap}, the corresponding quantities are
\[
\Delta_t^{\mathrm{gap}}(a)
:=
\mathbb{E}\!\left[
\Delta_{\boldsymbol{\theta}}(a)
\mid
\mathcal H_t
\right],
\qquad
I_t^{\mathrm{gap}}(a)
:=
I\!\left(
\boldsymbol{\theta};
\mathbf Y_{t,a}
\mid
\mathcal H_t,A_t=a
\right).
\]
The posterior expected gap is nonlinear in the complete collection of
mean reward vectors. In particular,
\[
\mathbb{E}\!\left[
\Delta_{\boldsymbol{\theta}}(a)
\mid
\mathcal H_t
\right]
\neq
\Delta_{\mathbb{E}[\boldsymbol{\theta}\mid\mathcal H_t]}(a)
\]
in general. Using the same \(N_{\mathrm{MC}}=100\) posterior samples,
we compute
\[
\Delta_{\boldsymbol{\theta}^{(m)}}(a)
=
\left(
\max_{b\in\mathcal A}
\min_{j\in[d]}
\left[
\mu_{\boldsymbol{\theta}^{(m)},j}(b)
-
\mu_{\boldsymbol{\theta}^{(m)},j}(a)
\right]
\right)_+
\]
and approximate
\[
\widehat{\Delta}_t^{\mathrm{gap}}(a)
=
\frac{1}{N_{\mathrm{MC}}}
\sum_{m=1}^{N_{\mathrm{MC}}}
\Delta_{\boldsymbol{\theta}^{(m)}}(a).
\]

\paragraph{General Monte Carlo parameter-information estimator.}
The parameter-level information gain can also be approximated directly
from posterior samples without requiring a model-specific closed form.
For a fixed action \(a\), let
\[
p_m^a(\mathbf y)
:=
p\!\left(
\mathbf y
\mid
A_t=a,\boldsymbol{\theta}^{(m)}
\right),
\qquad
m=1,\ldots,N_{\mathrm{MC}},
\]
and approximate the posterior predictive law by
\[
\widehat p_t^a(\mathbf y)
=
\frac{1}{N_{\mathrm{MC}}}
\sum_{\ell=1}^{N_{\mathrm{MC}}}
p_\ell^a(\mathbf y).
\]
Using
\[
I_t^{\mathrm{gap}}(a)
=
\mathbb{E}_{\boldsymbol{\theta}\mid\mathcal H_t}
\!\left[
D_{\mathrm{KL}}
\!\left(
p(\mathbf Y_{t,a}\mid
\boldsymbol{\theta},A_t=a)
\,\middle\|\,
p(\mathbf Y_{t,a}\mid
\mathcal H_t,A_t=a)
\right)
\right],
\]
we obtain the plug-in posterior-sampling approximation
\[
\widehat I_{t,\mathrm{MC}}^{\mathrm{gap}}(a)
=
\frac{1}{N_{\mathrm{MC}}}
\sum_{m=1}^{N_{\mathrm{MC}}}
D_{\mathrm{KL}}
\!\left(
p_m^a
\,\middle\|\,
\widehat p_t^a
\right).
\]

For discrete Bernoulli reward vectors,
\(\mathcal Y=\{0,1\}^d\), the KL divergence can be evaluated explicitly:
\[
\widehat I_{t,\mathrm{MC}}^{\mathrm{gap}}(a)
=
\frac{1}{N_{\mathrm{MC}}}
\sum_{m=1}^{N_{\mathrm{MC}}}
\sum_{\mathbf y\in\mathcal Y}
p_m^a(\mathbf y)
\log
\frac{
p_m^a(\mathbf y)
}{
\widehat p_t^a(\mathbf y)
}.
\]
Thus no sampling of the Bernoulli observation itself is required:
posterior parameters are sampled, while the sum over all
\(2^d\) possible reward vectors is evaluated explicitly.

For continuous observation models, the KL expectation may instead be
approximated using an additional observation-level Monte Carlo step.
For each posterior parameter sample, draw
\[
\mathbf Y_{t,a}^{(m,r)}
\sim
p_m^a,
\qquad
r=1,\ldots,R,
\]
and compute
\[
\widehat I_{t,\mathrm{MC}}^{\mathrm{gap}}(a)
=
\frac{1}{N_{\mathrm{MC}}R}
\sum_{m=1}^{N_{\mathrm{MC}}}
\sum_{r=1}^{R}
\log
\frac{
p_m^a(
\mathbf Y_{t,a}^{(m,r)})
}{
\widehat p_t^a(
\mathbf Y_{t,a}^{(m,r)})
}.
\]
This gives a general implementation of gap-based IDS even when no
closed-form parameter-information expression is available.

\paragraph{Closed-form parameter information for Beta--Bernoulli rewards.}
For Bernoulli rewards, suppose the posterior factorizes across
action--objective pairs:
\[
Y_{t,a,j}
\mid
\theta_{a,j}
\sim
\mathrm{Bernoulli}(\theta_{a,j}),
\qquad
\theta_{a,j}
\mid
\mathcal H_t
\sim
\mathrm{Beta}
\!\left(
\alpha_{t,a,j},
\beta_{t,a,j}
\right).
\]
Conditional on \(A_t=a\), the observation depends on the unknown
parameter only through
\[
\boldsymbol{\theta}_a
=
(\theta_{a,1},\ldots,\theta_{a,d}),
\]
and, by factorization across objectives,
\[
I_t^{\mathrm{gap}}(a)
=
\sum_{j=1}^{d}
I\!\left(
\theta_{a,j};
Y_{t,a,j}
\mid
\mathcal H_t,A_t=a
\right).
\]
Define
\[
\bar\theta_{t,a,j}
=
\frac{
\alpha_{t,a,j}
}{
\alpha_{t,a,j}+\beta_{t,a,j}
},
\qquad
h(p)
=
-p\log p-(1-p)\log(1-p).
\]
The posterior predictive distribution is
\[
Y_{t,a,j}
\mid
\mathcal H_t,A_t=a
\sim
\mathrm{Bernoulli}
(\bar\theta_{t,a,j}),
\]
so
\[
I\!\left(
\theta_{a,j};
Y_{t,a,j}
\mid
\mathcal H_t,A_t=a
\right)
=
h(\bar\theta_{t,a,j})
-
\mathbb E\!\left[
h(\theta_{a,j})
\mid
\mathcal H_t
\right].
\]
For
\[
\theta\sim\mathrm{Beta}(\alpha,\beta),
\]
the standard Beta identities give
\[
\mathbb E[\theta\log\theta]
=
\frac{\alpha}{\alpha+\beta}
\left[
\psi(\alpha+1)
-
\psi(\alpha+\beta+1)
\right],
\]
and
\[
\mathbb E[(1-\theta)\log(1-\theta)]
=
\frac{\beta}{\alpha+\beta}
\left[
\psi(\beta+1)
-
\psi(\alpha+\beta+1)
\right],
\]
where \(\psi(\cdot)\) denotes the digamma function. Hence
\[
\begin{aligned}
\mathbb E[
h(\theta_{a,j})
\mid
\mathcal H_t]
&=
-
\frac{\alpha_{t,a,j}}
{\alpha_{t,a,j}+\beta_{t,a,j}}
\left[
\psi(\alpha_{t,a,j}+1)
-
\psi(\alpha_{t,a,j}+\beta_{t,a,j}+1)
\right]
\\
&\quad
-
\frac{\beta_{t,a,j}}
{\alpha_{t,a,j}+\beta_{t,a,j}}
\left[
\psi(\beta_{t,a,j}+1)
-
\psi(\alpha_{t,a,j}+\beta_{t,a,j}+1)
\right],
\end{aligned}
\]
and therefore
\[
I_t^{\mathrm{gap}}(a)
=
\sum_{j=1}^{d}
\left[
h\!\left(
\frac{\alpha_{t,a,j}}
{\alpha_{t,a,j}+\beta_{t,a,j}}
\right)
-
\mathbb E[
h(\theta_{a,j})
\mid
\mathcal H_t]
\right].
\]
Thus, in the Beta--Bernoulli model the gap-based
parameter-information term also admits the exact expression above.
In our implementation we use the general posterior-sampling estimator,
while the closed form provides an equivalent model-specific alternative
that avoids Monte Carlo approximation of the information term. The
posterior expected Pareto gap itself is still estimated from posterior
samples.

\paragraph{Closed-form parameter information for a known-covariance Gaussian special case.}
For completeness, suppose the Gaussian observation covariance is known:
\[
\mathbf Y_{t,a}
\mid
\boldsymbol{\mu}_a
\sim
\mathcal N(
\boldsymbol{\mu}_a,\Sigma_a),
\qquad
\boldsymbol{\mu}_a
\mid
\mathcal H_t
\sim
\mathcal N(
\mathbf m_{t,a},V_{t,a}),
\]
where
\(\Sigma_a\succ0\).
Integrating out \(\boldsymbol{\mu}_a\) gives
\[
\mathbf Y_{t,a}
\mid
\mathcal H_t,A_t=a
\sim
\mathcal N(
\mathbf m_{t,a},
\Sigma_a+V_{t,a}).
\]
Since the observation under action \(a\) depends on the unknown
parameter only through \(\boldsymbol{\mu}_a\),
\[
I_t^{\mathrm{gap}}(a)
=
I\!\left(
\boldsymbol{\mu}_a;
\mathbf Y_{t,a}
\mid
\mathcal H_t,A_t=a
\right).
\]
Using
\[
I(X;Y)=H(Y)-H(Y\mid X)
\]
and the entropy formula for a multivariate Gaussian,
\[
\begin{aligned}
I_t^{\mathrm{gap}}(a)
&=
\frac12
\log
\frac{
\det(\Sigma_a+V_{t,a})
}{
\det(\Sigma_a)
}
\\
&=
\frac12
\log\det
\left(
I_d+
\Sigma_a^{-1/2}
V_{t,a}
\Sigma_a^{-1/2}
\right)
\\
&=
\frac12
\log\det
\left(
I_d+\Sigma_a^{-1}V_{t,a}
\right),
\end{aligned}
\]
where the last equality follows from
\(\det(I+AB)=\det(I+BA)\).

If the reward coordinates are conditionally independent, with
\[
Y_{t,a,j}
\mid
\mu_{a,j}
\sim
\mathcal N(
\mu_{a,j},
\sigma_{a,j}^2),
\qquad
\mu_{a,j}
\mid
\mathcal H_t
\sim
\mathcal N(
m_{t,a,j},
\tau_{t,a,j}^2),
\]
then
\[
V_{t,a}
=
\operatorname{diag}
(
\tau_{t,a,1}^2,\ldots,\tau_{t,a,d}^2
),
\qquad
\Sigma_a
=
\operatorname{diag}
(
\sigma_{a,1}^2,\ldots,\sigma_{a,d}^2
),
\]
and hence
\[
I_t^{\mathrm{gap}}(a)
=
\frac12
\sum_{j=1}^{d}
\log
\left(
1+
\frac{
\tau_{t,a,j}^2
}{
\sigma_{a,j}^2
}
\right).
\]
Under the Gaussian prior
\[
\mu_{a,j}
\sim
\mathcal N(
m_{0,a,j},
\tau_{0,a,j}^2),
\]
if action \(a\) has been selected \(N_{t,a}\) times before round \(t\),
then conjugate Gaussian updating gives
\[
\tau_{t,a,j}^2
=
\left(
\frac{1}{\tau_{0,a,j}^2}
+
\frac{N_{t,a}}{\sigma_{a,j}^2}
\right)^{-1}.
\]
Thus, when the observation covariance is known, the Gaussian
parameter-information term can be evaluated exactly from the current
Gaussian posterior. For Gaussian models with unknown observation
variance, including the more general setting considered in the
additional Gaussian experiments, we instead use the posterior-sampling
construction above; the known-covariance expression is included as a
closed-form special case.

\paragraph{Optimization of the empirical IDS ratios.}
For either regret criterion, let
\(\widehat{\Delta}_t(a)\) and
\(\widehat I_t(a)\) denote the corresponding estimated or closed-form
quantities above. For
\(\pi\in\Pi(\mathcal A)\), define
\[
\widehat{\Delta}_t(\pi)
=
\sum_{a\in\mathcal A}
\pi(a)\widehat{\Delta}_t(a),
\qquad
\widehat I_t(\pi)
=
\sum_{a\in\mathcal A}
\pi(a)\widehat I_t(a),
\]
and
\[
\widehat{\psi}_t(\pi)
=
\frac{
\widehat{\Delta}_t(\pi)^2
}{
\widehat I_t(\pi)
}.
\]
We use the extended-value convention
\[
\frac{0}{0}=0,
\qquad
\frac{x}{0}=+\infty
\quad\text{for }x>0.
\]

As in standard IDS implementations, an optimizer can be chosen with
support on at most two actions. We therefore search over all one- and
two-action mixtures. For each pair \(i,j\in\mathcal A\), let
\[
\pi_q
=
q e_i+(1-q)e_j,
\qquad
q\in[0,1],
\]
and write
\[
\delta_i
=
\widehat{\Delta}_t(i),
\qquad
\delta_j
=
\widehat{\Delta}_t(j),
\qquad
\iota_i
=
\widehat I_t(i),
\qquad
\iota_j
=
\widehat I_t(j).
\]
The corresponding one-dimensional objective is
\[
f_{ij}(q)
=
\frac{
\left(
q\delta_i+(1-q)\delta_j
\right)^2
}{
q\iota_i+(1-q)\iota_j
}.
\]
Writing
\[
d_\delta
=
\delta_i-\delta_j,
\qquad
d_\iota
=
\iota_i-\iota_j,
\]
we have, whenever the denominator is positive,
\[
f_{ij}(q)
=
\frac{
(\delta_j+qd_\delta)^2
}{
\iota_j+qd_\iota
}.
\]
Differentiating gives
\[
f_{ij}'(q)
=
\frac{
(\delta_j+qd_\delta)
\left[
2d_\delta(\iota_j+qd_\iota)
-
d_\iota(\delta_j+qd_\delta)
\right]
}{
(\iota_j+qd_\iota)^2
}.
\]
Apart from the degenerate case in which the numerator gap term
vanishes, an interior stationary point therefore satisfies
\[
2d_\delta
(\iota_j+qd_\iota)
=
d_\iota
(\delta_j+qd_\delta).
\]
Whenever
\[
d_\delta d_\iota\neq0,
\]
the corresponding candidate is
\[
q_{ij}^{\star}
=
\frac{
d_\iota\delta_j
-
2d_\delta\iota_j
}{
d_\delta d_\iota
}
=
\frac{
(\iota_i-\iota_j)\delta_j
-
2(\delta_i-\delta_j)\iota_j
}{
(\delta_i-\delta_j)(\iota_i-\iota_j)
}.
\]
For each pair, we evaluate \(f_{ij}\) at the endpoints
\[
q=0,
\qquad
q=1,
\]
and, whenever
\[
q_{ij}^{\star}\in(0,1)
\]
and
\[
q_{ij}^{\star}\iota_i
+
(1-q_{ij}^{\star})\iota_j
>0,
\]
also at \(q=q_{ij}^{\star}\). If
\(d_\delta=0\) or \(d_\iota=0\), no additional nondegenerate interior
candidate is required; an optimum is attained at an endpoint unless the
objective is constant over the pair. Cases with zero information
denominator are handled by the extended-value convention above.

Finally, the pair and mixing weight attaining the smallest empirical
information ratio define
\[
\widehat{\pi}_t^{\mathrm{IDS}},
\]
and the action \(A_t\) is sampled according to this distribution.

\subsection{Additional Gaussian Reward Experiments}
\label{app:gaussian_experiments}

\begin{figure*}[h]
\centering
\begin{subfigure}[b]{0.45\textwidth}
    \centering
    \includegraphics[height=3.3cm,width=\textwidth]{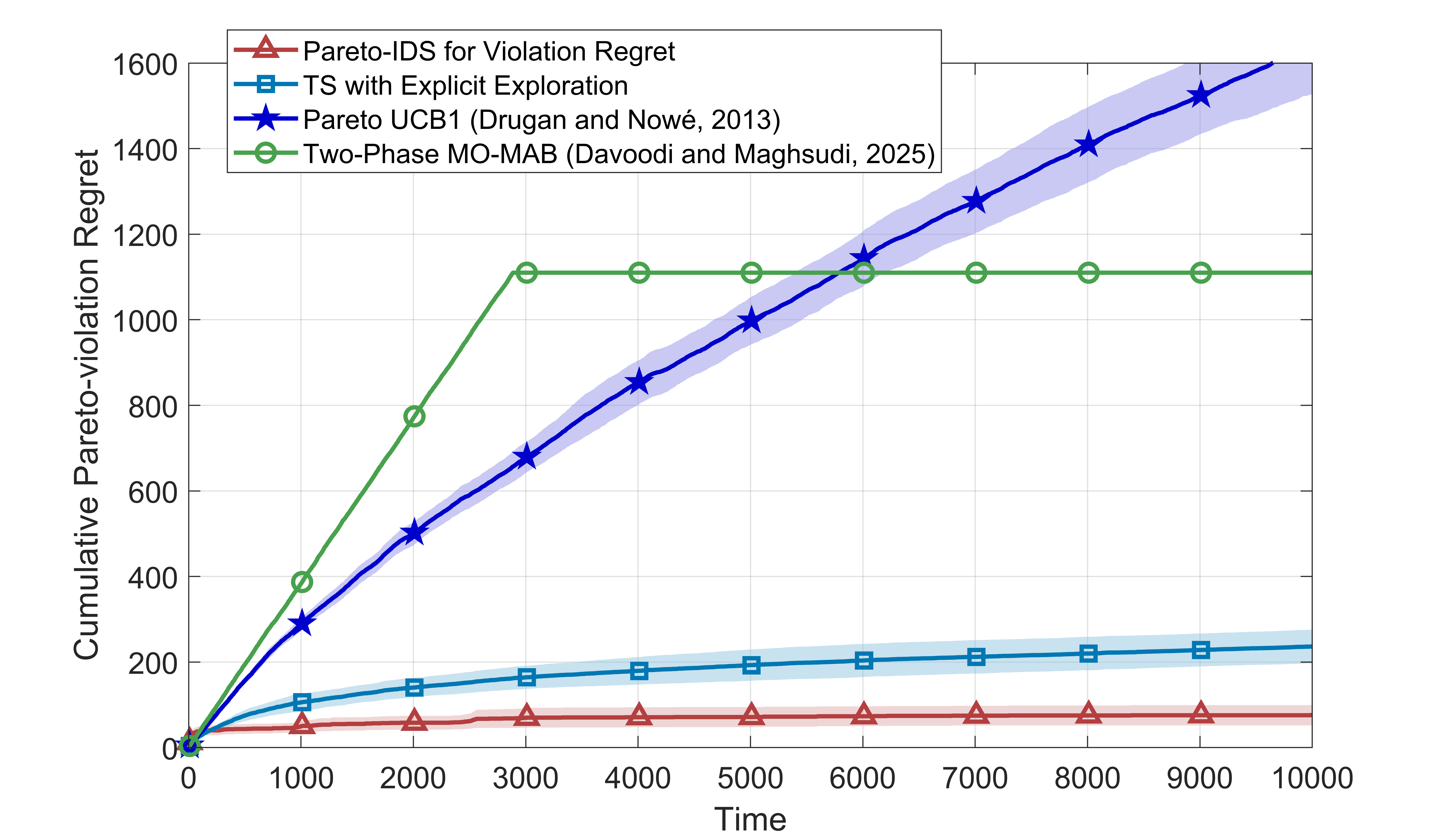}
    \subcaption{Pareto-violation regret}
    \label{fig:gaussian_violation}
\end{subfigure}
\hfill
\begin{subfigure}[b]{0.45\textwidth}
    \centering
    \includegraphics[height=3.3cm,width=\textwidth]{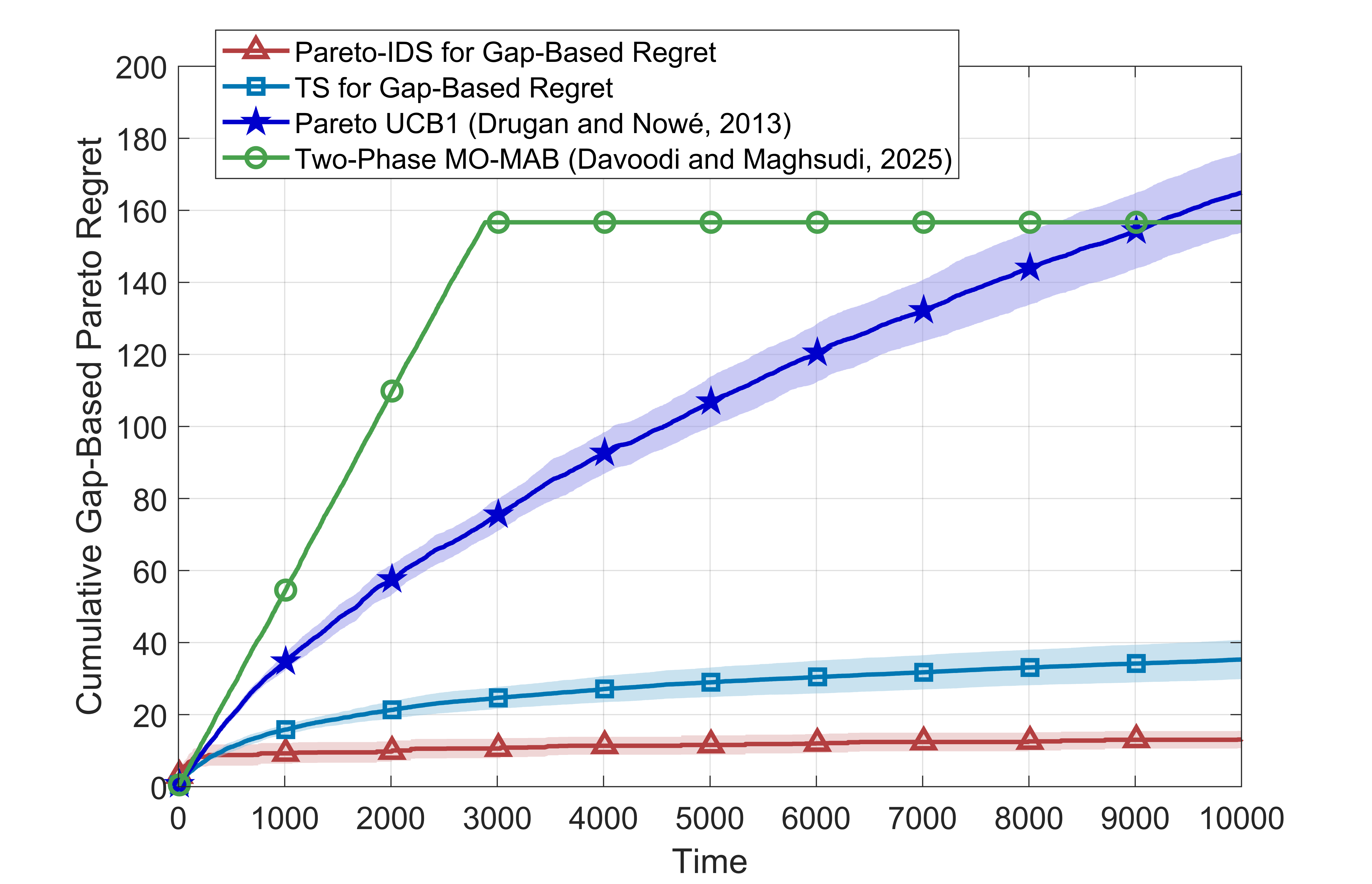}
    \subcaption{Gap-based Pareto regret}
    \label{fig:gaussian_gap}
\end{subfigure}
\caption{Additional Gaussian reward experiments with $|\mathcal A|=10$ actions and $d=4$ objectives. For each action and objective, the mean is sampled from $\mathrm{Unif}(0.05,0.95)$ and the variance from $\mathrm{Unif}(1,4)$. Rewards are generated independently across actions and objectives.}
\label{fig:gaussian_exp}
\end{figure*}

In addition to the Bernoulli experiments in Section~\ref{sec:experiments}, we evaluate the proposed methods on a continuous-reward Gaussian model. This experiment serves as a robustness check showing that the qualitative behavior of the algorithms is not specific to Bernoulli observations.

We use the same experimental setup as in the main paper, but replace Bernoulli rewards with Gaussian rewards. Specifically, we consider $|\mathcal A|=10$ actions and $d=4$ objectives. For each action $a\in\mathcal A$ and objective $j\in[d]$, the mean reward $\mu_j(a)$ is sampled independently from $\mathrm{Unif}(0.05,0.95)$, and the variance $\sigma_j^2(a)$ is sampled independently from $\mathrm{Unif}(1,4)$. Rewards are then generated independently across actions and objectives according to
\[
Y_{t,j}(a)\sim \mathcal N(\mu_j(a),\sigma_j^2(a)).
\]
Thus, both the mean and variance are action- and objective-specific, and both are unknown to the learner. The learner maintains the corresponding Gaussian posterior over the unknown reward parameters and uses this posterior in the TS and IDS computations.

We evaluate both regret criteria considered in the paper. For Pareto-violation regret, the loss at round $t$ is $\mathbbm{1}\{A_t\notin\mathcal S^*\}$, where $\mathcal S^*$ is the Pareto set induced by the true mean reward vectors. For gap-based Pareto regret, we use the same Pareto-gap definition as in the main experiments,
\[
\Delta(a)
=
\left(
\sup_{b\in\mathcal A}
\min_{j\in[d]}
\bigl(\mu_j(b)-\mu_j(a)\bigr)
\right)_+ .
\]
We compare the Gaussian versions of TS-Pareto and Pareto-IDS against the same baselines as in Section~\ref{sec:experiments}. For each method, we report the mean cumulative regret with shaded bands corresponding to one standard deviation across independent trials.

The results are shown in Figure~\ref{fig:gaussian_exp}. The qualitative ordering is consistent with the Bernoulli experiments: Pareto-IDS achieves the lowest cumulative regret, followed by TS-Pareto, while Pareto-UCB1 and the two-phase MO-MAB baseline incur larger regret. This supports the robustness of the main empirical message beyond Bernoulli rewards. In particular, IDS remains effective because it directly trades off posterior expected loss against information gain, while TS-Pareto benefits from posterior sampling but is less targeted than IDS.

\begin{figure*}[h]
\centering
\begin{subfigure}[b]{4.4cm}
\includegraphics[height=3.05cm,width=4.4cm]{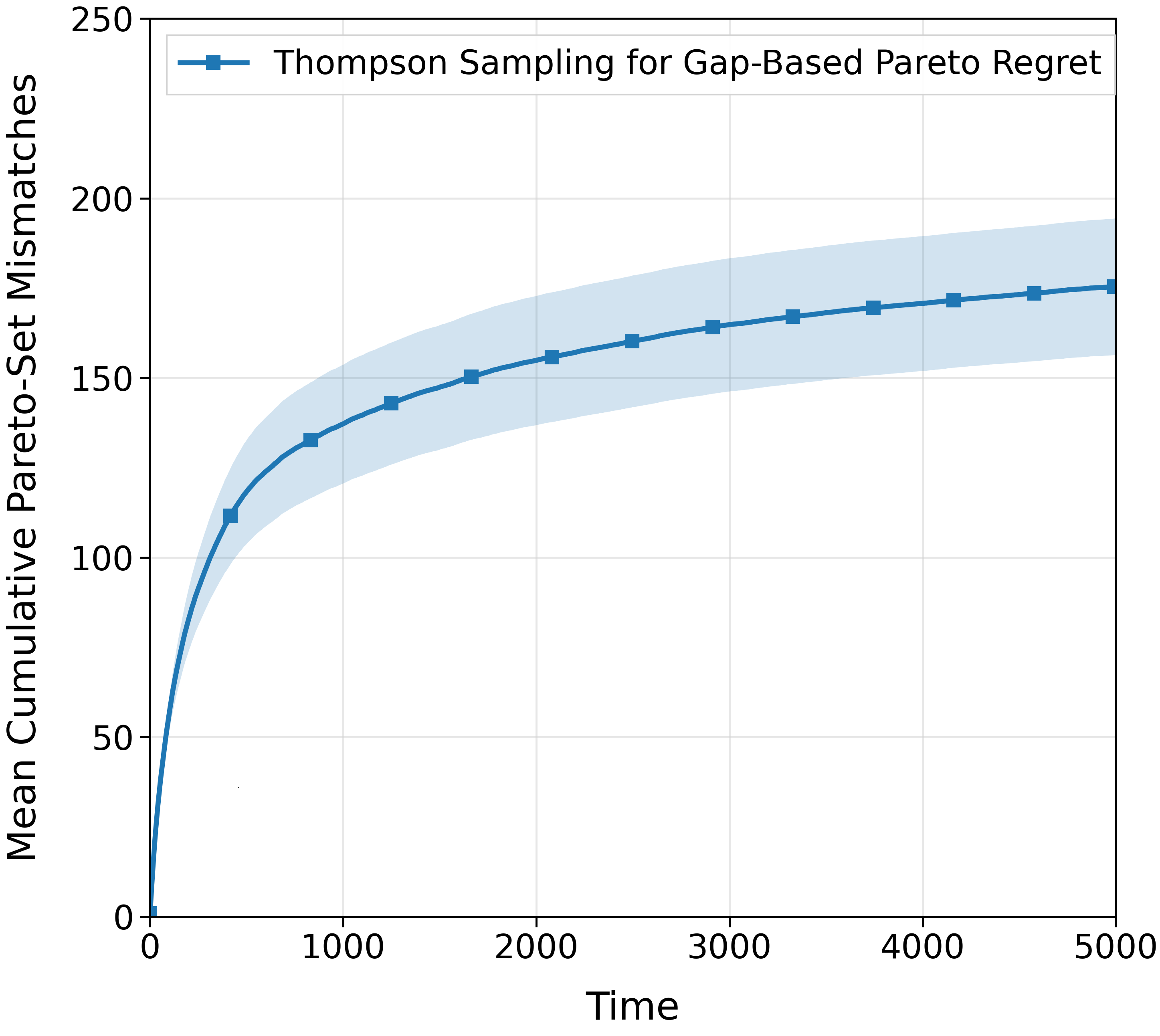}
\subcaption{$|\mathcal A|=5,\;d=2$}
\end{subfigure}
\enspace
\begin{subfigure}[b]{4.4cm}
\includegraphics[height=3.05cm,width=4.4cm]{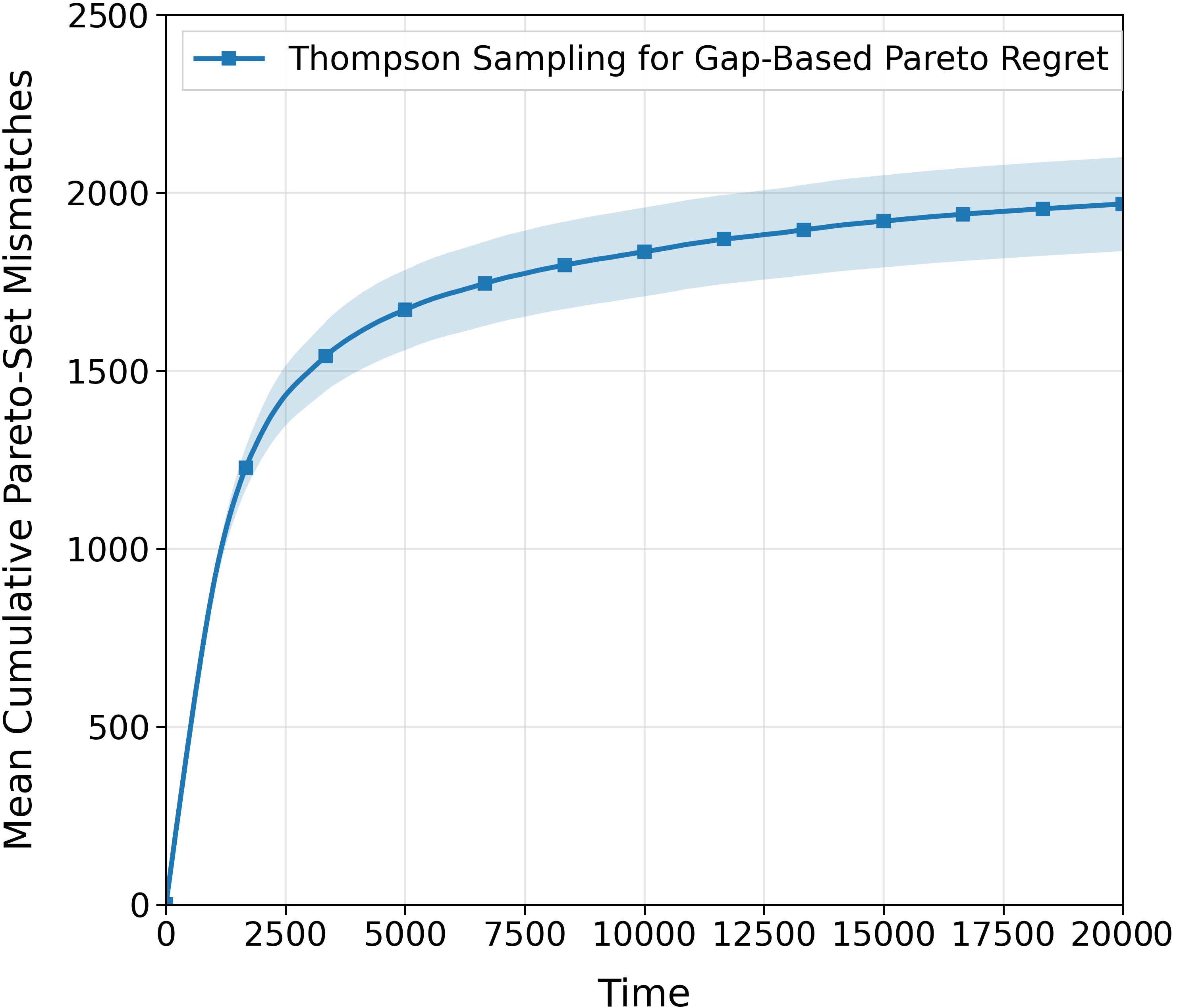}
\subcaption{$|\mathcal A|=10,\;d=4$}
\end{subfigure}
\enspace
\begin{subfigure}[b]{4.4cm}
\includegraphics[height=3.05cm,width=4.4cm]{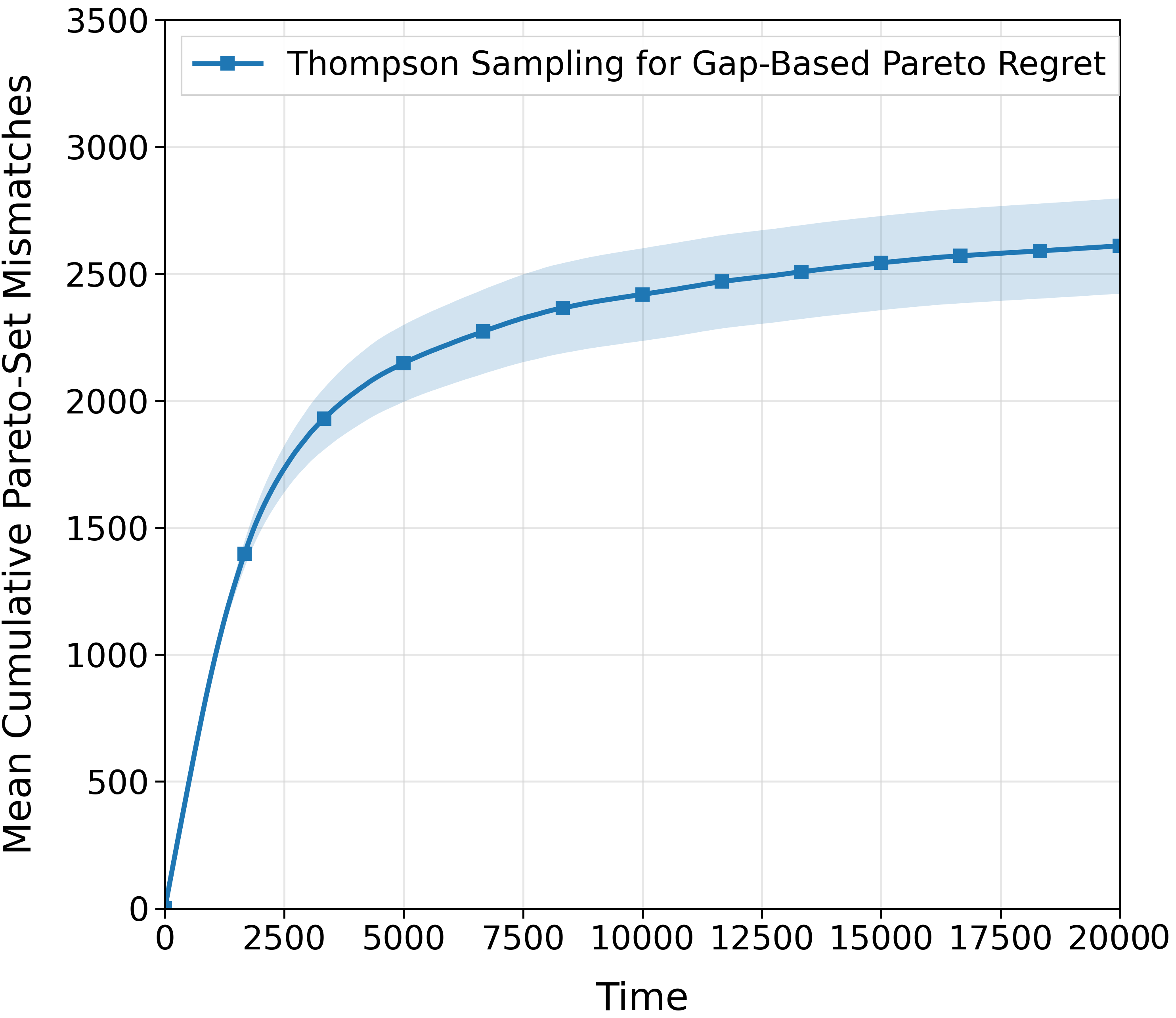}
\subcaption{$|\mathcal A|=15,\;d=5$}
\end{subfigure}

\caption{Mean cumulative Pareto-Set Mismatches on multi-objective Gaussian bandit instances}
\label{ptsxx1}

\end{figure*}
We repeat the TS-Pareto set-recovery experiment with independent Gaussian rewards, drawing the mean and variance for each action and objective from $\mathrm{Unif}(0.05,0.95)$ and $\mathrm{Unif}(1,4)$, respectively. Figure~\ref{ptsxx1} plots $\sum_{t=1}^{T}\mathbbm{1}\{S_t\ne\mathcal S^\ast\}$ averaged over $200$ independent trials per instance. Mismatches accumulate more slowly after the initial rounds and account for about $3.5\%$ of the first $5{,}000$ rounds in Figure~\ref{ptsxx1}(a) and $10\%$ and $13\%$ of the first $20{,}000$ rounds in Figures~\ref{ptsxx1}(b) and~\ref{ptsxx1}(c), respectively.

\subsection{Pareto Regret with Closely Spaced Bernoulli Means}
\begin{figure*}[h]
\centering
\begin{subfigure}[b]{4.4cm}
  \centering
  \includegraphics[height=3.05cm,width=\linewidth]{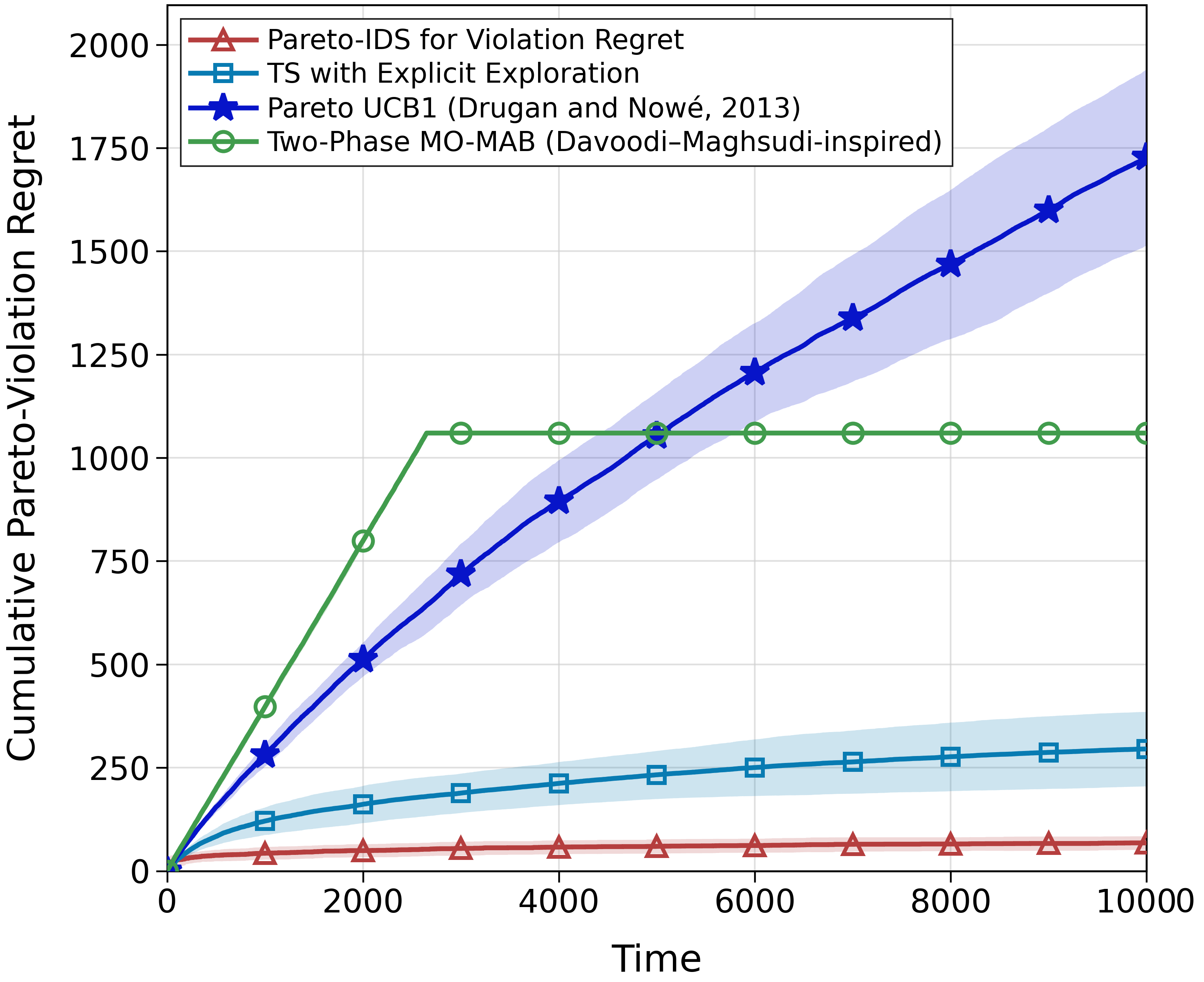}
  \subcaption{$[0.05,0.95]$, $T=10{,}000$}
\end{subfigure}
\enspace
\begin{subfigure}[b]{4.4cm}
  \centering
  \includegraphics[height=3.05cm,width=\linewidth]{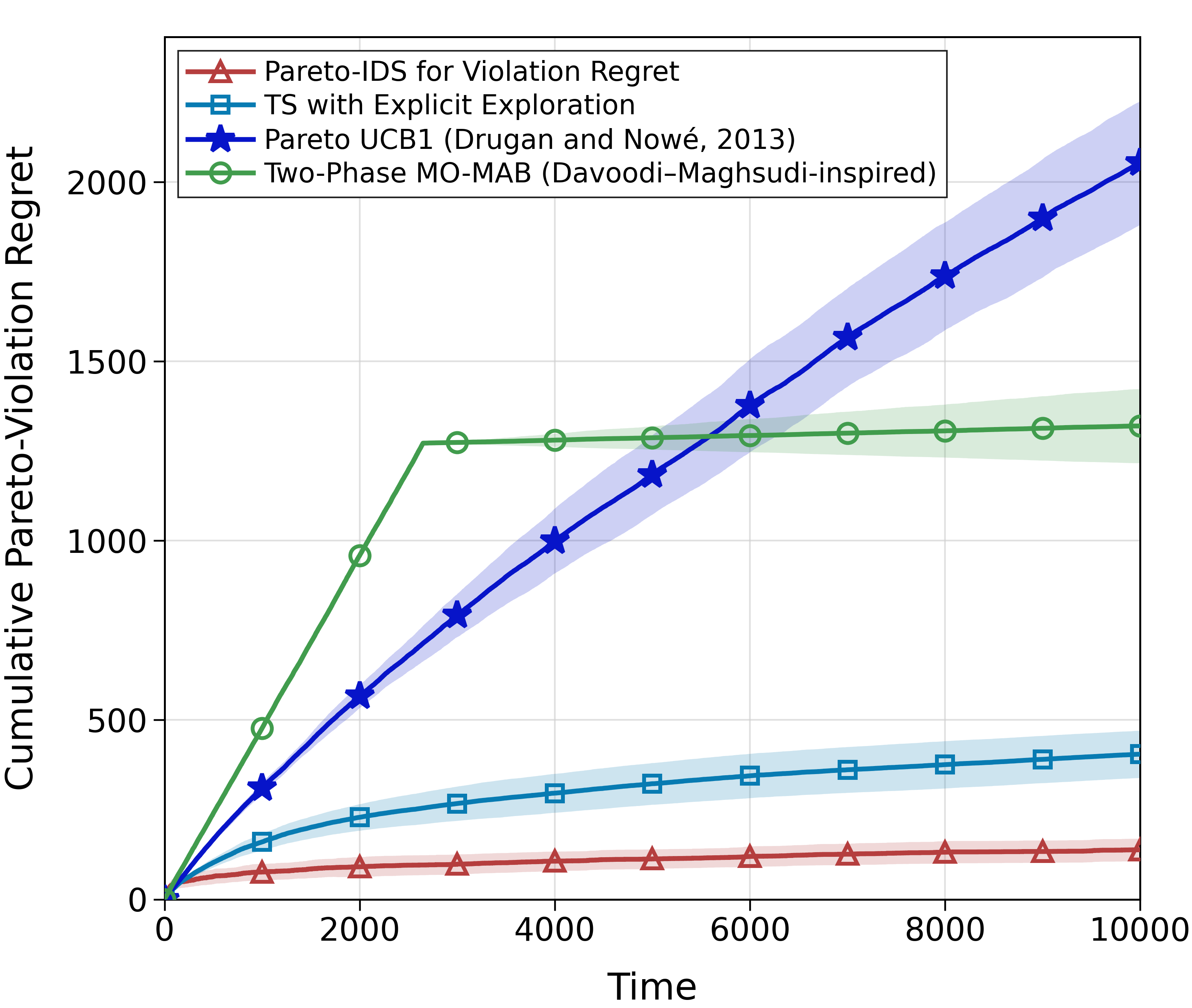}
  \subcaption{$[0.25,0.75]$, $T=10{,}000$}
\end{subfigure}
\enspace
\begin{subfigure}[b]{4.4cm}
  \centering
  \includegraphics[height=3.05cm,width=\linewidth]{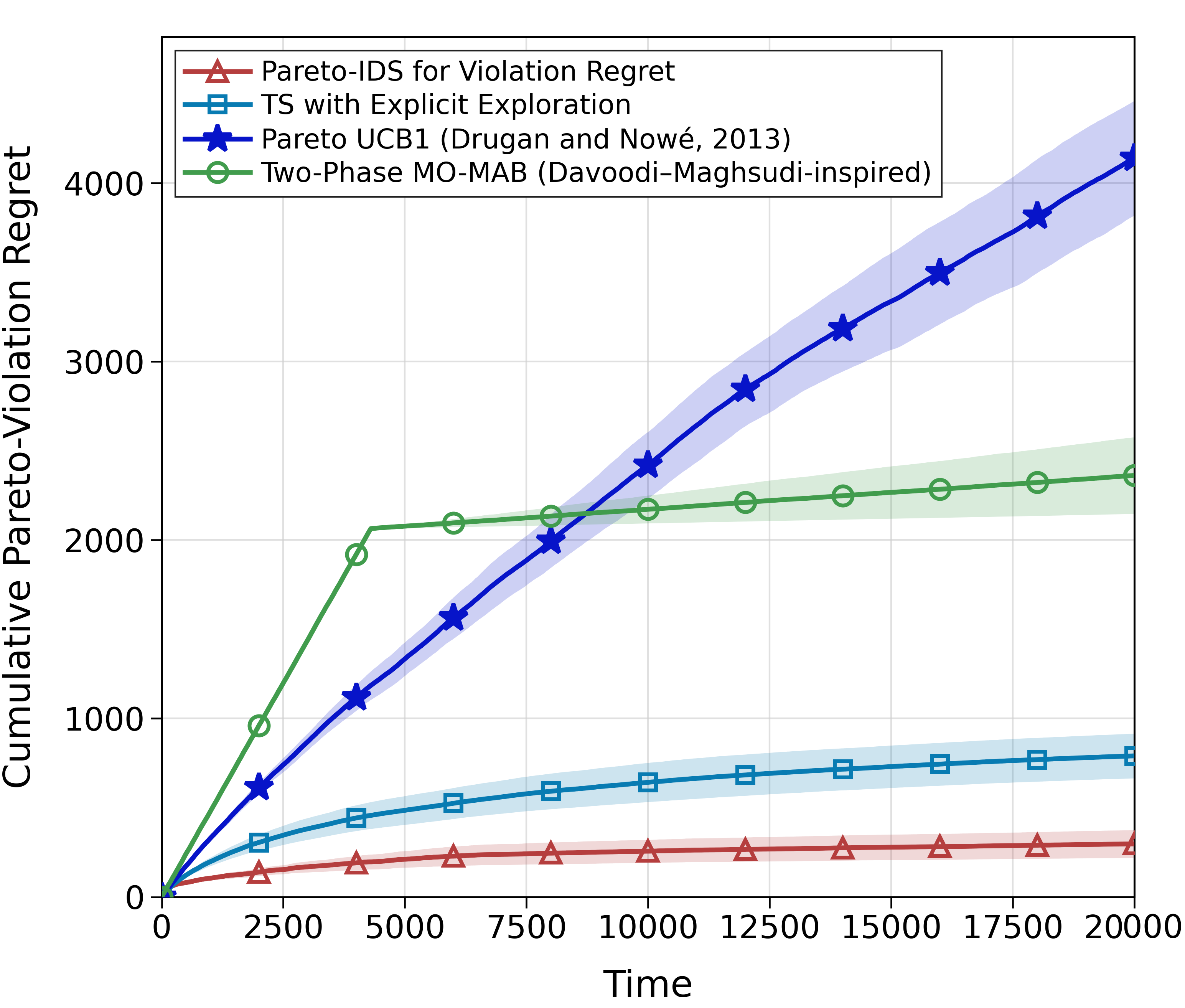}
  \subcaption{$[0.40,0.60]$, $T=20{,}000$}
\end{subfigure}

\caption{Cumulative Pareto-violation regret (Section~3) on Bernoulli
bandit instances with $|\mathcal A|=10$ and $d=4$. Each subcaption
gives the interval from which the means are sampled uniformly and
the run horizon $T$.}
\label{pts11}
\end{figure*}

\begin{figure*}[h]
\centering
\begin{subfigure}[b]{4.4cm}
  \centering
  \includegraphics[height=3.05cm,width=\linewidth]{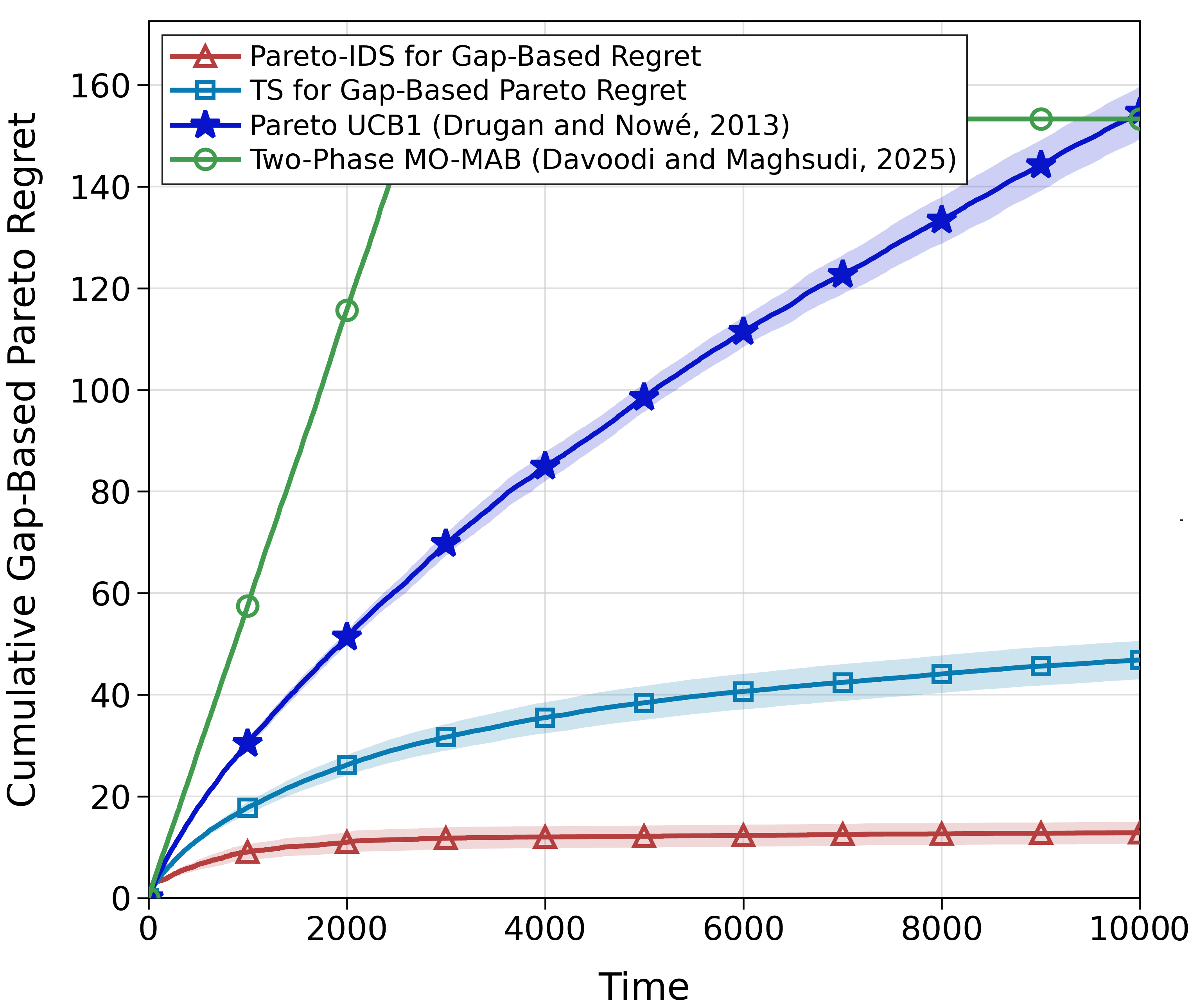}
  \subcaption{$[0.05,0.95]$, $T=10{,}000$}
\end{subfigure}
\enspace
\begin{subfigure}[b]{4.4cm}
  \centering
  \includegraphics[height=3.05cm,width=\linewidth]{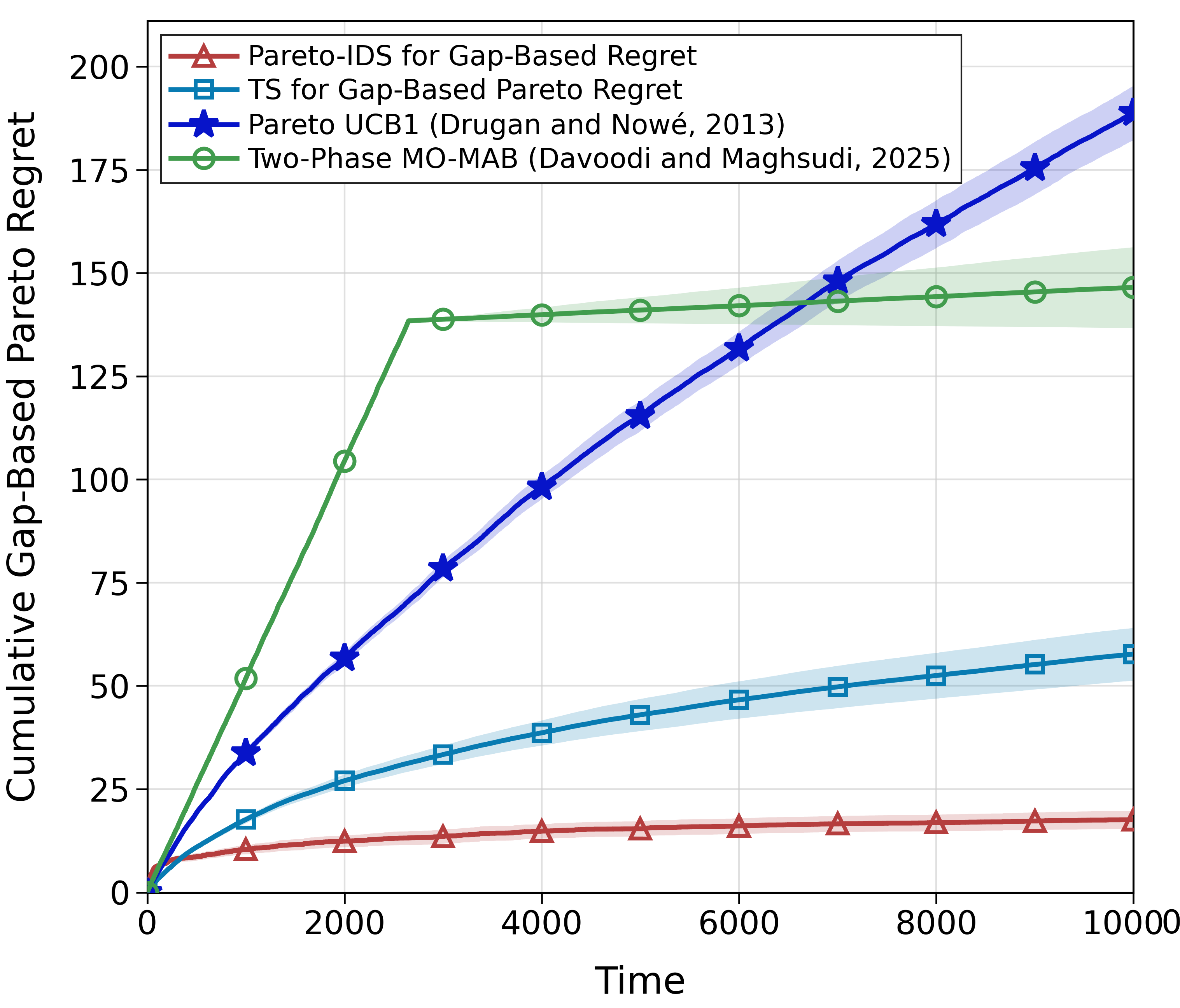}
  \subcaption{$[0.25,0.75]$, $T=10{,}000$}
\end{subfigure}
\enspace
\begin{subfigure}[b]{4.4cm}
  \centering
  \includegraphics[height=3.05cm,width=\linewidth]{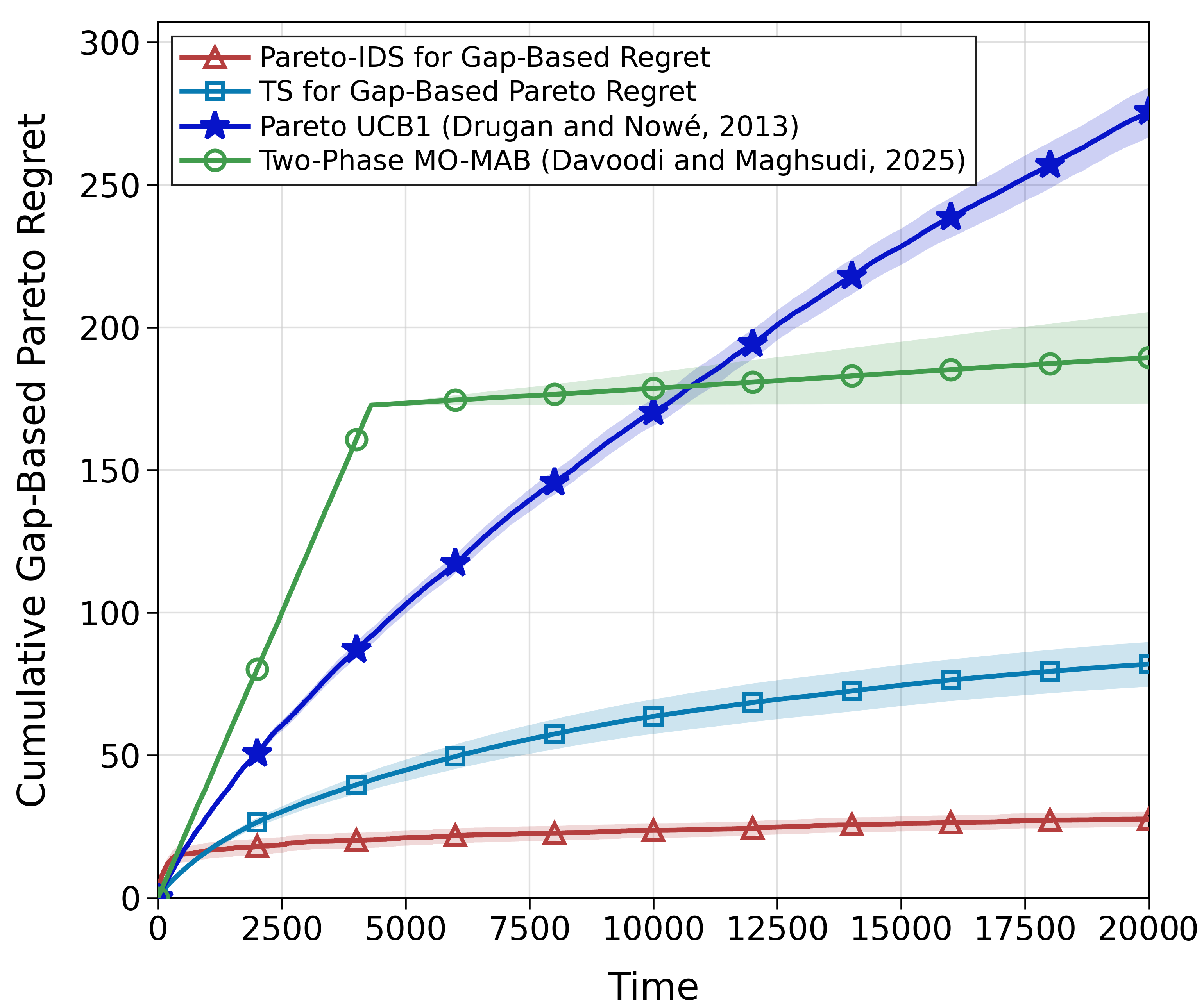}
  \subcaption{$[0.40,0.60]$, $T=20{,}000$}
\end{subfigure}

\caption{Cumulative gap-based Pareto regret (Section~4) on Bernoulli
bandit instances with $|\mathcal A|=10$ and $d=4$. Each subcaption
gives the interval from which the means are sampled uniformly and
the run horizon $T$.}
\label{fig:additional-bernoulli-gap-regret}

\end{figure*}

We test how the algorithms perform when the mean differences that
determine Pareto dominance are harder to resolve. We fix
$|\mathcal A|=10$ and $d=4$ and sample the Bernoulli mean for each arm
and objective uniformly from $[0.05,0.95]$, $[0.25,0.75]$, or
$[0.40,0.60]$, depending on the setting. Narrower intervals reduce
typical differences between arm means while concentrating them near
$1/2$, where Bernoulli rewards have the highest variance.
Figures~\ref{pts11} and~\ref{fig:additional-bernoulli-gap-regret}
compare the corresponding Pareto-IDS and TS-Pareto variants with
Pareto-UCB1 and two-phase MO-MAB under the two regret criteria.
At round $10{,}000$, all four methods have accumulated more Pareto
violations as the interval narrows. Pareto-IDS finishes with the fewest
violations in each setting, followed by TS-Pareto with explicit
exploration. The gap-based variants also finish first and second
under cumulative Pareto-gap regret. In the narrowest interval after
$20{,}000$ rounds, mean cumulative gap regret is approximately $25$
for Pareto-IDS and $80$ for TS-Pareto, compared with $190$ for
two-phase MO-MAB and $275$ for Pareto-UCB1.

\subsection{Experimental Compute Resources}
\label{compute_res}

All experiments were run on a server equipped with an AMD Ryzen Threadripper PRO 5995WX CPU, which has 64 cores and 128 threads, a base clock speed of 2.7 GHz, a maximum boost clock speed of 4.5 GHz, and 128 GB of RAM.

\end{document}